\documentclass{article} 
\usepackage{iclr2027_conference,times}

\usepackage{amsmath,amsfonts,bm}

\def\eqref#1{equation~\ref{#1}}

\def\1{\bm{1}}

\DeclareMathAlphabet{\mathsfit}{\encodingdefault}{\sfdefault}{m}{sl}
\SetMathAlphabet{\mathsfit}{bold}{\encodingdefault}{\sfdefault}{bx}{n}

\usepackage[utf8]{inputenc} 
\usepackage[T1]{fontenc}    
\usepackage{hyperref}       
\usepackage{url}            
\usepackage{booktabs}       
\usepackage{amsfonts}       
\usepackage{nicefrac}       
\usepackage{microtype}      
\usepackage{xcolor}         

\usepackage{algorithm}
\usepackage{algorithmic}

\usepackage{comment}   
\usepackage{amsmath, pifont}    
\usepackage{amssymb}
\usepackage{dsfont}    
\usepackage{graphicx}  
\usepackage{colortbl}  
\usepackage{multirow}  
\usepackage{multicol}  
\usepackage{wrapfig}  
\usepackage{rotating}   
\usepackage{tablefootnote} 
\usepackage{caption}     
\usepackage{subcaption}  
\usepackage{titletoc}  
\usepackage[title]{appendix}
\setcitestyle{square}  

\usepackage{multibib}   
\newcites{supp}{References}

\usepackage{caption}
\newcommand{\X}{\textbf{X}}  
\newcommand{\D}{\textbf{D}}  
\newcommand{\B}{\textbf{B}}  
\newcommand{\A}{\textbf{A}}  
\newcommand{\C}{\textbf{C}}  
\newcommand{\I}{\textbf{I}}  
\newcommand{\cmark}{\ding{51}} 
\newcommand{\xmark}{\ding{55}} 

\def\1{\mathbbm{1}}
\newcommand{\RR}{\mathbb{R}}
\newcommand{\ZZ}{\mathbb{Z}}

\newcommand{\NN}{\mathbb{N}}

\newcommand{\gray}[1]{\textcolor{gray}{#1}}

\definecolor{colorw}{RGB}{0,102,102}

\definecolor{colorh}{RGB}{102,0,102}
\newcommand{\colorh}[1]{\textcolor{colorh}{#1}}
\definecolor{colory}{RGB}{255,0,255}

\definecolor{colorx}{RGB}{0,0,128}

\definecolor{colorbetter}{RGB}{0,128,128}
\newcommand{\colorbetter}[1]{\textcolor{colorbetter}{#1}}
\definecolor{colorworse}{RGB}{153,0,153}
\newcommand{\colorworse}[1]{\textcolor{colorworse}{#1}}

\definecolor{colorhres}{RGB}{0,64,64}

\definecolor{shapecolor}{RGB}{102,0,102}

\definecolor{colorm}{RGB}{110,84,121}
\definecolor{colorf}{RGB}{80,22,74}

\arrayrulecolor{colorf} 

\newcommand{\XANA}{\textbf{X}_{ANA}}  
\newcommand{\XEMO}{\textbf{X}_{EMO}}  
\newcommand{\XIMERG}{\textbf{X}_{IMERG}}  
\newcommand{\XCPC}{\textbf{X}_{CPC}}  
\newcommand{\XHRES}{\textbf{X}_{HRES}}  
\newcommand{\XALPHA}{\textbf{X}_{AlphaE}}  
\newcommand{\XF}{\textbf{X}_{features}}  

\def\1{\mathbbm{1}}

\title{Distributed Hydrological Modeling in the Feature Space}

\author{
Mohamad Hakam Shams Eddin, Shijie Jiang, \& Markus Reichstein\\
Max Planck Institute for Biogeochemistry, Germany\\
\texttt{\{mshams,sjiang,mreichstein\}@bgc-jena.mpg.de} \\
\And
Maria Luisa Taccari \\
European Centre for Medium-Range Weather Forecasts (ECMWF), United Kingdom \\
\texttt{marialuisa.taccari@ecmwf.int} \\
\And
Juergen Gall \\
University of Bonn, Germany\\
Lamarr Institute for Machine Learning and Artificial Intelligence, Germany\\
\texttt{gall@iai.uni-bonn.de}\\
\And
Yikui Zhang\\
University of Bonn, Germany\\
Institute of Bio- and Geosciences Agrosphere (IBG-3), Research Centre J\"ulich, Germany\\
\texttt{yik.zhang@extern.fz-juelich.de}\\
}

\def\logo{\makebox[44pt][c]{\raisebox{-3.0ex}{\includegraphics[height=56pt]{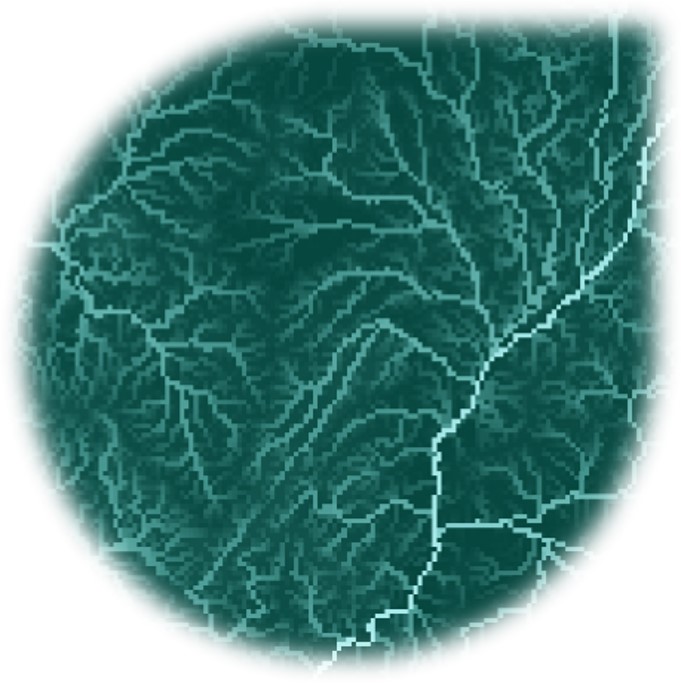}}}\hspace{20pt}}

\iclrfinalcopy 
\begin{document}

\maketitle


\begin{abstract}

Accurate forecasting of river discharge and floods is very challenging. 
River dynamics are affected by storage, meteorological forcing, and flow propagation at different spatial and temporal scales.
Forecasting thus requires a framework that considers the upstream-to-downstream flow through river networks across grid cells and catchments.
This modeling is known in hydrology as distributed modeling and routing. Existing deep learning approaches either ignore this topology, operate on lumped catchments, or route predicted physical quantities through a separate graph or physical routing model. We instead introduce feature-space routing: a topology-aware state-space operator embedded directly in the forecasting dynamics. At every forecast step, the operator gathers latent states from upstream grid cells and causally updates the downstream state according to the known river network. This preserves the physical connectivity of the river system while allowing the propagated state itself to be learned end-to-end and allows the model to predict river discharge considering both local dynamics and neighboring upstream contributions. To address uncertainty and provide probabilistic forecasts, we minimize the fair continuous ranked probability score (fCRPS) as a training objective.
Our experiments on the European Flood Awareness System (EFAS) and observational data for river discharge forecasting demonstrate that encoding the physical structure of river networks explicitly in the feature space substantially improves the forecasting skill, particularly in an ungauged setting. 
Our approach achieves state-of-the-art results on both reanalysis and observational data and is able to forecast maps of river discharge at 1 arcminute and 6-hourly resolution up to 10 days lead time.
\end{abstract}

\begin{figure}[!h]
  \centering
  \includegraphics[width=.9\textwidth]{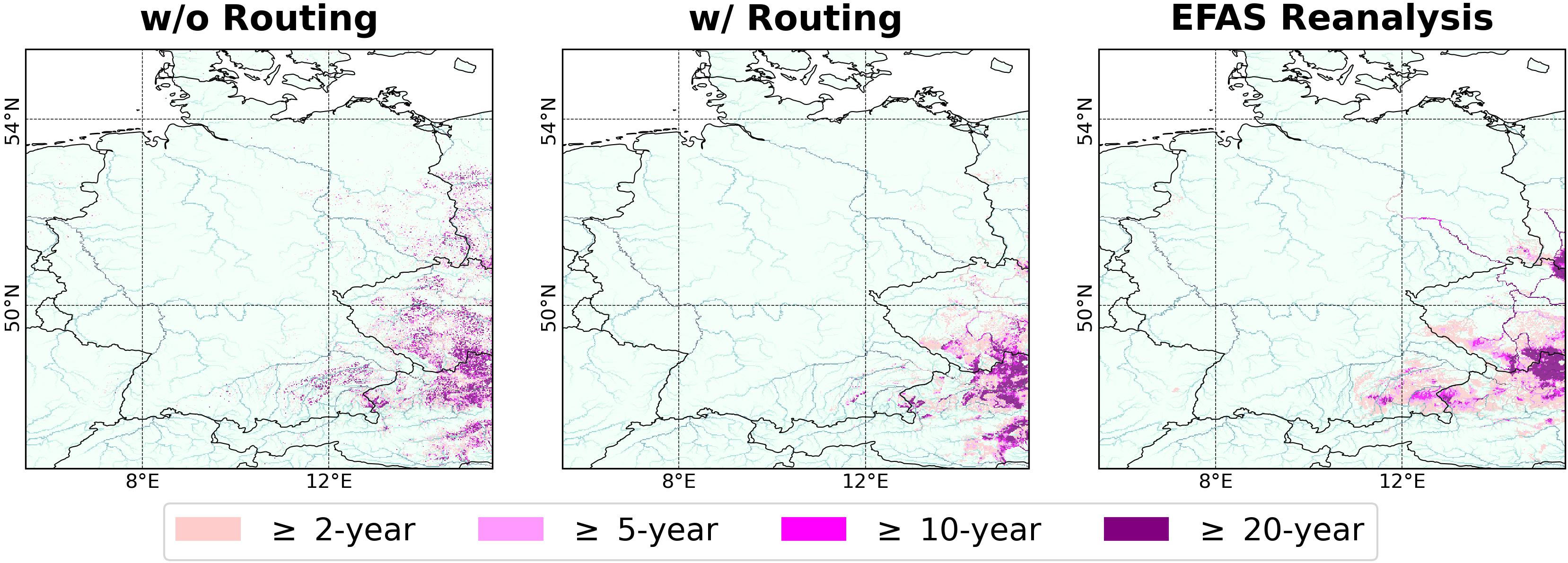}
  \caption{Example river discharge and flood forecasting over mid Europe in Sep.~2024. 
  In the middle is the forecast produced by our model with routing at 4-day lead time, compared to the same model but without routing mechanism on the left and EFAS reanalysis data as a reference on the right side. The flood categories are shown below, e.g., high river discharge with $20$-year return period.
  Without explicit distributed modeling, the routing-free model produces heterogeneous predictions in which neighboring pixels can have inconsistent discharge, leading to many false positives. Our routing module instead yields more spatially coherent predictions that better follow the river network.}
  \label{fig:teaser}
\end{figure}

\section{Introduction}
Accurate flood forecasting in rivers is essential for early warning and protecting citizens. 
In many operational systems, the forecast must infer the flow magnitude and propagation from uncertain meteorological forcing and limited hydrological initial conditions \citep{GloFAS_Forecast}. 
In a nutshell, river discharge forecasting combines two geometries: meteorological forcing and catchment properties are naturally represented as spatial fields, while water propagates through a sparse directed river network. 
Existing approaches typically privilege one of these representations. Classical hydrological models address this by parametrized calibration and forcing physical laws on the modeling of runoff generation and river routing. 
However, physics-based models are expensive to run and require extensive calibration efforts.
For instance, operating the European Flood Awareness System (EFAS) for real-time operations is computationally expensive and it is only provided for authorized users~\citep{EFAS}. 
Therefore, deep learning approaches have become more attractive for early warning systems of floods~\citep{nearing2021role}.
While they achieve promising results, most existing models are lumped models which simplify the modeling and treat the whole catchment as a single entity~\citep{Google_nature}. Yet, this neglects key drivers needed for accurate flood prediction. 
Only few works touched the distributed modeling and routing with deep learning at large-scale~\citep{RiverMamba, yang2025global}. 
These approaches, however, either do not achieve the required spatio-temporal resolution required for early warnings or are limited in their capabilities in modeling the flow propagation within and through catchments. For instance, \citet{yang2025global} use a physics-based routing module but the approach is limited to 0.25$^\circ$, not accounting for rapidly evolving dynamics and fine-scale characteristics of the catchments. The routing is also a post-processing step and not part of the training of the LSTM runoff model. While \citep{RiverMamba} is much more efficient and capable of operating on 0.05$^\circ$ and daily resolution, it does not explicitly represent the directional flow connectivity between upstream and downstream locations. 

In this work, we thus propose to encode the routing explicitly in the model by applying a State Space Model (SSM)~\citep{mamba2} directly on the graph structure of the river networks. 
Using a specialized algorithm to convert the grid points into 1D sequences, the SSM-based routing module propagates upstream-to-downstream information in the feature space while the direction of propagation is constrained by the known river topology. This novel hybrid modeling approach has the advantage that it remains very efficient while capturing flow propagation in a more realistic way than purely machine learning approaches. While the dense grid representation is retained, the learning of the latent representation and its dynamics is guided by the directional flow connectivity, which improves the forecast as shown in Fig.~\ref{fig:teaser}. Building upon the proposed routing module, we introduce the first deep learning approach that is capable of forecasting high-resolution probabilistic river discharge and floods at $1$-arcminute and sub-daily resolution up to $10$ days at continental-scale and in an operationally compatible setting. 
To address uncertainty, we optimize the model with the fair Continuous Ranked Probability Score (fCRPS) as a training objective. 
Our contributions can be summarized as follows:
\begin{itemize}
    \item We introduce a novel feature-space routing, a directed state-space operator that injects known network topology into dense auto-regressive forecasting by propagating learned latent states from upstream to downstream grid points. 
    
    \item We show through controlled ablations against no routing and standard GNN aggregation that this structural inductive bias substantially improves discharge prediction, flood detection, and particularly spatial generalization to ungauged locations.
    
    \item We introduce the first operational deep learning approach that is optimized with a score-based objective function and capable of providing probabilistic maps of river discharge forecast at $1$ arcminute and $6$-hourly resolution, and up to $10$ days lead time over Europe.

\end{itemize}
\section{related works}
\label{sec:related_works}
\textbf{River Discharge and flood forecasting.}
We are interested in floods that occur in rivers. Generally, such floods are termed as \textit{Fluvial} or \textit{riverine} flood. They occur when the river flow increases and the water level rises high enough to overflow onto the surrounding land~\citep{Bomers_2023}. Such flood events can be detected from river discharge itself, whenever the flow exceeds a defined threshold. 
Throughout this work, flood severity refers to exceedance of discharge return-period thresholds instead of inundation depth or extent~\citep{Jiang2024}. 
Deep learning has become common for forecasting river discharge in hydrology~\citep{Slater2025}. Most existing approaches rely on a lumped model trained jointly across many catchments, for instance an LSTM~\citep{Kratzert_2018, EA-LSTM, hess-28-4187-2024, ruparell2024hydra}.
A key limitation here is that a lumped model represents an entire catchment as one undifferentiated unit, using averaged meteorological forcing and landscape features over the catchment.
Yet catchments are internally heterogeneous, both in their hydrological processes and in the structure of their river networks. This induces spatio-temporal causal dependencies within and across catchments~\citep{Merz2022, stein2025causalrivers}. 
Therefore, some studies start adapting Graph Neural Networks (GNNs) to model the topology along river networks~\citep{Sun_2021}. But, they are still trained on small scale river networks and limited in term of generalizability~\citep{kirschstein2024merit}. 
Most recently, RiverMamba proposed a new methodology based on State Space Model (SSM) to model large spatio-temporal contexts and forecast river discharge at grid-scale~\citep{RiverMamba}. 
However, RiverMamba has limited forecast horizon due to the fixed number of forecasting heads i.e., it has no auto-regressive mechanism and it does not encode the upstream-to-downstream flow propagation. 
This process is known in hydrology as distributed modeling and river routing.

In this work, we propose a distributed modeling mechanism that is encoded explicitly inside the forecasting engine. In addition, we build our model on SSM and based the forecast on rollout concept to achieve long-term forecast at sub-daily resolution.

\textbf{Distributed hydrological modeling and routing.}
Classical hydrological models represent routing and spatial heterogeneity using semi- and fully-distributed formulations~\citep{Freeze1969}. Semi-distributed models split the catchment into sub-catchments and perform routing along the river network, whereas fully-distributed models such as LISFLOOD~\citep{LISFLOOD} discretize the catchment into grid cells and solve physics-based governing equations for routing at each grid point. Dedicated distributed routing models such as CaMa-Flood~\citep{Yamazaki2011} similarly propagate runoff through prescribed river networks. 
Such models, however, remain expensive to run at high resolution and require extensive calibration~\citep{hirpa2018calibration}.
There have been attempts in hydrology to design deep learning models capable of fully spatially distributed modeling. Typical approaches estimate runoff, then route it downstream using either a physics-based routing module~\citep{yang2025global, kraft2025drop} or another neural
network~\citep{Vischer_2025, Mosaffa_2026}. 
\citet{Hascoet_2026} instead reformulate linear time-invariant routing as a block-sparse convolution, keeping it physically constrained but differentiable and trainable end-to-end with a runoff model. 
\citet{Scholz_2025} propose an LSTM coupled with a CNN to learn runoff embeddings, which a GNN then propagates over the river network topology to route flows between stations. 
\citet{Li_2026} apply the same approach but use CNNs to represent the spatial runoff. 
Other works use hybrid frameworks to learn the parametrization of physical equations including river routing~\citep{Wang_2024, Bindas_2024, Song_2025}. 

In our work, we propose a novel method in hydrology based on SSMs to perform fully spatially distributed modeling. Rather than coupling a grid-based runoff model with a physics-based routing or a separated neural network operating on a coarser river graph, we use a single SSM architecture to both process the spatio-temporal grid and route high-dimensional features directly at the native grid resolution within one forecasting engine.

\textbf{Probabilistic river discharge forecasting.} 
Reliable flood prediction depends on uncertainty estimation~\citep{Baste_2026}. 
This includes, aleatoric and epistemic uncertainties because measurements and initial conditions are inherently uncertain.
Physics-based systems such as GloFAS~\citep{GloFAS_Forecast} and EFAS perturb their inputs using ensemble weather forecast to account for these uncertainties. 
In deep learning for hydrology, \citet{Klotz_2022} introduced mixture density networks. 
This approach is also adopted by Google's flood forecasting systems~\citep{Google_nature}. Another approach is to model the distribution of discharge with denoising diffusion probabilistic models~\citep{DRUM_2025, wang2025hydrodiffusion} or variational auto-encoder~\citep{JAHANGIR2024130498}. 
More recently, \citet{Kraft_2026} and \citet{ruparell2026ai} proposed a score-based generative approaches and paired it with an LSTM for rainfall-runoff prediction. 
\citet{ruparell2026joint} demonstrate that moving from lumped to distributed probabilistic hydrology demands explicit modeling of the spatial joint structure underlying runoff uncertainty. 

In our work, inspired by recent advances in weather forecasting research~\citep{alet2025skillful, lang2026aifs}, we propose the fair Continuous Ranked Probability Score (fCRPS) as a training objective. 
To the best of our knowledge, ours is the first deep learning approach optimized with fCRPS in a fully spatially distributed modeling framework.
\vspace{-1mm}
\section{Method}
\label{sec:method}
In this work, we present a novel approach for distributed hydrological modeling along with the first deep learning model that is able of generating probabilistic river discharge and flood forecasting up to $10$ days at $1$-arcminute and $6$-hourly spatio-temporal resolution (Fig.~\ref{fig:teaser}). This is a very challenging task since it requires an approach that should not only learn the catchment dynamics locally, but also considers the upstream-to-downstream flow propagation through river networks to resolve heterogeneity at fine scale. We thus propose a novel feature-space routing. It consists of a directed state-space operator that injects known network topology into a network by propagating learned latent states from upstream to downstream nodes. 
We will first briefly describe the Structured State Space Duality (SSD) in Sec.~\ref{sec:ssd} and the routing in Sec.~\ref{sec:routing}. In Sec.~\ref{sec:model_components}, we describe the entire network.               

\vspace{-1mm}
\subsection{Structured State Space Duality (SSD)}
\label{sec:ssd}

Mamba2 or SSD~\citep{mamba2} is a state space model that transforms a 1D sequence of states $x(t)$ into another representation $y(t)$ through an implicit hidden latent state $h(t)$:
\begin{align}
    h_t = \bar{\A}h_{t-1} + \bar{\B}x_t\,, ~~~~ y_t = \C h_{t} + \D x_t\,, \label{eq:ssd}\\
    \bar{\A} = e^{(\Delta\textbf{A})}\,, ~~~~ \bar{\B} = (\Delta \A)^{-1} (e^{(\Delta\textbf{A})}-\I)\Delta \B \,, \label{eq:ssd_disc}
\end{align}
where $\A, \B, \C,$ and $\D$ are parametrized matrices discretized with a timescale parameter $\Delta$.
$\B(x)$, $\C(x)$, and $\Delta(x)$ depend on the input state $x$ which makes the system selective time-variant, i.e., it is able to control whether to keep or ignore some information along the sequence. 
SSD is considered as an efficient alternative to Transformers~\citep{Transformer} for processing long sequences of points since it has a linear complexity w.r.t.\ the input tokens. In SSD, $\A$ is restricted into a scalar-times identity, $\A = a\,\I$. This restriction enables matrix multiplication and the recurrence in Eq.~\ref{eq:ssd} to be state-space dual, equivalently to causal linear attention. 
In Eq.~\ref{eq:ssd}, $x$ can be high dimensional and the dynamic of SSD can be used independently per group of channels (head). This is analogous to multi-head attention in Transformer~\citep{Transformer}. 
The sequence transformation can thus be computed equivalently either as a sequential scan or as a chunked matrix multiplication over blocks of timesteps.

\vspace{-1mm}
\subsection{Routing}
\label{sec:routing}
In hydrological models, channel routing propagates discharge through the river network, with locally generated surface and subsurface runoff entering the channel as lateral inflow. Each grid point receives upstream contributions, combines them with local inflow, and propagates the resulting flow downstream. We adopt the same structure in our model, but route learned high dimensional \emph{features} $x_p$ rather than physical quantities. For each point $p$, a routing module aggregates the features of $p$'s upstream neighbors and processes them jointly with $p$'s own features via a selective State Space Duality (SSD). 
\begin{figure}[!t]
\centering
\includegraphics[width=.99\linewidth]{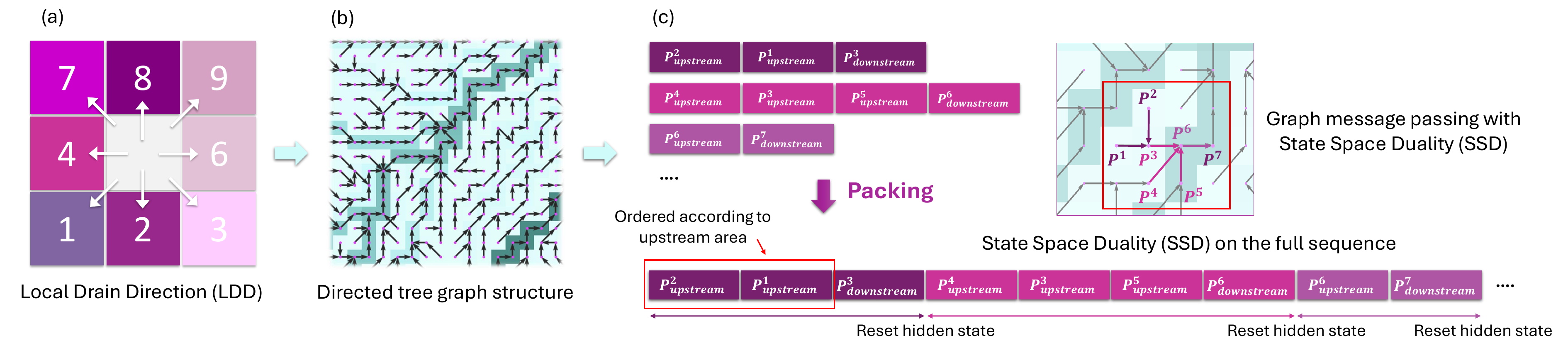}
\caption{Overview of the routing module. (a) Shows the Local Drain Direction (LDD) which is used to generate the graph of river network (b). Each downstream point forms a sequence with its upstream neighbors and routing is conducted via SSD on concatenated point sequences at once (c).}\label{fig:routing}
\end{figure}
An overview is shown in Fig.~\ref{fig:routing}. 
The routing structure comes from the Local Drain Direction (LDD), a static field derived from terrain and hydrographic information that assigns each point exactly one downstream direction i.e., one of $9$ values (Fig.~\ref{fig:routing}~(a)).  
Inverting this map gives, for every point, up to $8$ upstream neighbors, one per compass direction, which together define a directed drainage graph representing the river network (Fig.~\ref{fig:routing}~(b)). 
We propose to conduct routing over this graph as message passing implemented via SSD. For each point $p$, we form a short sequence of $p$'s upstream neighbors followed by $p$ itself. Formally, for point $p$ with $j(p)$ upstream neighbors: 
\begin{align}
    s(p) &= \big(n_1(p), n_2(p), \ldots, n_{j(p)}(p),\, p\big)\,, \label{eq:routing_seq} \\
    d_i(p) &=
    \begin{cases}
        \text{LDD}\big(n_i(p)\big)\,, & i \le j(p) \\
        5\,, & i = j(p)+1\,,
    \end{cases} \label{eq:routing_slot} \\
    u_i(p) &= x_{s(p)_i} + PE_{d_i(p)}\,, \qquad i = 1, \ldots, j(p)+1\,, \label{eq:routing_input} \\
    x_p &\leftarrow x_p + \Big[\text{SSD}_\theta\big(u_1(p), \ldots, u_{j(p)+1}(p)\big)\Big]_{j(p)+1}\,, \label{eq:routing_ssm}
\end{align}
where $n_i(p)$ is the $i$-th upstream neighbor of $p$ in ascending order of upstream area ($j(p)\le 8$), $s(p)_i$ is the $i$-th entry of $s(p)$, $x_{s(p)_i}$ is the feature vector of the $i$-th entry, and $u_i(p)$ is the input token of the $i$-th entry which will be given to the SSD. 
$d_i(p)$ is the direction code of the $i$-th token. For an upstream neighbor it is the neighbor's own LDD value, i.e.~the direction in which it drains into $p$ (one of the $8$ codes $\{1,2,3,4,6,7,8,9\}$), while $p$ itself receives code $5$, which marks it as the sequence's downstream point rather than a neighbor. $PE_{d} \in \mathbb{R}^{K}$ is a learned direction encoding, added to the raw feature $x_{s(p)_i}$ to form the SSD input token $u_i(p)$. 
Only $\text{SSD}_\theta$'s output at the final position, corresponding to $p$ itself, is added residually to $x_p$. 
Since the sequence length $j(p){+}1$ varies between points, the sequences of all points are packed into one long sequence rather than run one at a time as illustrated in Fig.~\ref{fig:routing}~(c).  
To prevent state leakage across segments, every token is tagged with the index of its own segment, $\text{seg}(u_i(p)) = p$, and the hidden state is reset whenever segment changes:
\begin{align}
    h_i = \mathds{1}\big[\text{seg}(u_i) = \text{seg}(u_{i-1})\big]\, \bar{\A}\, h_{i-1} + \bar{\B}\, u_i\,,
    \label{eq:routing_reset}
\end{align}
where the indicator evaluates to $0$ at the first token of every point's sequence and to $1$ elsewhere. 
This makes $\text{SSD}_\theta$ applied to the packed sequence exactly equivalent to running it independently on each point's sequence, while naturally handling the variable sequence lengths. 
 The routing update is local and can be evaluated for all grid points in parallel, repeated application at each auto-regressive forecast step allows routed information to propagate downstream over time. A point with no upstream neighbors simply forms a sequence of length one. Because Mamba's scan is causal, a point is only influenced by points earlier in the sequence, so we place $p$ at the end of the sequence which give it a receptive field over all upstream neighbors. 
Moreover, since the SSD's hidden state decays with distance, neighbors placed closer to $p$ \emph{in the sequence} have a stronger, less-decayed effect on $p$'s aggregated representation than neighbors placed further away. 
We use upstream drainage area as a hydrologically informed ordering proxy for a neighbor's importance to $p$, and order neighbors by ascending area, so that the largest upstream area point sits immediately before $p$ and receives the strongest influence. 
Unlike permutation-invariant neighbour pooling, the SSD operator maintains a learned recurrent state over upstream contributions. Ordering the sequence by upstream area permits to learn nonlinear, context-dependent feature aggregation rather than reducing upstream information through a fixed sum or mean. 
Note that the LDD we used here is the same as in the LISFLOOD hydrological model, where each point has one downstream direction. Our formulation is nonetheless general and applies unchanged if points were instead assigned multiple downstream directions or a dynamic LDD.
We use this proposed routing operator in the forecasting engine of our model to propagate spatio-temporal routing latent states auto-regressively (Sec.~\ref{sec:model_components}).

\subsection{Model architecture}
\label{sec:model_components}

\begin{figure}[!t]
  \centering
  \includegraphics[width=.98\textwidth]{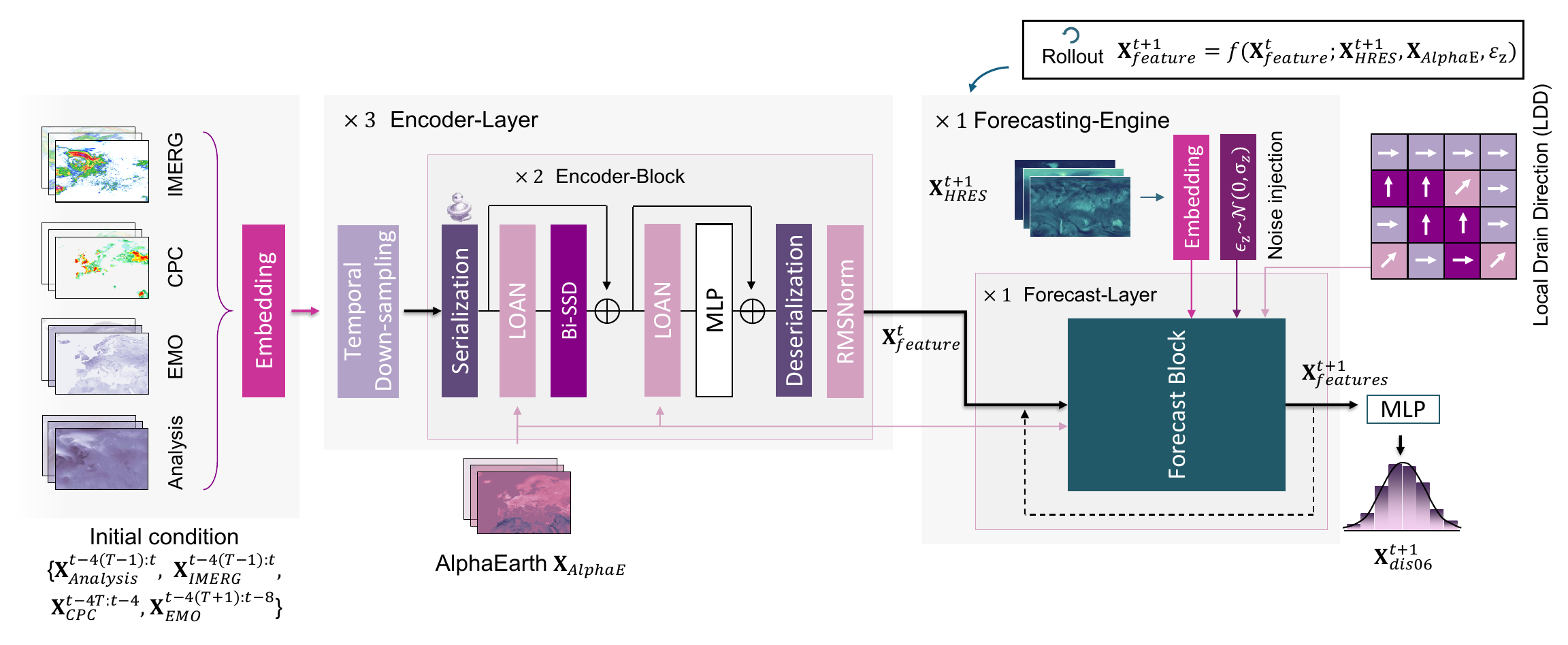}
  \caption{An overview of the proposed model for river discharge forecasting. 
  The model initializes the forecast at time $t$ and forecasts high-resolution river discharge maps $\X_{dis06}^{t:t+L}$ from the initial conditions ($\XANA^{t-4(T-1):t}$, $\XIMERG^{t-4(T-1):t}$, $\XCPC^{t-4T:t-4}$, $\XEMO^{t-4(T+1):t-8}$), AlphaEarth embedding ($\XALPHA$), and weather forecasts ($\XHRES^{t+1:t+L}$), conditioned on a noise vector $\epsilon_z$ to generate probabilistic forecast. 
  }\label{fig:Model}
\end{figure}

An overview of our entire model, which includes the routing (described in Section \ref{sec:routing}), is shown in Fig.~\ref{fig:Model}. It integrates as initial conditions different sources at daily temporal resolution. We start the forecast at time $t$ at 00:00 UTC and use Analysis data from the ECMWF Integrated Forecast System (IFS) $\{\XANA^{t-4i}\}_{i=0}^{T-1}$, the IMERG data $\{\XIMERG^{t-4i}\}_{i=0}^{T-1}$ as a satellite-based daily precipitation estimate~\citep{IMERG}, CPC data as a gauge-based daily precipitation $\{\XCPC^{t-4i}\}_{i=1}^{T}$~\citep{xie2010cpc}, and the EMO gridded ground weather stations $\{\XEMO^{t-4i}\}_{i=2}^{T+1}$~\citep{thiemig2022emo5}, where $T=4$ is the number of daily samples and $t$ is at $6$-hourly temporal resolution. Note that each data has a different latency according to their near real-time acquisition.

Additionally, we include AlphaEarth~\citep{AlphaEarth} embeddings $\XALPHA$ as static features. 
Furthermore, since forecasting river discharge is highly dependent on the meteorological forcing leading the forecast, we use medium-range and high-resolution weather forecast from IFS up to $10$ days ($t=40$) and denoted as $\{\XHRES^{i}\}_{i=t+1}^{t+L}$, where $L=40$ denotes the lead time of the forecast. 
More details about the input data are described in the suppl.\ material. 
The model encodes these initial conditions into $\XF^{t}$ which is then given to the forecasting engine $f$ that generates the next step of probabilistic river discharge forecast at $6$-hourly interval conditioned on the weather forecast $\XHRES$, static attributes $\XALPHA$, and noise vector $\epsilon_z\sim\NN(0, \sigma_z)$ for probabilistic forecast. The forecasting engine rolls out up to $40$ steps ($10$ days) in the future including nowcasting ($t=0$). 
The auto-regressive forecast for $1$ step can be formulated as $\X_{dis06}^{t+1} = f\left(\X_{feature}^{t};\, \XHRES^{t+1},\, \XALPHA,\, \epsilon_z\right)$, where $\X_{dis06}^{t+1}$ is the averaged river discharge over $6$ hours between $t$ and $t+1$. Long-term forecast is generated with rollout. 
In what follows, we describe the components of the model in details.

\textbf{Input embedding.} First, each grid point $p$ of an input modality is projected into a higher feature dimension with a linear layer and followed with a hyperbolic tangent activation (Tanh). After projection all modalities are combined and normalized:
\begin{align}
    \X_{embedding}^{t}(p) = \text{RMSNorm}\Bigl(\text{Tanh}\Bigl(\text{Concat}\bigl(\text{Linear}(\XANA^t(p)), \text{Linear}(\XIMERG^t(p)), \notag \\
    \text{Linear}(\XCPC^{t-4}(p)), \text{Linear}(\XEMO^{t-8}(p))\bigr)\Bigr)\Bigr) + \X_{pos}\,, \label{eq:05}
\end{align} where RMSNorm is the root mean square layer normalization, Linear is a projection layer, $\X_{pos} \in \RR^{B \times T \times 1 \times K}$ is a learnable positional embedding, $\X_{embedding}^{t}(p) \in \RR^{B \times T \times P \times K}$ is the input embedding with $K=192$, $B$ is the batch size, and $P$ is the number of points (tokens) in the domain. 
Analysis data have the dimensionality $\XANA \in \RR^{B \times T \times P \times V_{a}}$ with $V_{a}=14$, projected to $96$, IMERG and CPC both have $\XIMERG, \XCPC \in \RR^{B \times T \times P \time 1}$, projected to $16$ each, and $\XEMO \in \RR^{B \times T \times P \times V_{e}}$ is projected to a $64$d space. 

\textbf{Encoder.} The encoder is adapted from RiverMamba~\citep{RiverMamba}. It is used to encode the initial condition into a representation $\XF^{t} \in \RR^{B \times 1 \times P \times K}$ via spatio-temporal modeling inside Mamba modules. 
The encoder consists of $3$ encoder layers and each one of them consists of $2$ encoder blocks. The first layer models the full temporal resolution of $T=4$. Afterward, $T$ is down-sampled to $T/2$ using a linear layer at the beginning of each subsequent layer. Unlike bidirectional Mamba blocks in RiverMamba, where the forward and backward passes use two separately-parametrized SSMs, we use a shared SSM for bi-directionality. Each block incorporates further AlphaEarth embedding as static features via Location-aware Adaptive Normalization (LOAN)~\citep{loan}. The output $\XF^{t}$ is then given to the forecasting engine.

\begin{figure}[!t]
  \centering
  \includegraphics[width=.98\textwidth]{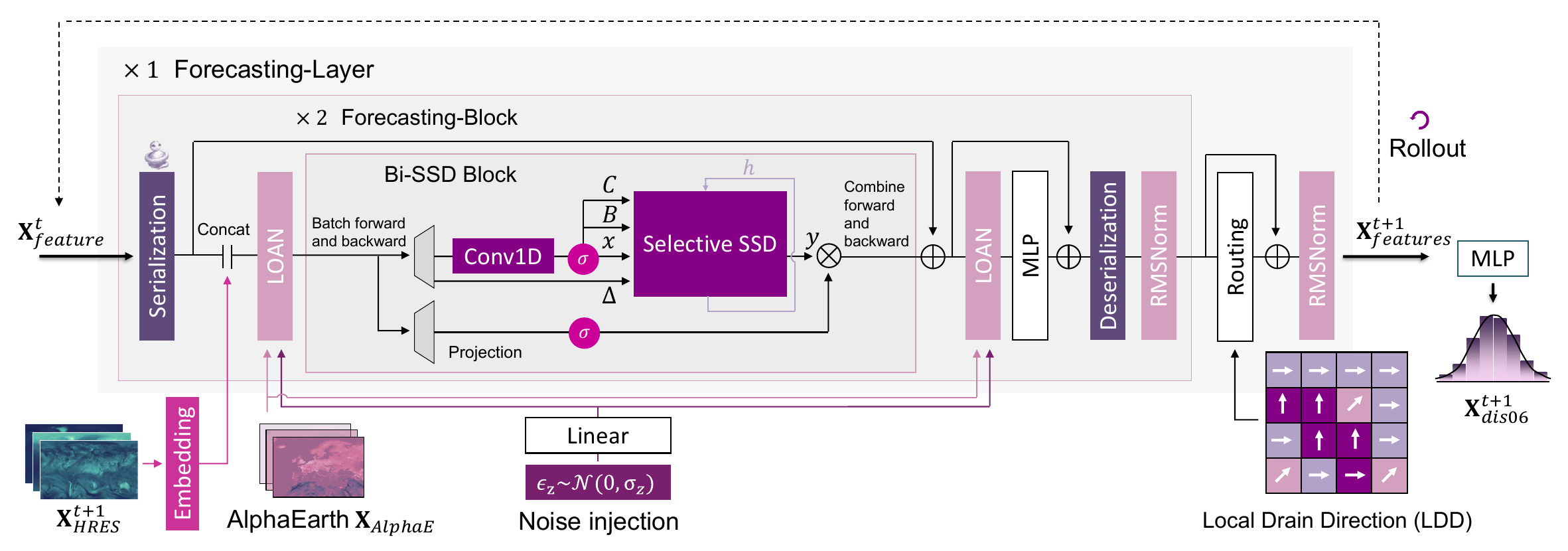}
  \caption{
  The structure of the forecast layer. It has a similar structure as the encoder layer, but it additionally incorporates meteorological forecasts (HRES) by concatenation, the noise $\epsilon_z$, and the routing module. The bi-directional SSD block is shown in more details.
  }\label{fig:Forecasting_block}
\end{figure}

\textbf{Forecasting Engine.}
The forecasting engine takes $\XF^{t}$ and predicts the next $\XF^{t+1}$ auto-regressively and conditioned on the weather forecast $\XHRES^{t+1}$, static attributes $\XALPHA$, and noise vector $\epsilon_z$, as shown in Fig.~\ref{fig:Model}. 
The structure of the block is similar to the encoder block but has $3$ more processes (Fig.~\ref{fig:Forecasting_block}). First, it incorporates the weather forecast $\XHRES^{t+1}$ by projecting it to $64$d, serializing it and concatenating it with $\XF^t$.
Second, it uses noise injection with $\epsilon_z$ to generate an ensemble of predictions. 
Finally, the routing module operates in the end of the forecasting layer to distribute the learned spatio-temporal features through the river network. The output $\XF^{t+1}$ for each lead time $l$ produced by the forecasting engine is then given to an MLP which projects $\XF^{t+1}$ to $64$d and then $1$d to estimate the river discharge $\X^{t+1}_{dis06}$. 

\textbf{Noise injection and training.} 
The initial condition and weather forecast have systematic biases and are inherently uncertain. To model uncertainty, we use a score-based function that trains the model only on the marginal distribution (i.e., optimizing on few ensemble members), however, still be able to approximate the joint distribution over the predicted discharge. 
For this we incorporated a learned perturbation via noise injection in the forecasting engine. 
The noise is sampled from a Gaussian distribution $\epsilon_z\sim \NN(0, \sigma_z=0.5)^{32}$ with $32$ dimension and mapped via $2$ linear layers to $\epsilon_{z1} \in \RR^{B \times 1 \times 1 \times K}$ and $\epsilon_{z2} \in \RR^{B \times 1 \times 1 \times K}$.  
These perturbation are then added to both LOAN layers in the forecasting block, e.g., for the first part of the block $\text{LOAN}(\X) ~+~ \epsilon_{z1}$. Note that we share the noise vector across all grid points and lead time.  
As a score-based function, we use the fair Continuous Ranked Probability Score (CRPS) as an objective~\citep{CRPS}. For $\X_{dis06}^{1:M}$ samples, an unbiased estimate~\citep{fCRPS} can be written as:
\begin{equation}
    \mathcal{L} = \frac{1}{BLP} \sum_{b,t,p} \left[ \frac{1}{M} \sum_{m=1}^{M} \left| x_{\text{dis06}}^{m,b,t}(p) - \hat{x}_{\text{dis06}}^{b,t}(p) \right| - \frac{1}{2M(M-1)} \sum_{m,m'=1}^{M} \left| x_{\text{dis06}}^{m,b,t}(p) - x_{\text{dis06}}^{m',b,t}(p) \right| \right]\,, \label{eq:fcrps}
\end{equation}
where $M$ is the number of ensemble members, $x_{dis06}^{e,b,t}(p)$ is the prediction, and
$\hat{x}_{dis06}^{b,t}(p)$ is the ground truth. During training, we set $M=4$ ensemble members per GPU node. For experiments in a deterministic setting, we used the mean squared error as an objective function.
All target river discharge values are transformed with the formula $\text{sign}(\hat{x}) \text{log}(1+\lvert \hat{x} \rvert)$ as in RiverMamba. 

We first pretrain on EFAS reanalysis $6$-hourly river discharge data ($\X_{dis06}$) and later fine-tune on two sparse daily observational datasets, the GRDC and CAMELS datasets. 
For fine-tuning on observations, we keep the number of input $P$ and optimize only on points with available observations. Note that our model still produces $6$-hourly output for both reanalysis and observational data. During fine-tuning on observational data, we average the prediction over $4$ six-hour steps to match the daily resolution before computing the loss. We assume that the learned sub-daily dynamics primarily originates from the pretraining stage. 
For inference, we can produce any number of ensemble flood forecasting and densely as in Fig.~\ref{fig:teaser}.

\textbf{Warmup and nowcasting.} For the first forecast step, $\XF^{t}$ is produced by the encoder, whereas at every subsequent step, $\XF^{t+1:t+L}$ is the forecasting engine's own previous prediction fed back as input. 
This auto-regressive feedback can induce a distribution shift. Inspired by physics-based systems, we use warmup to stabilize the forecast. Similar to spin-up in numerical weather and hydrological models, we run the forecasting engine for $4$ steps before the actual forecast horizon. The first three warmup steps are driven by an earlier-initialized IFS-HRES forecast cycle at its short lead times, i.e., the same near-term forecast that would genuinely be available in an operational setting for that recent past. The fourth and final warmup step is driven by the latest available analysis. Only the internal state after this final warmup step is kept to initialize the rollout over $\XF^{t+1:t+L}$. 
\section{Experimental results}
Data for pretraining on river discharge are obtained from EFAS reanalysis. 
The EFAS dataset is provided as a 6-hourly discharge over Europe at $1$ arcminute resolution. We use the EFAS reanalysis as a target discharge for training and testing in Sec.~\ref{sec:results_reanalysis}. For fine-tuning on observations, we test the model on observational GRDC data in Sec.~\ref{sec:results_obs} and CAMELS in the suppl.\ material. 
We train up to 40 steps (10 days) on the years 1992-2021, validate on 2022-2023, and test on 2024-2025. 
We evaluate the performance on both EFAS reanalysis and observations, where diagnostic stations are available ($1971$ GRDC and $2206$ CAMELs stations). 
Metrics are calculated by taking the median on individual grid points.
We compare our model to the state-of-the-art diffusion-based DRUM \citep{DRUM_2025}, Encoder-Decoder LSTM (ED-LSTM) \citep{Google_nature}, Mean-Embedding LSTM (ME-LSTM) \cite{Google_neu} models, 
RiverMamba \citep{RiverMamba}, and the operational EFAS forecasting system operated by ECMWF \cite{EFAS}. 
We used $M=12$ ensemble members for all probabilistic models (RiverMamba is deterministic). 
More details about dataset, evaluation metrics, and baselines are in the suppl.\ material. 
\subsection{Experiments on EFAS river discharge reanalysis}
\label{sec:results_reanalysis}
\begin{figure}[h!]
\centering
\includegraphics[width=.99\linewidth]{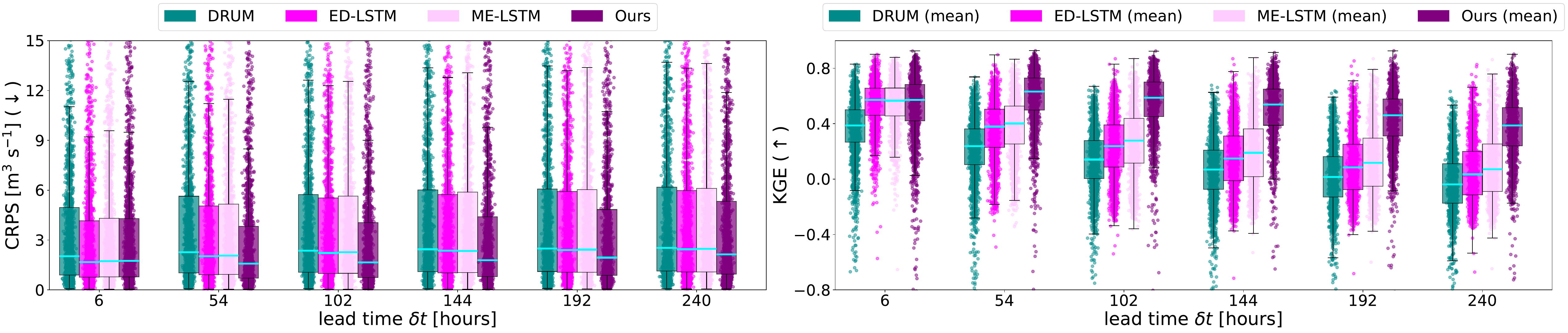}
\caption{Results on EFAS reanalysis with lead time.}\label{fig:reanalysis_baselines}
\end{figure}

Fig.~\ref{fig:reanalysis_baselines} plots the median of CRPS and KGE metrics with $6$ selected lead time. Boxes represent distribution quartiles, with evaluation grid points plotted along the y-axis.  
LSTM has a competitive performance but its prediction skill drops with long-term forecast. 
Our model achieves the best results with big margins. We attribute this to the spatio-temporal modeling and routing in the proposed model. 
Table~\ref{table:reanalysis_baselines} shows that our model achieves also better results quantitatively. 
Table \ref{table:rivermamba_0} shows a comparison with RiverMamba in a deterministic setting over the Mid Europe domain (281K points) for 16 steps lead time. 
Our model consistently outperforms RiverMamba on F1 score.
The RiverMamba baseline has no explicit routing mechanism inside the model. It relies on the spatio-temporal modeling of SSM without considering local drain directions, while our model explicitly models this within the routing module which improves flood detection.
More results are in the suppl.\ material. In the following, we discuss the main ablation studies that were performed over a mid European domain for 16 steps lead time in a deterministic setting.
\begin{table}[!t]
  \caption{Results on EFAS Reanalysis (CAMELS split).}
  \label{table:reanalysis_baselines}
  \centering
  \small
  \tabcolsep=3.0pt\relax
  \setlength\extrarowheight{0pt}
  \begin{tabular}{c r *{8}l}
  \toprule
  & & & \multicolumn{3}{c}{Validation (2022-2023)} & & \multicolumn{3}{c}{Test (2024-2025)} \\
    \midrule
    & Model & & R2$^{r}$ {\scriptsize($\uparrow$)} & KGE$^{r}$ {\scriptsize($\uparrow$)} & CRPS$^{r}$ {\scriptsize($\downarrow$)} & & R2$^{r}$ {\scriptsize($\uparrow$)} & KGE$^{r}$ {\scriptsize($\uparrow$)} & CRPS$^{r}$ {\scriptsize($\downarrow$)} \\
    \midrule
    \multirow{4}{*}{\rotatebox[origin=c]{90}{\scriptsize{Probabilistic}}} & DRUM (mean) & & 0.1496 & 0.1505 & 2.0872 & & 0.0984 & 0.1352 & 2.3581 \\
       
    & ED-LSTM (mean) & & 0.1726 & 0.2431 & 1.9121 & & 0.1347 & 0.2333 & 2.1971 \\
    
    & ME-LSTM (mean) & & 0.2093 & 0.3071 & 1.8395 & & 0.1210 & 0.2640 & 2.2249 \\
    
    & Ours (mean) & & \textbf{0.4515} & \textbf{0.5751} & \textbf{1.4720} & & \textbf{0.3907} & \textbf{0.5474} & \textbf{1.7619} \\
  \bottomrule
\end{tabular}
\end{table}
\begin{table}[!b]
  \vspace{-3mm}
  \caption{Ablation studies on the validation set over Mid Europe. Shown are KGE on reanalysis (KGE$^{r}$) and observational (KGE$^{o}$) CAMELS data along with F1$^{r}$@(1.5-500) for reanalysis over the full domain ($281$K points).}
  \label{table:ablation}
  \centering
  \small
  \tabcolsep=1.0pt\relax
  \setlength\extrarowheight{0pt}
  \begin{tabular}{*{4}clcclc}
  
  \multicolumn{3}{l}{(a) Routing and Warmup} & & \multicolumn{2}{l}{(b) Routing Module} & & \multicolumn{2}{l}{(c) Upstream Aggregation}\\
      \cmidrule[\heavyrulewidth]{1-3} \cmidrule[\heavyrulewidth]{5-6} \cmidrule[\heavyrulewidth]{8-9}
      
      Routing & Warmup & KGE$^{r}$ | KGE$^{o}$ | F1$^{r}$($\uparrow$) & & Module & KGE$^{r}$ | KGE$^{o}$ | F1$^{r}$($\uparrow$) & & Type & KGE$^{r}$ | KGE$^{o}$ | F1$^{r}$($\uparrow$) \\

      \cmidrule[\heavyrulewidth]{1-3} \cmidrule[\heavyrulewidth]{5-6} \cmidrule[\heavyrulewidth]{8-9}
      
      \xmark & \xmark & 0.666 | 0.692 | 0.096 & & GS-mean & 0.680 | 0.721 | 0.310  & & Summation & 0.738 | 0.682 | 0.335 \\
      
      \cmark & \xmark & 0.672 | 0.739 | 0.114 & & GS-sum & 0.702 | 0.702 | 0.285 & & Mean & 0.730 | 0.739 | 0.332 \\

      \xmark & \cmark & 0.687 | 0.721 | 0.162 & & GAT & 0.714 | 0.633 | 0.268 & & GraphMamba & 0.722 | 0.720 | 0.276 \\
      
      \cmark & \cmark & \textbf{0.740} | \textbf{0.747} | \textbf{0.339} & & SSM & \textbf{0.740} | \textbf{0.747} | \textbf{0.339} & & Ours & \textbf{0.740} | \textbf{0.747} | \textbf{0.339} \\

  \cmidrule[\heavyrulewidth]{1-3} \cmidrule[\heavyrulewidth]{5-6} \cmidrule[\heavyrulewidth]{8-9}
  \end{tabular}
\end{table}

\textbf{Routing.}~In Table \ref{table:ablation} (a), we evaluate the impact of adding explicit routing (Eq.~\ref{eq:routing_input}-\ref{eq:routing_ssm}) in the forecasting engine. In fact, routing introduces a hydrologically informed structural prior into the model which improves the forecasting skill substantially on both reanalysis and observational data. Without explicit routing, the model produces heterogeneous predictions in which neighboring pixels can have inconsistent
discharge. The routing module instead yields more homogeneous, spatially clustered predictions (Fig.~\ref{fig:teaser}). 
\begin{wraptable}[13]{r}{0.42\textwidth}
  \centering
  \small
  \caption{F1 comparison to RiverMamba.}
  \label{table:rivermamba_0}
  \tabcolsep=2.0pt\relax
  \setlength\extrarowheight{0pt}
  \begin{tabular}{*{3}c}
    \toprule
    & Val & Test \\
    \midrule
    RiverMamba & 0.141{\tiny\gray{$\pm$0.01}} & 0.138{\tiny\gray{$\pm$0.01}} \\
    Ours & \textbf{0.334}{\tiny\gray{$\pm$0.02}} & \textbf{0.314}{\tiny\gray{$\pm$0.01}} \\
    \bottomrule
  \end{tabular}
  \vspace{3mm}
  \caption{Ungauged results on Reanalysis.}
  \label{table:ungauged_main}
  \tabcolsep=2.0pt\relax
  \setlength\extrarowheight{0pt}
  \begin{tabular}{c c c c}
    \toprule
    & R2$^{r}$ & KGE$^{r}$ & F1$^{r}$ \\
    \midrule
    \xmark~Routing & -0.501 & -0.002 & 0.015 \\
    \cmark~Routing & \textbf{0.202} & \textbf{0.338} & \textbf{0.105} \\
    \bottomrule
  \end{tabular}
\end{wraptable}
\vspace{3mm}
In Table~\ref{table:ungauged_main}, we show that routing has even a bigger positive impact on spatial generalization for unseen catchments during training (ungauged setting). On average it improves R2 form -0.5 to +0.2.

In Table \ref{table:ablation} (b), we compare our proposed SSM message passing for routing with GNNs. 
GS uses the GraphSAGE model~\citep{GraphSAGE}. While GS-mean uses mean aggregation over upstream neighbors, GS-sum uses summation over the neighbors. GAT~\citep{GAT} is a graph transformer. As can be seen, our SSM message passing achieves the best result compared to the GNN variants. 

In Table \ref{table:ablation} (c), we show the impact of different aggregation techniques for routing. In the first row, we sum up the upstream features and give them as input to the SSD as a sequence of length $2$ for each point. In the second row, we do the same but take the mean instead of summation. GraphMamba~\citep{GraphMamba} sorts the tokens by their feature norms where the neighbor with highest norm is placed last in the sequence. Our technique works the best. 

\textbf{Warmup.}~In Table \ref{table:ablation} (a), we also show the benefit of doing warmup on the forecast. As can be seen, warmup improves the results since it closes the distribution shift between $\XF^{t}$ and $\XF^{t+1:t+L}$ making the forecast more stable. More ablation studies are in the suppl.\ material. 

\begin{table}[!t]
  \caption{Results on gauged GRDC observations.}
  \label{table:obs_baselines}
  \centering
  \small
  \tabcolsep=3.0pt\relax
  \setlength\extrarowheight{0pt}
  \begin{tabular}{c r *{8}l}
  \toprule
  & & & \multicolumn{3}{c}{Validation (2022-2023)} & & \multicolumn{3}{c}{Test (Jun 2024- Dec 2025)} \\
    \midrule
    & Model & & R2$^{o}$ {\scriptsize($\uparrow$)} & KGE$^{o}$ {\scriptsize($\uparrow$)} & CRPS$^{o}$ {\scriptsize($\downarrow$)} & & R2$^{o}$ {\scriptsize($\uparrow$)} & KGE$^{o}$ {\scriptsize($\uparrow$)} & CRPS$^{o}$ {\scriptsize($\downarrow$)} \\
    \midrule
     & EFAS (control run) & & - & - & - & & 0.0410 & 0.3964 & - \\
    \midrule
     \multirow{4}{*}{\rotatebox[origin=c]{90}{\scriptsize{Probabilistic}}} & DRUM (mean) & & 0.1750 & 0.2834 & 4.5375 & & -0.0089 & 0.2355 & 4.3474 \\
    & ED-LSTM (mean) & & 0.2841 & 0.3957 & 3.9024 & & 0.1214 & 0.3788 & 3.6818 \\
    & ME-LSTM (mean) & & 0.3339 & 0.4771 & 3.4928 & & 0.0913 & 0.3843 & 3.7962 \\
    & Ours (mean) & & \textbf{0.6042} & \textbf{0.6689} & \textbf{2.4592} & & \textbf{0.3693} & \textbf{0.5520} & \textbf{2.9708} \\
  \bottomrule
\end{tabular}
\end{table}
\begin{figure}[h!]
\centering
\includegraphics[width=.99\linewidth]{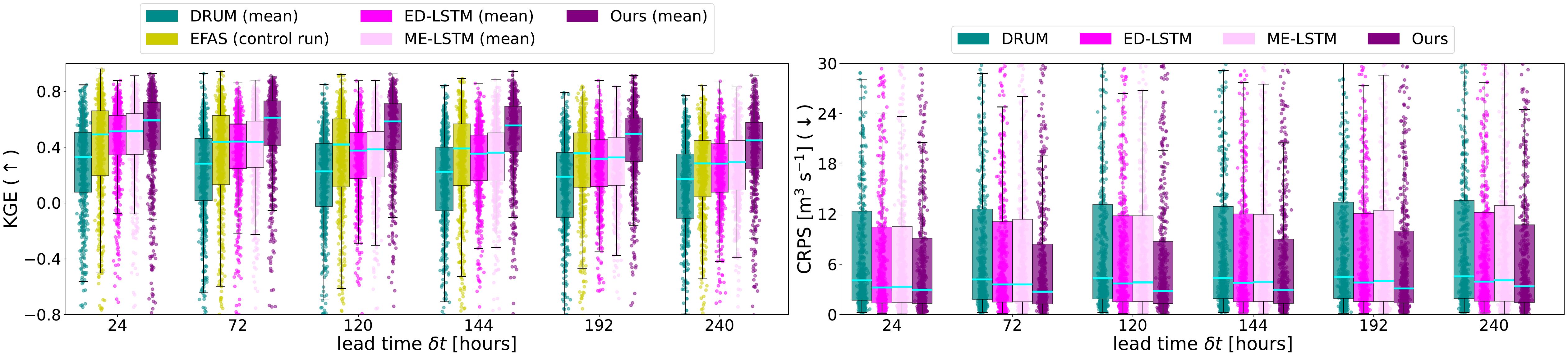}
\caption{Results on GRDC observations with lead time.}\label{fig:obs_baselines}
\end{figure}
\subsection{Experiments on observational river discharge data}
\label{sec:results_obs}
Performance on gauged GRDC river discharge observations is summarized in Table~\ref{table:obs_baselines}, which includes the EFAS model with explicit physics-based routing with forecast starting from mid 2024. As before, Fig.~\ref{fig:obs_baselines} evaluates these models across lead times. The diffusion-based DRUM model performs poorly, whereas EFAS performs on par with the ED-LSTM and ME-LSTM baselines despite its advantage of using discharge as initial condition. Nevertheless, EFAS has lower R2 values, likely because it models naturalized discharge without considering realistic human interventions such as dams and reservoirs. Our model consistently outperforms the other approaches for all metrics on observational data. More results are in the suppl.\ material.
\vspace{-1mm}

\section{Conclusions and limitations}
We presented a novel deep learning approach for fully spatially distributed modeling and routing in hydrology. 
Our approach leverages State Space Duality (SSD) for upstream-to-downstream flow modeling. 
It packs local upstream-neighbor sequences into one long segmented sequence and then adapts the SSD to route high-dimensional features rather than physical quantities. For each point, a routing SSD module aggregates the learned upstream features from neighbors and processes them jointly with the downstream point's features. 
We showed that explicitly encoding the river network in the feature space substantially improves forecast skill. 
Built upon this, we introduced an operational model that forecasts gridded river discharge at very high spatio-temporal resolution for up to $10$ days, achieving new state-of-the-art results. 
While this is a major advancement, our approach still has some limitations. 
The local drain directions are inferred from topology and we use a static directed graph in this work. The routing therefore does not represent backwater effects or flow reversals. 
Furthermore, satellite observations could be helpful to define a more accurate and dynamic graph. We also do not consider human interventions with natural flow such as reservoirs, abstractions, and diversions. 
Despite these, our model showed a very promising step towards accurate and high-resolution distributed river discharge forecasting. Which is very important for actionable responses and flood early warning.

\newpage

\subsubsection*{Acknowledgments}

This work was supported by Carl Zeiss Foundation for the project GENAI-X under grant no.\ P2024-11-042 and by the WeatherGenerator project under grant no.\ 101187947. The WeatherGenerator project is funded by the European Union. Views and opinions expressed are however those of the authors only and do not necessarily reflect those of the European Union or the Commission. Neither the European Union nor the granting authority can be held responsible for them.

We acknowledge EuroHPC for awarding us access to Leonardo at CINECA, Italy, and JUPITER Booster, Germany. through EuroHPC Regular Access Call - proposal No. AIFAC-5C0-154 and AIF-2026SC02-173. The authors also gratefully acknowledge the granted access to the DAIS computing cluster hosted by Max Planck Computing and Data Facility.

\bibliography{ref}
\bibliographystyle{iclr2027_conference}

\newpage

\appendix
\label{sec:Appendix}

\begin{center}
{\huge
\bfseries
Distributed Hydrological Modeling in the Feature Space} \\
\vspace{0.4cm}
{\Large
\logo
Technical Appendix and Supplementary Material \\
}
\end{center}

\vspace{0.5cm}
{\Large Table of Contents}
\vspace{0.3cm}

\startcontents[sections]
\printcontents[sections]{l}{1}{\setcounter{tocdepth}{3}}
\clearpage

\section{Ablation Studies}

For the ablation studies, we conducted experiments over a Mid European domain ($55.5^\circ$N $46.5^\circ$S, $5.5^\circ$E $15.5^\circ$E) which has $280,820$ points on land and includes $480$ GRDC and $892$ CAMELS stations for evaluation. Fig.~\ref{fig:efas_domain_germany} gives an overview of the domain. For the experiments, we run only deterministic model for $L=16$ lead time including training on nowcasting ($t$=0) and evaluate on the validation set 2022-2023.

\subsection{Feature importance}

In Table \ref{table:ablation_feature_importance}, we study the impact of different input features on the performance of our model. In general, removing any input feature degrades the performance. However, the model is still functional whenever we remove one of the input. Removing the observational IMERG has less effect than removing CPC. Although removing the IFS-HRES weather forecast has the biggest impact, the model is still produces a skillful forecast estimating only from the initial conditions. Finally, the best performance is achieved when we use all input variables.

\begin{table}[h!]
  \caption{Ablation studies for feature importance on the validation set over Europe. 
  Shown are KGE for CAEMLS (KGE$_{c}^{r}$) and GRDC (KGE$_{g}^{r}$) splits on reanalysis. along with the F1$^{r}$@(1.5-500) for reanalysis over the full domain ($281$K points). In the second half of the table, we report the results of R2$_{c}^{o}$, KGE$_{c}^{o}$, and F1$_{c}^{o}$@(1.5-20) on the fine-tunning on CAMELS observations.
  }
  \label{table:ablation_feature_importance}
  \centering
  \tabcolsep=4.0pt\relax
  \setlength\extrarowheight{0pt}
  \begin{tabular}{*{7}c}
  
    \toprule
    \multicolumn{7}{l}{Results on EFAS Reanalysis} \\
    \midrule
    Static & CPC & IMERG & EMO & Analysis & IFS-HRES & R2$_{c}^{r}$ | KGE$_{c}^{r}$ | R2$_{g}^{r}$ | KGE$_{g}^{r}$ | F1$^{r}$($\uparrow$) \\
      
    \cmidrule[\heavyrulewidth]{1-7}
    
    \xmark & \cmark & \cmark & \cmark & \cmark & \cmark & 0.6963 | 0.7338 | 0.6919 | 0.7443 | 0.2734 \\
        
    \cmark & \xmark & \cmark & \cmark & \cmark & \cmark & 0.6741 | 0.6972 | 0.6674 | 0.7145 | 0.3053 \\
    
    \cmark & \cmark & \xmark & \cmark & \cmark & \cmark & 0.7021 | 0.7303 | 0.7041 | 0.7411 | 0.3100 \\

    \cmark & \cmark & \cmark & \xmark & \cmark & \cmark & 0.6863 | \textbf{0.7474} | 0.6637 | 0.7116 | 0.3101 \\
    
    \cmark & \cmark & \cmark & \cmark & \xmark & \cmark & 0.6869 | 0.6979 | 0.6835 | 0.7107 | 0.3069 \\

    \cmark & \cmark & \cmark & \cmark & \cmark & \xmark & 0.5065 | 0.5817 | 0.5262 | 0.6125 | 0.2549 \\

    \cmark & \cmark & \cmark & \cmark & \cmark & \cmark & \textbf{0.7061} | 0.7396 | \textbf{0.7043} | \textbf{0.7493} | \textbf{0.3394} \\
    
    \midrule
    \multicolumn{7}{l}{Results on CAMELS observational data} \\
    \midrule
    Static & CPC & IMERG & EMO & Analysis & IFS-HRES & R2$_{c}^{o}$ | KGE$_{c}^{o}$ | F1$_{c}^{o}$($\uparrow$)  \\
      
    \cmidrule[\heavyrulewidth]{1-7}
    
    \xmark & \cmark & \cmark & \cmark & \cmark & \cmark & 0.6863 | 0.7163 | 0.2691 \\
        
    \cmark & \xmark & \cmark & \cmark & \cmark & \cmark & 0.6668 | 0.7308 | 0.3069 \\
    
    \cmark & \cmark & \xmark & \cmark & \cmark & \cmark & \textbf{0.7006} | \textbf{0.7493} | 0.2857 \\

    \cmark & \cmark & \cmark & \xmark & \cmark & \cmark & 0.6964 | 0.7100 | 0.2101 \\
    
    \cmark & \cmark & \cmark & \cmark & \xmark & \cmark & 0.6704 | 0.7157 | 0.2133 \\

    \cmark & \cmark & \cmark & \cmark & \cmark & \xmark & 0.5376 | 0.5646 | 0.1008 \\

    \cmark & \cmark & \cmark & \cmark & \cmark & \cmark & 0.6990 | 0.7472 | \textbf{0.3109} \\
    
    \bottomrule
  \end{tabular}
\end{table}

\subsection{LISFLOOD vs AlphaEarth}

In this experiment, we study the impact of replacing LISFLOOD static river attribute features (Sec.~\ref{sec:Appendix_LISFLOOD_static_data}) by the embedding of AlphaEarth from Google DeepMind.
LISFLOOD is considered an important source for modeling in hydrology and it is used in state-of-the-art physics-based EFAS and GloFAS systems. In contrast to a learned embedding like AlphaEarth, LISFLOOD static features have physical meaning. We report the result in Table \ref{table:ablation_static}. In the first row, we remove all static features and keep only the Cartesian coordinates as a positional encoding (Sec.~\ref{sec:Appendix_LISFLOOD_static_data}). removing the static features reduces the performance for flood detection the most and has a bigger impact on fine-tunning on observations. More interesting is that AlphaEarth embedding can be used to replace the LISFLOOD static maps in our model. The advantage is that, AlphaEarth embedding is learned from radar, satellites, and 3D laser measurements. It exists globally at $10$ meter resolution and does not need the efforts to generate over $100$ static maps and calibration parameters globally. As a comparison, in this experiment, we used all $64$ features from AlphaEarth compared to $119$ LISFLOOD static features. 
 
\begin{table}[h!]
  \caption{Ablation studies for static feature importance on the validation set over Europe. 
  Shown are KGE for CAEMLS (KGE$_{c}^{r}$) and GRDC (KGE$_{g}^{r}$) splits on reanalysis. along with the F1$^{r}$@(1.5-500) for reanalysis over the full domain ($281$K points). In the second half of the table, we report the results of R2$_{c}^{o}$, KGE$_{c}^{o}$, and F1$_{c}^{o}$@(1.5-20) on the fine-tunning on CAMELS observations.
  }
  \label{table:ablation_static}
  \centering
  \tabcolsep=4.0pt\relax
  \setlength\extrarowheight{0pt}
  \begin{tabular}{*{3}c}
  
    \toprule
    \multicolumn{3}{l}{Results on EFAS Reanalysis} \\
    \midrule
    LISFLOOD Static & AlphaEarth & R2$_{c}^{r}$ | KGE$_{c}^{r}$ | R2$_{g}^{r}$ | KGE$_{g}^{r}$ | F1$^{r}$($\uparrow$) \\
      
    \cmidrule[\heavyrulewidth]{1-3}
    
    \xmark & \xmark & 0.6963 | 0.7338 | 0.6919 | 0.7443 | 0.2734 \\
    \cmark & \xmark & 0.7045 | \textbf{0.7404} | 0.6943 | 0.7462 | 0.3360 \\
    \xmark & \cmark & \textbf{0.7061} | 0.7396 | \textbf{0.7043} | \textbf{0.7493} | \textbf{0.3394} \\
    
    \midrule
    \multicolumn{3}{l}{Results on CAMELS observational data} \\
    \midrule
    LISFLOOD Static & AlphaEarth & R2$_{c}^{o}$ | KGE$_{c}^{o}$ | F1$_{c}^{o}$($\uparrow$)  \\
      
    \cmidrule[\heavyrulewidth]{1-3}

    \xmark & \xmark & 0.6863 | 0.7163 | 0.2691 \\

    \cmark & \xmark & 0.6821 | 0.7101 | 0.2059 \\

    \xmark & \cmark & \textbf{0.6990} | \textbf{0.7472} | \textbf{0.3109} \\

    \bottomrule
  \end{tabular}
\end{table}

\subsection{Routing}

In Table \ref{table:ablation_bi_ssd}, we show if an explicit routing can replace the learned features by Bi-SSD in the forecasting block. Note that in this experiment, we keep the Bi-SSD in the encoder and remove only the spatio-temporal modeling in the forecasting block by removing Bi-SSD in the forecasting block.
From the Table \ref{table:ablation_bi_ssd}, we can see a clear benefit to keep the spatio-temporal modeling with Bi-SSD even if we have an explicit routing module. The reason could be that Bi-SSD is useful to capture features over the whole domain which improves the forecasting over long horizon and high-resolution i.e., Bi-SSD can capture if a heavy rain could happen in the next steps based on its large receptive field. This spatio-temporal information over the domain is only captured on a small-scale with routing.

\begin{table}[h!]
  \caption{Ablation studies for the importance of Bi-SSD in the forecasting block. 
  Shown are KGE for CAEMLS (KGE$_{c}^{r}$) and GRDC (KGE$_{g}^{r}$) splits on reanalysis. along with the F1$^{r}$@(1.5-500) for reanalysis over the full domain ($281$K points). We also report the results of R2$_{c}^{o}$, KGE$_{c}^{o}$, and F1$_{c}^{o}$@(1.5-20) on the fine-tunning on CAMELS observations.
  }
  \label{table:ablation_bi_ssd}
  \centering
  \tabcolsep=4.0pt\relax
  \setlength\extrarowheight{0pt}
  \begin{tabular}{*{3}c}
  
    \toprule
    & EFAS Reanalysis & CAMELS Observations \\
    \midrule
    Bi-SSD & R2$_{c}^{r}$ | KGE$_{c}^{r}$ | R2$_{g}^{r}$ | KGE$_{g}^{r}$ | F1$^{r}$($\uparrow$) & R2$_{c}^{o}$ | KGE$_{c}^{o}$ | F1$_{c}^{o}$($\uparrow$) \\
      
    \cmidrule[\heavyrulewidth]{1-3}
    
    \xmark & 0.6715 | 0.7057 | 0.6307 | 0.6915 | 0.3225 & 0.6793 | 0.7324 | 0.2947 \\
    \cmark & \textbf{0.7061} | \textbf{0.7396} | \textbf{0.7043} | \textbf{0.7493} | \textbf{0.3394} & \textbf{0.6990} | \textbf{0.7472} | \textbf{0.3109} \\

    \bottomrule
  \end{tabular}
\end{table}

In Table \ref{table:ablation_number_of_routing}, we study the impact of having more than one routing module or having multi-step routing in the forecasting block. We noticed that adding more than one routing module reduces generalization of the model and a better performance was obtained by using only one routing module for each forecasting step i.e., doing routing or distributed modeling every $6$ hours.

\begin{table}[h!]
  \caption{Ablation studies for the number of routing module/distributed steps in the forecasting block. 
  Shown are KGE for CAEMLS (KGE$_{c}^{r}$) and GRDC (KGE$_{g}^{r}$) splits on reanalysis. along with the F1$^{r}$@(1.5-500) for reanalysis over the full domain ($281$K points).}
  \label{table:ablation_number_of_routing}
  \centering
  \tabcolsep=4.0pt\relax
  \setlength\extrarowheight{0pt}
  \begin{tabular}{*{2}c}
  
    \toprule
    \multicolumn{2}{l}{Results on EFAS Reanalysis} \\
    \midrule
    \# Routing modules & R2$_{c}^{r}$ | KGE$_{c}^{r}$ | R2$_{g}^{r}$ | KGE$_{g}^{r}$ | F1$^{r}$($\uparrow$) \\
    \midrule
    $\times1$ & \textbf{0.7061} | \textbf{0.7396} | \textbf{0.7043} | \textbf{0.7493} | \textbf{0.3394} \\
    $\times2$ & 0.6547 | 0.7057 | 0.6289 | 0.6650 | 0.2839 \\
    \bottomrule
  \end{tabular}
\end{table}

\begin{figure}[!h]
  \centering
  \includegraphics[width=.9\textwidth]{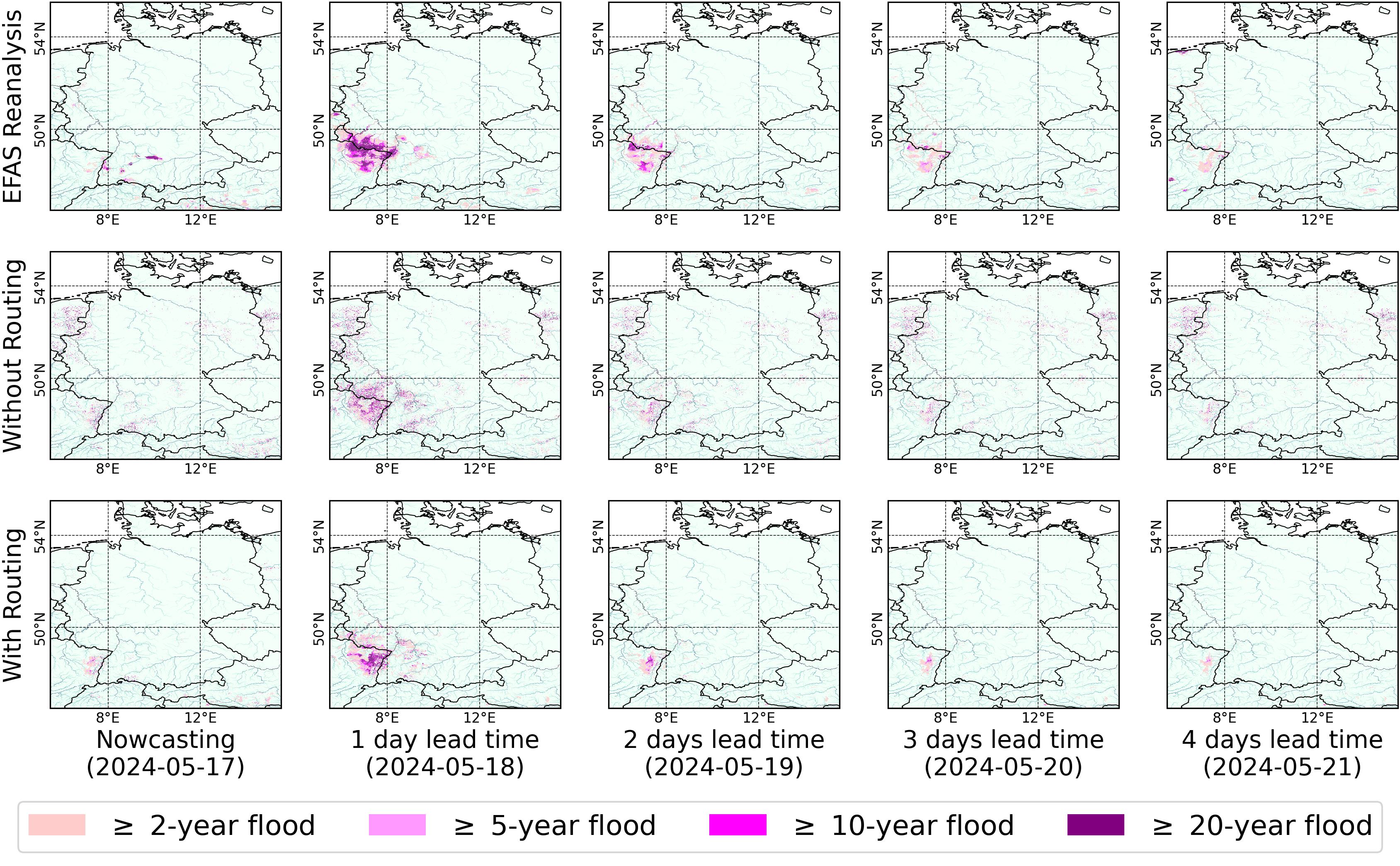}
  \caption{Comparison between EFAS reanalysis (top, used as a reference), a model without routing (middle row), a model with routing (bottom row) during an extreme flood in Saarland Germany , mid May 2024. Shown are flood severity maps at 1 arcminute resolution where each panel shows flood extent at different lead times from nowcasting to 4 days lead time. Without explicit distributed modeling, the routing-free model produces heterogeneous predictions in which neighboring pixels can have inconsistent discharge, leading to many false positives. Our routing module instead yields more homogeneous and spatially clustered predictions. In addition, the routing module makes the predicted river discharge consistent with topology across the river skeletons (Sec.~\ref{sec:appendix_consistency_topology}).}
  \label{fig:appendix_qualitative_routing}
\end{figure}

In Table \ref{table:ablation_upstream_agg}, we show supplementary results for Table \ref{table:ablation} (c) with different upstream aggregation techniques. In "Connection per Neighbor", we run SSD from each upstream point to its downstream (SSD on small blocks with a sequence of $2$ each) e.g., a point with, $3$ valid upstream neighbors appears as the token in $3$ separate $2$-token sequences. The point final routing vector is the mean of its self token outputs across all its incoming upstream edges. In "Binary Mask", we study if routing propagates useful upstream-to-downstream information or if it just suppresses points with no upstream to reduce false positives. For this in the second row of Table \ref{table:ablation_upstream_agg}, we check if a point has at least $1$ upstream neighbor, we double its own feature vector i.e., no information is propagated from upstream neighbors. If a point has no neighbors, we do not add anything to the feature vector. Like this "Binary Mask" adds an inductive bias to the model informing it explicitly about the topology without propagating information between grids. As can be seen from the first row, better results are achieved when we use all upstream points and sort them with upstream area as a one long sequence. The second row experiment verifies that the routing in the model propagates useful upstream-to-downstream information between the gird points and learns features beyond the topology of the graph.
In Fig.~\ref{fig:appendix_qualitative_routing}, we show the impact of adding routing on the flood severity detection qualitatively.

\begin{table}[h!]
  \caption{Ablation studies for different upstream aggregation techniques as a supplementary to Table \ref{table:ablation} (c). Shown are KGE for CAEMLS (KGE$_{c}^{r}$) and GRDC (KGE$_{g}^{r}$) splits on reanalysis. along with the F1$^{r}$@(1.5-500) for reanalysis over the full domain ($281$K points). We also report the results of R2$_{c}^{o}$, KGE$_{c}^{o}$, and F1$_{c}^{o}$@(1.5-20) on the fine-tunning on CAMELS observations.
  }
  \label{table:ablation_upstream_agg}
  \centering
  \tabcolsep=4.0pt\relax
  \setlength\extrarowheight{0pt}
  \begin{tabular}{*{3}c}
  
    \toprule
    & EFAS Reanalysis & CAMELS Observations \\
    \midrule
    Type & R2$_{c}^{r}$ | KGE$_{c}^{r}$ | R2$_{g}^{r}$ | KGE$_{g}^{r}$ | F1$^{r}$($\uparrow$) & R2$_{c}^{o}$ | KGE$_{c}^{o}$ | F1$_{c}^{o}$($\uparrow$) \\
      
    \cmidrule[\heavyrulewidth]{1-3}
    
    Connection per Neighbor &  0.7027 | 0.7264 | \textbf{0.7064} | 0.7456 | 0.2950 & 0.6936 | 0.7318 | 0.2461 \\

    Binary Mask &  0.6622 | 0.6816 | 0.6404 | 0.6751 | 0.1122 & 0.6245 | 0.7158 | 0.2954 \\

    All$+$Sort(area) & \textbf{0.7061} | \textbf{0.7396} | 0.7043 | \textbf{0.7493} | \textbf{0.3394} & \textbf{0.6990} | \textbf{0.7472} | \textbf{0.3109} \\

    \bottomrule
  \end{tabular}
\end{table}

\subsection{Bi-directional Mamba}

In this experiment, we compare the Mamba as in RiverMamba~\citepsupp{RiverMamba} to the Mamba2 variant used in this work.
Bi-directional Mamba blocks in RiverMamba use two separately-parametrized SSMs for the forward and backward passes. In our work, we use a shared SSM for bi-directionality.
Inside each block, we build a second copy of the sequence by reversing the order of points along the space-filling curve, batch it together with the forward-order sequence, and pass both through the same SSD block in one call. 
Table~\ref{table:appendix_ablation_bidirectionality} compares the two approaches. From the table we can see that the SSD algorithm of Mamba2 utilizes the GPU more efficiently. With Mamba2, we can train and run inference faster than RiverMamba. This is important since we have a rollout mechanism in the forecasting engine and we need an efficient algorithm to run the forward and backward passes simultaneously.

\begin{table}[h!]
  \caption{Ablation studies for bi-directional Mamba in comparison to RiverMamba. 
  Shown are training time for in epoch in minutes and inference time in seconds.}
  \label{table:appendix_ablation_bidirectionality}
  \centering
  \small
  \tabcolsep=3.0pt\relax
  \setlength\extrarowheight{0pt}
  \begin{tabular}{l*{2}c}
    \toprule
    \multicolumn{3}{l}{Results on EFAS Reanalysis} \\
    \midrule
    Bi-directionality & Training time* ($\downarrow$) & Inference time* ($\downarrow$) \\
     & 1 epoch 16 steps & 280,820 points \\
    \midrule
    $\times2$~Mamba (RiverMamba) & $\sim$159 min & 6.20{\scriptsize\gray{$\pm$0.005}} sec \\
    $\times1$~Mamba2 (Ours) & $\sim$49 min & 4.10{\scriptsize\gray{$\pm$0.008}} sec \\
    \bottomrule
  \multicolumn{3}{l}{\footnotesize$^*$ tested on NVIDIA A40 GPU.}\\
  \end{tabular}
\end{table}

\section{Consistency with River Network Topology}
\label{sec:appendix_consistency_topology}

In this experiments, we study the effect of adding explicit routing to the model on having consistent forecast along the river network topology. For this, we train a deterministic model with and without routing on the Mid European domain for $16$ time steps + $1$ nowcasting. 

It is assumed that every upstream pixel should have less or equal river discharge value than its downstream direction since catchment area only grows moving downstream. 
For every pixel with a defined local drain direction (LDD) direction, we evaluate this constraint on the forecast river discharge value.
We perform the test over $2024$-$2025$ with $728$ forecasting issued dates for $277{,}805$ checkable pixel pairs. We also consider a small tolerance of $5\%$ relative $+\,0.1\,\mathrm{m^3 s^{-1}}$ absolute per pixel pair. Table~\ref{table:ldd_consistency} shows that adding the routing improves consistency at every lead time. 
The improvement when adding routing is up to $+0.54\%$ at $96$ h. 

\begin{figure}[!h]
  \centering
  \includegraphics[width=.98\textwidth]{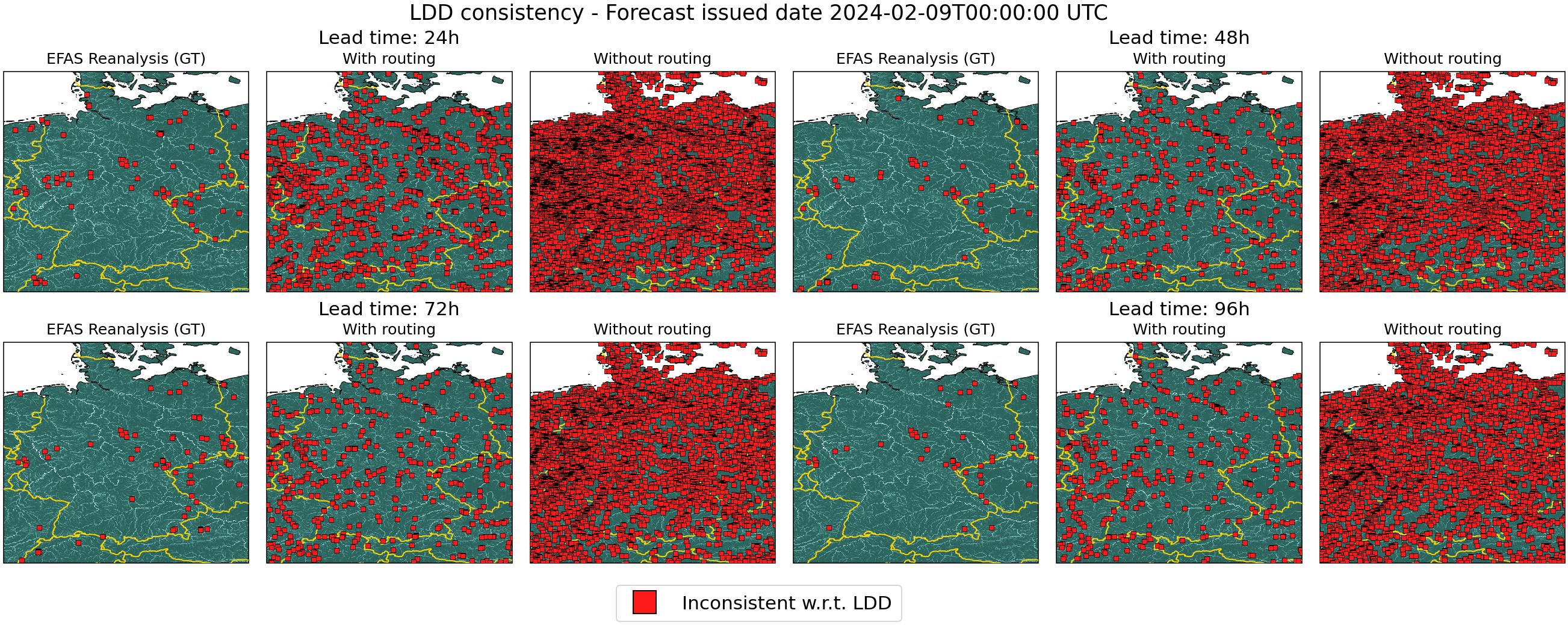}
  \caption{Local Drain Direction (LDD) consistency check (downstream $\geq$ upstream) for the forecast issued on 2024-02-09T00:00:00~UTC, at four lead times ($24$, $48$, $72$, and $96\,\mathrm{h}$. The magenta (GT) or red (model) markers flag pixels where the LDD upstream point carries more discharge than its downstream neighbour.}
  \label{fig:appendix_ldd_consistency_1}
\end{figure}

In Fig.~\ref{fig:appendix_ldd_consistency_1}, we show the spatial distribution of constraint violation over the Mid European domain for a random date. As can be seen the routing provides more consistent forecast along the topology of LDD. It is also notable that the errors of the model without routing are not spread uniformly but concentrate at specific, presumably topologically difficult regions like river confluences, narrow channels, or steep local gradients. 

\begin{table}[h!]
\centering
\caption{LDD consistency rate (ratio of pixel pairs satisfying downstream $\geq$ upstream) for the EFAS reanalysis as a reference and the two forecast models. Evaluation is averaged for the years $2024$-$2025$ over the Mid European domain for $16$ time steps and $1$ nowcasting step.}
\label{table:ldd_consistency}
\begin{tabular}{lccccc}
\toprule
& \multicolumn{1}{c}{Overall} & \multicolumn{4}{c}{Lead Time $t$} \\
\cmidrule(lr){2-2} \cmidrule(lr){3-6}
Source & Mean & $24\,\mathrm{h}$ & $48\,\mathrm{h}$ & $72\,\mathrm{h}$ & $96\,\mathrm{h}$ \\
\midrule
EFAS Reanalysis (GT)     & 99.97\%  & -   & -   & -   & -      \\
\midrule
\xmark~~routing & 99.09\% & 99.19\% & 99.34\% & 99.37\% & 99.35\% \\
\cmark~routing    & \textbf{99.62\%} & \textbf{99.73\%} & \textbf{99.87\%} & \textbf{99.89\%} & \textbf{99.89\%} \\
\midrule
$\Delta$ Benefit of adding routing  & \colorbetter{+0.53\%} & \colorbetter{+0.54\%} & \colorbetter{+0.53\%} & \colorbetter{+0.52\%} & \colorbetter{+0.54\%} \\
\bottomrule
\end{tabular}
\end{table}

Note that even the EFAS reanalysis which is produced by a physics-based hydrological model (LISFLOOD) and uses routing with the same LDD is $99.97\%$ consistent with the strict (downstream $\geq$ upstream) rule at a single $6$-hourly timestep.
Although, the difference in magnitude is typically small (median $\approx 3\%$ of the upstream value) and constraint violations are not spread uniformly across the domain. 
We suspect that due to the landscape i.e., degree of slope and the finite propagation speed of the kinematic wave scheme between adjacent grid cells ($\sim 1.5\,\mathrm{km}$ apart at $1$-arcminute EFAS resolution). At a single instant in time, a flood may have reached an upstream cell but not yet fully propagated into its downstream neighbour. Second Reservoirs can act as a temporary storage that can delay discharge propagation. Finally, the evaporation, irrigation, and withdrawals can also play different roles along the flow path.


\section{Performance Distribution of distributed modeling}
\label{sec:appendix_distribution_routing}

\begin{figure}[!h]
  \centering
  \includegraphics[width=.9\textwidth]{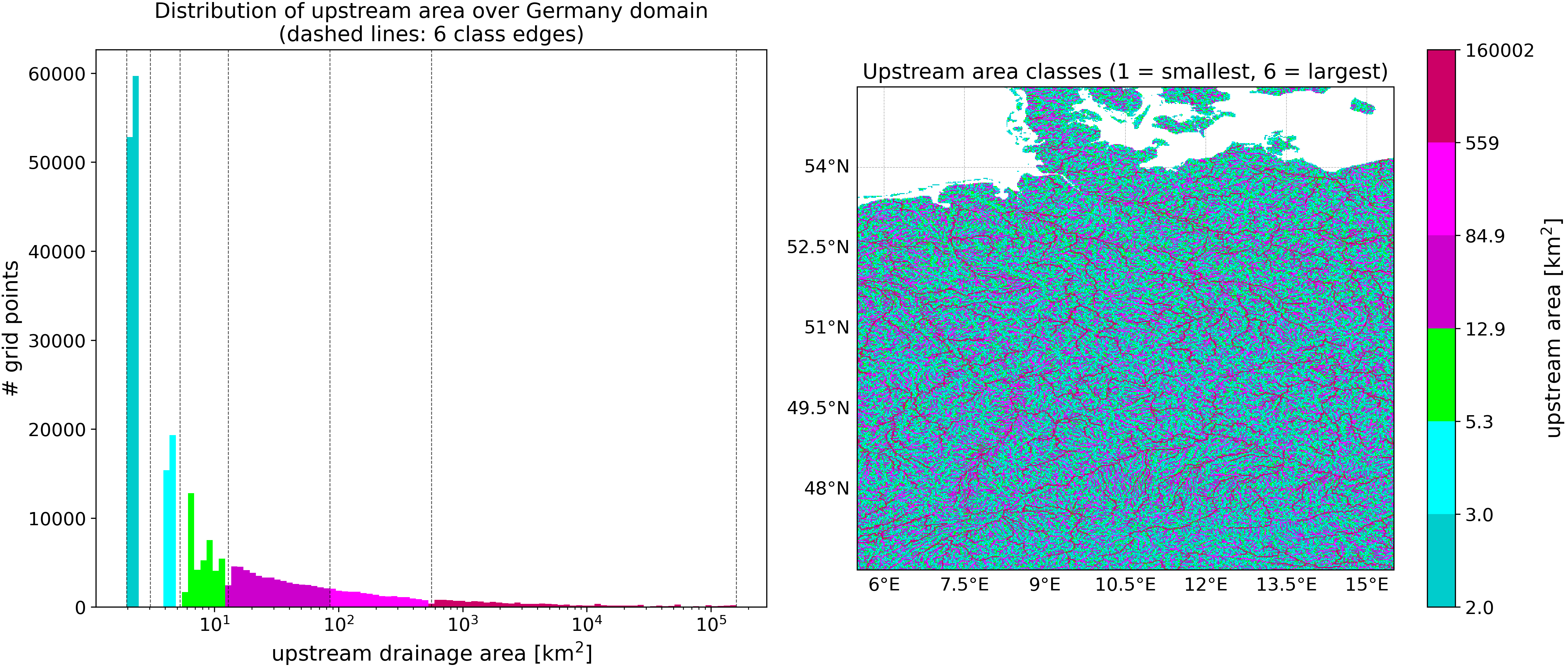}
  \caption{Upstream drainage area distribution over the Mid European domain with $280,820$ EFAS grid points and the $6$ classification classes we used in Sec.~\ref{sec:appendix_distribution_routing}, Table \ref{table:ablation_distributed_classes_1}, and Fig.~\ref{fig:distribution_1}.
  On the left side is the histogram of upstream area.
  On the right is the spatial map of the assigned class at every grid point starting from small headwater streams (class 1) to the largest rivers (class 6).}
  \label{fig:distribution_classes}
\end{figure}

In this experiment, we study the impact of routing spatially in a gauged setting. We trained for $16$ steps + $1$ step nowcasting and evaluated on $1458$ daily forecasts spanning the years $2022$-$2025$ against EFAS reanalysis as a reference.
Each grid point is assigned an upstream-drainage-area class from the EFAS upstream area static field (Fig.~\ref{fig:distribution_classes}). 
At every grid point and lead time we compute on the $4$-years time series the metrics MAE, RMSE, R2, KGE, and F1.

\begin{figure}[!h]
  \centering
  \includegraphics[width=.98\textwidth]{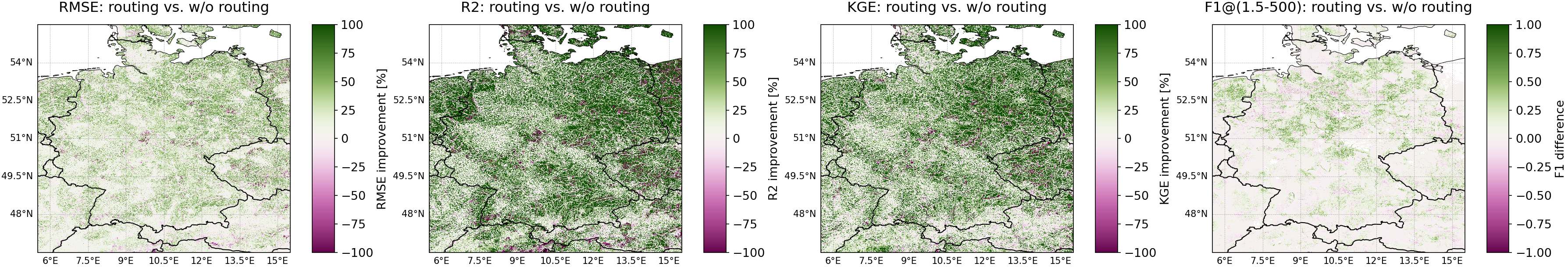}
  \caption{Per pixel percentage improvement from the model with routing over a model without routing, for RMSE$^r$, R2$^r$, KGE$^r$ and F1$^r$@(1.5-500). For each grid point, the percentage improvement is computed as
  $ 
  \Delta{\text{RMSE}} = \frac{\text{RMSE}_{\text{(w/o routing)}} - \text{RMSE}_{\text{(routing)}}}{\text{RMSE}_{\text{(w/o routing)}}} \times 100,~
    \Delta{\text{R2}} = \frac{\text{R2}_{\text{(routing)}} - \text{R2}_{\text{(w/o routing)}}}{|\text{R2}_{\text{(w/o routing)}}|} \times 100,~ \Delta{\text{KGE}} = \frac{\text{KGE}_{\text{(routing)}} - \text{KGE}_{\text{(w/o routing)}}}{|\text{KGE}_{\text{(w/o routing)}}|} \times 100 \,,
    $ where subscripts "routing" and "w/o routing" denote a model with routing and without it. 
    F1 is computed per pixel and the plotted quantity is the plain difference $
    \Delta{\text{F1}} = \text{F1}_{\text{(routing)}} - \text{F1}_{\text{(w/o routing)}}.$ The values are bounded in $[-1, 1]$.
    Positive (green) values indicate routing improves the metric at that pixel and negative (magenta) values indicate it is worse.}
  \label{fig:distribution_1}
\end{figure}

Table \ref{table:ablation_distributed_classes_1} reports the performance per class as the median over the class's grid points and averaged across lead times. 
Cells are colored \colorbetter{Cyan} where adding routing is better than without routing and \colorworse{Purple} where it is worse. Routing improves every metric in every class except one the MAE for the largest basins (class 6) which is marginally worse with routing (7.916 $\rightarrow$ 7.930), even though RMSE, R2 and KGE all improve for that class. The gain is also uneven across classes. It is largest for the smallest headwater cells (class $1$: R2 $0.016 \rightarrow 0.417$, KGE $0.236 \rightarrow 0.532$) and generally shrinks through the mid-sized classes, reaching its minimum at class 5 (R2 +$0.029$, KGE +$0.041$) before a small uptick at class 6 (R2 +$0.060$, KGE +$0.052$). For F1, routing improves detection at essentially every class and return period combination. 
Fig.~\ref{fig:distribution_1} shows the spatial impact of the routing module on the forecasting skill.

\begin{table}[!h]
 \caption{Gauged performance of the model with routing against a model without it by upstream-area class. Evaluation is done over $280,820$ EFAS grid points for the validation and test sets combined ($2022$-$2025$).
 For each class and metric, values are the class-median of the per-pixel MAE/RMSE/R2/KGE and averaged across the $17$ lead times. \colorbetter{Cyan} marks cells where routing improves the metric and \colorworse{Purple} marks cells where it is worse.}
 \label{table:ablation_distributed_classes_1}
 \small
 \centering
 \setlength\tabcolsep{2pt}
 \begin{tabular}{l *{6}l}
 \toprule
 Class (Uparea km$^2$) & MAE ($\downarrow$) & RMSE ($\downarrow$) & R2 ($\uparrow$) & KGE ($\uparrow$) & F1@(1.5-500) ($\uparrow$) \\
\midrule
1: 1.95-3.04 & 0.0143 $\rightarrow$ \colorbetter{0.0078} & 0.0213 $\rightarrow$ \colorbetter{0.0133} & 0.016 $\rightarrow$ \colorbetter{0.417} & 0.236 $\rightarrow$ \colorbetter{0.532} & 0.1218 $\rightarrow$ \colorbetter{0.1279} \\
2: 3.04-5.28 & 0.0180 $\rightarrow$ \colorbetter{0.0144} & 0.0306 $\rightarrow$ \colorbetter{0.0244} & 0.287 $\rightarrow$ \colorbetter{0.447} & 0.429 $\rightarrow$ \colorbetter{0.554} & 0.0687 $\rightarrow$ \colorbetter{0.1249} \\
3: 5.28-12.88 & 0.0349 $\rightarrow$ \colorbetter{0.0257} & 0.0589 $\rightarrow$ \colorbetter{0.0437} & 0.243 $\rightarrow$ \colorbetter{0.476} & 0.399 $\rightarrow$ \colorbetter{0.581} & 0.0471 $\rightarrow$ \colorbetter{0.1215} \\
4: 12.88-84.87 & 0.0960 $\rightarrow$ \colorbetter{0.0860} & 0.1681 $\rightarrow$ \colorbetter{0.1542} & 0.422 $\rightarrow$ \colorbetter{0.521} & 0.540 $\rightarrow$ \colorbetter{0.626} & 0.0578 $\rightarrow$ \colorbetter{0.1071} \\
5: 84.87-559.26 & 0.5460 $\rightarrow$ \colorbetter{0.5327} & 1.0069 $\rightarrow$ \colorbetter{0.9718} & 0.559 $\rightarrow$ \colorbetter{0.588} & 0.630 $\rightarrow$ \colorbetter{0.671} & 0.0503 $\rightarrow$ \colorbetter{0.0851} \\
6: 559.26-160002 & 7.9160 $\rightarrow$ \colorworse{7.9300} & 14.408 $\rightarrow$ \colorbetter{14.003} & 0.557 $\rightarrow$ \colorbetter{0.617} & 0.641 $\rightarrow$ \colorbetter{0.693} & 0.0410 $\rightarrow$ \colorbetter{0.0840} \\
\bottomrule
\end{tabular}
\end{table}

\clearpage
\newpage

\section{Performance in an Ungauged Setting}

\subsection{Impact of distributed modeling on ungauged EFAS reanalysis}
\label{sec:appendix_ungauged_reanalysis}

\begin{figure}[!h]
  \centering
  \includegraphics[width=.5\textwidth]{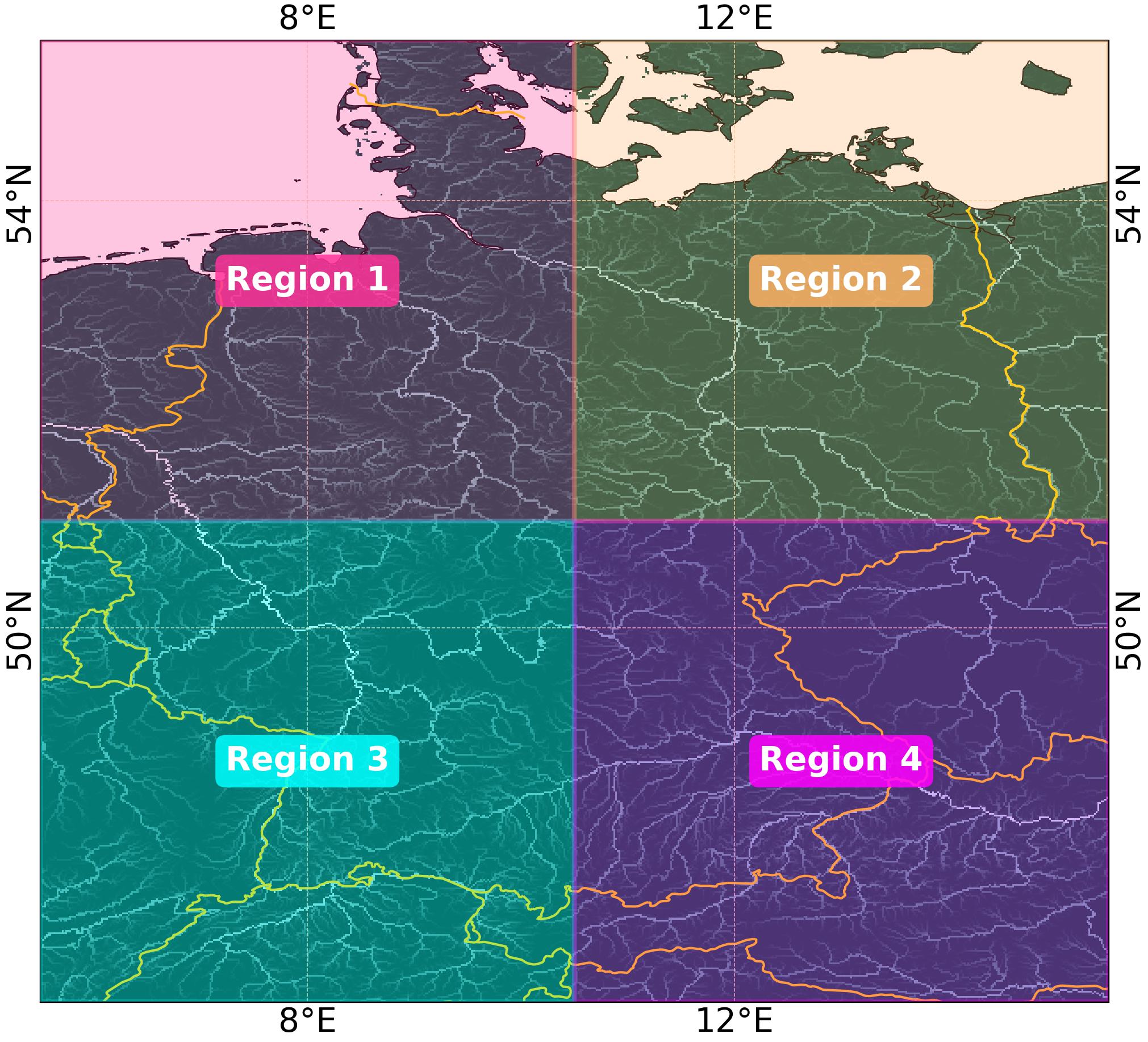}
  \caption{The $4$ spatial-holdout regions used to construct the ungauged evaluation in Table \ref{table:ablation_ungauged_1} and Fig.~\ref{fig:ungauged_1}. 
  The Mid European domain is split into $4$ equal quadrants at its geometric midpoint (51$^\circ$N, 10.5$^\circ$E). Region $1$ (top-left), Region $2$ (top-right), Region $3$ (bottom-left) and Region $4$ (bottom-right).}\label{fig:ungauged_regions}
\end{figure}

We repeat the routing ablation from Sec.~\ref{sec:appendix_distribution_routing} in a fully ungauged setting, using the same $6$ classification classes of upstream area (Fig.~\ref{fig:distribution_classes}). The Mid European domain is split into $4$ quadrants (top-left, top-right, bottom-left, and bottom-right). For each experiments (with routing and without routing), we train $4$ models with one holdout region. Training is done until $2021$ so every point in the final map is genuinely out-of-sample in space and time in an ungauged setting on reanalysis. For evaluation, we evaluate on all $280,820$ grid points for $16$ lead time + $1$ nowcasting for the years $2022$-$2025$. Table~\ref{table:ablation_ungauged_1} reports the results for MAE, RMSE, R2, KGE, and F1 per class. Colored \colorbetter{Cyan} indicates where a model with routing is better than a model without routing and \colorworse{Purple} where it is worse. The effect of routing is far larger here than in the gauged setting (Sec.~\ref{sec:appendix_distribution_routing}). 

Without routing, the ungauged model is inaccurate.
Adding routing turns R2 positive for classes 1-5 (e.g. class 1: $-6.758 \rightarrow 0.089$; class 3: $-1.034 \rightarrow 0.109$) and makes KGE positive for every class, including class 6 ($-0.036 \rightarrow 0.239$). 
MAE and RMSE improve at every class with no exceptions.
For F1, routing improves detection at every class.

\begin{figure}[!h]
  \centering
  \includegraphics[width=.98\textwidth]{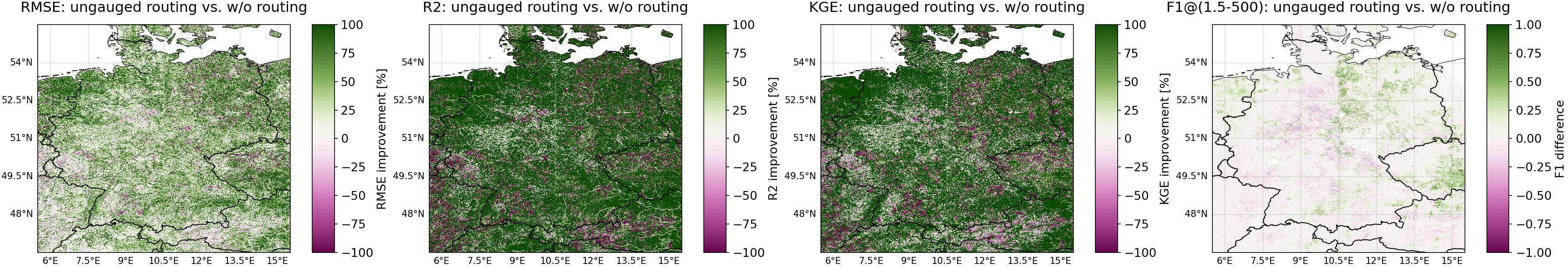}
  \caption{Per pixel percentage improvement for routing over without routing in the ungauged (spatial-holdout) setting. Metrics RMSE$^r$, R2$^r$, KGE$^r$, and F1$^r$@(1.5-500) are averaged over lead time for the years $2022$-$2025$.
  For each grid point, the percentage improvement is computed as
  $ 
  \Delta{\text{RMSE}} = \frac{\text{RMSE}_{\text{(w/o routing)}} - \text{RMSE}_{\text{(routing)}}}{\text{RMSE}_{\text{(w/o routing)}}} \times 100,~
    \Delta{\text{R2}} = \frac{\text{R2}_{\text{(routing)}} - \text{R2}_{\text{(w/o routing)}}}{|\text{R2}_{\text{(w/o routing)}}|} \times 100,~ \Delta{\text{KGE}} = \frac{\text{KGE}_{\text{(routing)}} - \text{KGE}_{\text{(w/o routing)}}}{|\text{KGE}_{\text{(w/o routing)}}|} \times 100 \,,
    $ where subscripts "routing" and "w/o routing" denote a model with routing and without it. 
    F1 is computed per pixel and the plotted quantity is the plain difference $
    \Delta{\text{F1}} = \text{F1}_{\text{(routing)}} - \text{F1}_{\text{(w/o routing)}}. $ The F1 values are bounded in $[-1, 1]$. Positive (green) values indicate routing improves the metric at that pixel and negative (magenta) values indicate it is worse.}\label{fig:ungauged_1}
\end{figure}

Fig.~\ref{fig:ungauged_1} shows the spatial impact of the routing module on the forecasting skill in an ungauged setting.

\begin{table}[!h]
\small
 \caption{Ungauged (spatial-holdout) performance of routing against without routing by upstream-area class. The evaluation is at all $280,820$ EFAS grid points over the years $2022$-$2025$. For each class and metric, values are the class-median of the per pixel metric and averaged across the $17$ lead times. \colorbetter{Cyan} marks points where routing improves the metric and \colorworse{Purple} marks points where it is worse.}\label{table:ablation_ungauged_1}
 \centering
 \setlength\tabcolsep{2pt}
 \begin{tabular}{l *{6}l}
 \toprule
 Class (Uparea km$^2$) & MAE ($\downarrow$) & RMSE ($\downarrow$) & R2 ($\uparrow$) & KGE ($\uparrow$) & F1@(1.5-500) ($\uparrow$)  \\
\midrule
1: 1.95-3.04 & 0.0346 $\rightarrow$ \colorbetter{0.0126} & 0.0536 $\rightarrow$ \colorbetter{0.0183} & -6.758 $\rightarrow$ \colorbetter{0.089} & -1.159 $\rightarrow$ \colorbetter{0.237} & 0.0539 $\rightarrow$ \colorbetter{0.0543} \\
2: 3.04-5.28 & 0.0502 $\rightarrow$ \colorbetter{0.0228} & 0.0835 $\rightarrow$ \colorbetter{0.0338} & -2.085 $\rightarrow$ \colorbetter{0.105} & -0.349 $\rightarrow$ \colorbetter{0.268} & 0.0391 $\rightarrow$ \colorbetter{0.0519} \\
3: 5.28-12.88 & 0.0843 $\rightarrow$ \colorbetter{0.0409} & 0.1348 $\rightarrow$ \colorbetter{0.0616} & -1.034 $\rightarrow$ \colorbetter{0.109} & -0.098 $\rightarrow$ \colorbetter{0.291} & 0.0334 $\rightarrow$ \colorbetter{0.0531} \\
4: 12.88-84.87 & 0.2624 $\rightarrow$ \colorbetter{0.1469} & 0.4032 $\rightarrow$ \colorbetter{0.2264} & -0.633 $\rightarrow$ \colorbetter{0.119} & -0.031 $\rightarrow$ \colorbetter{0.300} & 0.0288 $\rightarrow$ \colorbetter{0.0454} \\
5: 84.87-559.26 & 1.4830 $\rightarrow$ \colorbetter{0.9720} & 2.1890 $\rightarrow$ \colorbetter{1.4710} & -0.602 $\rightarrow$ \colorbetter{0.143} & -0.044 $\rightarrow$ \colorbetter{0.319} & 0.0266 $\rightarrow$ \colorbetter{0.0390} \\
6: 559.26-160002 & 20.564 $\rightarrow$ \colorbetter{14.658} & 28.943 $\rightarrow$ \colorbetter{22.410} & -0.918 $\rightarrow$ \colorbetter{-0.144} & -0.036 $\rightarrow$ \colorbetter{0.239} & 0.0186 $\rightarrow$ \colorbetter{0.0223} \\
\bottomrule
\end{tabular}
\end{table}

\subsection{performance on ungauged CAMELS observations}
\label{sec:appendix_ungauged_camels}
\begin{figure}[!h]
  \centering
  \includegraphics[width=.4\textwidth]{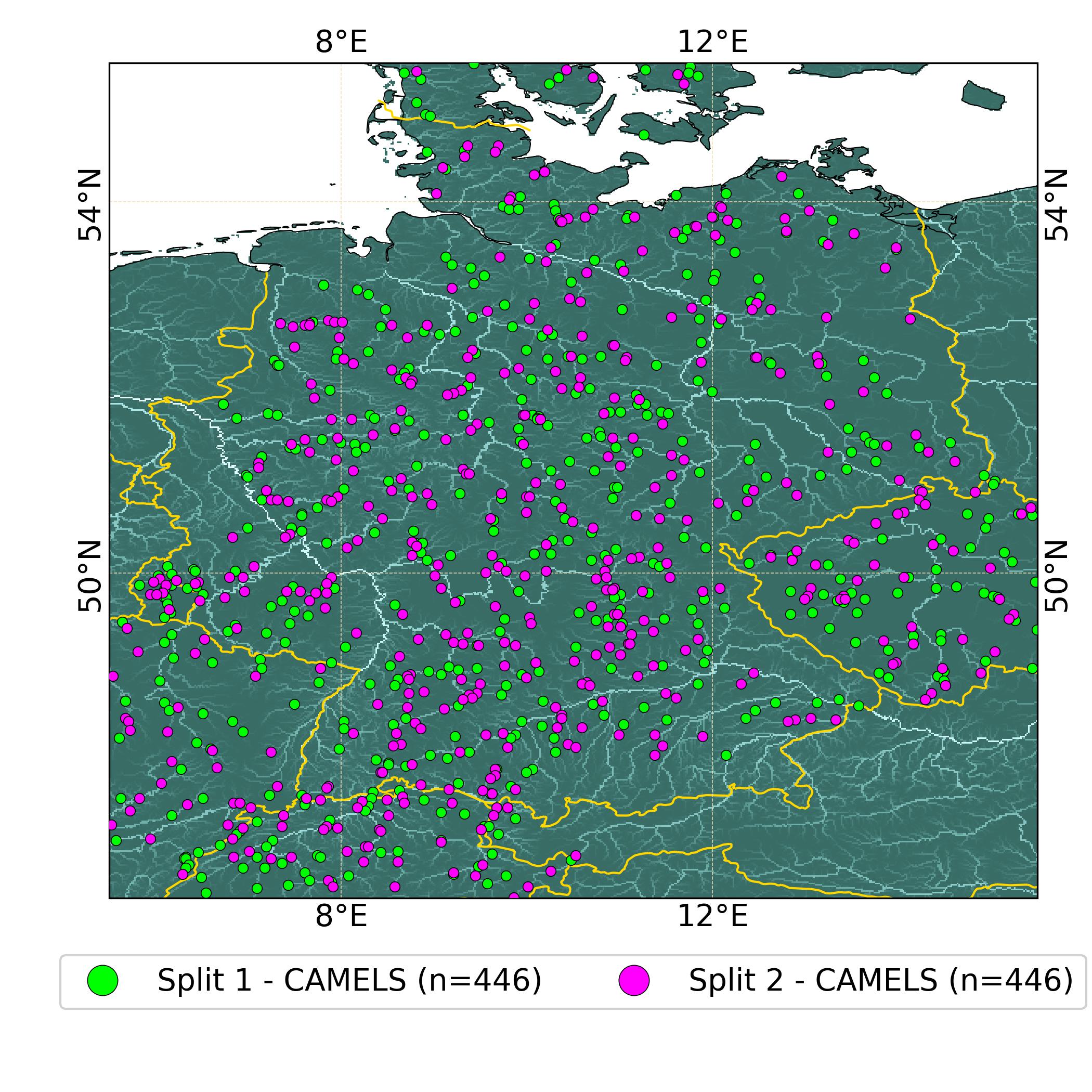}
  \caption{The two random holdout splits of the $892$ CAMELS stations ($446$ each) used for the fully ungauged evaluation.}\label{fig:ungauged_camels_splits}
\end{figure}

\begin{figure}[!h]
  \centering
  \includegraphics[width=.98\textwidth]{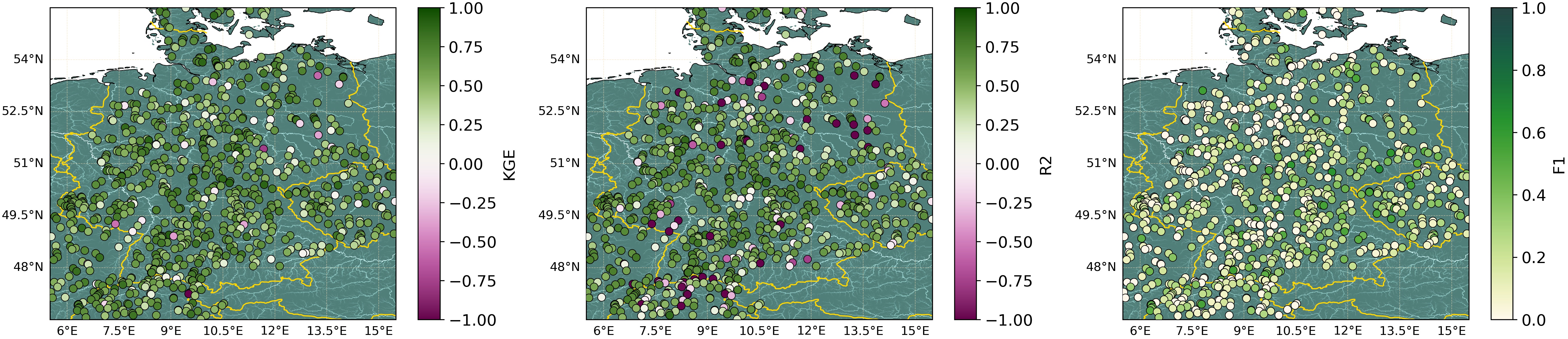}
  \caption{Per station KGE, R2, and F1 of the ungauged CAMELS prediction. Evaluation is done for the years $2010$ to $2024$ and for $4$ days lead time + $1$ nowcasting.}\label{fig:ungauged_camels_map}
\end{figure}

We further test the fully ungauged setting against real station observations rather than EFAS reanalysis. 
The $892$ CAMELS stations are split into two random halves $446$ stations each (Fig.~\ref{fig:ungauged_camels_splits}).
We trained $2$ models on each half of the stations for $4$ days lead time + $1$ nowcasting step against real observed discharge. 
For every station, we then keep only the prediction from the model that never saw it during training and evaluate for the years $2010$ to $2024$. 
Table~\ref{table:ungauged_obs_camels} and
Fig.~\ref{fig:ungauged_camels_map} show that the model transfers well to unseen gauges.
KGE is positive at the large majority of stations (mean $0.563$, median $0.602$), R2 is positive at most stations (mean $0.316$, median $0.581$), and floods are detected well across all return periods (F1 mean $0.185$).
\begin{table}[!h]
\centering
\caption{Per station skill of the ungauged CAMELS prediction against observations.}\label{table:ungauged_obs_camels}
\begin{tabular}{lccc}
\toprule
 & KGE ($\uparrow$) & R2 ($\uparrow$) & F1@(1.5-500) ($\uparrow$)\\
\midrule
mean   & 0.563 & 0.316 & 0.185 \\
median & 0.602 & 0.581 & 0.159 \\
\bottomrule
\end{tabular}
\end{table}

\subsection{performance on ungauged GRDC observations}
\label{sec:appendix_ungauged_grdc}

\begin{figure}[!h]
  \centering
  \includegraphics[width=.4\textwidth]{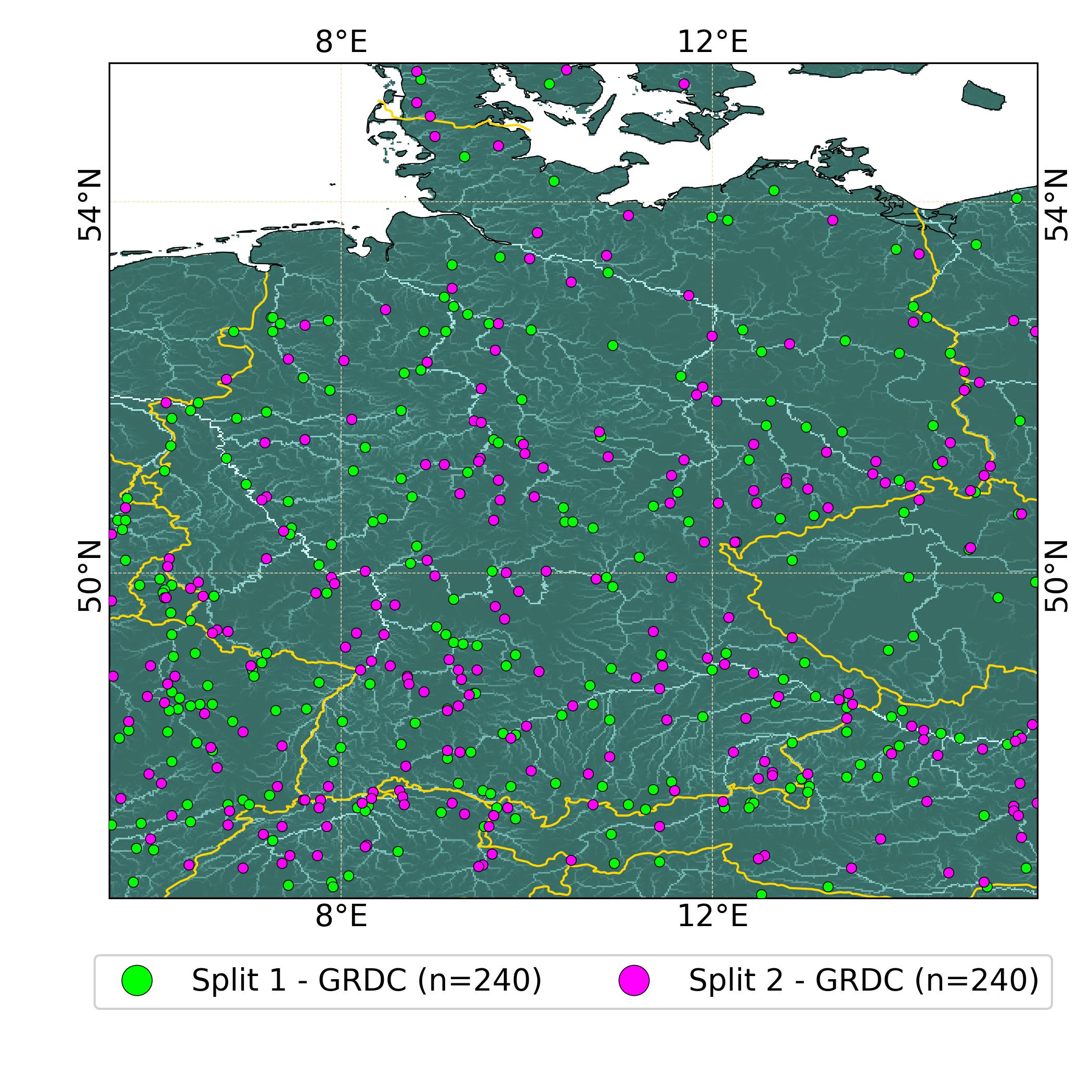}
  \caption{The two random holdout splits of the $480$ GRDC stations ($240$ each) used for the fully ungauged evaluation.}\label{fig:ungauged_grdc_splits}
\end{figure}

\begin{figure}[!h]
  \centering
  \includegraphics[width=.98\textwidth]{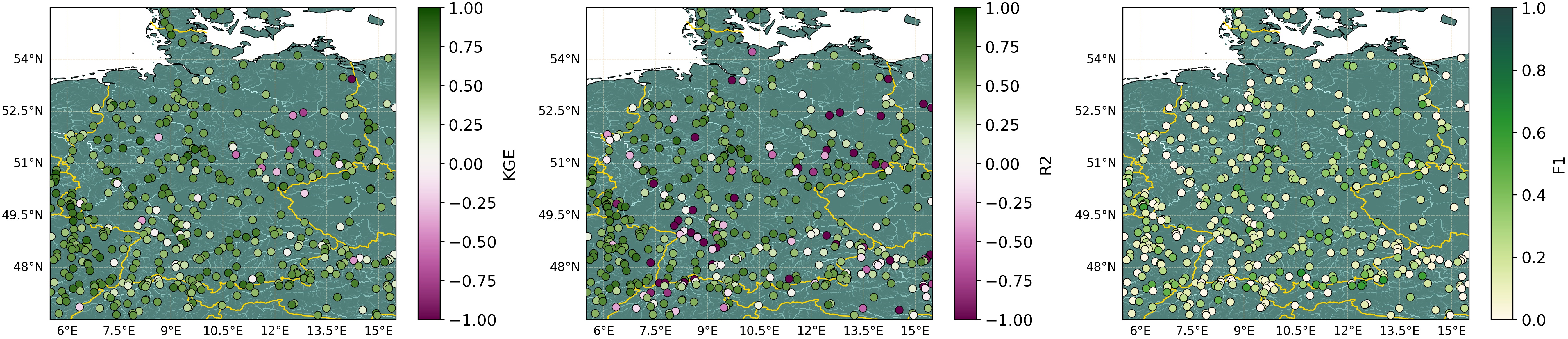}
  \caption{Per station KGE, R2, and F1 of the ungauged GRDC prediction. Evaluation is done for the years $2010$ to $2025$ and for $4$ days lead time + $1$ nowcasting.}\label{fig:ungauged_grdc_map}
\end{figure}

We repeat the same protocol in Sec.~\ref{sec:appendix_ungauged_camels} for the $480$ GRDC stations.
The stations are split into $2$ random halves $240$ each (Fig.~\ref{fig:ungauged_grdc_splits}),
We trained $2$ models on each half of the stations for $4$ days lead time + $1$ nowcasting step against real observed discharge. 
We evaluate on years $2010$ to $2025$, such that every station is evaluated by the model that did not see it during training.
The model also generalizes well to unseen GRDC gauges. Table~\ref{table:ungauged_obs_grdc} and Fig.~\ref{fig:ungauged_grdc_map} show the results. 
Mean KGE is $0.542$ (median $0.595$), mean R2 is
$0.202$ (median $0.545$), and F1 remains skillful across return periods (mean $0.200$).
\begin{table}[!h]
\centering
\caption{Per station skill of the ungauged GRDC prediction against observations.}\label{table:ungauged_obs_grdc}
\begin{tabular}{lccc}
\toprule
 & KGE ($\uparrow$) & R2 ($\uparrow$) & F1@(1.5-500) ($\uparrow$)\\
\midrule
mean   & 0.542 & 0.202 & 0.200 \\
median & 0.595 & 0.545 & 0.175 \\
\bottomrule
\end{tabular}
\end{table}

\newpage
\clearpage

\section{Comparison to RiverMamba}
\label{sec:appendix_rivermamba}

\subsection{Comparison to RiverMamba on gauged EFAS Reanalysis}
\label{sec:appendix_rivermamba_reanalysis}
\begin{figure}[h!]
  \centering
  \includegraphics[width=0.9\textwidth]{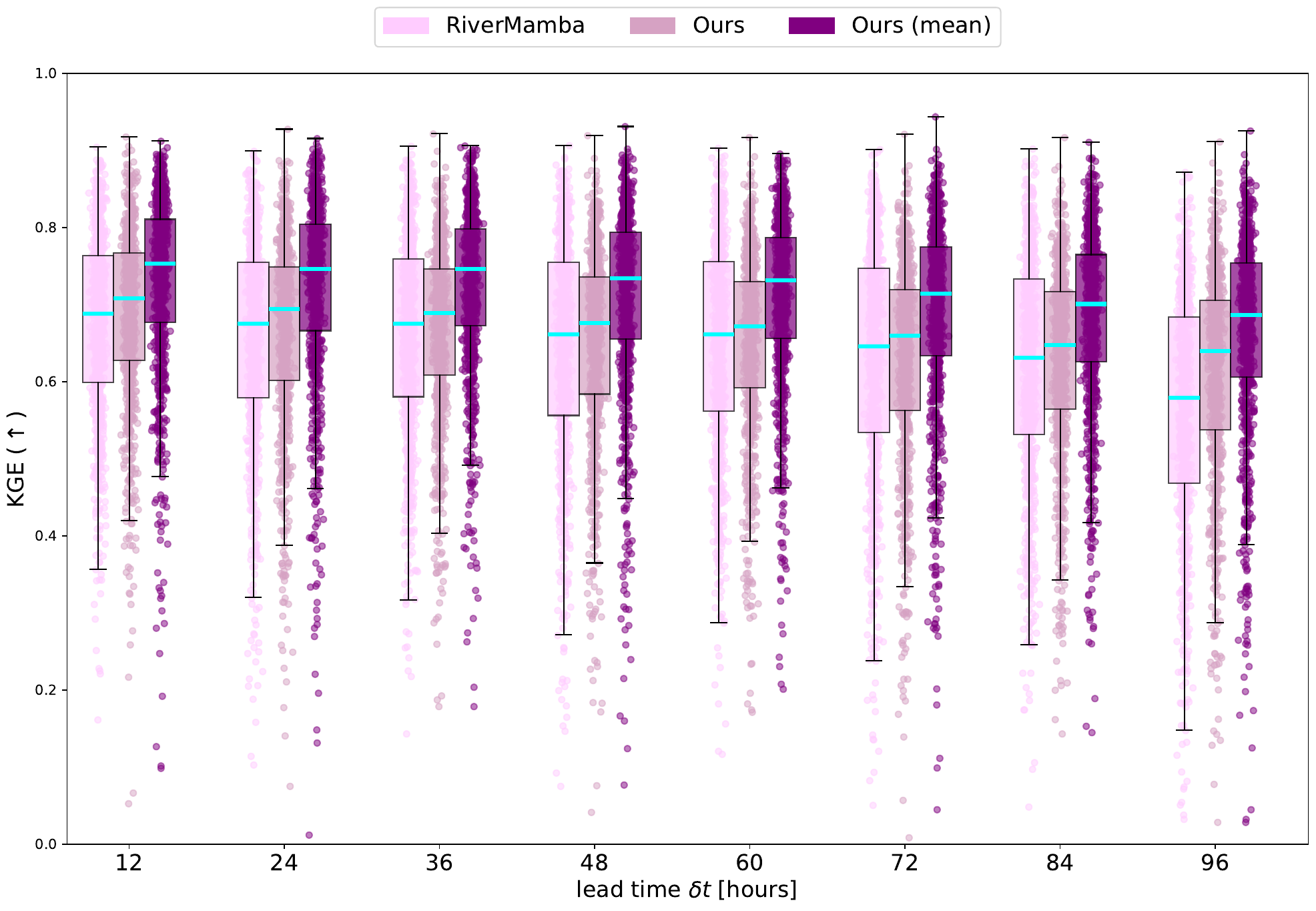}
  \caption{KGE of the river discharge forecasting with different lead time on CAMELS reanalysis split (test set $2024$-$2025$ temporally out-of-sample).  Distribution quartiles are displayed in boxes, and the entire range excluding outliers is displayed in whiskers. The median score for the model is shown by the cyan line in the box.}
  \label{fig:appendix_rivermamba_reanalysis_camels_kge}
\end{figure}

\begin{figure}[h!]
  \centering
  \includegraphics[width=0.9\textwidth]{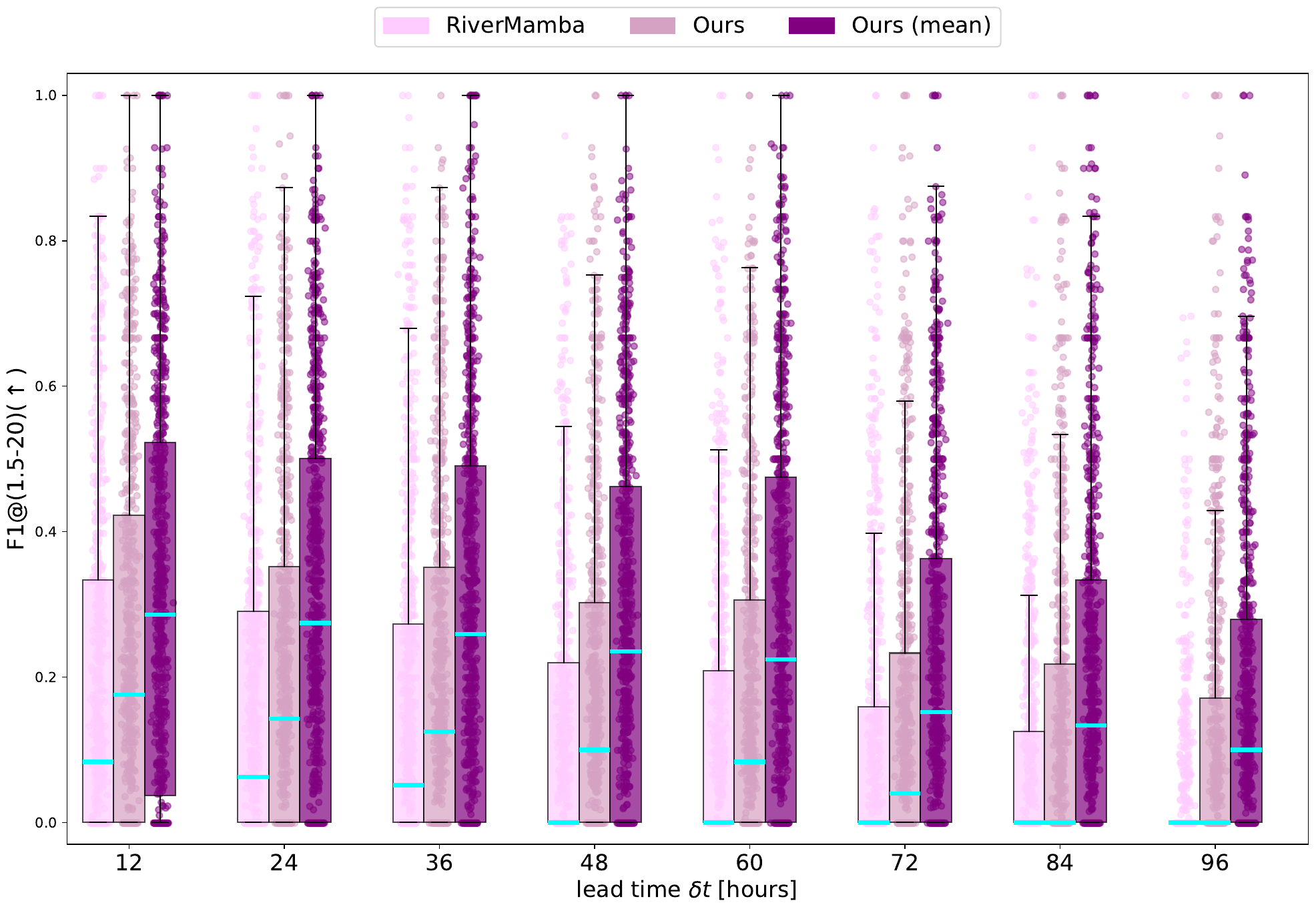}
  \caption{Averaged F1-score for RPs 1.5-20 with lead time on CAMELS reanalysis split (test set $2024$-$2025$ temporally out-of-sample). Distribution quartiles are displayed in boxes, and the entire range excluding outliers is displayed in whiskers. The median score for the model is shown by the cyan line in the box.}
  \label{fig:appendix_rivermamba_reanalysis_camels_f1}
\end{figure}

In this experiment, we compare to RiverMamba developed by \citetsupp{RiverMamba}. RiverMamba does not provide forecast for EFAS domain. In addition, it was trained with different spatio-temporal resolution and input data. For this, we train RiverMamba on our dataset over EFAS domain and with the same input data. For the expieremnt, we train over the Mid European domain up to $16$ lead time steps ($96$ hours). 
Note that in principle RiverMamba is not scalable to long-term forecast. It has fixed learned horizon for the forecast (fixed lead time). Another difference is that RiverMamba does not provide probabilistic forecast nor it provides nowcasting. Our model in contrast, uses rollout for the forecast and can be scaled up to longer lead time. We could even increase the rollout step during inference compared to the training setting. In addition, our model provide nowcasting and probabilistic forecast (Table~\ref{table:rivermamba_compare}). Tables~\ref{table:rivermamba_1} and \ref{table:rivermamba_2} show the results for EFAS reanalysis on both CAMELS and GRDC splits. As can be seen, our probabilistic model achieves the best results overall, while our deterministic model is consistently outperforming RiverMamba on F1 score. This is notable with the evaluation on the full domain with $280,820$ points (Table~\ref{table:rivermamba_3}). The RiverMamba baseline has no explicit routing mechanism inside the model. It relies mainly on the spatio-temporal modeling of SSM to solve the routing implicitly, while our model does this explicitly with the routing module which improves flood detection. This is also consistent with the results in Table~\ref{table:ablation} (a) for the study of the routing impact. 
More results are provided in Figs.~\ref{fig:appendix_rivermamba_reanalysis_camels_kge}-\ref{fig:appendix_rivermamba_reanalysis_camels_f1}.

\begin{table}[h!]
 \caption[Comparison to RiverMamba]{Comparison to RiverMamba~\citepsupp{RiverMamba}.}  \label{table:rivermamba_compare}
  \centering
  \tabcolsep=4.0pt\relax
  \setlength\extrarowheight{0pt}
  \begin{tabular}{r*{4}c}
  
    \toprule
    Model & \# Param. & auto-regressive & Probabilistic & Nowcasting \\
    \midrule
    RiverMamba & 4.4 million & \xmark & \xmark & \xmark \\
    \midrule
    Ours & 2.3 million & \cmark & \cmark & \cmark \\
    \bottomrule
  \end{tabular}
\end{table}

\begin{table}[!h]
  \caption{Results on EFAS Reanalysis (CAMELS split). \gray{($\pm$)} denotes the standard deviation for 3 runs.}
  \label{table:rivermamba_1}
  \centering
  \small
  \tabcolsep=2.0pt\relax
  \setlength\extrarowheight{0pt}
  \begin{tabular}{c r *{8}l}
  \toprule
  & & & \multicolumn{3}{c}{Validation (2022-2023)} & & \multicolumn{3}{c}{Test (2024-2025)} \\
    \midrule
    & Model & & R2 {\scriptsize($\uparrow$)} & KGE {\scriptsize($\uparrow$)} & F1{\scriptsize@(1.5-20) ($\uparrow$)} & & R2 {\scriptsize($\uparrow$)} & KGE {\scriptsize($\uparrow$)} & F1{\scriptsize@(1.5-20) ($\uparrow$)}  \\
    \midrule
   \multirow{2}{*}{\rotatebox[origin=c]{90}{\tiny{Determ}}} & RiverMamba & & 0.7075{\scriptsize\color{gray}{$\pm$0.0138}} & 0.7255{\scriptsize\gray{$\pm$0.0077}} & 0.1242{\scriptsize\gray{$\pm$0.0160}} & & 0.5806{\scriptsize\gray{$\pm$0.0221}} & 0.6580{\scriptsize\gray{$\pm$0.0118}} & 0.0611{\scriptsize\gray{$\pm$0.0104}} \\
  & Ours & & 0.7075{\scriptsize\color{gray}{$\pm$0.0056}} & 0.7410{\scriptsize\gray{$\pm$0.0134}} & 0.1627{\scriptsize\gray{$\pm$0.0143}} & & 0.5676{\scriptsize\gray{$\pm$0.0097}} & 0.6589{\scriptsize\gray{$\pm$0.0162}} & 0.0841{\scriptsize\gray{$\pm$0.0037}} \\
 \midrule
   \multirow{1}{*}{\rotatebox[origin=c]{90}{\tiny{Prob}}} & Ours (mean) & & \textbf{0.7247} & \textbf{0.7695} & \textbf{0.2129} & & \textbf{0.6280} & \textbf{0.7284} & \textbf{0.2266} \\
  \bottomrule
\end{tabular}
\end{table}

\begin{table}[!h]
\vspace{-2mm}
  \caption{Results on EFAS Reanalysis (GRDC split). \gray{($\pm$)} denotes the standard deviation for 3 runs.}
  \label{table:rivermamba_2}
  \centering
  \small
  \tabcolsep=2.0pt\relax
  \setlength\extrarowheight{0pt}
  \begin{tabular}{c r *{8}l}
  \toprule
  & & & \multicolumn{3}{c}{Validation (2022-2023)} & & \multicolumn{3}{c}{Test (2024-2025)} \\
    \midrule
    & Model & & R2 {\scriptsize($\uparrow$)} & KGE {\scriptsize($\uparrow$)} & F1{\scriptsize@(1.5-20) ($\uparrow$)} & & R2 {\scriptsize($\uparrow$)} & KGE {\scriptsize($\uparrow$)} & F1{\scriptsize@(1.5-20) ($\uparrow$)}  \\
    \midrule
   \multirow{2}{*}{\rotatebox[origin=c]{90}{\tiny{Determ}}} & RiverMamba & & 0.7326{\scriptsize\color{gray}{$\pm$0.0099}} & 0.7579{\scriptsize\gray{$\pm$0.0052}} & 0.1106{\scriptsize\gray{$\pm$0.0125}} & & 0.6161{\scriptsize\gray{$\pm$0.0222}} & 0.6927{\scriptsize\gray{$\pm$0.0141}} & 0.0990{\scriptsize\gray{$\pm$0.0151}} \\
  & Ours & & 0.6957{\scriptsize\color{gray}{$\pm$0.0273}} & 0.7507{\scriptsize\gray{$\pm$0.0273}} & 0.1326{\scriptsize\gray{$\pm$0.0272}} & & 0.5631{\scriptsize\gray{$\pm$0.0237}} & 0.6645{\scriptsize\gray{$\pm$0.0241}} & 0.1132{\scriptsize\gray{$\pm$0.0136}} \\
 \midrule
   \multirow{1}{*}{\rotatebox[origin=c]{90}{\tiny{Prob}}} & Ours (mean) & & \textbf{0.7359} & \textbf{0.7876} & \textbf{0.1998} & & \textbf{0.6505} & \textbf{0.7552} & \textbf{0.2125} \\
  \bottomrule
\end{tabular}
\end{table}

\begin{table}[!h]
  \caption{Results on EFAS Reanalysis (Full mid Europe domain). \gray{($\pm$)} denotes the standard deviation for 3 runs.}
  \label{table:rivermamba_3}
  \centering
  \small
  \tabcolsep=2.0pt\relax
  \setlength\extrarowheight{0pt}
  \begin{tabular}{c r *{6}l}
  \toprule
  & & & \multicolumn{2}{c}{Validation (2022-2023)} & & \multicolumn{2}{c}{Test (2024-2025)} \\
    \midrule
    & Model & & F1{\scriptsize@(1.5-20) ($\uparrow$)} & F1{\scriptsize@(1.5-500) ($\uparrow$)} & & F1{\scriptsize@(1.5-20) ($\uparrow$)} & F1{\scriptsize@(1.5-500) ($\uparrow$)}  \\
    \midrule
   \multirow{2}{*}{\rotatebox[origin=c]{90}{\tiny{Determ}}} & RiverMamba & & 0.1670{\scriptsize\color{gray}{$\pm$0.0122}} & 0.1407{\scriptsize\gray{$\pm$0.0099}} & & 0.1713{\scriptsize\gray{$\pm$0.0134}} & 0.1379{\scriptsize\gray{$\pm$0.0108}} \\
  & Ours & & \textbf{0.3542}{\scriptsize\color{gray}{$\pm$0.0199}} & \textbf{0.3337}{\scriptsize\gray{$\pm$0.0176}} & & \textbf{0.3183}{\scriptsize\gray{$\pm$0.0191}} & \textbf{0.3143}{\scriptsize\gray{$\pm$0.0129}} \\
  \bottomrule
\end{tabular}
\end{table}


\clearpage
\newpage

\begin{figure}[h!]
  \centering
  \includegraphics[width=0.98\textwidth]{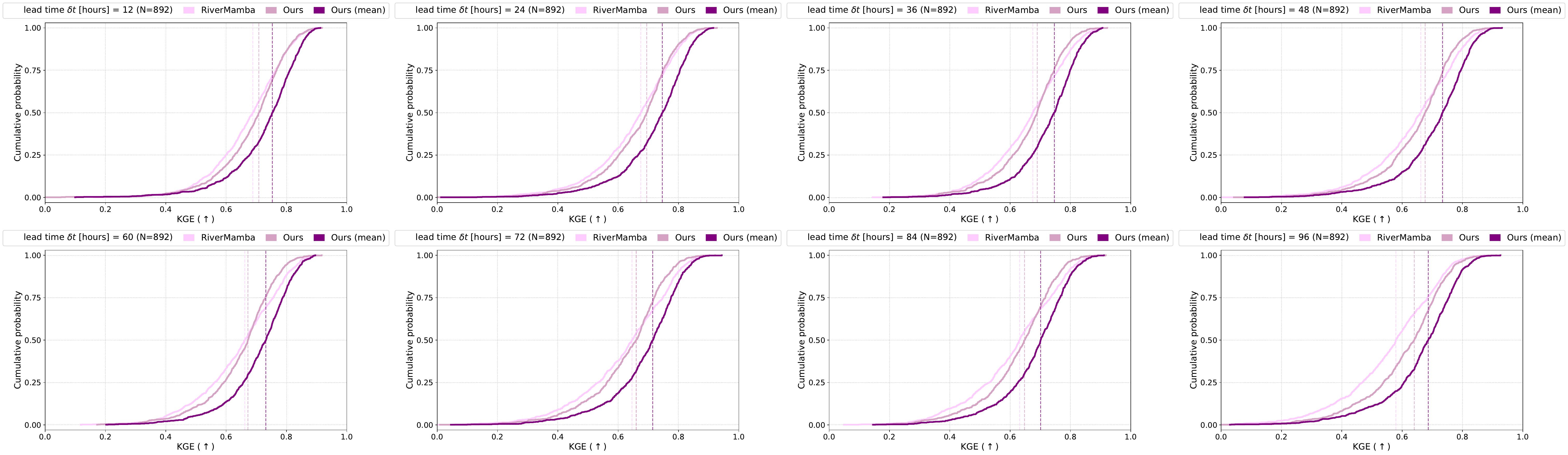}
  \caption{Cumulative distribution function (CDF) of station-level KGE on the CAMELS reanalysis split (N=892), across lead times. Dashed vertical lines shows each model's media.}
  \label{fig:appendix_rivermamba_camels_cdf_kge}
\end{figure}

\begin{figure}[h!]
  \centering
  \includegraphics[width=0.98\textwidth]{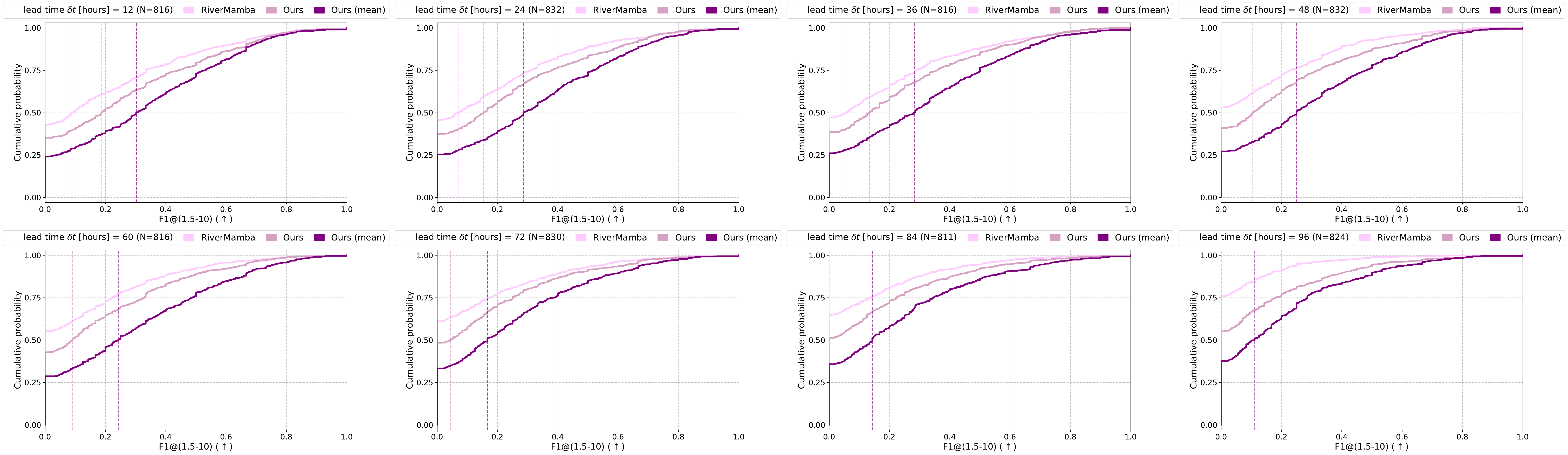}
  \caption{Cumulative distribution function (CDF) of station-level F1@(1.5-10) on the CAMELS reanalysis split (N=892), across lead times. Dashed vertical lines shows each model's media.}
  \label{fig:appendix_rivermamba_camels_cdf_f1}
\end{figure}

In Figs.~\ref{fig:appendix_rivermamba_camels_cdf_kge}-\ref{fig:appendix_rivermamba_camels_cdf_f1}, we show the empirical cumulative distribution functions (CDFs) of KGE and F1-score@(1.5-10) across lead time for both CAMELS (N=892) and GRDC (N=480) reanalysis splits. 
A curve shifted to the right indicates a better performing model. Ours (mean) model's curve lies uniformly to the right of both the deterministic model (Ours) and the RiverMamba baseline. This is consistent across lead times, metrics, and splits. This indicates that the gain is broad across the catchments rather than concentrated in a few ones.

\begin{figure}[h!]
  \centering
  \includegraphics[width=\textwidth]{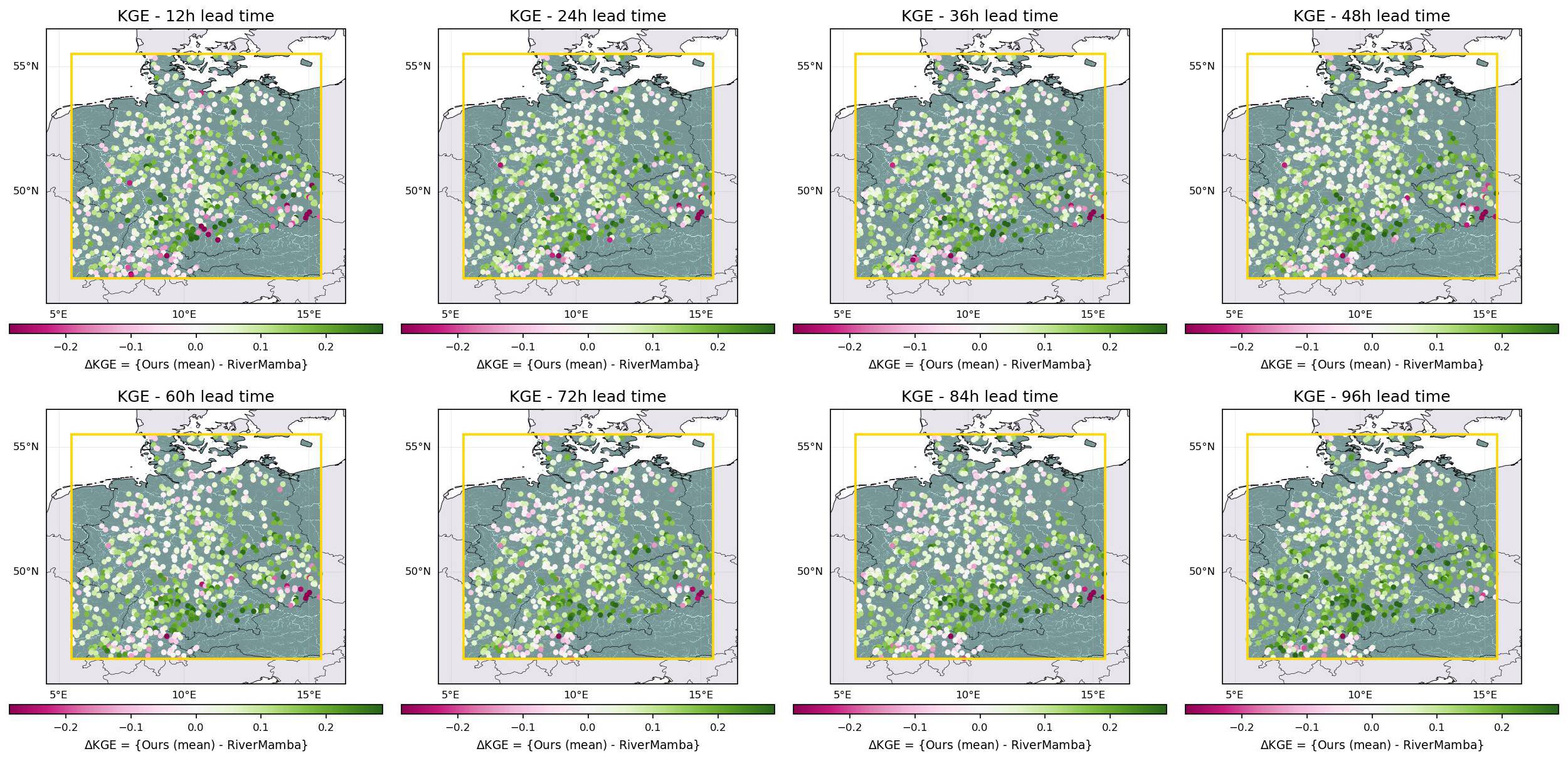}
  \caption{Comparison of KGE between Ours (mean) and RiverMamba on CAMELS EFAS reanalysis split across lead times (test set $2024$-$2025$ temporally out-of-sample).}
  \label{fig:appendix_rivermamba_camels_map_kge_ours}
\end{figure}

\begin{figure}[h!]
  \centering
  \includegraphics[width=\textwidth]{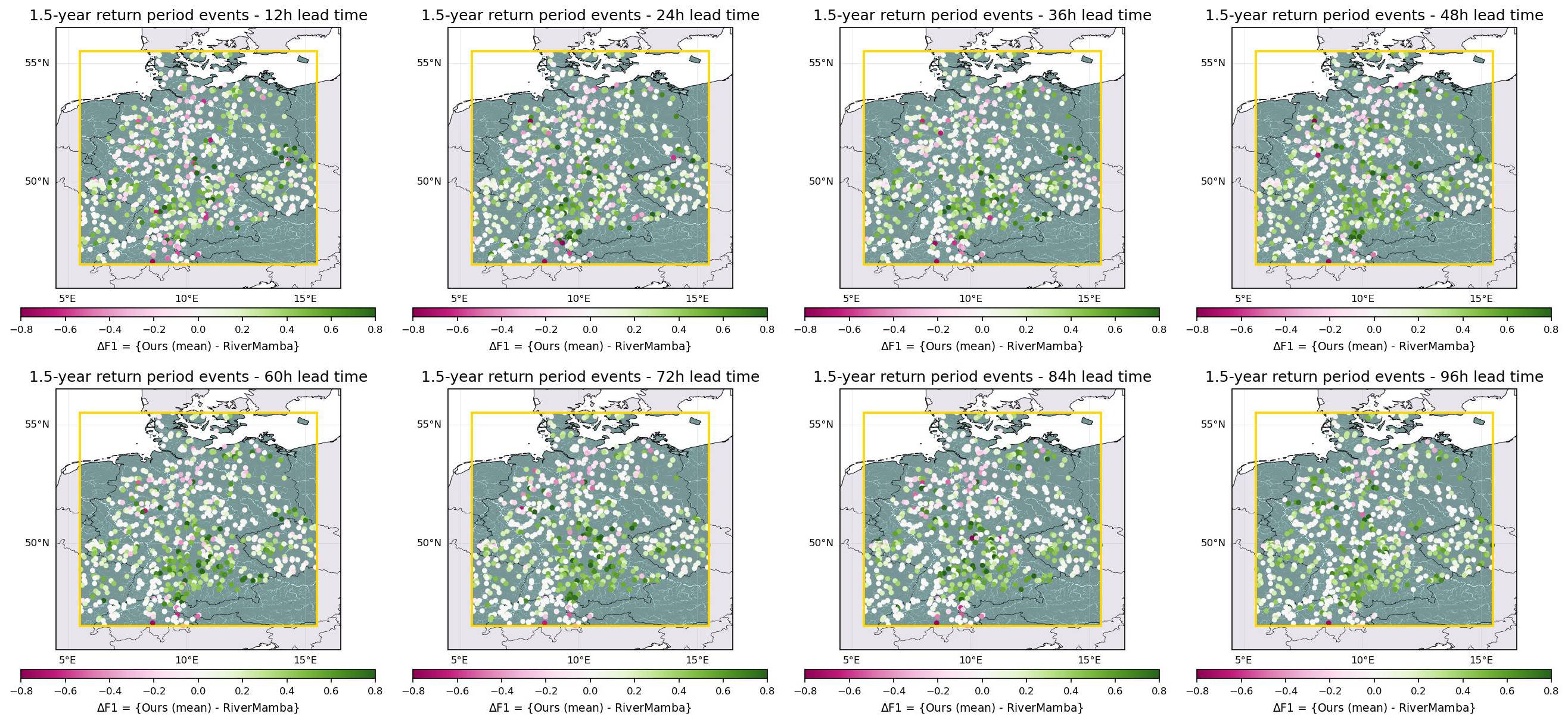}
  \caption{Comparison of F1-score between Ours (mean) and RiverMamba on CAMELS reanalysis EFAS split for the 1.5-year return period events across lead times (test set $2024$-$2025$ temporally out-of-sample).}
  \label{fig:appendix_rivermamba_camels_map_f1_our}
\end{figure}

In Figs.~\ref{fig:appendix_rivermamba_camels_map_kge_ours}-\ref{fig:appendix_rivermamba_camels_map_f1_our}, we compute the per-station difference in KGE and F1@(1.5) between each of our two model variants, the deterministic model (Ours) and the probabilistic model (Ours (mean)) and the RiverMamba baseline. 
Positive (\textcolor{green!60!black}{green}) values indicate our model outperforms the baseline, while negative (\textcolor{magenta}{magenta}) values indicate the opposite.


\subsection{Comparison to RiverMamba on gauged CAMELS Observations}

\begin{figure}[h!]
  \centering
  \includegraphics[width=0.9\textwidth]{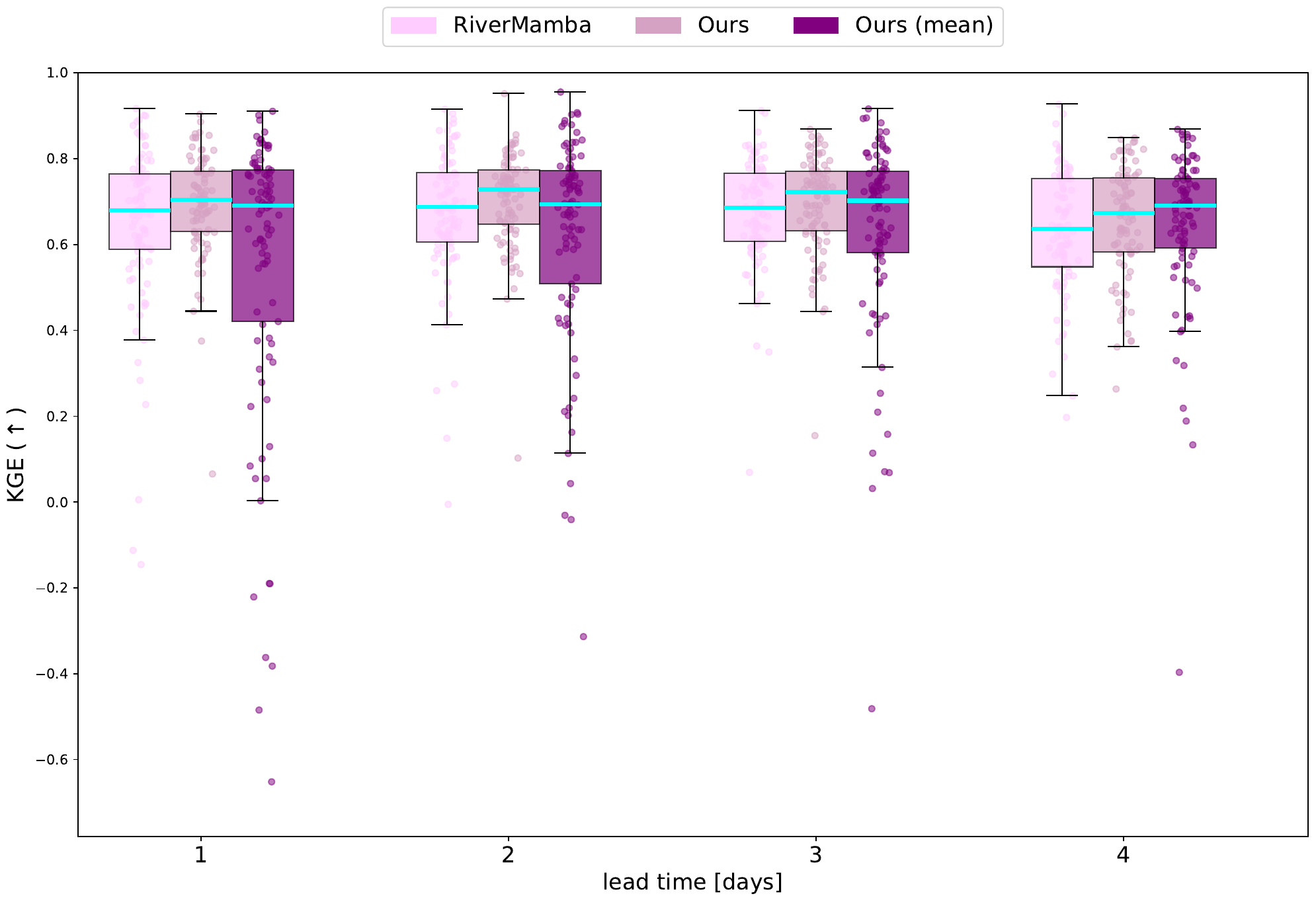}
  \caption{KGE of the river discharge forecasting with different lead time on CAMELS observations (test set $2024$ temporally out-of-sample). Distribution quartiles are displayed in boxes, and the entire range excluding outliers is displayed in whiskers. The median score for the model is shown by the cyan line in the box.}
  \label{fig:appendix_rivermamba_obs_camels_kge}
\end{figure}

\begin{figure}[h!]
  \centering
  \includegraphics[width=0.9\textwidth]{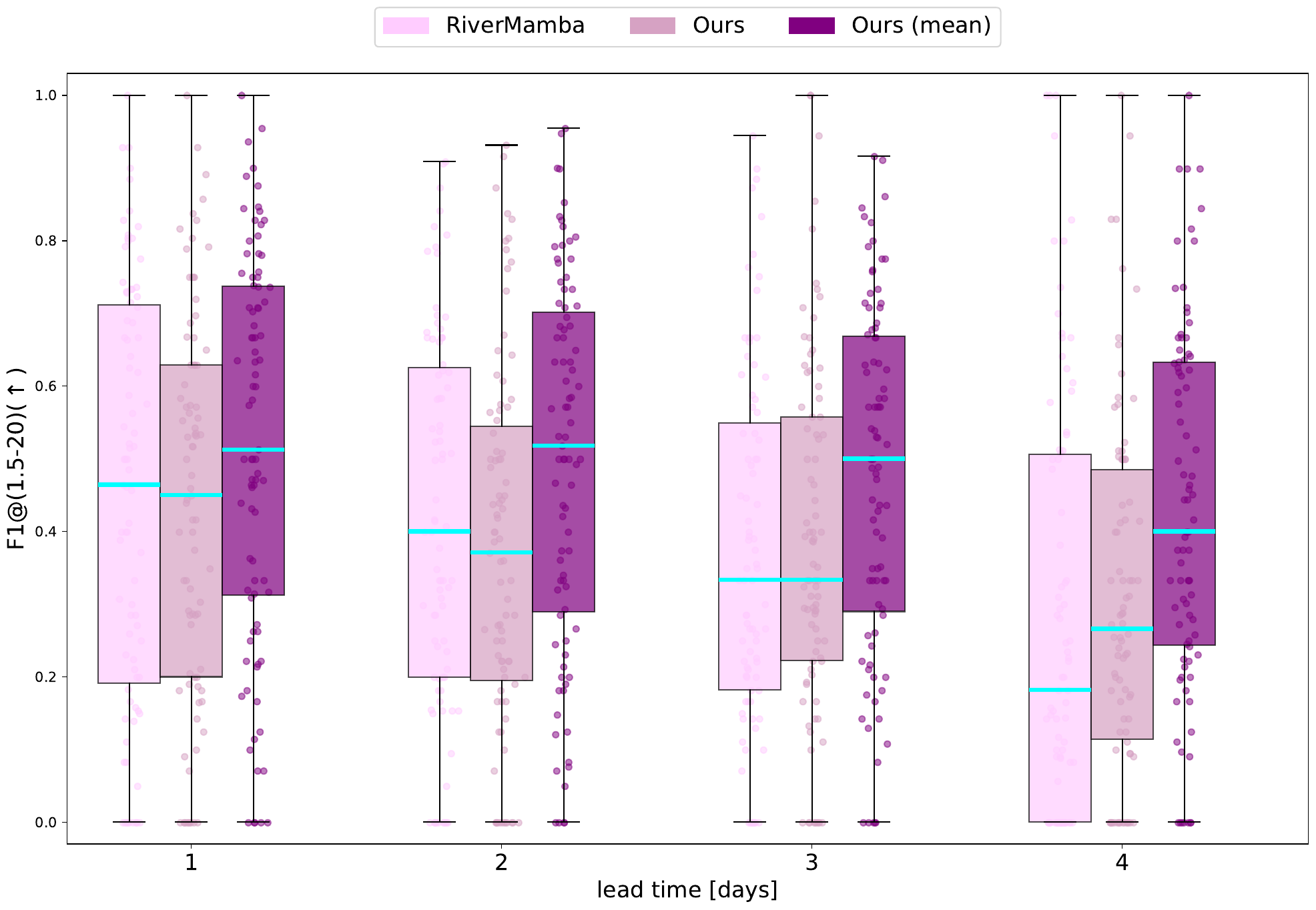}
  \caption{Averaged F1-score for 1.5-20 year return periods with lead time on CAMELS observations (test set $2024$ temporally out-of-sample). Distribution quartiles are displayed in boxes, and the entire range excluding outliers is displayed in whiskers. The median score for the model is shown by the cyan line in the box.}
  \label{fig:appendix_rivermamba_obs_camels_f1}
\end{figure}

In this section, we compare to RiverMamba on observational CAMELS data.
For this experiments, we used the models trained on reanalysis in Sec.~\ref{sec:appendix_rivermamba_reanalysis} and finetune them CAMELS observations for $4$ days over the Mid European domain. In Table~\ref{table:rivermamba_6} we show the quantitative results and more results are provided in Figs.~\ref{fig:appendix_rivermamba_obs_camels_kge} and \ref{fig:appendix_rivermamba_obs_camels_f1}.

\begin{table}[!h]
  \caption{Results on gauged CAMELS Observations. \gray{($\pm$)} denotes the standard deviation for 3 runs.}
  \label{table:rivermamba_6}
  \centering
  \small
  \tabcolsep=2.0pt\relax
  \setlength\extrarowheight{0pt}
  \begin{tabular}{c r *{8}l}
  \toprule
  & & & \multicolumn{3}{c}{Validation (2022-2023)} & & \multicolumn{3}{c}{Test (2024)} \\
    \midrule    
    & Model & & R2 {\scriptsize($\uparrow$)} & KGE {\scriptsize($\uparrow$)} & F1{\scriptsize@(1.5-20) ($\uparrow$)} & & R2 {\scriptsize($\uparrow$)} & KGE {\scriptsize($\uparrow$)} & F1{\scriptsize@(1.5-20) ($\uparrow$)}  \\
    \midrule
   \multirow{2}{*}{\rotatebox[origin=c]{90}{\tiny{Determ}}} & RiverMamba & & 0.7236{\scriptsize\color{gray}{$\pm$0.0000}} & 0.7563{\scriptsize\gray{$\pm$0.0113}} & \textbf{0.3181}{\scriptsize\gray{$\pm$0.0041}} & & 0.5428{\scriptsize\gray{$\pm$0.0151}} & 0.6257{\scriptsize\gray{$\pm$0.0235}} & 0.2864{\scriptsize\gray{$\pm$0.0675}} \\
  & Ours & & \textbf{0.7286}{\scriptsize\color{gray}{$\pm$0.0011}} & 0.7716{\scriptsize\gray{$\pm$0.0065}} & 0.2883{\scriptsize\gray{$\pm$0.0008}} & & \textbf{0.6216}{\scriptsize\gray{$\pm$0.0313}} & 0.6761{\scriptsize\gray{$\pm$0.0256}} & 0.2805{\scriptsize\gray{$\pm$0.0107}} \\
 \midrule
   \multirow{1}{*}{\rotatebox[origin=c]{90}{\tiny{Prob}}} & Ours (mean) & & 0.7175 & \textbf{0.7744} & 0.3151 & & 0.6134 & \textbf{0.6941} & \textbf{0.3580} \\
  \bottomrule
\end{tabular}
\end{table}

\clearpage
\newpage

\section{Comparison to operational EFAS forecasts}
\label{sec:appendix_comparison_efas}
\subsection{Comparison to EFAS control run (deterministic forecast)}

\begin{figure}[h!]
  \centering
  \includegraphics[width=0.99\textwidth]{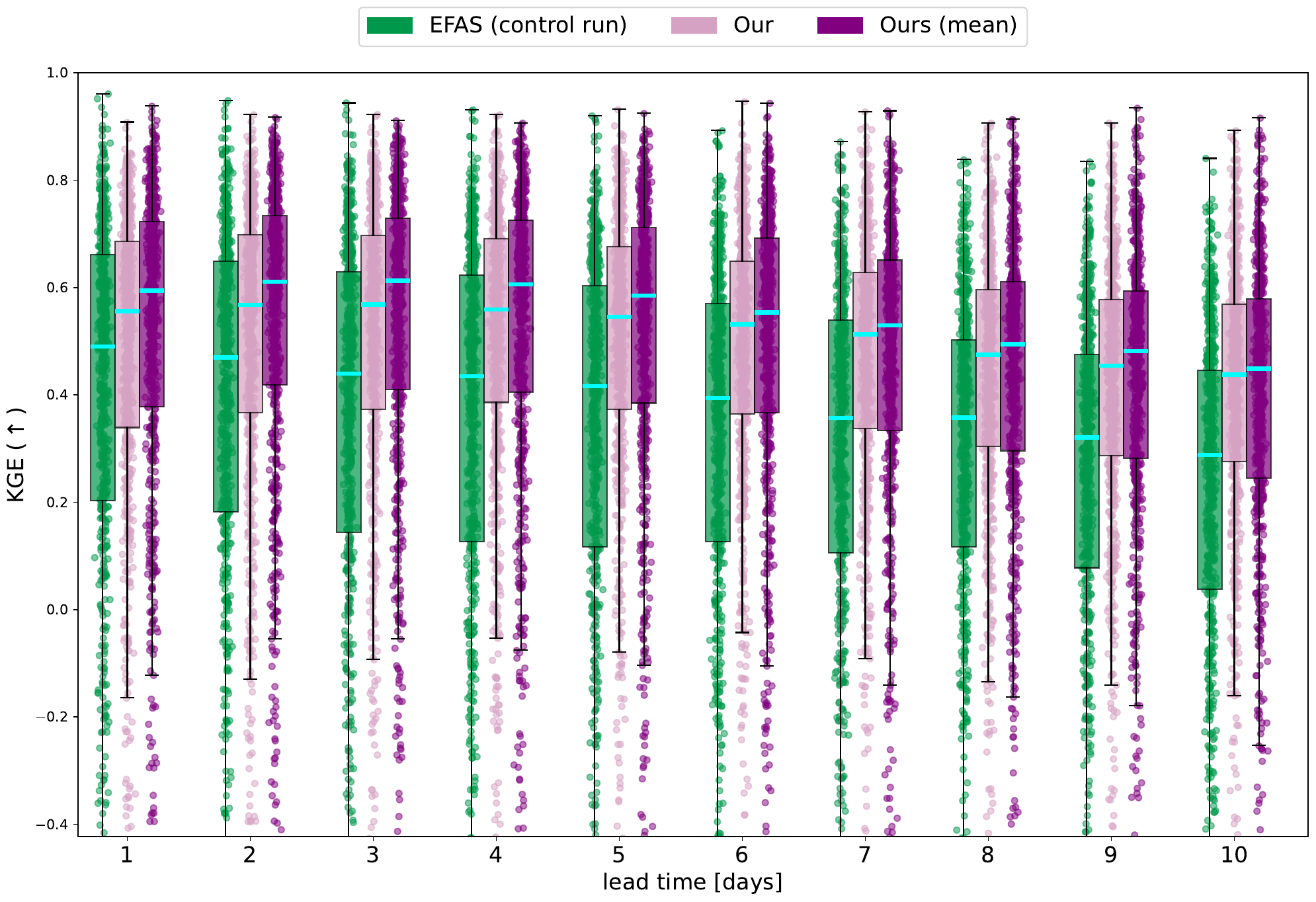}
  \caption{KGE of the river discharge forecasting with different lead time on GRDC observations (test set (2024-05-30 to 2025-12-21) temporally out-of-sample). Distribution quartiles are displayed in boxes, and the entire range excluding outliers is displayed in whiskers. The median score for the model is shown by the cyan line in the box.}
  \label{fig:appendix_efas_obs_grdc_kge}
\end{figure}

\begin{figure}[h!]
  \centering
  \includegraphics[width=0.98\textwidth]{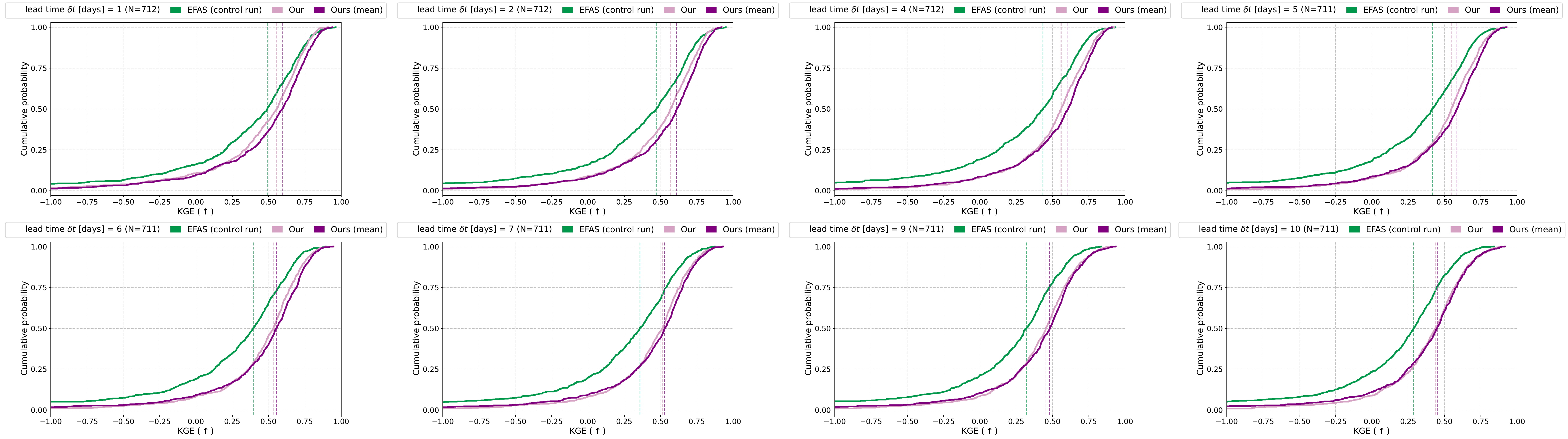}
  \caption{Cumulative distribution function (CDF) of station-level KGE on the GRDC observations, across lead times. Dashed vertical lines shows each model's media.}
  \label{fig:appendix_efs_grdc_cdf_kge}
\end{figure}

In this section, we compare to the operational EFAS forecasts (Sec.~\ref{sec:Appendix_baselines_efas_forecast}).  
We use the deterministic forecast from EFAS starting from mid May 2024 until end of 2025 and we validate on GRDC observations. Note that EFAS is a very strong baseline over the European domain. EFAS uses LISFLOOD hydrological model~\citepsupp{LISFLOOD} which has explicit kinematic wave equations for routing. It uses also river discharge from reanalysis as an initial condition while in our work we do not use any past river state as input to the model. The quantitative results are shown in Table~\ref{table:efas_deterministic} and qualitative results are shown in Fig.~\ref{fig:appendix_efas_obs_grdc_kge} and \ref{fig:appendix_efs_grdc_cdf_kge}. A comparsion to the probabilsitic forecast from EFAS is shown in Sec.~\ref{sec:appendix_efas_prob}.

\begin{table}[!h]
  \caption{Results on gauged GRDC Observations. \gray{($\pm$)} denotes the standard deviation for 3 runs.}
  \label{table:efas_deterministic}
  \centering
  \small
  \tabcolsep=3.0pt\relax
  \setlength\extrarowheight{0pt}
  \begin{tabular}{c r *{6}l}
  \toprule
  & & & \multicolumn{5}{c}{Test set (2024-05-30 to 2025-12-21)} \\
    \midrule    
    & Model & & RMSE {\scriptsize($\downarrow$)} & R2 {\scriptsize($\uparrow$)} & KGE {\scriptsize($\uparrow$)} & F1{\scriptsize@(1.5-20) ($\uparrow$)} & F1{\scriptsize@(1.5-500) ($\uparrow$)} \\
    \midrule
   \multirow{2}{*}{\rotatebox[origin=c]{90}{\tiny{Determ}}} & EFAS (control run) & & 9.8394 & 0.0481 & 0.3967 & 0.0945 & 0.0525 \\
  & Ours & & 7.2889{\scriptsize\color{gray}{$\pm$0.1628}} & 0.3413{\scriptsize\gray{$\pm$0.0135}} & 0.5180{\scriptsize\gray{$\pm$0.0073}} & 0.0703{\scriptsize\gray{$\pm$0.0159}} & 0.0512{\scriptsize\gray{$\pm$0.0175}}\\
 \midrule
   \multirow{1}{*}{\rotatebox[origin=c]{90}{\tiny{Prob}}} & Ours (mean) & & \textbf{7.2860} & \textbf{0.3673} & \textbf{0.5515} & \textbf{0.1015} & \textbf{0.1515} \\
  \bottomrule
\end{tabular}
\end{table}

\clearpage
\newpage

\subsection{Comparison to EFAS ensemble run (probabilistic forecast)}
\label{sec:appendix_efas_prob}

\begin{table}[!h]
  \caption{Results on gauged GRDC Observations.}
  \label{table:rivermamba_4}
  \centering
  \small
  \tabcolsep=3.0pt\relax
  \setlength\extrarowheight{0pt}
  \begin{tabular}{c r *{7}l}
  \toprule
  & & & \multicolumn{5}{c}{Test set (2024-06-01 to 2024-08-31)} \\
    \midrule    
    & Model & & RMSE {\scriptsize($\downarrow$)} & R2 {\scriptsize($\uparrow$)} & KGE {\scriptsize($\uparrow$)} & CRPS {\scriptsize($\downarrow$)} & CRPS$_{\scriptsize{mae}}$ {\scriptsize($\downarrow$)} & CRPS$_{\scriptsize{spread}}$ {\scriptsize($\uparrow$)} \\
    \midrule
   \multirow{2}{*}{\rotatebox[origin=c]{90}{\tiny{Prob}}} & EFAS (mean of 51 ens members) & & 6.8122 & -0.8517 & 0.1667 & 3.8862 & 5.4043 & 1.5180 \\
   & Ours (mean of 12 ens members) & & \textbf{4.5627} & \textbf{0.0519} & \textbf{0.3718} & \textbf{2.1599} & \textbf{4.1280} & \textbf{1.9682} \\
  \bottomrule
\end{tabular}
\end{table}

\begin{figure}[h!]
  \centering
  \includegraphics[width=0.99\textwidth]{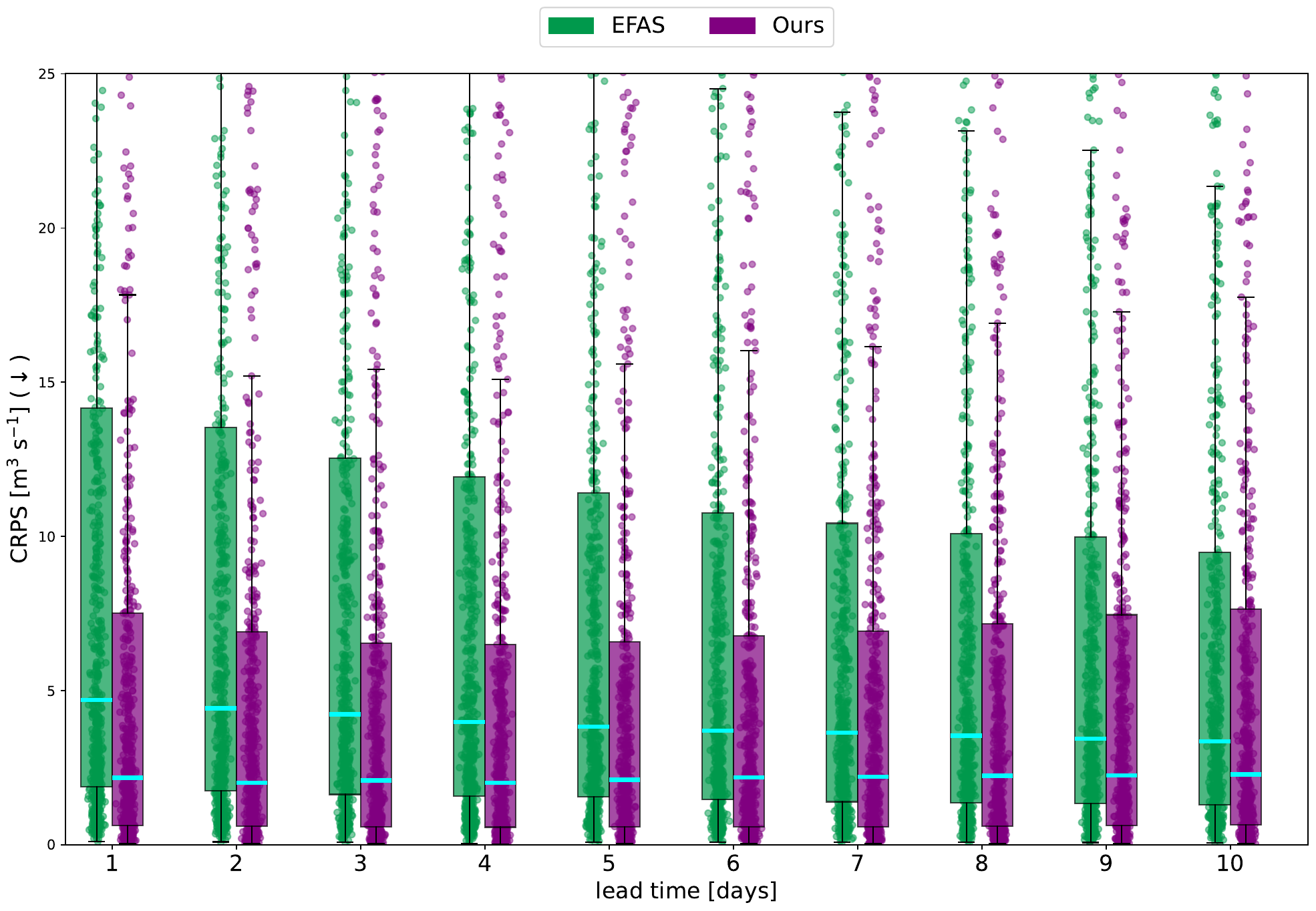}
  \caption{CRPS of the river discharge forecasting with different lead time on GRDC observations (test set (2024-06-01 to 2024-08-31) temporally out-of-sample). Distribution quartiles are displayed in boxes, and the entire range excluding outliers is displayed in whiskers. The median score for the model is shown by the cyan line in the box.}
  \label{fig:appendix_efas_obs_grdc_crps}
\end{figure}

\begin{figure}[h!]
  \centering
  \includegraphics[width=0.99\textwidth]{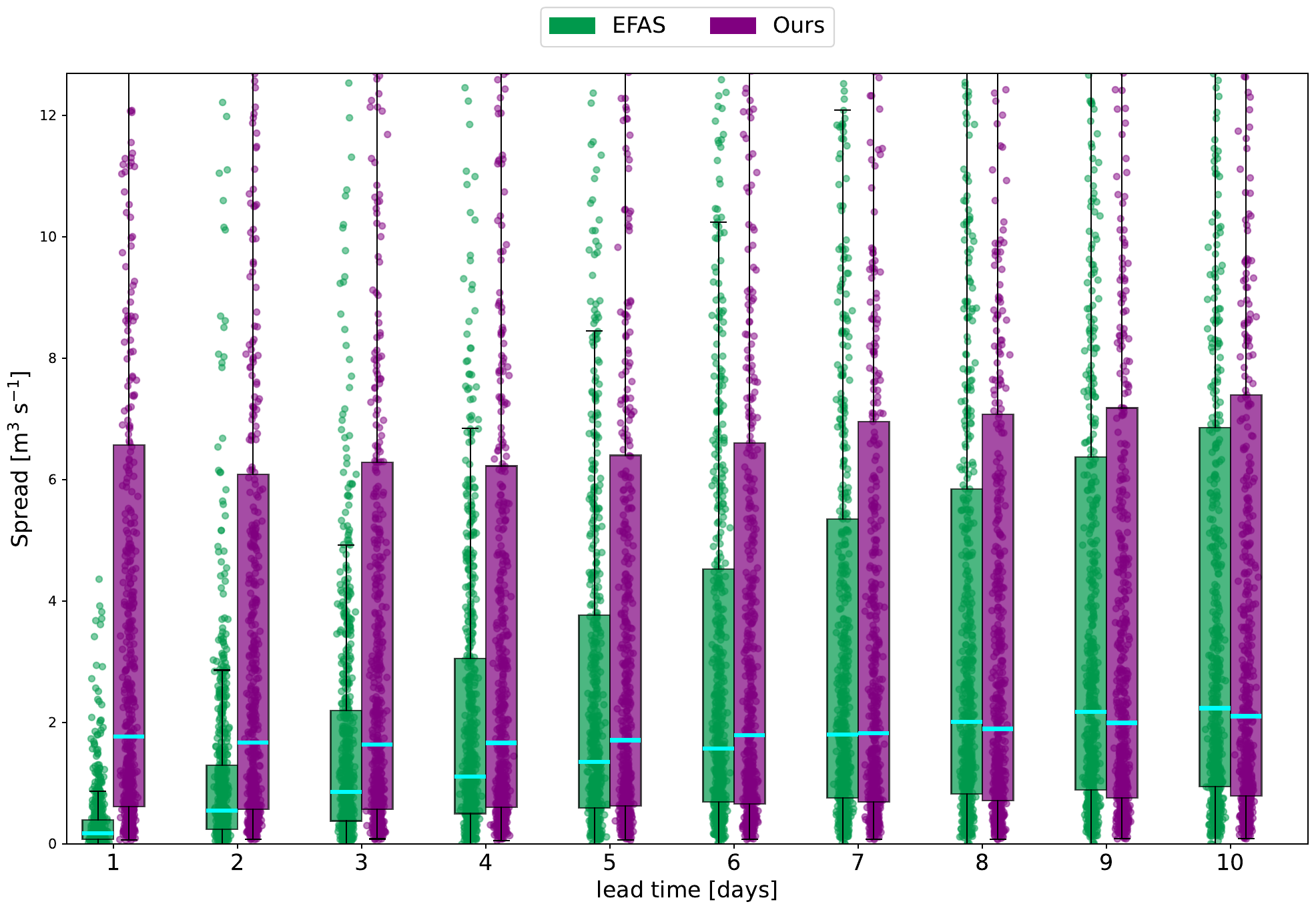}
  \caption{Ensemble spread of the river discharge forecasting with different lead time on GRDC observations (test set (2024-06-01 to 2024-08-31) temporally out-of-sample). Distribution quartiles are displayed in boxes, and the entire range excluding outliers is displayed in whiskers. The median score for the model is shown by the cyan line in the box.}
  \label{fig:appendix_efas_obs_grdc_spread}
\end{figure}

\begin{figure}[h!]
  \centering
  \includegraphics[width=0.98\textwidth]{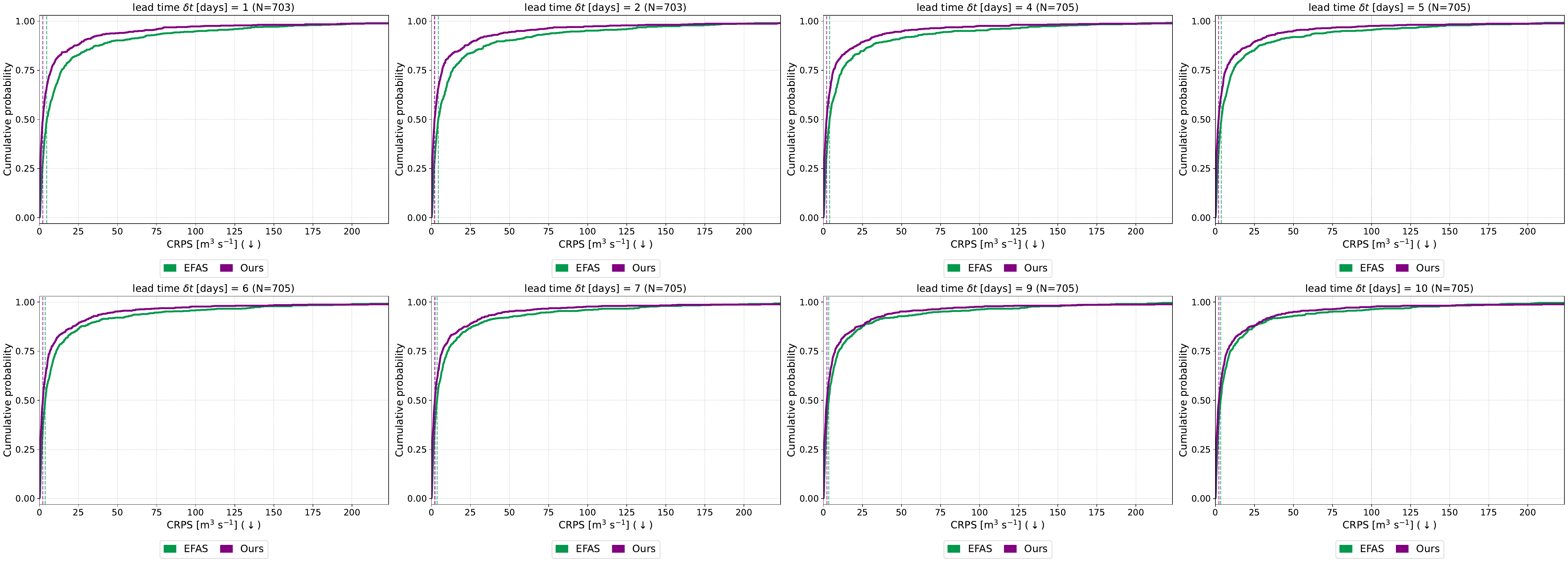}
  \caption{Cumulative distribution function (CDF) of station-level CRPS on the GRDC observations, across lead times. Dashed vertical lines shows each model's media.}
  \label{fig:appendix_efs_grdc_cdf_crps}
\end{figure}

\clearpage
\newpage


\section{Hydrographs}

\subsection{Example hydrographs from training on EFAS Reanalysis data}

\begin{figure}[h!]
  \centering
  \includegraphics[width=0.99\textwidth]{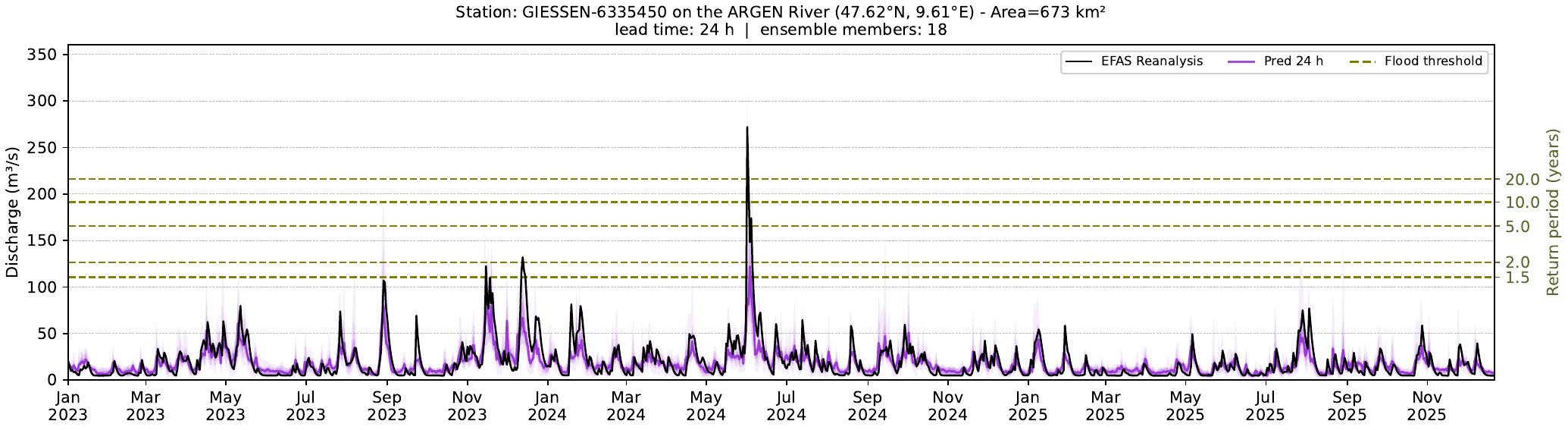}
  \caption{Ensemble forecast against EFAS reanalysis at Giessen (GRDC 6335450) on the Argen River ($47.62^\circ$N, $9.61^\circ$E, catchment area $673$ km$^2$), for lead time $24$\,h. For forecast, the ensemble median (solid line) is shown with nested uncertainty bands spanning the $25$th–$75$th percentile (darkest), $10$th–$90$th percentile, and min–max range (lightest) across all $18$ ensemble members, against the EFAS reanalysis (black). Dashed olive lines are the EFAS flood return period thresholds ($1.5$-$20.0$ years).} 
  \label{fig:appendix_hydrograph_reanalysis_1}
\end{figure}

\begin{figure}[h!]
  \centering
  \includegraphics[width=0.99\textwidth]{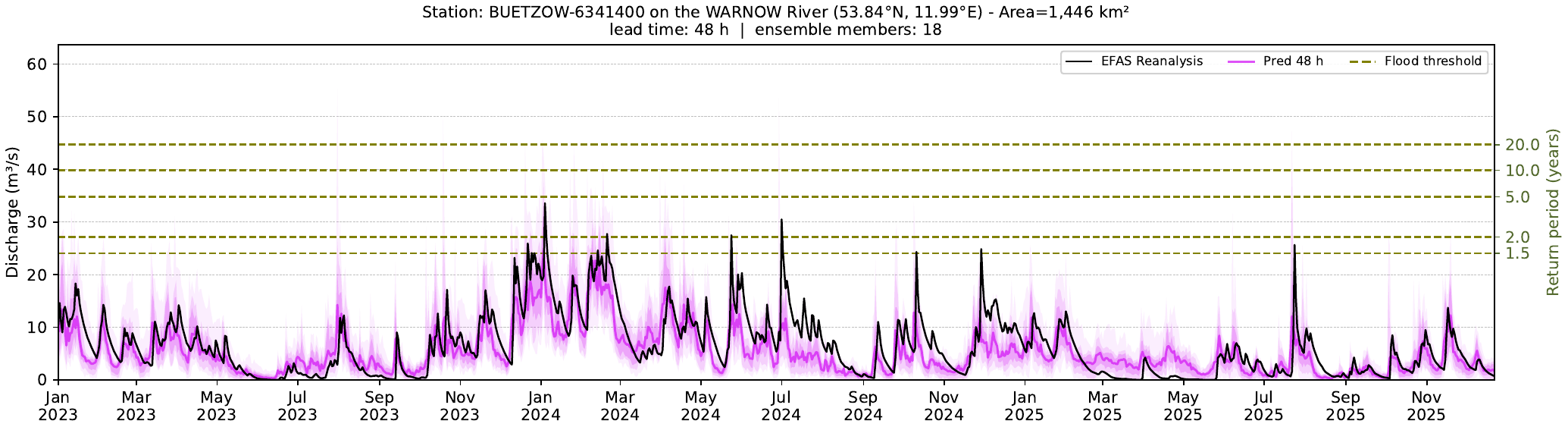}
  \caption{Ensemble forecast against EFAS reanalysis at Buetzow (GRDC 6341400) on the Warnow River ($53.84^\circ$N, $11.99^\circ$E, catchment area $1,446$ km$^2$), for lead time $48$\,h. For forecast, the ensemble median (solid line) is shown with nested uncertainty bands spanning the $25$th–$75$th percentile (darkest), $10$th–$90$th percentile, and min–max range (lightest) across all $18$ ensemble members, against the EFAS reanalysis (black). Dashed olive lines are the EFAS flood return period thresholds ($1.5$-$20.0$ years).}
  \label{fig:appendix_hydrograph_reanalysis_2}
\end{figure}

\begin{figure}[h!]
  \centering
  \includegraphics[width=0.99\textwidth]{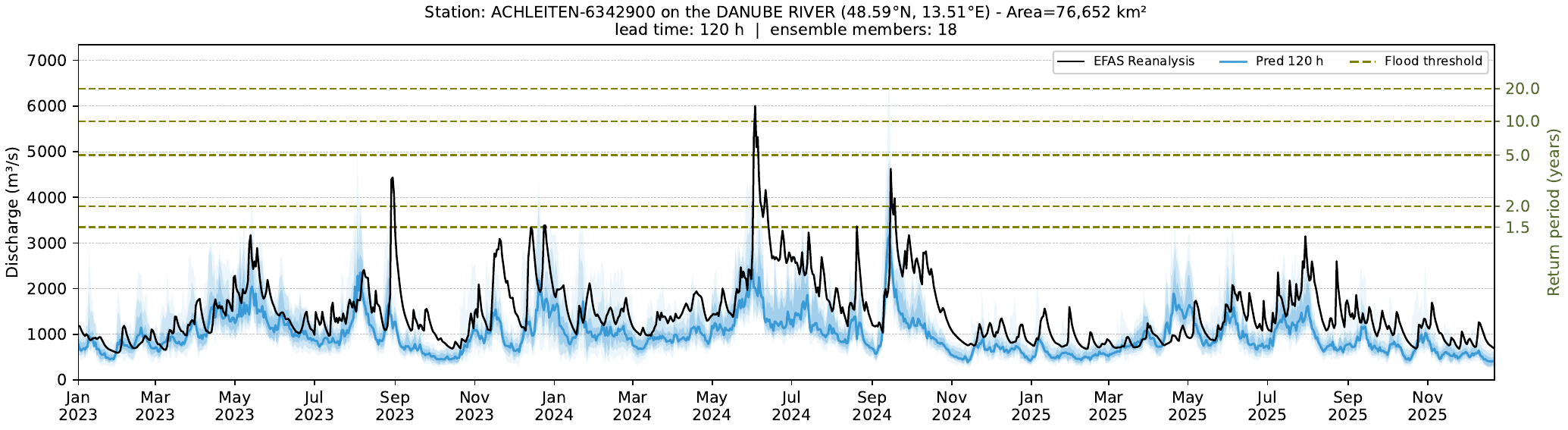}
  \caption{Ensemble forecast against EFAS reanalysis at Achleiten (GRDC 6342900) on the Danube River ($48.59^\circ$N, $13.51^\circ$E, catchment area $76,652$ km$^2$), for lead time $120$\,h. For forecast, the ensemble median (solid line) is shown with nested uncertainty bands spanning the $25$th–$75$th percentile (darkest), $10$th–$90$th percentile, and min–max range (lightest) across all $18$ ensemble members, against the EFAS reanalysis (black). Dashed olive lines are the EFAS flood return period thresholds ($1.5$-$20.0$ years).} 
  \label{fig:appendix_hydrograph_reanalysis_4}
\end{figure}

\clearpage
\newpage

\begin{figure}[h!]
  \centering
  \includegraphics[width=0.99\textwidth]{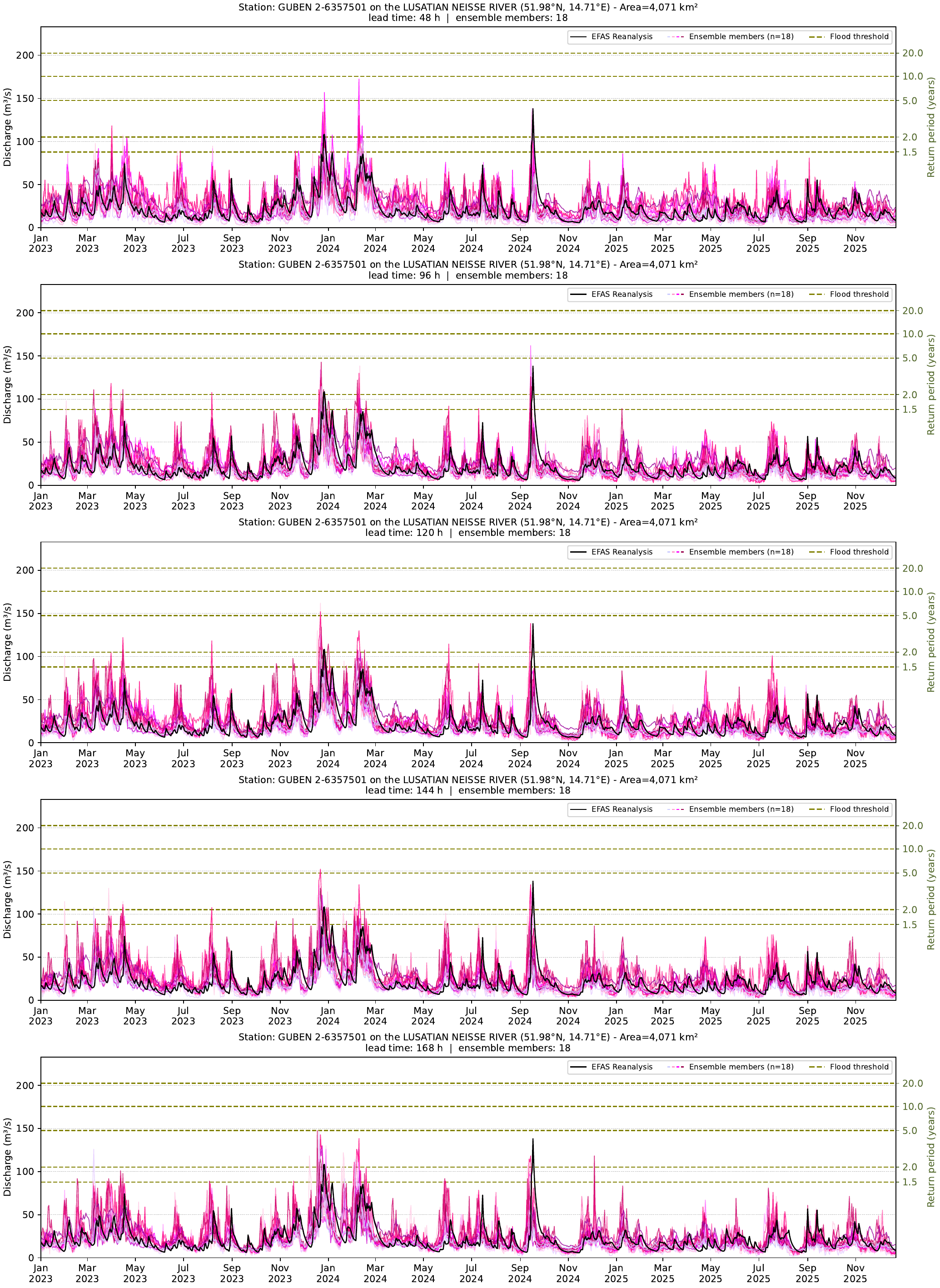}
  \caption{Ensemble forecasts against EFAS reanalysis at Guben (GRDC 6357501) on the Lech River ($51.98^\circ$N, $14.71^\circ$E, catchment area $1,184$ km$^2$), for five lead times ($48$ h, $96$ h, $120$ h, $144$ h, $168$). Each row shows the EFAS reanalysis discharge (black) against individual ensemble members (colored lines) for that lead time. Dashed olive lines are the EFAS flood return period thresholds ($1.5$-$20.0$ years).}
  \label{fig:appendix_hydrograph_reanalysis_8}
\end{figure}

\clearpage
\newpage

\subsection{Example hydrographs from fine-tuning on CAMELS observational data}

\begin{figure}[h!]
  \centering
  \includegraphics[width=0.99\textwidth]{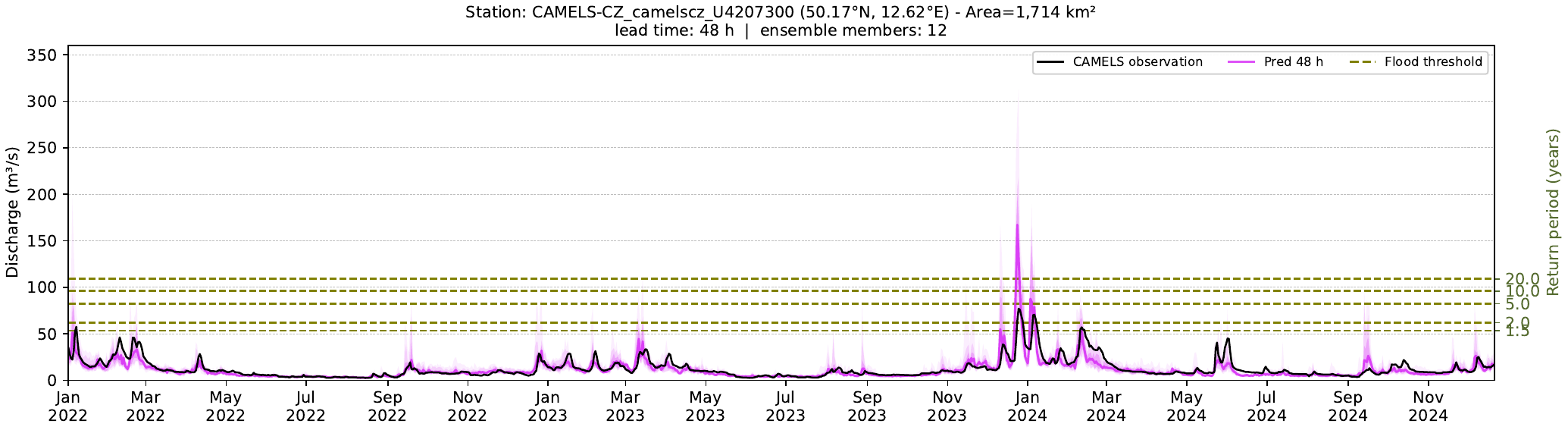}
  \caption{Ensemble forecast against CAMELS-CZ in Czechia for the station U420730 on the Oh\v{r}e River ($50.17^\circ$N, $12.62^\circ$E, catchment area $1,714$ km$^2$), for lead time $48$\,h. For forecast, the ensemble median (solid line) is shown with nested uncertainty bands spanning the $25$th–$75$th percentile (darkest), $10$th–$90$th percentile, and min–max range (lightest) across all $12$ ensemble members, against the CAMELS-CZ observations (black). Dashed olive lines are the flood return period thresholds ($1.5$-$20.0$ years) computed from observations.} 
  \label{fig:appendix_hydrograph_obs_camels_1}
\end{figure}

\begin{figure}[h!]
  \centering
  \includegraphics[width=0.99\textwidth]{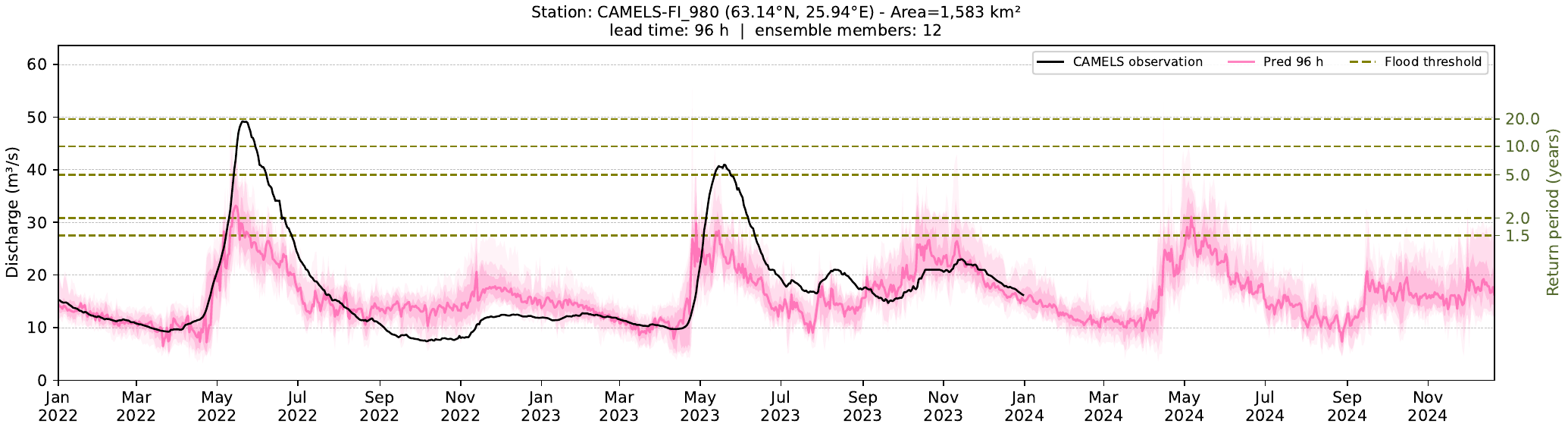}
  \caption{Ensemble forecast against CAMELS-FI in Finland for the station K\"arn\"aj\"arvi-Kellankoski (FI\_980) at the lake Kymijoki ($63.14^\circ$N, $25.94^\circ$E, catchment area $1,583$ km$^2$), for lead time $96$\,h. For forecast, the ensemble median (solid line) is shown with nested uncertainty bands spanning the $25$th–$75$th percentile (darkest), $10$th–$90$th percentile, and min–max range (lightest) across all $12$ ensemble members, against the CAMELS-FI observations (black). Dashed olive lines are the flood return period thresholds ($1.5$-$20.0$ years) computed from observations.} 
  \label{fig:appendix_hydrograph_obs_camels_2}
\end{figure}

\clearpage
\newpage

\begin{figure}[h!]
  \centering
  \includegraphics[width=0.99\textwidth]{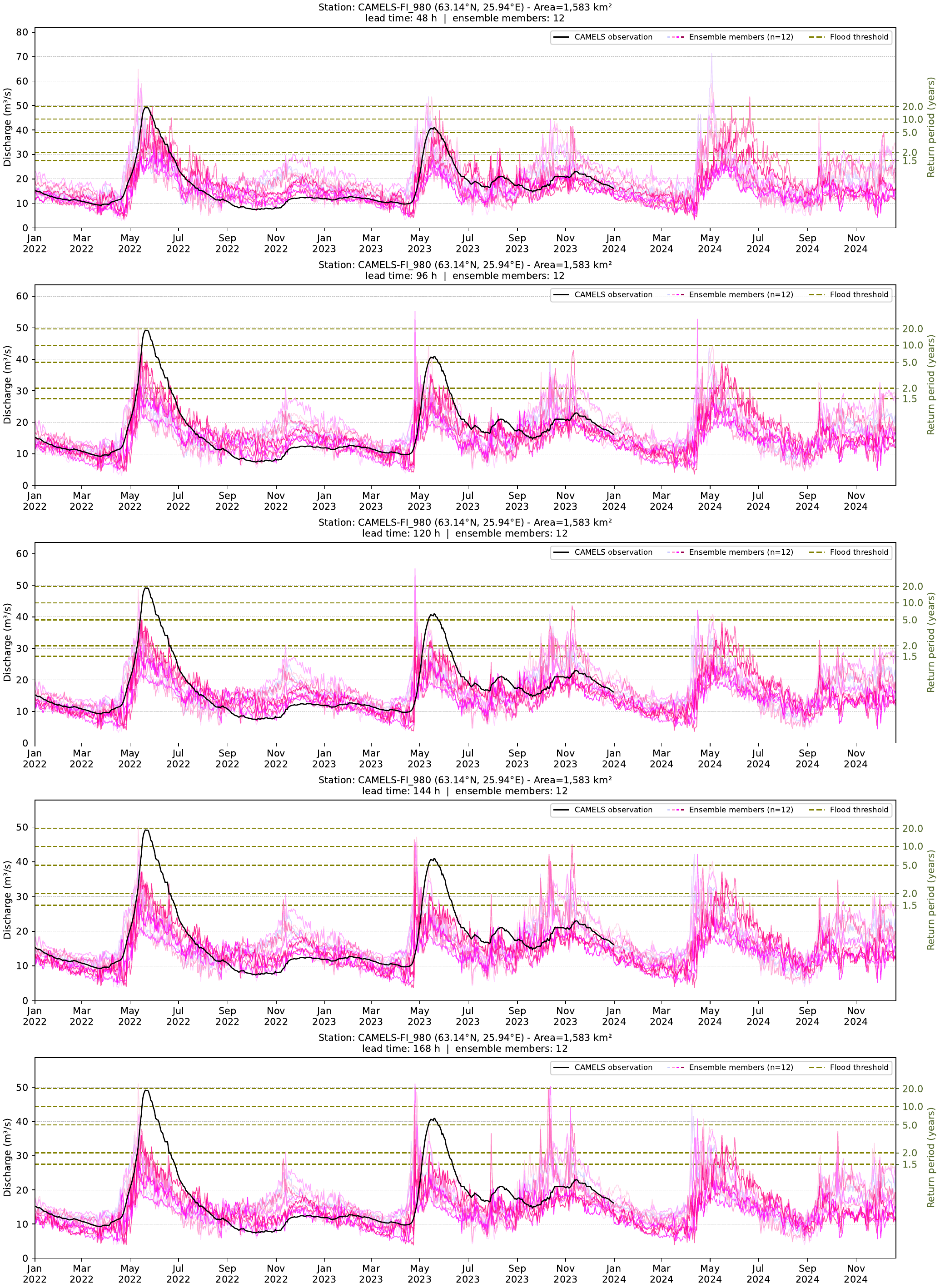}
  \caption{Ensemble forecasts against CAMELS-FI in Finland for the station K\"arn\"aj\"arvi-Kellankoski (FI\_980) at the lake Kymijoki ($63.14^\circ$N, $25.94^\circ$E, catchment area $1,583$ km$^2$), for five lead times ($48$ h, $96$ h, $120$ h, $144$ h, $168$). Each row shows the CAMELS-CZ observational discharge (black) against individual ensemble members (colored lines) for that lead time. Dashed olive lines are the flood return period thresholds ($1.5$-$20.0$ years) computed from observations.}
  \label{fig:appendix_hydrograph_obs_camels_4}
\end{figure}

\clearpage
\newpage

\subsection{Example hydrographs from fine-tuning on GRDC observational data}

\begin{figure}[h!]
  \centering
  \includegraphics[width=0.7\textwidth]{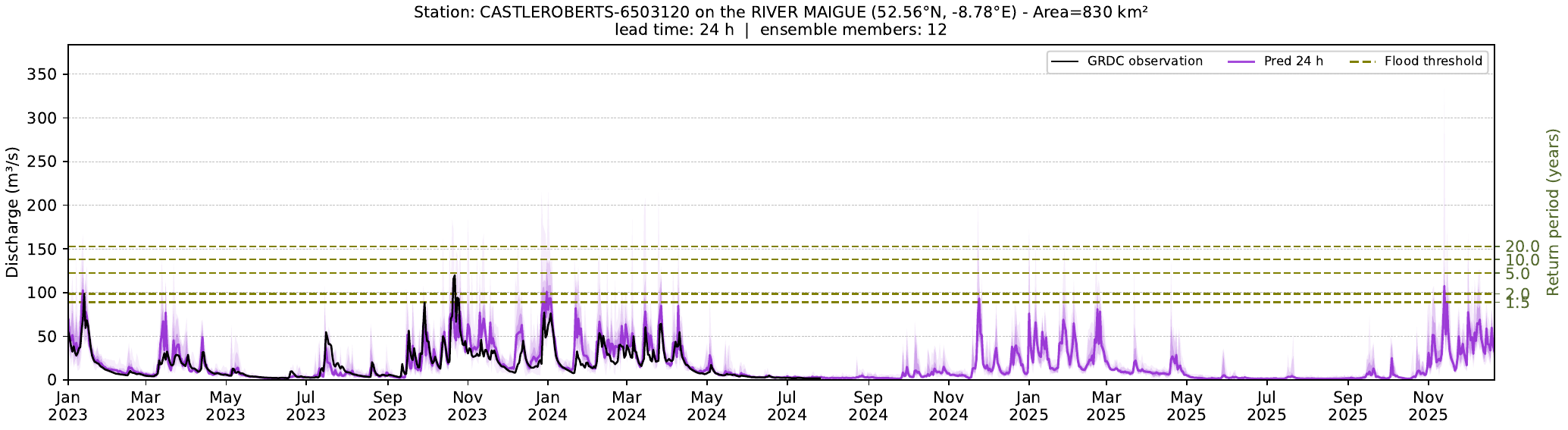}
  \caption{Ensemble forecast against GRDC observations at Castle Roberts (GRDC 6503120) on the Maigue River ($52.56^\circ$N, $8.78^\circ$W, catchment area $830$ km$^2$), for lead time $24$\,h. For forecast, the ensemble median (solid line) is shown with nested uncertainty bands spanning the $25$th–$75$th percentile (darkest), $10$th–$90$th percentile, and min–max range (lightest) across all $18$ ensemble members, against the EFAS reanalysis (black). Dashed olive lines are the flood return period thresholds ($1.5$-$20.0$ years) computed from observations.} 
  \label{fig:appendix_hydrograph_obs_grdc_1}
\end{figure}

\begin{figure}[h!]
  \centering
  \includegraphics[width=0.7\textwidth]{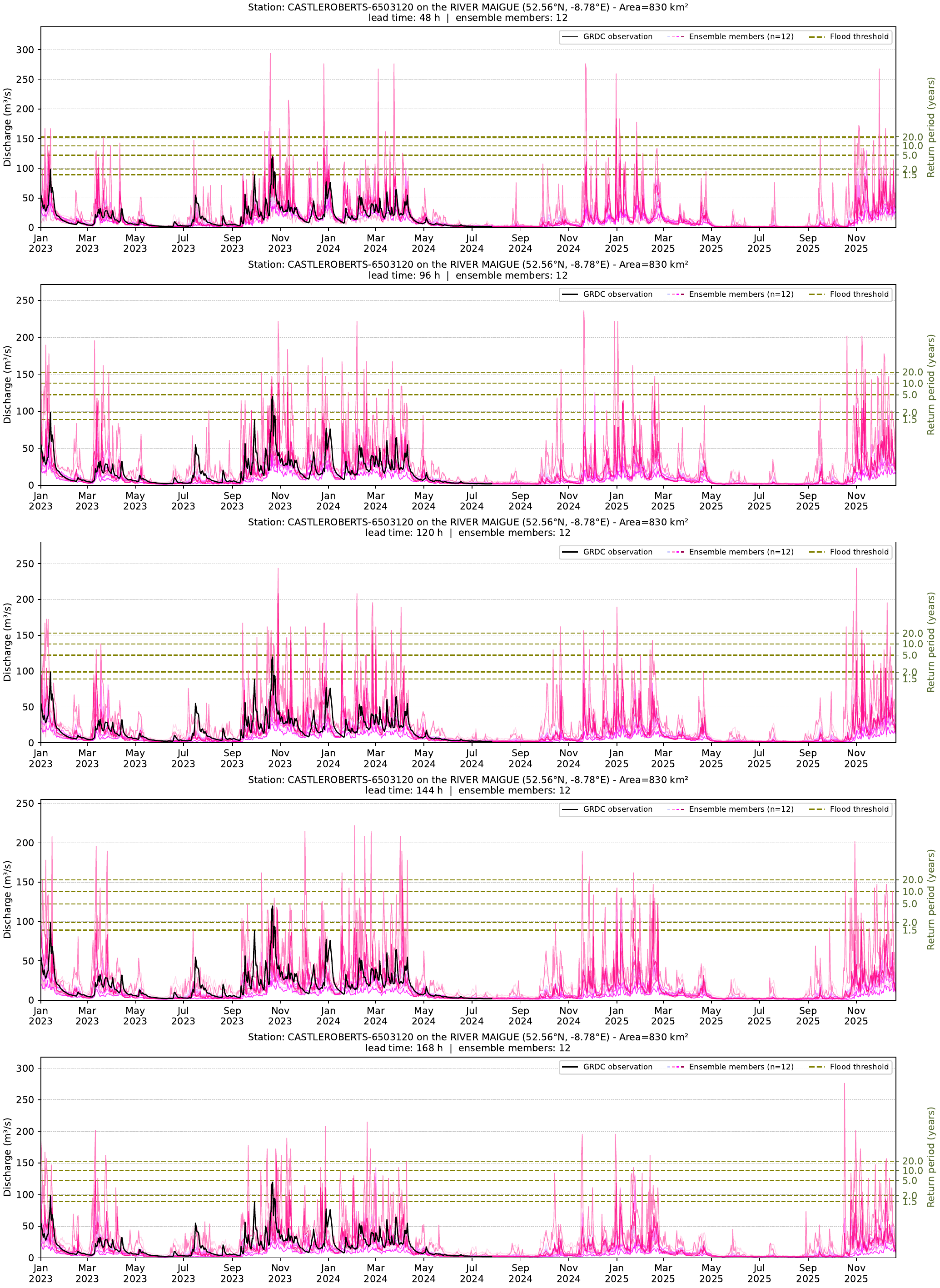}
  \caption{Ensemble forecasts against GRDC observations at Castle Roberts (GRDC 6503120) on the Maigue River ($52.56^\circ$N, $8.78^\circ$W, catchment area $830$ km$^2$), for five lead times ($48$ h, $96$ h, $120$ h, $144$ h, $168$). Each row shows the GRDC observational discharge (black) against individual ensemble members (colored lines) for that lead time. Dashed olive lines are the flood return period thresholds ($1.5$-$20.0$ years) computed from observations.}
  \label{fig:appendix_hydrograph_obs_grdc_2}
\end{figure}

\clearpage
\newpage

\section{Case Studies of Floods Across Europe}

\subsection{2024 Central European floods}

\begin{figure}[h!]
  \centering
  \includegraphics[width=0.99\textwidth]{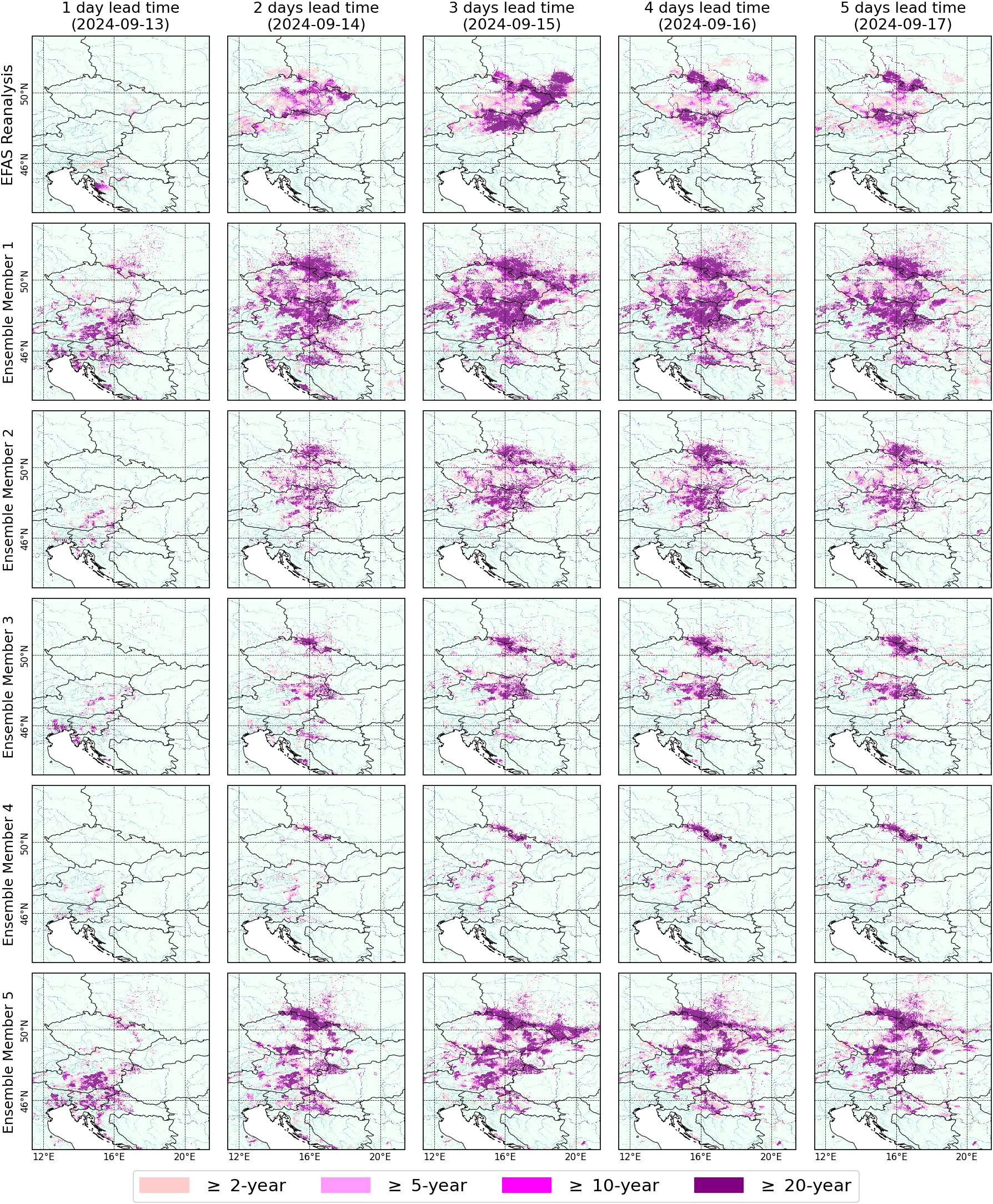}
  \caption{Comparison between EFAS reanalysis (first row, used as a reference) and our ensemble forecast (second to sixth rows) during the Central European flood, September 2024. Shown are flood severity maps at $1$-arcminute resolution where each panel shows flood extent at different lead times from $1$ to $5$ days.}
  \label{fig:appendix_2024_central_europe_flood}
\end{figure}

\begin{figure}[h!]
  \centering
  \includegraphics[width=0.99\textwidth]{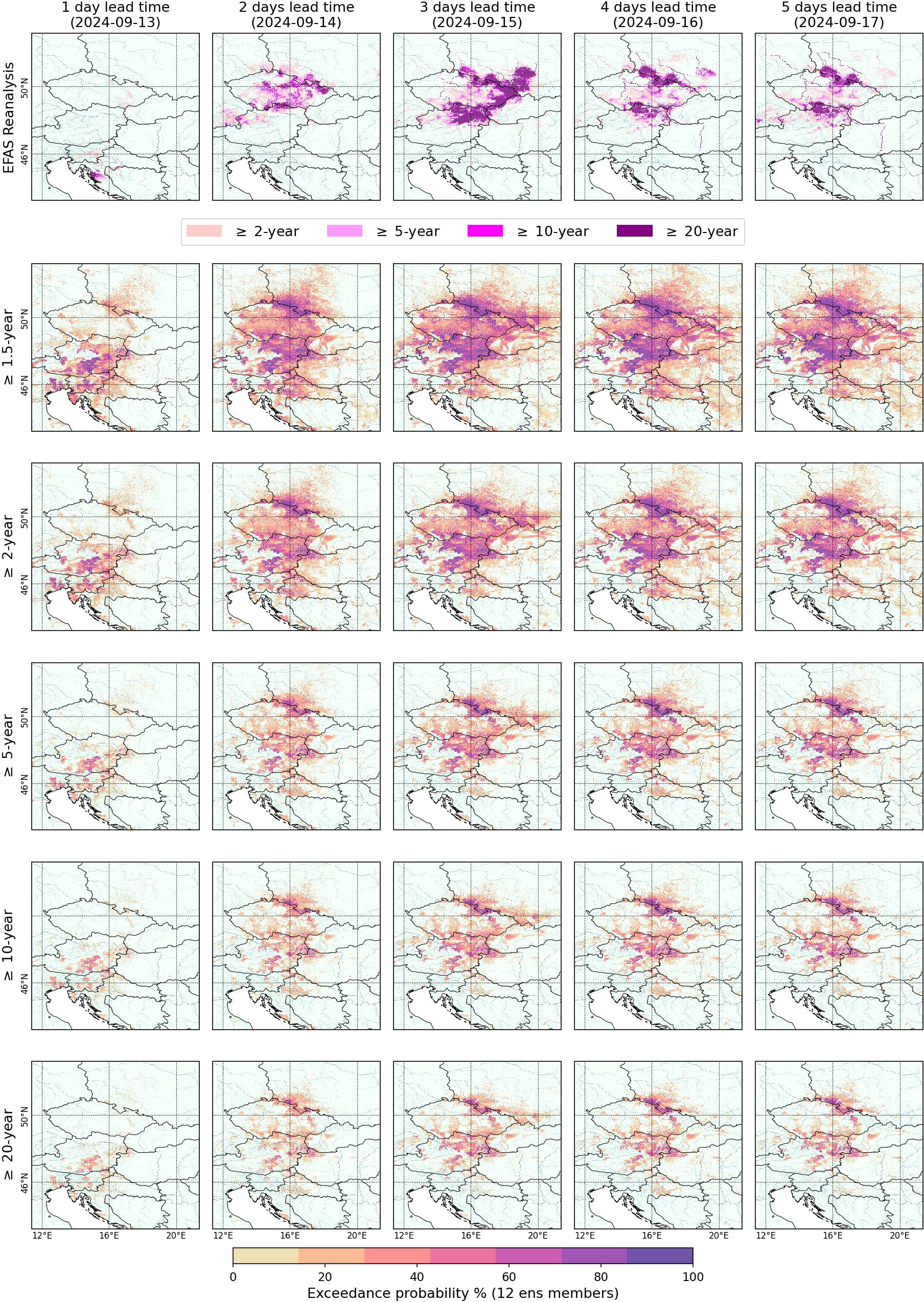}
  \caption{Comparison between EFAS reanalysis (first row, used as a reference) and our ensemble forecast (second to sixth rows) during the Central European flood, September 2024.
  First row shows flood severity maps at $1$-arcminute resolution where each panel shows flood extent at different lead times from $1$ to $5$ days.
  Rows $2$ to $6$ show the exceedance probabilities for flood return periods ($1.5$-$20.0$) computed from $12$ ensemble members.
  } 
  \label{fig:appendix_2024_central_europe_flood_prob}
\end{figure}

\clearpage
\newpage

\subsection{2024 Spanish floods}

\begin{figure}[h!]
  \centering
  \includegraphics[width=0.99\textwidth]{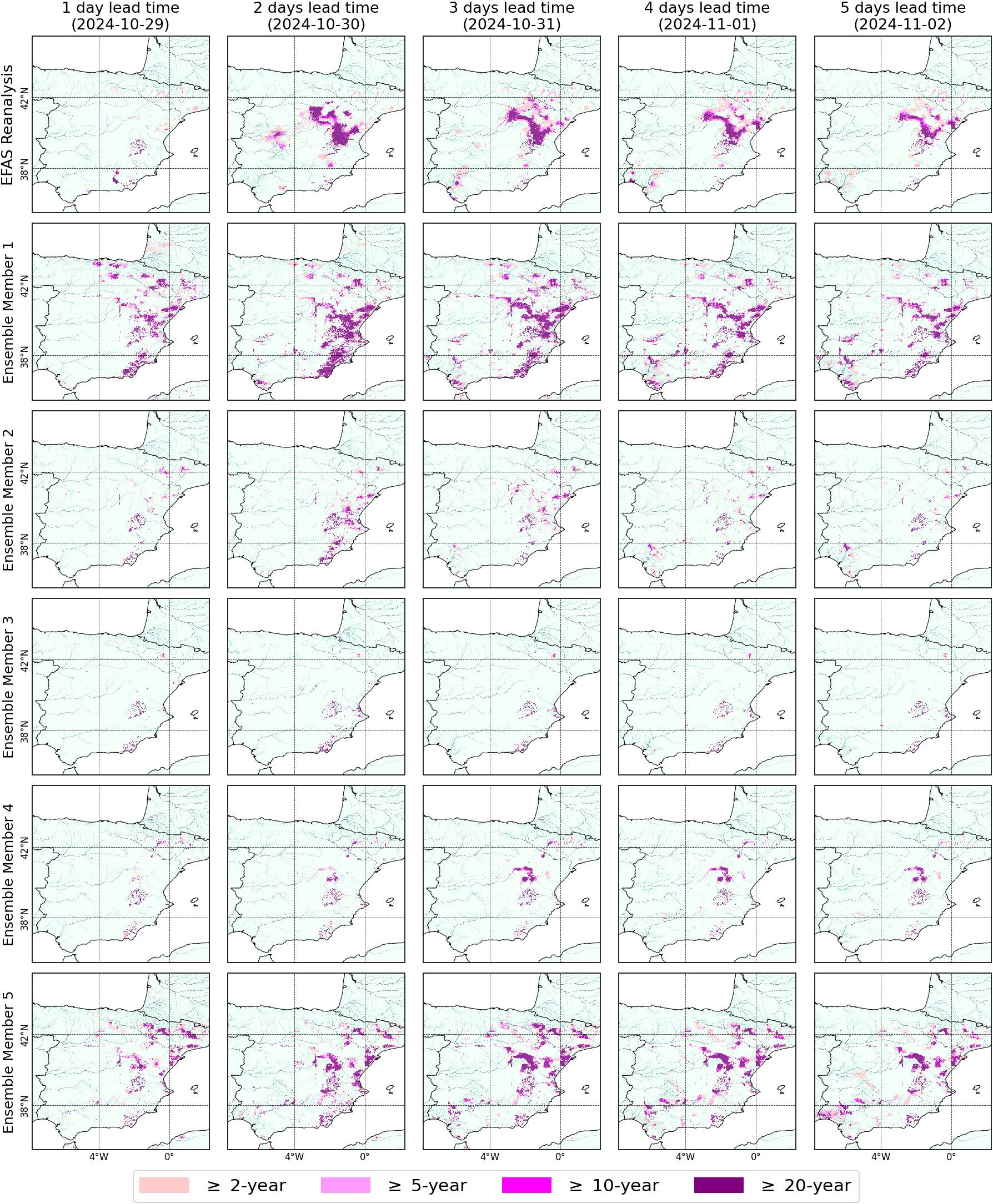}
  \caption{Comparison between EFAS reanalysis (first row, used as a reference) and our ensemble forecast (second to sixth rows) during the Spanish flood, October 2024. Shown are flood severity maps at $1$-arcminute resolution where each panel shows flood extent at different lead times from $1$ to $5$ days.}
  \label{fig:appendix_2024_spanish_flood}
\end{figure}

\begin{figure}[h!]
  \centering
  \includegraphics[width=0.99\textwidth]{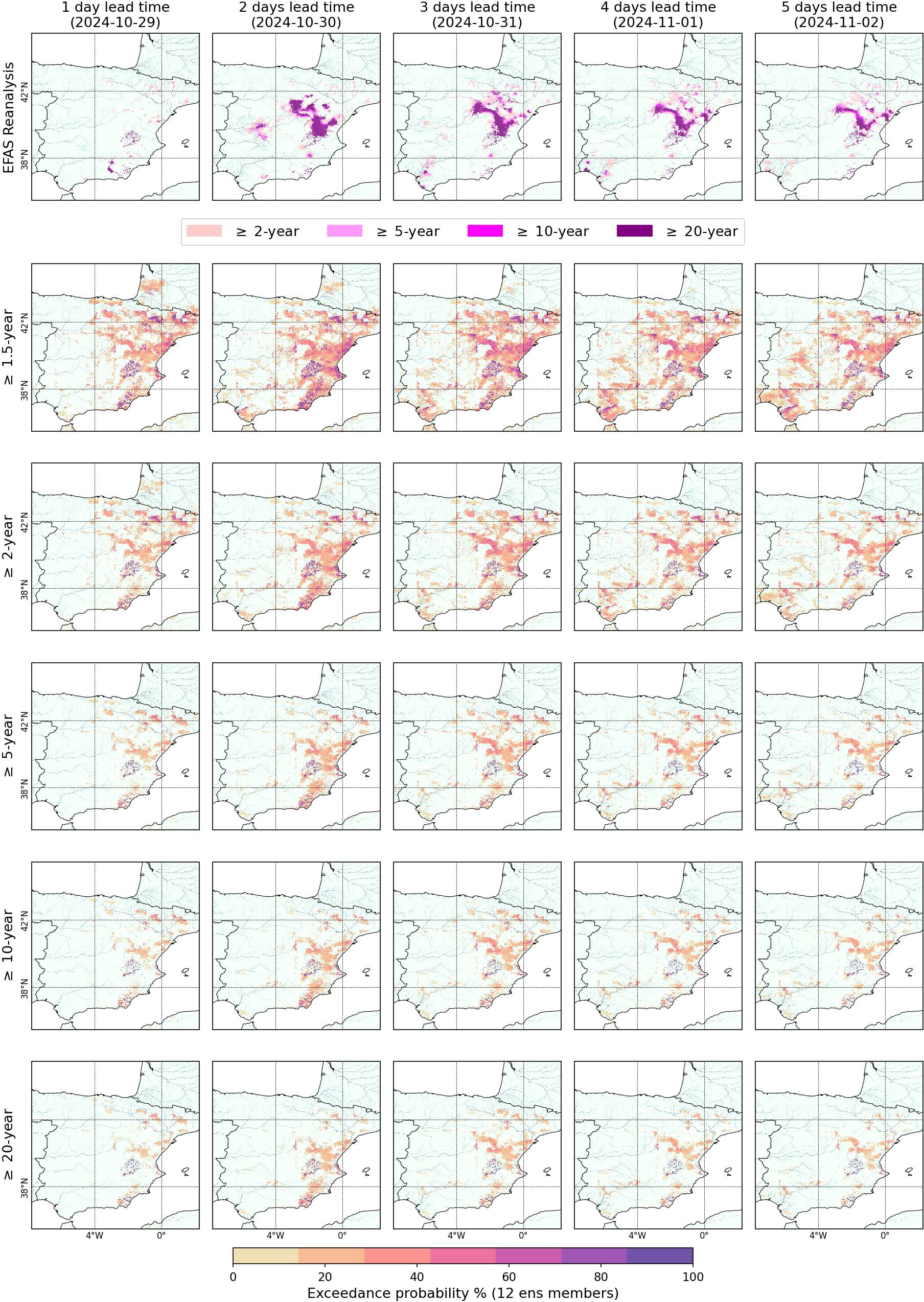}
  \caption{Comparison between EFAS reanalysis (first row, used as a reference) and our ensemble forecast (second to sixth rows) during the Spanish flood, October 2024.
  First row shows flood severity maps at $1$-arcminute resolution where each panel shows flood extent at different lead times from $1$ to $5$ days.
  Rows $2$ to $6$ show the exceedance probabilities for flood return periods ($1.5$-$20.0$) computed from $12$ ensemble members.
  } 
  \label{fig:appendix_2024_spanish_flood_prob}
\end{figure}

\clearpage
\newpage

\subsection{2025 United Kingdom Floods}

\begin{figure}[h!]
  \centering
  \includegraphics[width=0.99\textwidth]{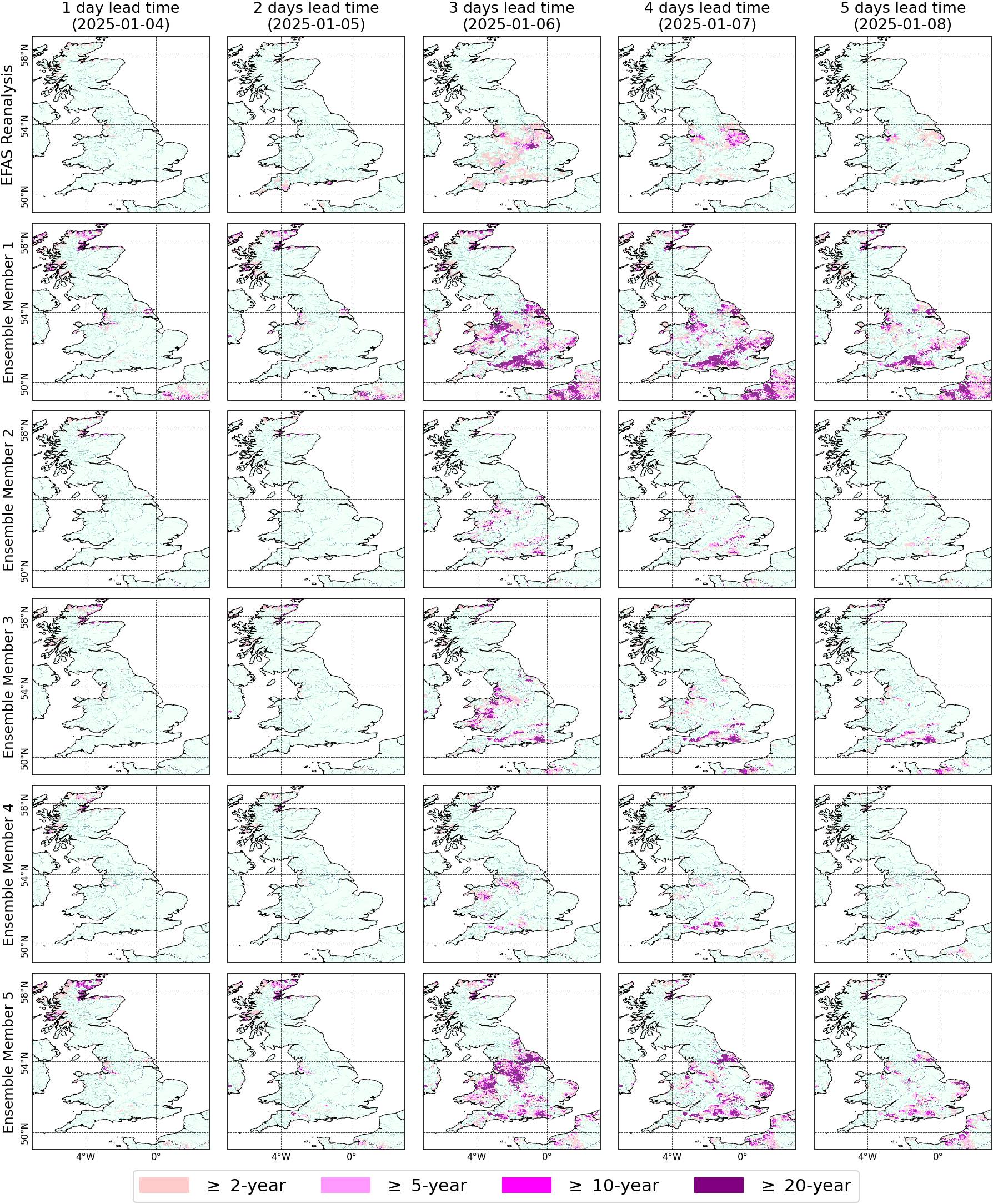}
  \caption{Comparison between EFAS reanalysis (first row, used as a reference) and our ensemble forecast (second to sixth rows) during the United Kingdom flood, January 2025. Shown are flood severity maps at $1$-arcminute resolution where each panel shows flood extent at different lead times from $1$ to $5$ days.}
  \label{fig:appendix_2025_uk_flood}
\end{figure}

\begin{figure}[h!]
  \centering
  \includegraphics[width=0.99\textwidth]{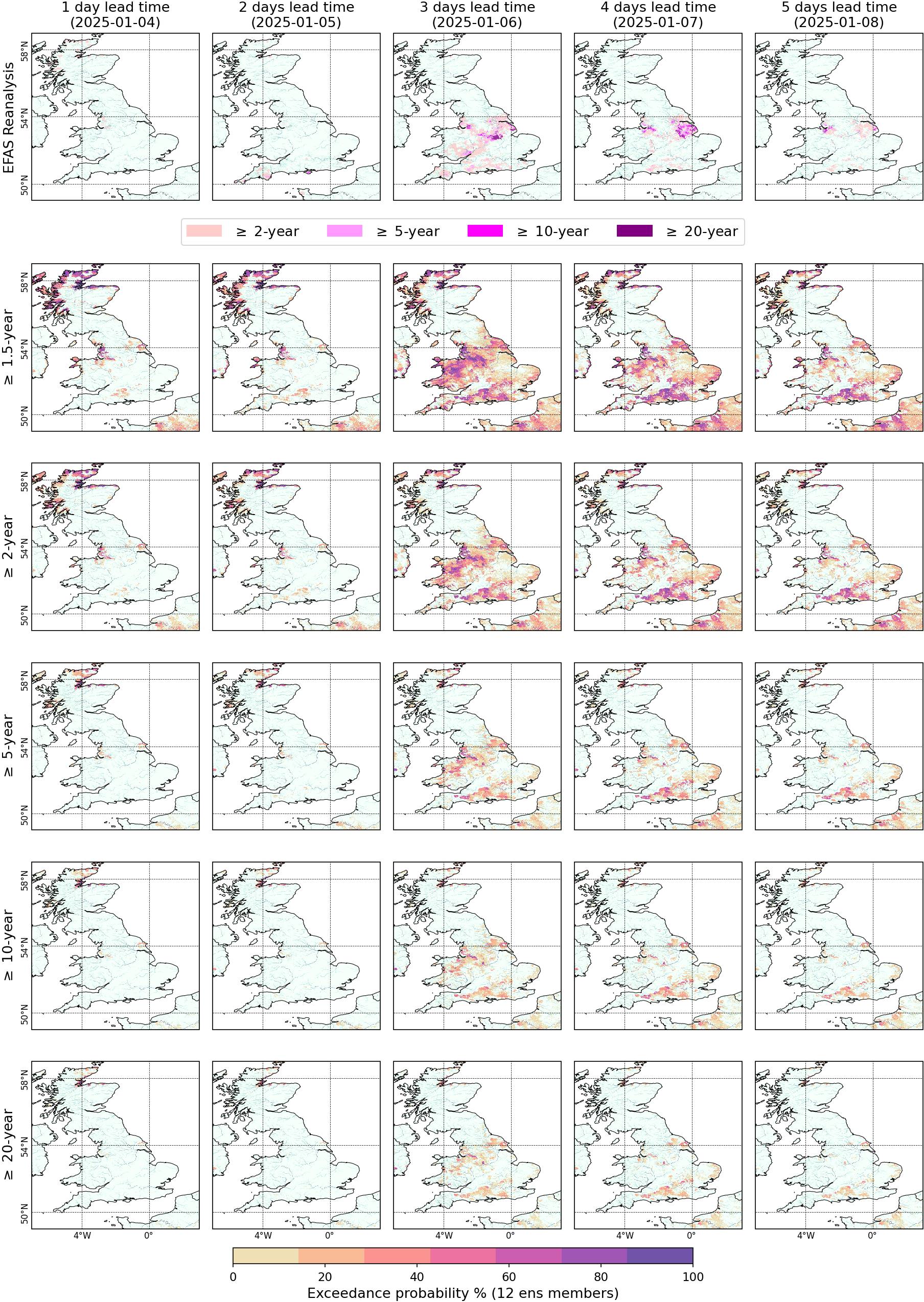}
  \caption{Comparison between EFAS reanalysis (first row, used as a reference) and our ensemble forecast (second to sixth rows) during the United Kingdom flood, January 2025.
  First row shows flood severity maps at $1$-arcminute resolution where each panel shows flood extent at different lead times from $1$ to $5$ days.
  Rows $2$ to $6$ show the exceedance probabilities for flood return periods ($1.5$-$20.0$) computed from $12$ ensemble members.
  } 
  \label{fig:appendix_2025_uk_flood_prob}
\end{figure}

\clearpage
\newpage

\subsection{2025 Balkan Rivers Floods}

\begin{figure}[h!]
  \centering
  \includegraphics[width=0.99\textwidth]{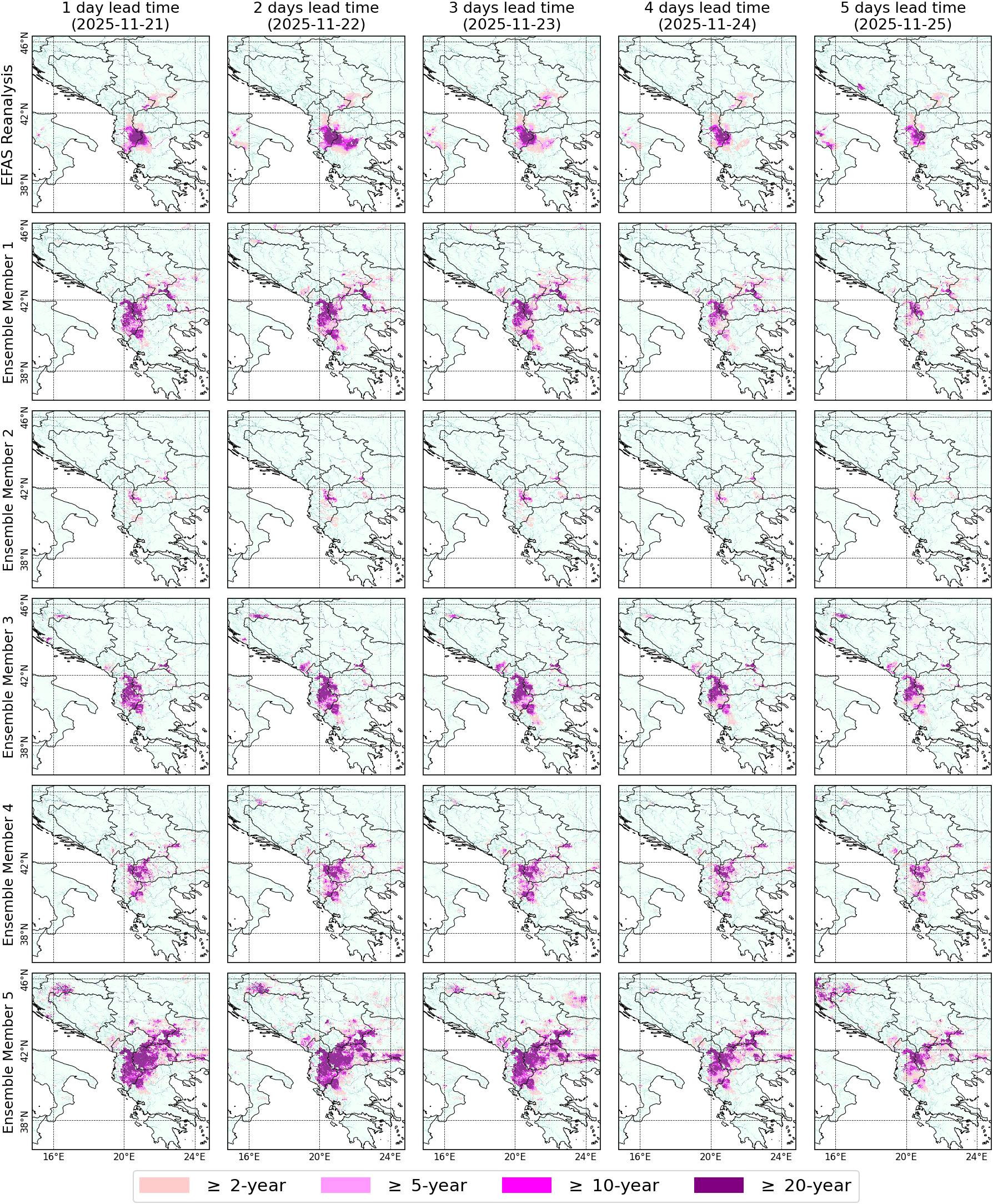}
  \caption{Comparison between EFAS reanalysis (first row, used as a reference) and our ensemble forecast (second to sixth rows) during the Balkan Rivers flood, November 2025. Shown are flood severity maps at $1$-arcminute resolution where each panel shows flood extent at different lead times from $1$ to $5$ days.}
  \label{fig:appendix_2025_balkan_flood}
\end{figure}

\begin{figure}[h!]
  \centering
  \includegraphics[width=0.99\textwidth]{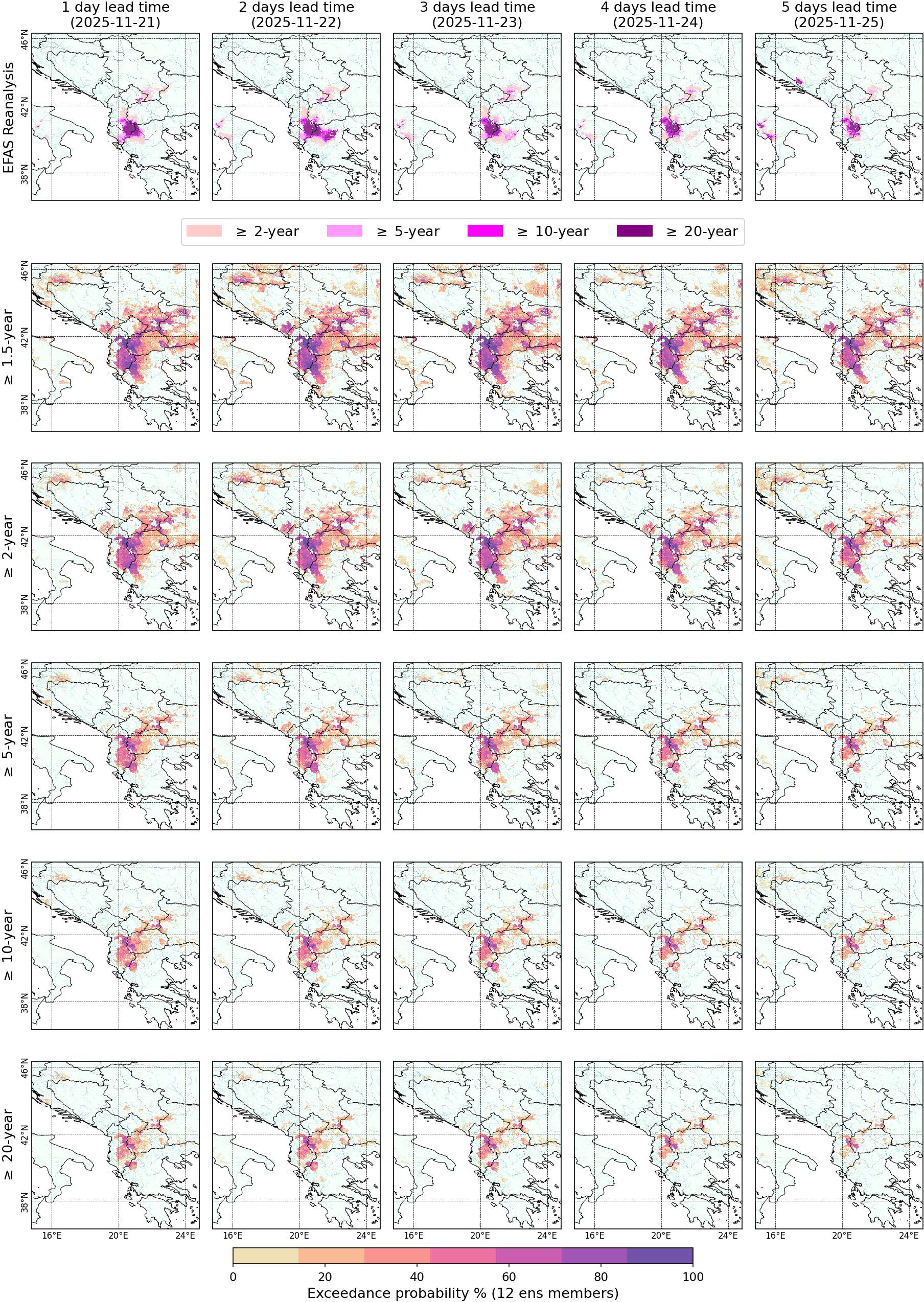}
  \caption{Comparison between EFAS reanalysis (first row, used as a reference) and our ensemble forecast (second to sixth rows) during the Balkan Rivers flood, November 2025.
  First row shows flood severity maps at $1$-arcminute resolution where each panel shows flood extent at different lead times from $1$ to $5$ days.
  Rows $2$ to $6$ show the exceedance probabilities for flood return periods ($1.5$-$20.0$) computed from $12$ ensemble members.
  } 
  \label{fig:appendix_2025_balkan_flood_prob}
\end{figure}

\clearpage
\newpage

\section{Non-linear reservoir}
\label{sec:reservoir}

In this experiment, we test whether the learned routing features behave as a non-linear reservoir i.e.\ an internal state that is carried forward and updated over time rather than a static representation per location.
For this we inspect the internal state of the routing module in the forecasting block for random dates from the years 2022-2025.

The routing SSD block produces, at every grid point $p$, lead time $l$, and forecast time $t$, a routing feature vector $x(p)^{t}_{l} \in \mathbb{R}^{192}$ (the residual in Eq.~\ref{eq:routing_ssm}). This is the state that the SSM propagates across the auto-regressive rollout. As a scalar proxy for the magnitude of this state, we use its L$2$-norm, $\bar{x}(p)^{t}_{l} = \|x(p)^{t}_{l}\|_2$.

\begin{figure}[!h]
  \centering
  \includegraphics[width=.98\textwidth]{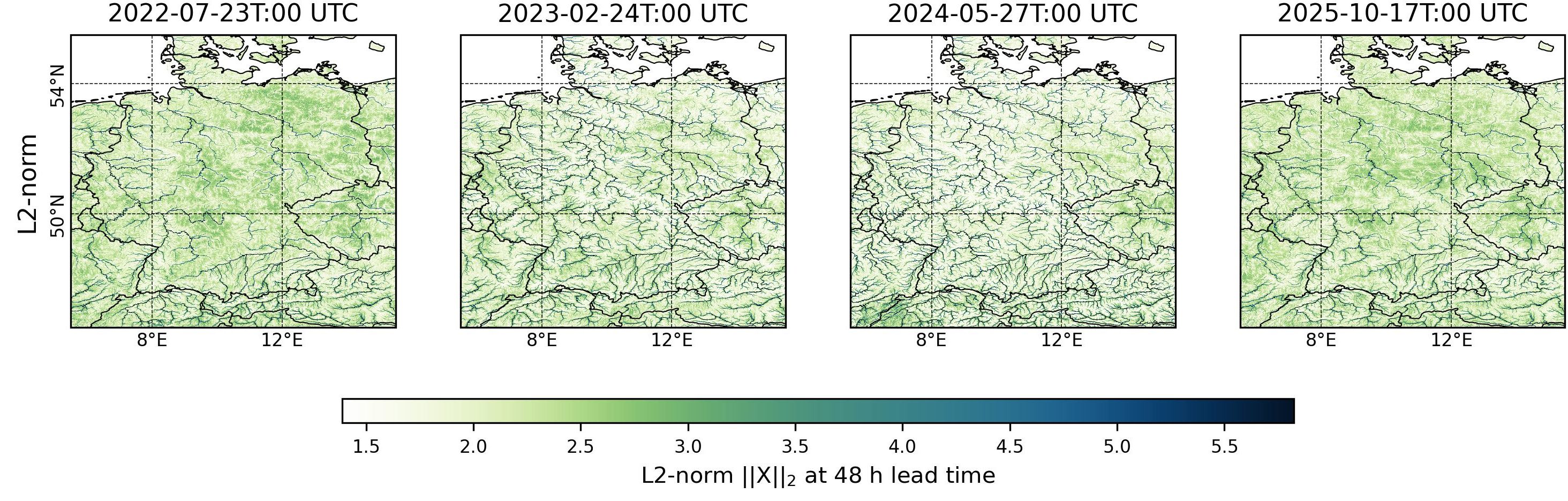}
  \caption{Routing state at a fixed $48\,$h lead time across four independent forecasts, one randomly sampled per year from 2022 to 2025. Top row shows the L$2$-norm of the $192$-dimensional routing feature vector across the years.}\label{fig:reservoir_2}
\end{figure}

In Fig.~\ref{fig:reservoir_2}, we show how state evolves across independent events. For this we fix the lead time at $48$\,h and compare the routing state across independent forecast times. 
We selected four random times for the forecast $2022$-$07$-$23$, $2023$-$02$-$24$, $2024$-$05$-$27$, and $2025$-$10$-$17$. 
Each subplot shows the norm vector $\bar{x}^t(p)$. 
All four subplots share the same coarse spatial structure of the river network. This shows, that the learned state is spatially organized like a hydrological routing system and it is not static but varies with time.

\clearpage
\newpage

\section{Additionl Results}
\subsection{Experiments on EFAS Reanalysis}

\subsubsection{Experiments on EFAS Reanalysis for CAMELS split}

\begin{table}[!h]
  \caption{Results on EFAS Reanalysis (CAMELS split). \gray{($\pm$)} denotes the standard deviation for 3 runs.}
  \label{table:appendix_reanalysis_baselines_1}
  \centering
  \small
  \tabcolsep=3.0pt\relax
  \setlength\extrarowheight{0pt}
  \begin{tabular}{c r *{8}l}
  \toprule
  & & & \multicolumn{3}{c}{Validation (2022-2023)} & & \multicolumn{3}{c}{Test (2024-2025)} \\
    \midrule
    & Model & & RMSE {\scriptsize($\downarrow$)} & R2 {\scriptsize($\uparrow$)} & KGE {\scriptsize($\uparrow$)} & & RMSE {\scriptsize($\downarrow$)} & R2 {\scriptsize($\uparrow$)} & KGE {\scriptsize($\uparrow$)} \\
    \midrule
   
    \multirow{4}{*}{\rotatebox[origin=c]{90}{\scriptsize{Deterministic}}} & Persistence & & 4.949 & 0.086 & 0.533 & & 6.013 & -0.118 & 0.456 \\

    & ED-LSTM & & 4.622{\scriptsize\color{gray}{$\pm$0.016}} & 0.229{\scriptsize\gray{$\pm$0.007}} & 0.300{\scriptsize\gray{$\pm$0.009}} & & 5.316{\scriptsize\gray{$\pm$0.010}} & 0.157{\scriptsize\gray{$\pm$0.007}} & 0.264{\scriptsize\gray{$\pm$0.013}} \\
    
    & ME-LSTM & & 4.830{\scriptsize\color{gray}{$\pm$0.034}} & 0.181{\scriptsize\gray{$\pm$0.011}} & 0.286{\scriptsize\gray{$\pm$0.018}} & & 5.584{\scriptsize\gray{$\pm$0.042}} & 0.095{\scriptsize\gray{$\pm$0.014}} & 0.235{\scriptsize\gray{$\pm$0.020}} \\
    
    & Ours & & \textbf{3.972}{\scriptsize\color{gray}{$\pm$0.052}} & 0.428{\scriptsize\gray{$\pm$0.015}} & 0.534{\scriptsize\gray{$\pm$0.011}} & & 4.747{\scriptsize\gray{$\pm$0.040}} & 0.333{\scriptsize\gray{$\pm$0.011}} & 0.481{\scriptsize\gray{$\pm$0.008}} \\
 \midrule
    \multirow{5}{*}{\rotatebox[origin=c]{90}{\scriptsize{Probabilistic}}} & Clima (median) & & 5.382 & -0.005 & -0.035 & & 6.025 & -0.068 & -0.074 \\
        
    & DRUM (mean) & & 4.985 & 0.150 & 0.151 & & 5.519 & 0.098 & 0.135 \\
    
    & ED-LSTM (mean) & & 4.782 & 0.173 & 0.243 & & 5.367 & 0.135 & 0.233 \\
    
    & ME-LSTM (mean) & & 4.682 & 0.209 & 0.307 & & 5.436 & 0.121 & 0.264 \\
    
    & Ours (mean) & & 3.991 & \textbf{0.452} & \textbf{0.575} & & \textbf{4.655} & \textbf{0.391} & \textbf{0.547} \\
  \bottomrule
\end{tabular}
\end{table}

\begin{figure}[h!]
  \centering
  \includegraphics[width=0.99\textwidth]{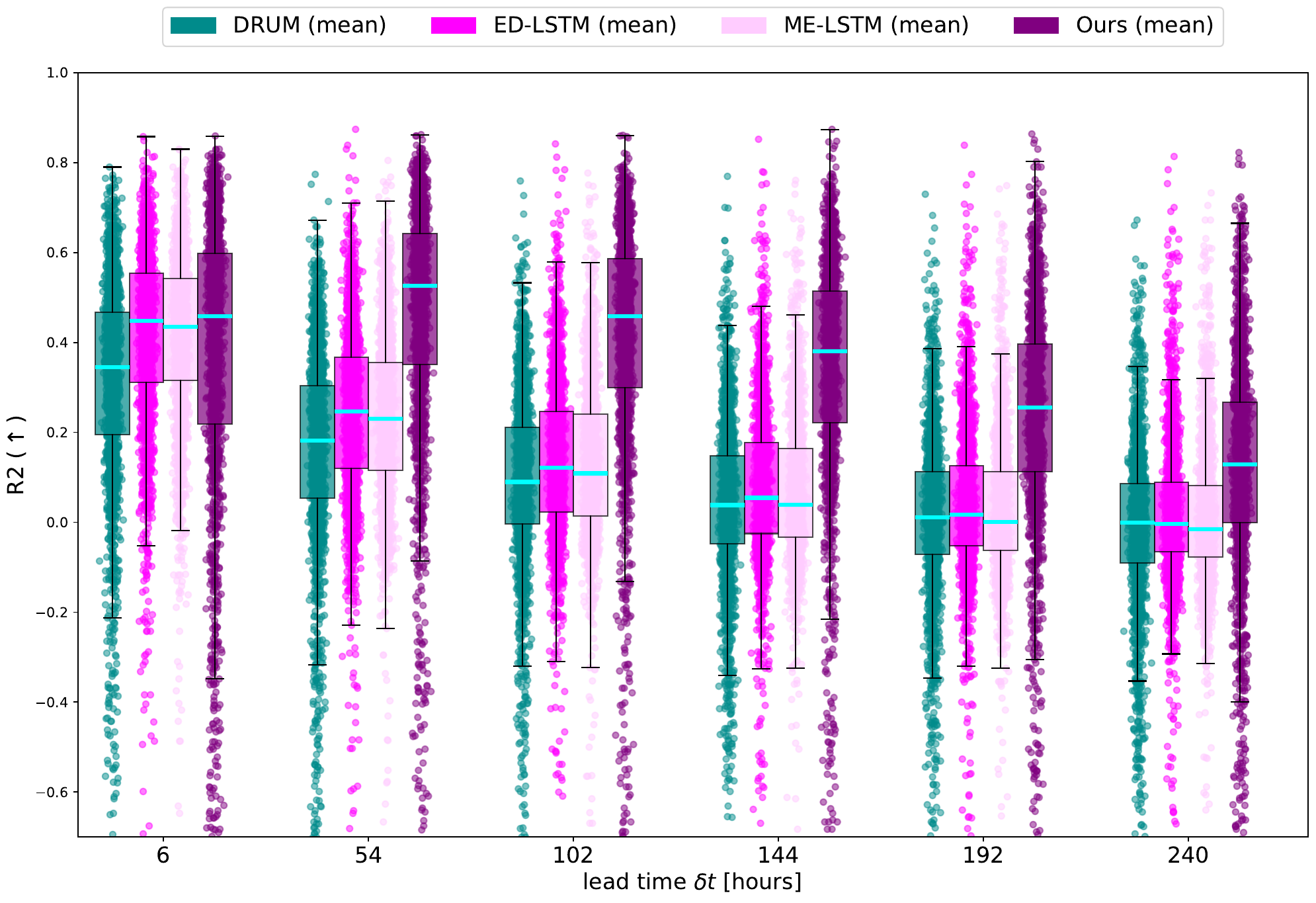}
  \caption{R2 of the river discharge forecasting with different lead time on reanalysis CAMELS split (test set 2024-2025, temporally out-of-sample). Distribution quartiles are displayed in boxes, and the entire range excluding outliers is displayed in whiskers. The median score for the model is shown by the cyan line in the box.}
  \label{fig:appendix_reanalysis_camels_r2}
\end{figure}

\begin{figure}[h!]
  \centering
  \includegraphics[width=0.99\textwidth]{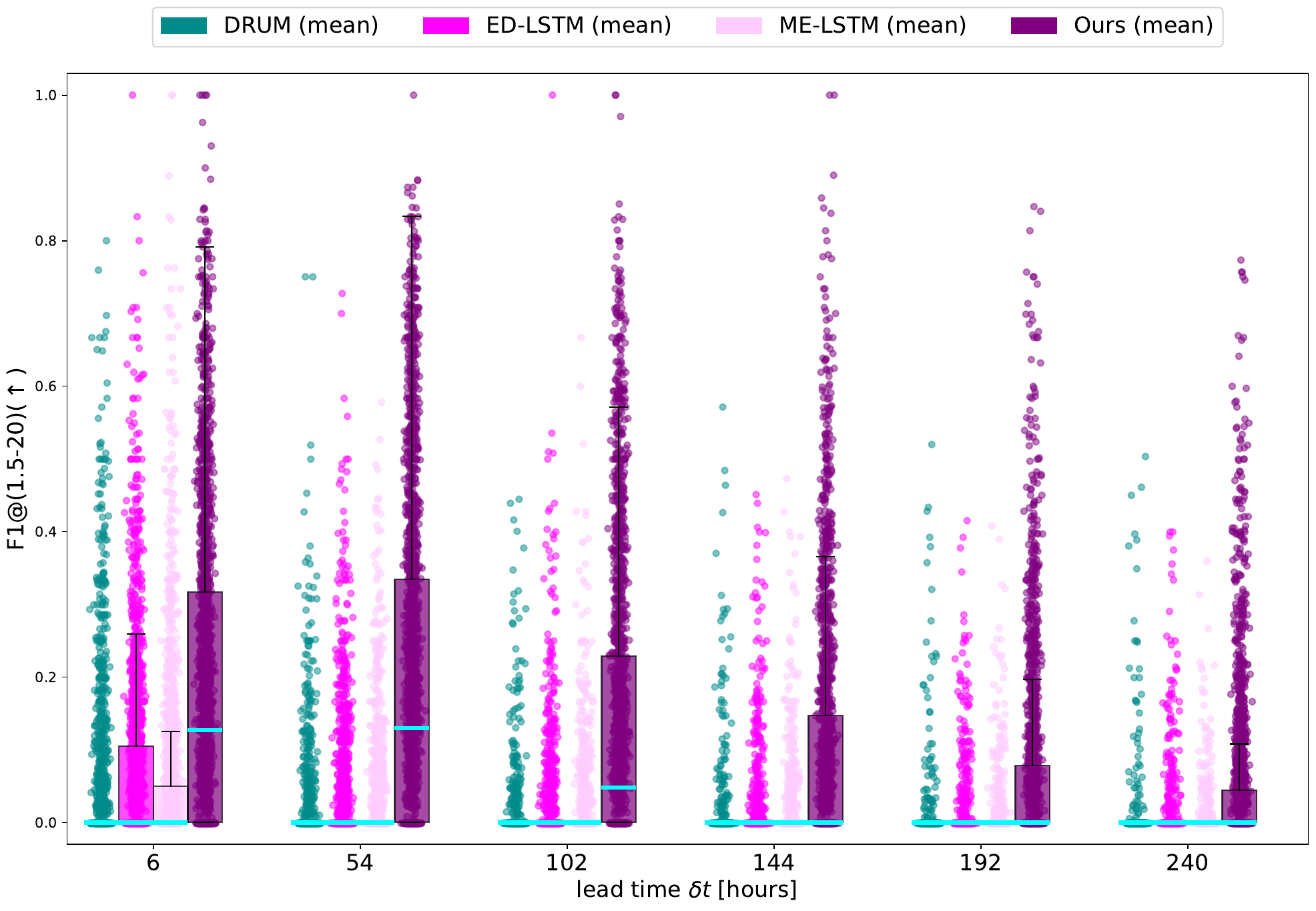}
  \caption{Mean F1@(1.5-20) of the river discharge forecasting with different lead time on reanalysis CAMELS split (test set 2024-2025, temporally out-of-sample). Distribution quartiles are displayed in boxes, and the entire range excluding outliers is displayed in whiskers. The median score for the model is shown by the cyan line in the box.}
  \label{fig:appendix_reanalysis_camels_f1}
\end{figure}

\begin{figure}[h!]
  \centering
  \includegraphics[width=0.98\textwidth]{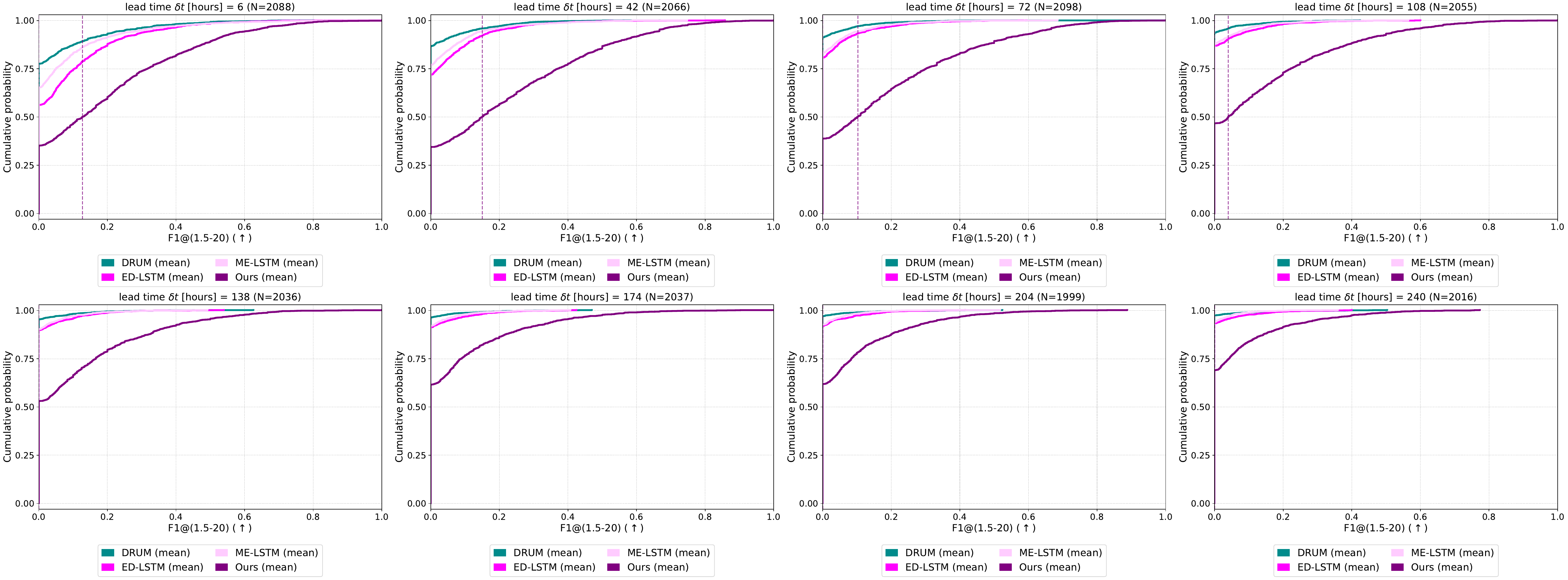}
  \caption{Cumulative distribution function (CDF) of station-level F1@(1.5-20) on the CAMELS reanalysis split (N=2206), across lead times. Dashed vertical lines shows each model's media.}
  \label{fig:appendix_reanalysis_camels_cdf_f1}
\end{figure}

\begin{figure}[h!]
  \centering
  \includegraphics[width=0.98\textwidth]{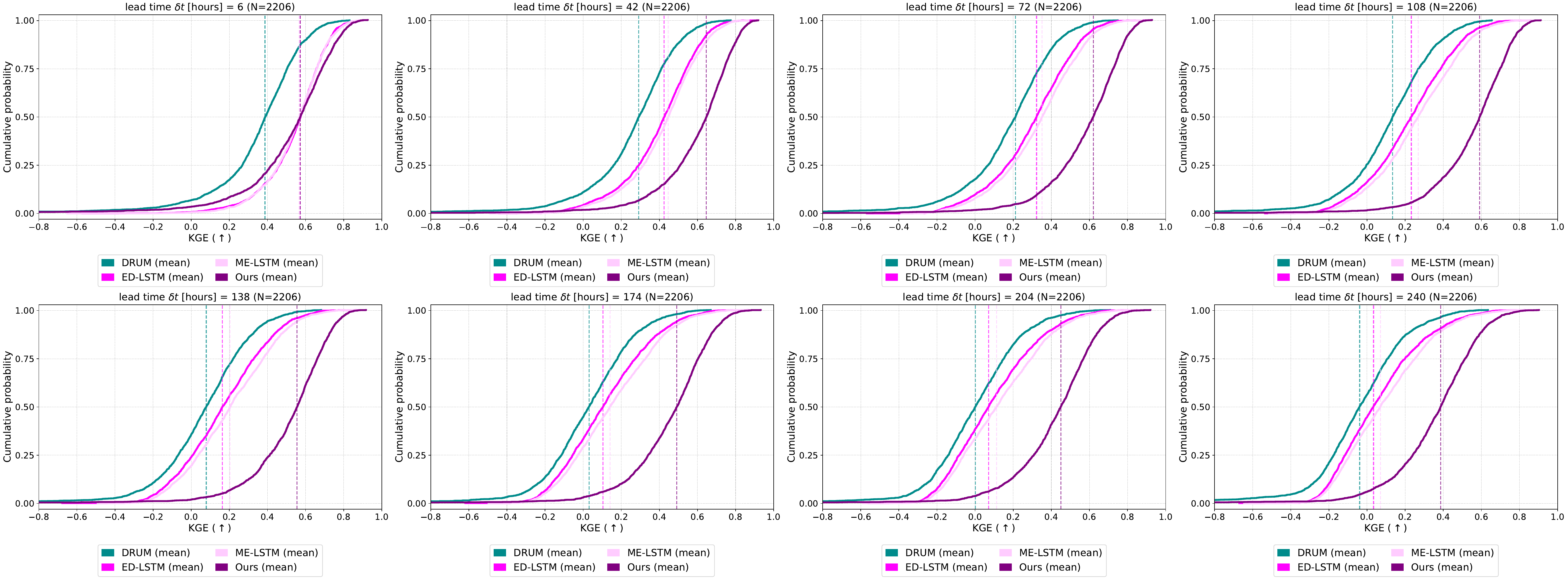}
  \caption{Cumulative distribution function (CDF) of station-level KGE on the CAMELS reanalysis split (N=2206), across lead times. Dashed vertical lines shows each model's media.}
  \label{fig:appendix_reanalysis_camels_cdf_kge}
\end{figure}

\begin{figure}[h!]
  \centering
  \includegraphics[width=0.98\textwidth]{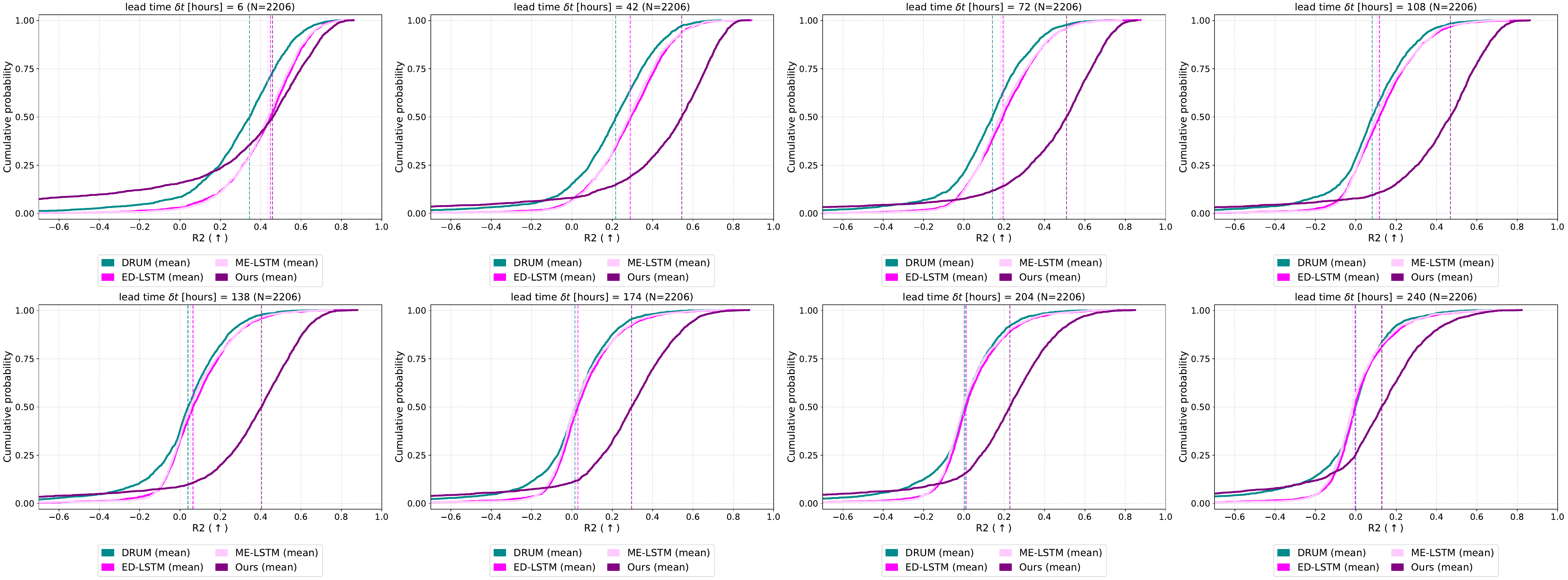}
  \caption{Cumulative distribution function (CDF) of station-level R2 on the CAMELS reanalysis split (N=2206), across lead times. Dashed vertical lines shows each model's media.}
  \label{fig:appendix_reanalysis_camels_cdf_r2}
\end{figure}

\clearpage
\newpage

\subsubsection{Experiments on EFAS Reanalysis for GRDC split}

\begin{table}[!h]
  \caption{Results on EFAS Reanalysis (GRDC split). \gray{($\pm$)} denotes the standard deviation for 3 runs.}
  \label{table:appendix_reanalysis_baselines_1}
  \centering
  \small
  \tabcolsep=3.0pt\relax
  \setlength\extrarowheight{0pt}
  \begin{tabular}{c r *{8}l}
  \toprule
  & & & \multicolumn{3}{c}{Validation (2022-2023)} & & \multicolumn{3}{c}{Test (2024-2025)} \\
    \midrule
    & Model & & RMSE {\scriptsize($\downarrow$)} & R2 {\scriptsize($\uparrow$)} & KGE {\scriptsize($\uparrow$)} & & RMSE {\scriptsize($\downarrow$)} & R2 {\scriptsize($\uparrow$)} & KGE {\scriptsize($\uparrow$)} \\
    \midrule
   
    \multirow{4}{*}{\rotatebox[origin=c]{90}{\scriptsize{Deterministic}}} & Persistence & & 10.417 & 0.096 & 0.542 & & 12.056 & -0.076 & 0.468 \\

    & ED-LSTM & & 9.739{\scriptsize\color{gray}{$\pm$0.065}} & 0.239{\scriptsize\gray{$\pm$0.009}} & 0.318{\scriptsize\gray{$\pm$0.009}} & & 10.975{\scriptsize\gray{$\pm$0.005}} & 0.151{\scriptsize\gray{$\pm$0.004}} & 0.269{\scriptsize\gray{$\pm$0.013}} \\
    
    & ME-LSTM & & 10.242{\scriptsize\color{gray}{$\pm$0.067}} & 0.186{\scriptsize\gray{$\pm$0.010}} & 0.321{\scriptsize\gray{$\pm$0.020}} & & 11.515{\scriptsize\gray{$\pm$0.062}} & 0.085{\scriptsize\gray{$\pm$0.013}} & 0.249{\scriptsize\gray{$\pm$0.021}} \\
    
    & Ours & & 8.494{\scriptsize\color{gray}{$\pm$0.138}} & 0.420{\scriptsize\gray{$\pm$0.019}} & 0.531{\scriptsize\gray{$\pm$0.014}} & & 9.851{\scriptsize\gray{$\pm$0.074}} & 0.314{\scriptsize\gray{$\pm$0.010}} & 0.474{\scriptsize\gray{$\pm$0.006}} \\
 \midrule
   \multirow{5}{*}{\rotatebox[origin=c]{90}{\scriptsize{Probabilistic}}} & Clima (median) & & 11.014 & 0.002 & -0.019 & & 12.169 & -0.062 & -0.058 \\

    & DRUM (mean) & & 10.461 & 0.148 & 0.169 & & 11.495 & 0.095 & 0.148 \\
    
    & ED-LSTM (mean) & & 10.092 & 0.178 & 0.255 & & 11.218 & 0.125 & 0.237 \\
    
    & ME-LSTM (mean) & & 9.882 & 0.211 & 0.331 & & 11.098 & 0.114 & 0.278 \\

    & Ours (mean) & & \textbf{8.375} & \textbf{0.441} & \textbf{0.567} & & \textbf{9.569} & \textbf{0.360} & \textbf{0.530} \\
  \bottomrule
\end{tabular}
\end{table}

\begin{table}[!h]
  \caption{Results for probabilistic models on EFAS Reanalysis (GRDC split).}
  \label{table:appendix_reanalysis_baselines_2}
  \centering
  \small
  \tabcolsep=6.0pt\relax
  \setlength\extrarowheight{0pt}
  \begin{tabular}{*{7}c}
  \toprule
   \multicolumn{7}{l}{Results on the Validation set (2022-2023)} \\
    \midrule
     & & Clima & DRUM & ED-LSTM & ME-LSTM & Ours \\
    \midrule
  CRPS {\scriptsize($\downarrow$)} & & 4.3328 &  4.5104 & 4.1379 & 3.9776 & \textbf{3.1837} \\
    \midrule
  \multicolumn{7}{l}{Results on the Test set (2024-2025)} \\
    \midrule
    & & Clima & DRUM & ED-LSTM & ME-LSTM & Ours \\
    \midrule
    CRPS {\scriptsize($\downarrow$)} & & 4.6667 & 4.9376 & 4.6384 & 4.6492 & \textbf{3.7604} \\
  \bottomrule
\end{tabular}
\end{table}

\begin{figure}[h!]
  \centering
  \includegraphics[width=0.8\textwidth]{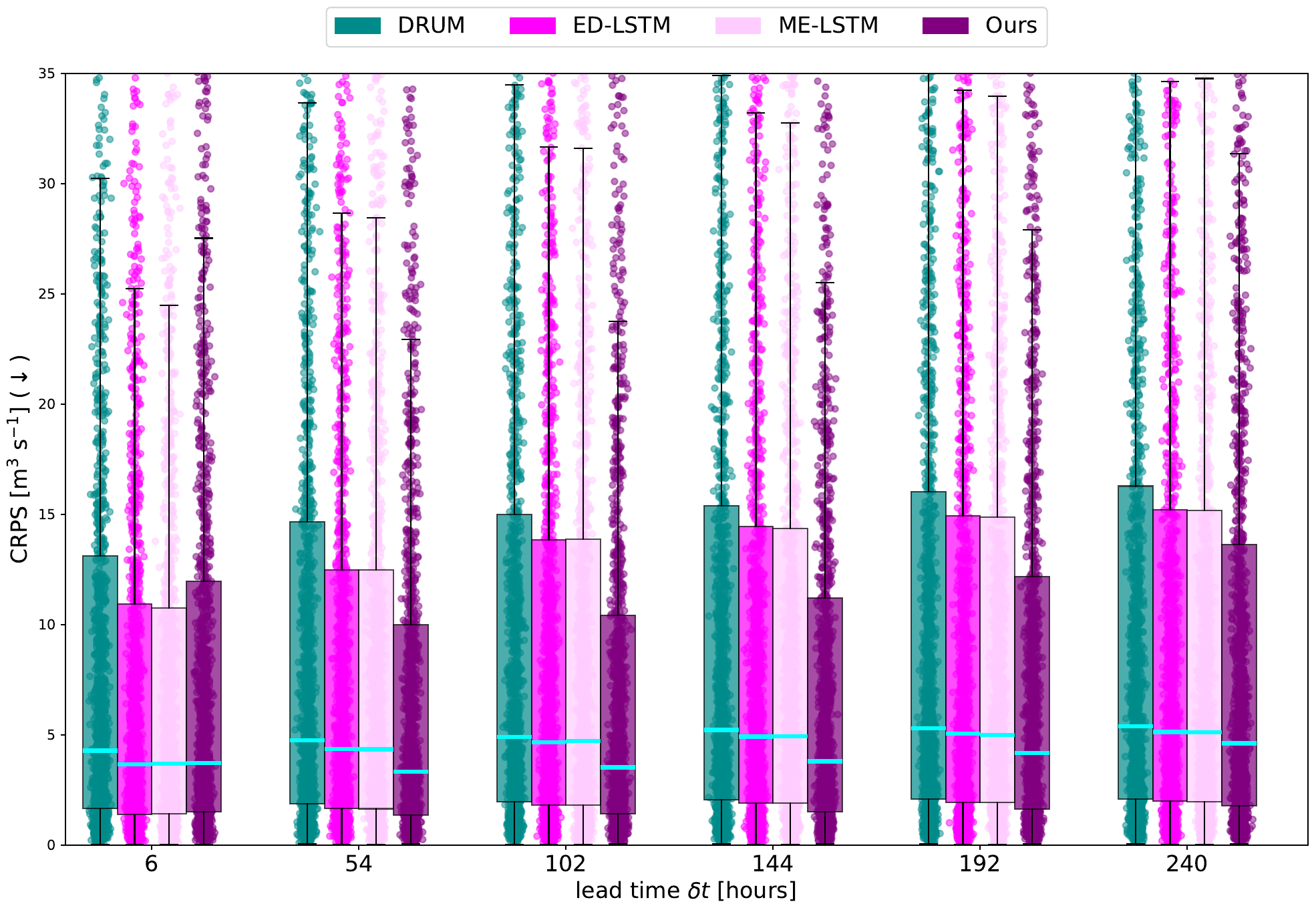}
  \caption{CRPS of the river discharge forecasting with different lead time on reanalysis GRDC split (test set 2024-2025, temporally out-of-sample). Distribution quartiles are displayed in boxes, and the entire range excluding outliers is displayed in whiskers. The median score for the model is shown by the cyan line in the box.}
  \label{fig:appendix_reanalysis_grdc_crps}
\end{figure}

\begin{figure}[h!]
  \centering
  \includegraphics[width=0.9\textwidth]{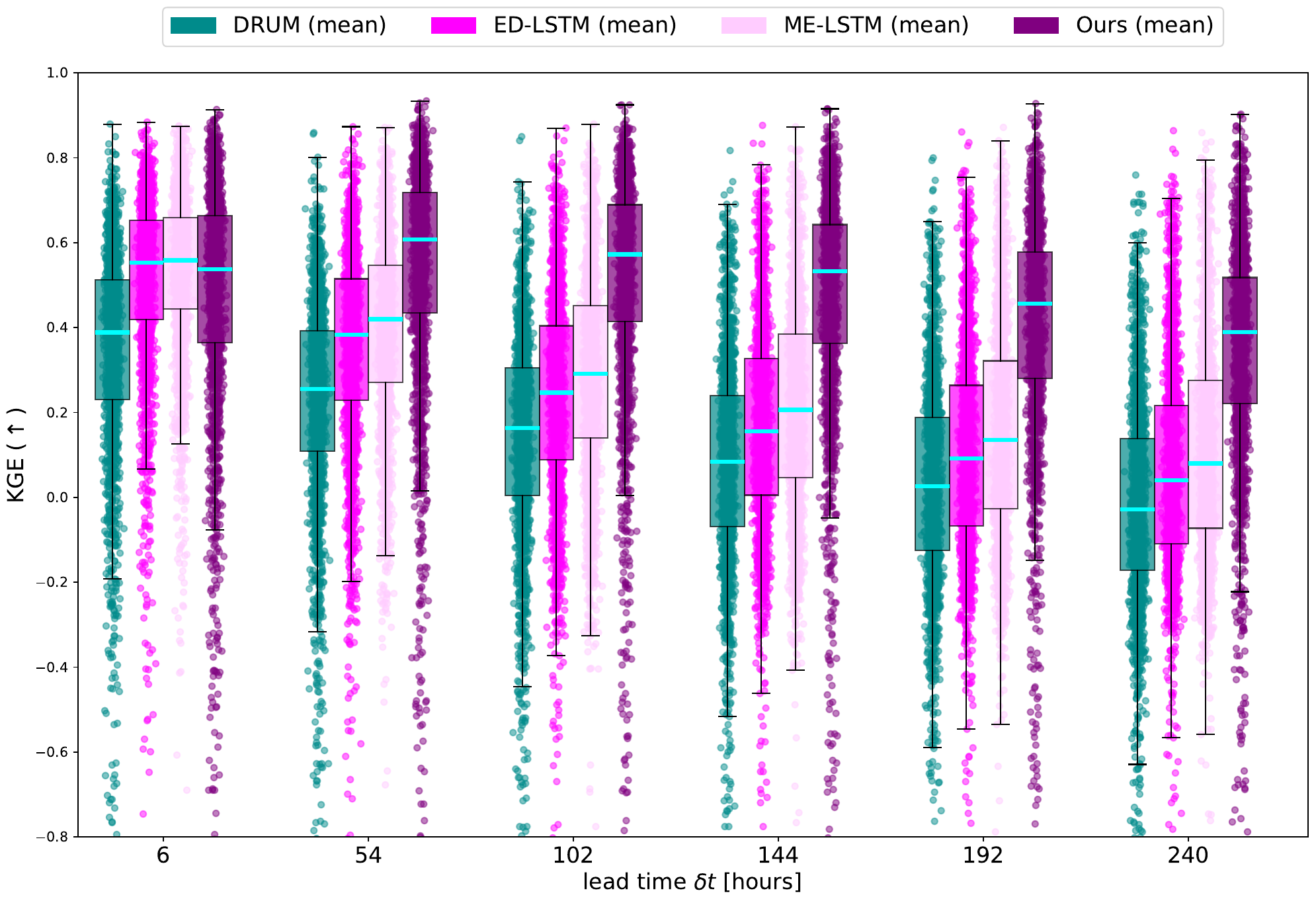}
  \caption{KGE of the river discharge forecasting with different lead time on reanalysis GRDC split (test set 2024-2025, temporally out-of-sample). Distribution quartiles are displayed in boxes, and the entire range excluding outliers is displayed in whiskers. The median score for the model is shown by the cyan line in the box.}
  \label{fig:appendix_reanalysis_grdc_kge}
\end{figure}

\begin{figure}[h!]
  \centering
  \includegraphics[width=0.9\textwidth]{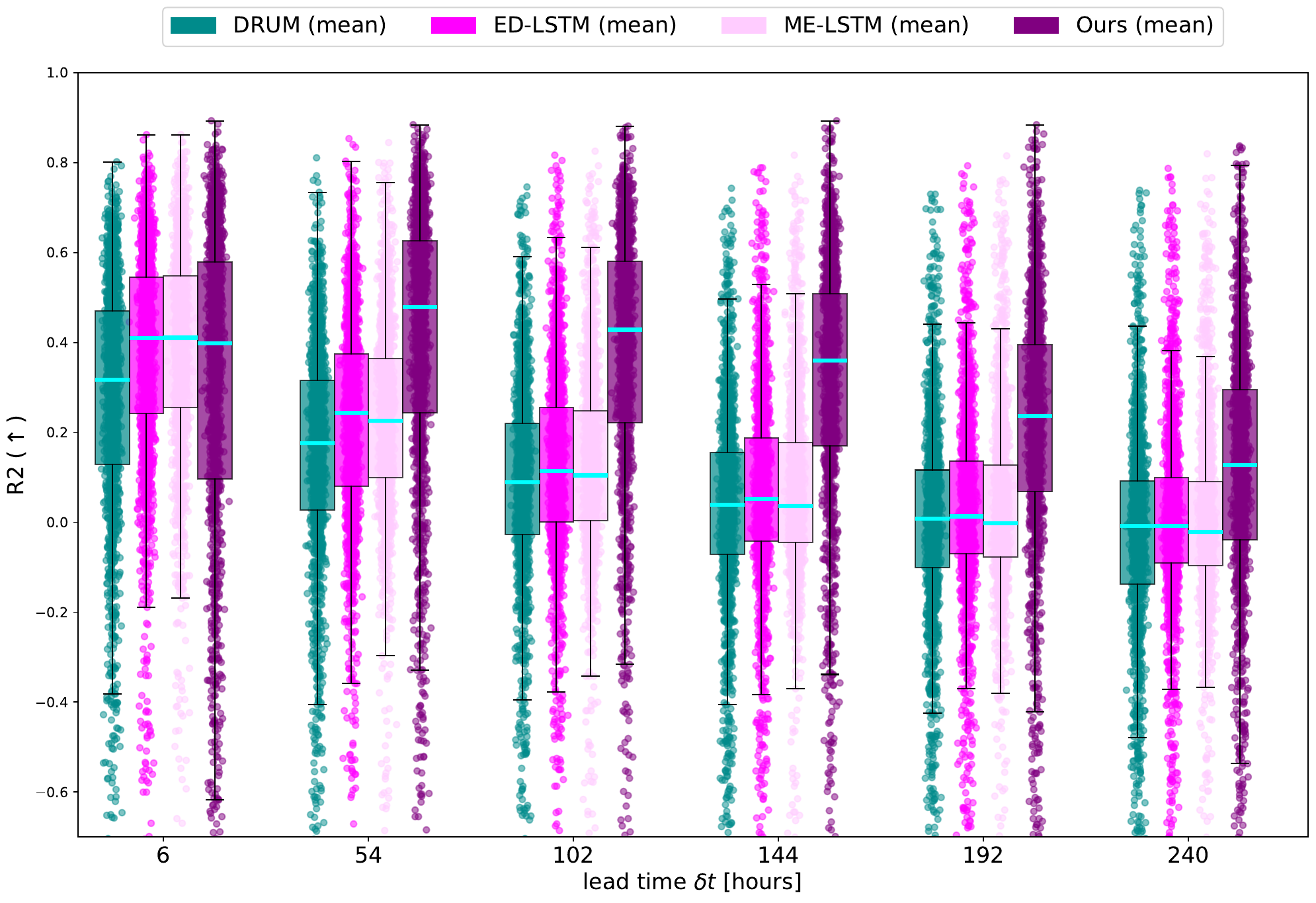}
  \caption{R2 of the river discharge forecasting with different lead time on reanalysis GRDC split (test set 2024-2025, temporally out-of-sample). Distribution quartiles are displayed in boxes, and the entire range excluding outliers is displayed in whiskers. The median score for the model is shown by the cyan line in the box.}
  \label{fig:appendix_reanalysis_grdc_r2}
\end{figure}

\begin{figure}[h!]
  \centering
  \includegraphics[width=0.99\textwidth]{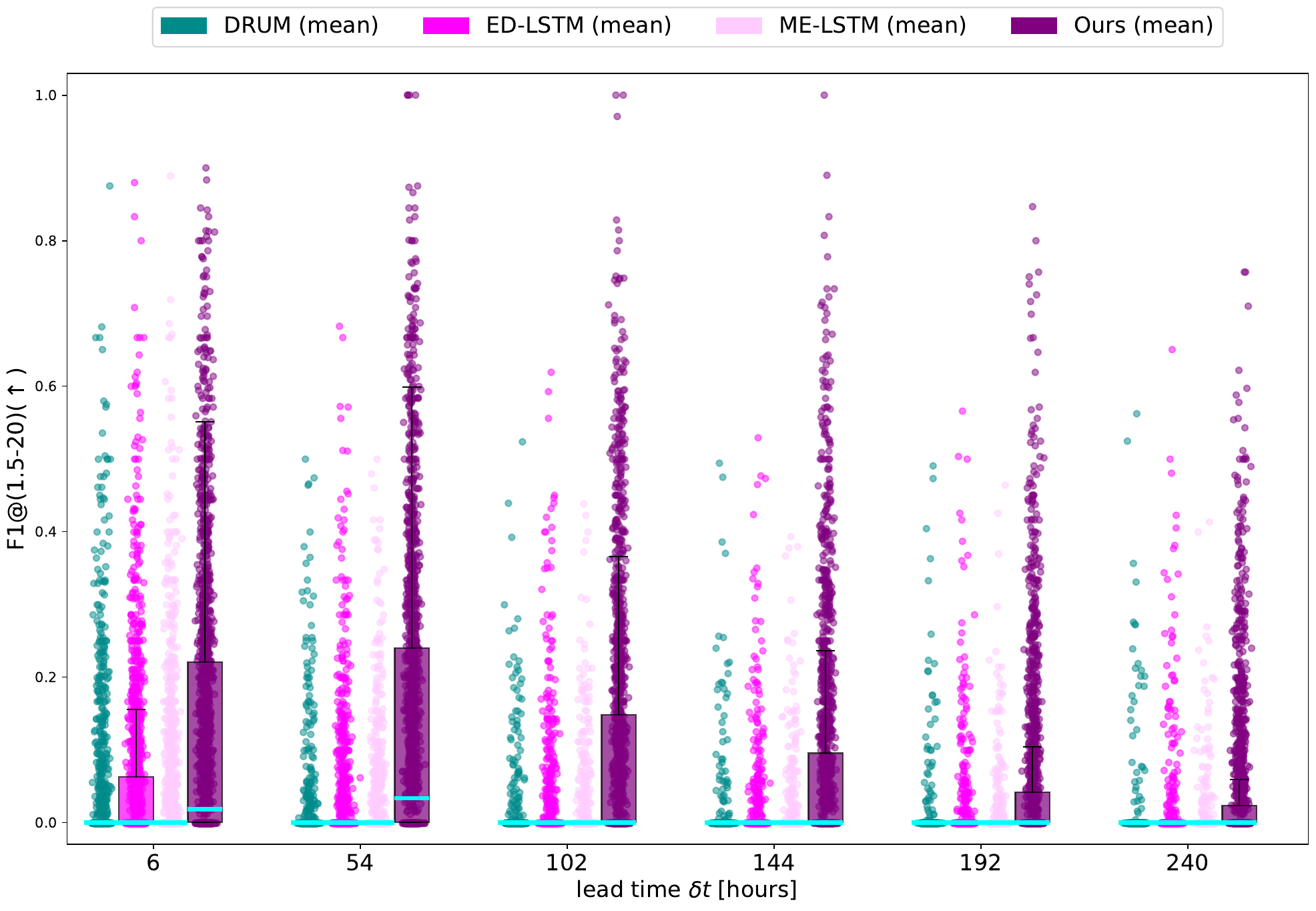}
  \caption{Mean F1@(1.5-20) of the river discharge forecasting with different lead time on reanalysis GDRC split (test set 2024-2025, temporally out-of-sample). Distribution quartiles are displayed in boxes, and the entire range excluding outliers is displayed in whiskers. The median score for the model is shown by the cyan line in the box.}
  \label{fig:appendix_reanalysis_grdc_f1}
\end{figure}

\begin{figure}[h!]
  \centering
  \includegraphics[width=0.98\textwidth]{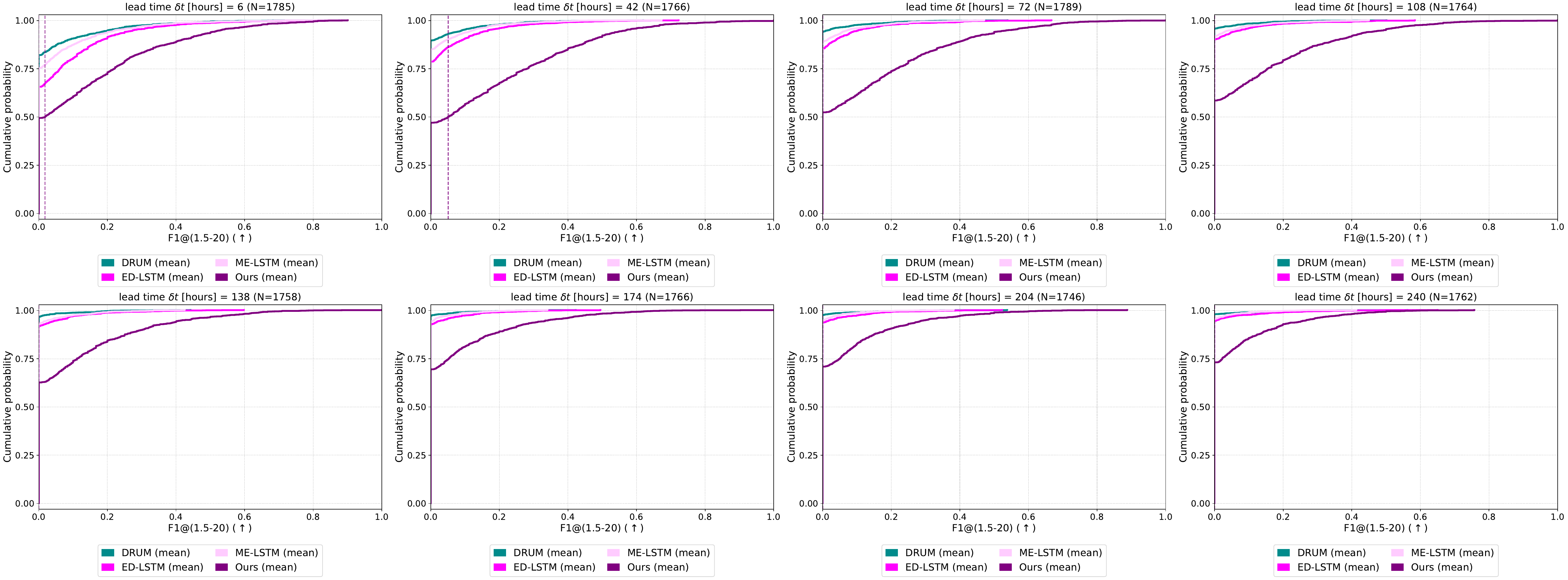}
  \caption{Cumulative distribution function (CDF) of station-level F1@(1.5-20) on the GRDC reanalysis split (N=1971), across lead times. Dashed vertical lines shows each model's media.}
  \label{fig:appendix_reanalysis_grdc_cdf_f1}
\end{figure}

\begin{figure}[h!]
  \centering
  \includegraphics[width=0.98\textwidth]{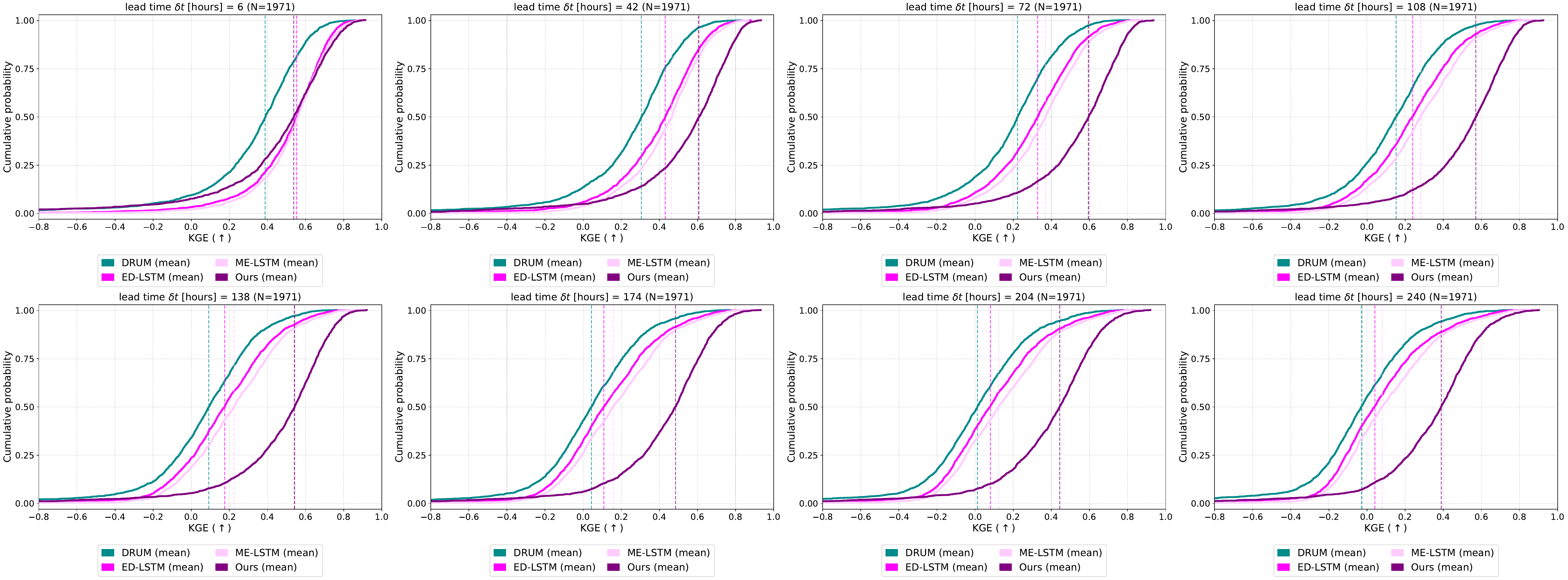}
  \caption{Cumulative distribution function (CDF) of station-level KGE on the GRDC reanalysis split (N=1971), across lead times. Dashed vertical lines shows each model's media.}
  \label{fig:appendix_reanalysis_grdc_cdf_kge}
\end{figure}

\begin{figure}[h!]
  \centering
  \includegraphics[width=0.98\textwidth]{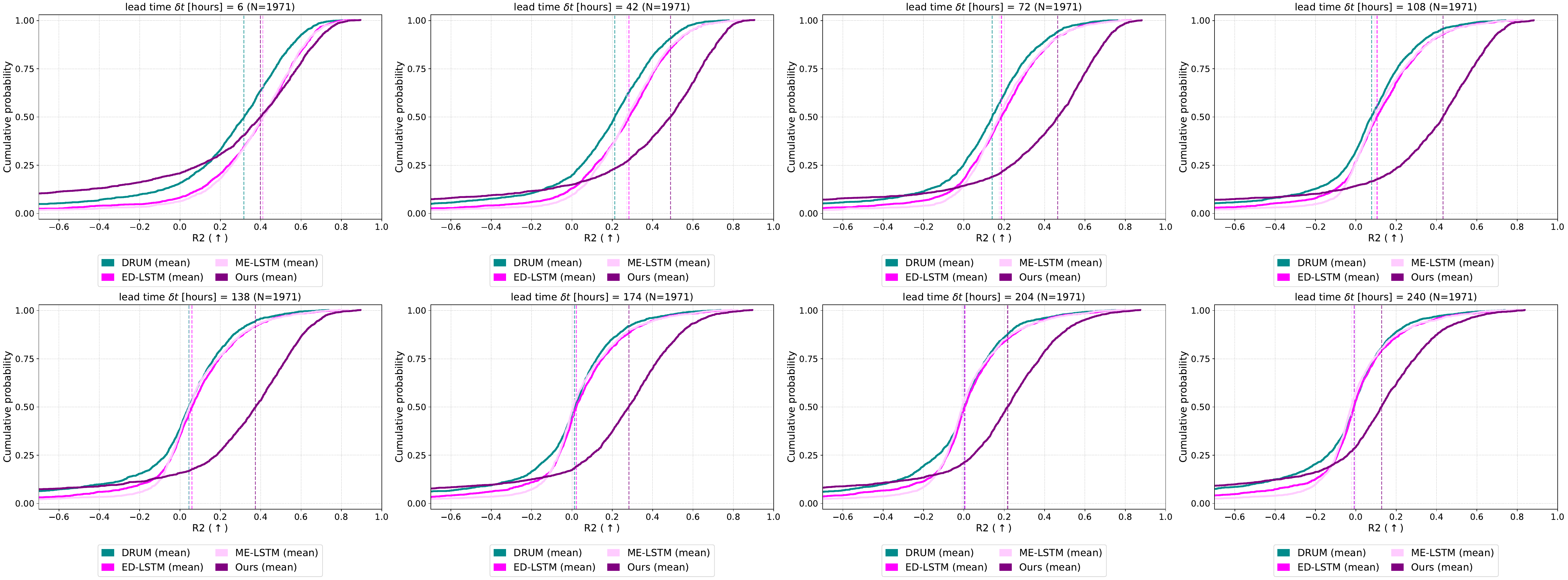}
  \caption{Cumulative distribution function (CDF) of station-level R2 on the GRDC reanalysis split (N=1971), across lead times. Dashed vertical lines shows each model's media.}
  \label{fig:appendix_reanalysis_grdc_cdf_r2}
\end{figure}

\clearpage
\newpage

\subsection{Experiments on GRDC Observations}

\begin{table}[!h]
  \caption{Results on GRDC observations. \gray{($\pm$)} denotes the standard deviation for 3 runs.}
  \label{table:appendix_reanalysis_baselines_1}
  \centering
  \small
  \tabcolsep=3.0pt\relax
  \setlength\extrarowheight{0pt}
  \begin{tabular}{c r *{8}l}
  \toprule
  & & & \multicolumn{3}{c}{Validation (2022-2023)} & & \multicolumn{3}{c}{Test (2024-2025)} \\
    \midrule
    & Model & & RMSE {\scriptsize($\downarrow$)} & R2 {\scriptsize($\uparrow$)} & KGE {\scriptsize($\uparrow$)} & & RMSE {\scriptsize($\downarrow$)} & R2 {\scriptsize($\uparrow$)} & KGE {\scriptsize($\uparrow$)} \\
    \midrule

    \multirow{3}{*}{\rotatebox[origin=c]{90}{\tiny{Deterministic}}} & ED-LSTM & & 6.0542{\scriptsize\color{gray}{$\pm$0.0082}} & 0.2705{\scriptsize\gray{$\pm$0.0012}} & 0.3491{\scriptsize\gray{$\pm$0.0214}} & & 11.3616{\scriptsize\gray{$\pm$0.1632}} & 0.2366{\scriptsize\gray{$\pm$0.0164}} & 0.4501{\scriptsize\gray{$\pm$0.0097}} \\
    
    & ME-LSTM & & 5.8810{\scriptsize\color{gray}{$\pm$0.0206}} & 0.2993{\scriptsize\gray{$\pm$0.0044}} & 0.3550{\scriptsize\gray{$\pm$0.0083}} & & 11,4754{\scriptsize\gray{$\pm$0.1683}} & 0.2384{\scriptsize\gray{$\pm$0.0246}} & 0.4330{\scriptsize\gray{$\pm$0.0138}} \\
    
    & Ours & & 4.6795{\scriptsize\color{gray}{$\pm$0.0681}} & 0.5603{\scriptsize\gray{$\pm$0.0095}} & 0.5890{\scriptsize\gray{$\pm$0.0038}} & & 9.0163{\scriptsize\gray{$\pm$0.0952}} & 0.4576{\scriptsize\gray{$\pm$0.0066}} & 0.5839{\scriptsize\gray{$\pm$0.0076}} \\
 \midrule

    \multirow{4}{*}{\rotatebox[origin=c]{90}{\tiny{Probabilistic}}} & DRUM (mean) & & 7.0236 & 0.1514 & 0.2177 & & 12.6358 & 0.1153 & 0.3059 \\
        
    & ED-LSTM (mean) & & 6.4217 & 0.2004 & 0.2780 & & 11.5629 & 0.2308 & 0.4266 \\
    
    & ME-LSTM (mean) & & 6.0267 & 0.2682 & 0.3921 & & 11.6064 & 0.2084 & 0.4478 \\
    
    & Ours (mean) & & \textbf{4.3823} & \textbf{0.6343} & \textbf{0.7021} & & \textbf{4.7963} & \textbf{0.6174} & \textbf{0.6893} \\
  \bottomrule
\end{tabular}
\end{table}

\begin{table}[!h]
  \caption{Results for probabilistic models on GRDC observations.}
  \label{table:appendix_reanalysis_baselines_2}
  \centering
  \small
  \tabcolsep=6.0pt\relax
  \setlength\extrarowheight{0pt}
  \begin{tabular}{*{6}c}
  \toprule
   \multicolumn{6}{l}{Results on the Validation set (2022-2023)} \\
    \midrule
     & & DRUM & ED-LSTM & ME-LSTM & Ours \\
    \midrule
  CRPS {\scriptsize($\downarrow$)} & & 4.5375 & 3.9024 & 3.4928 & \textbf{2.4592} \\
    \midrule
  \multicolumn{6}{l}{Results on the Test set (2024-2025)} \\
    \midrule
    & & DRUM & ED-LSTM & ME-LSTM & Ours \\
    \midrule
    CRPS {\scriptsize($\downarrow$)} & & 5.6934 & 5.0091 & 5.0361 & \textbf{3.5194} \\
  \bottomrule
\end{tabular}
\end{table}

\begin{figure}[h!]
  \centering
  \includegraphics[width=0.99\textwidth]{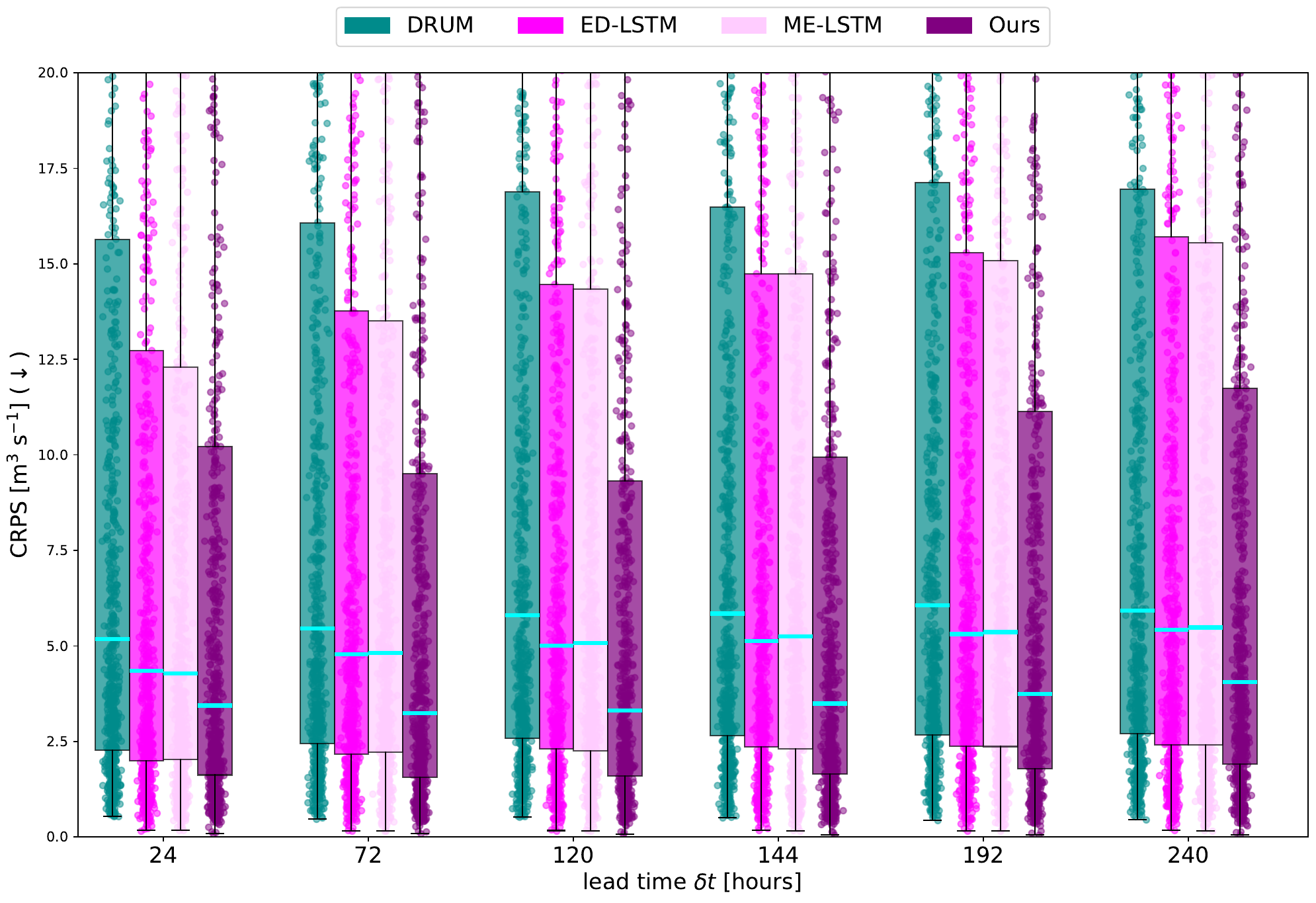}
  \caption{CRPS of the river discharge forecasting with different lead time on GRDC observations (test set 2024-2025, temporally out-of-sample). Distribution quartiles are displayed in boxes, and the entire range excluding outliers is displayed in whiskers. The median score for the model is shown by the cyan line in the box.}
  \label{fig:appendix_obs_grdc_crps}
\end{figure}

\begin{figure}[h!]
  \centering
  \includegraphics[width=0.9\textwidth]{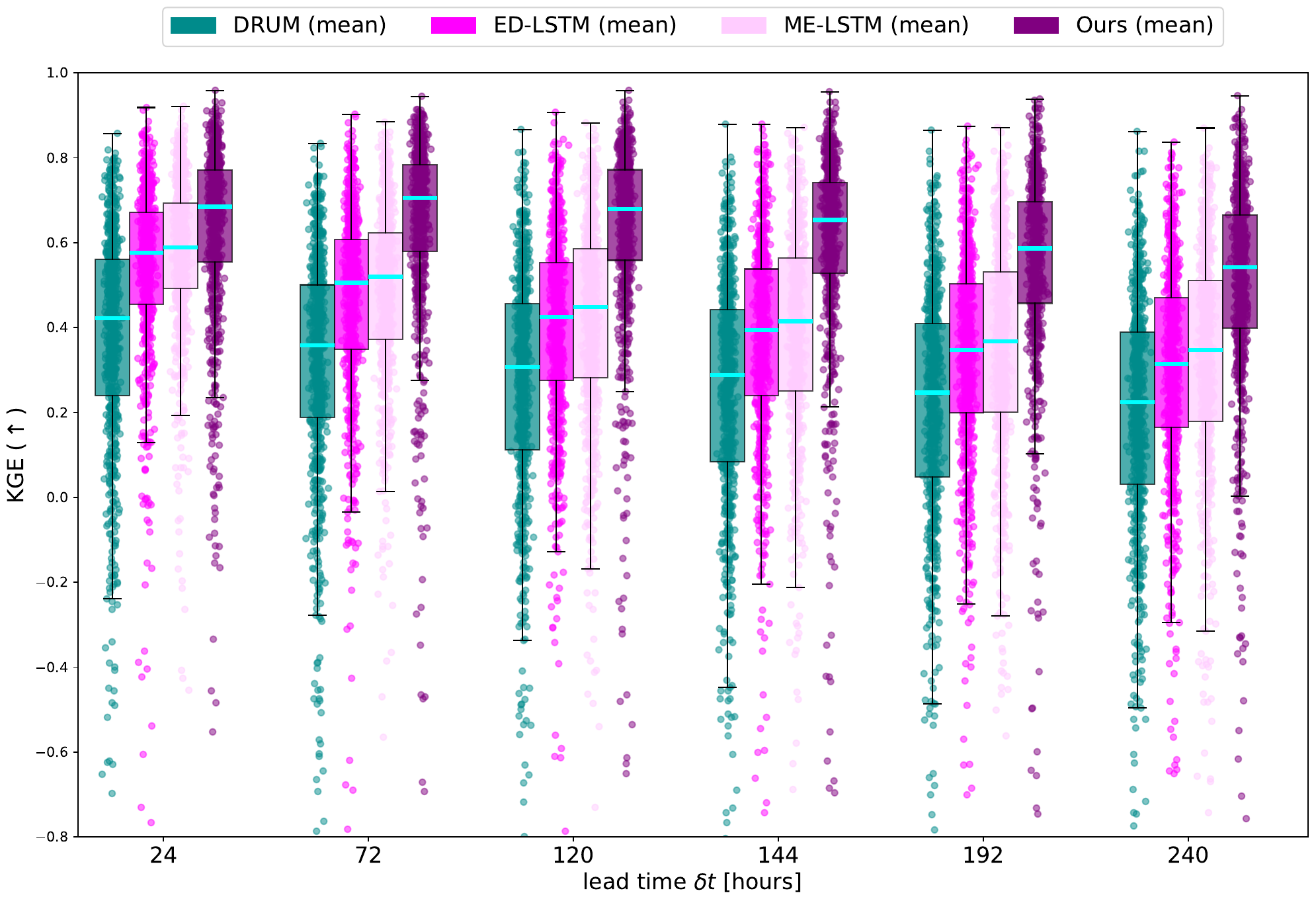}
  \caption{KGE of the river discharge forecasting with different lead time on GRDC observations (test set 2024-2025, temporally out-of-sample). Distribution quartiles are displayed in boxes, and the entire range excluding outliers is displayed in whiskers. The median score for the model is shown by the cyan line in the box.}
  \label{fig:appendix_obs_grdc_kge}
\end{figure}

\begin{figure}[h!]
  \centering
  \includegraphics[width=0.9\textwidth]{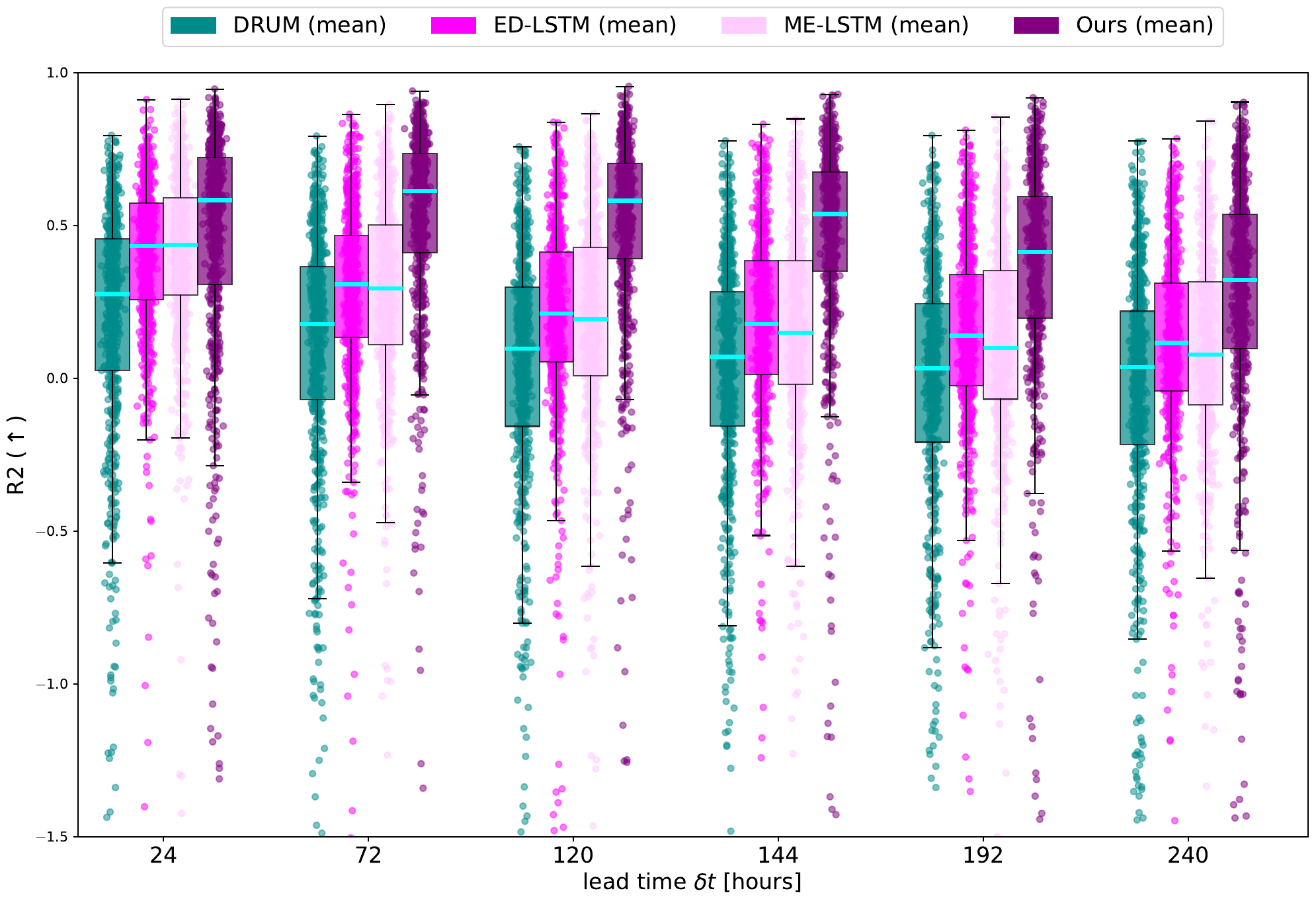}
  \caption{R2 of the river discharge forecasting with different lead time on GRDC observations (test set 2024-2025, temporally out-of-sample). Distribution quartiles are displayed in boxes, and the entire range excluding outliers is displayed in whiskers. The median score for the model is shown by the cyan line in the box.}
  \label{fig:appendix_obs_grdc_r2}
\end{figure}

\begin{figure}[h!]
  \centering
  \includegraphics[width=0.99\textwidth]{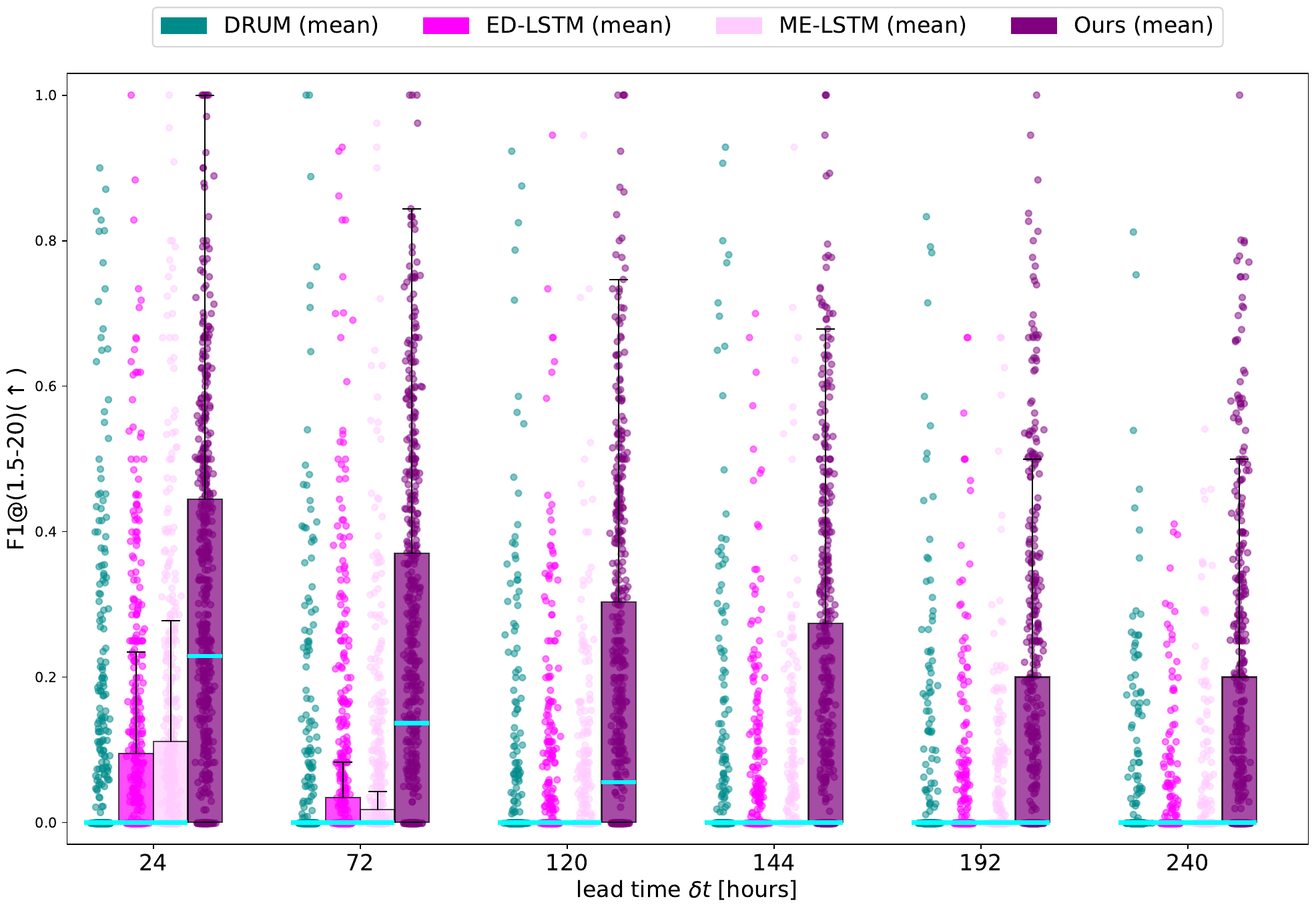}
  \caption{mean F1@(1.5-20) of the river discharge forecasting with different lead time on GRDC observations (test set 2024-2025, temporally out-of-sample). Distribution quartiles are displayed in boxes, and the entire range excluding outliers is displayed in whiskers. The median score for the model is shown by the cyan line in the box.}
  \label{fig:appendix_obs_grdc_f1}
\end{figure}

\begin{figure}[h!]
  \centering
  \includegraphics[width=0.98\textwidth]{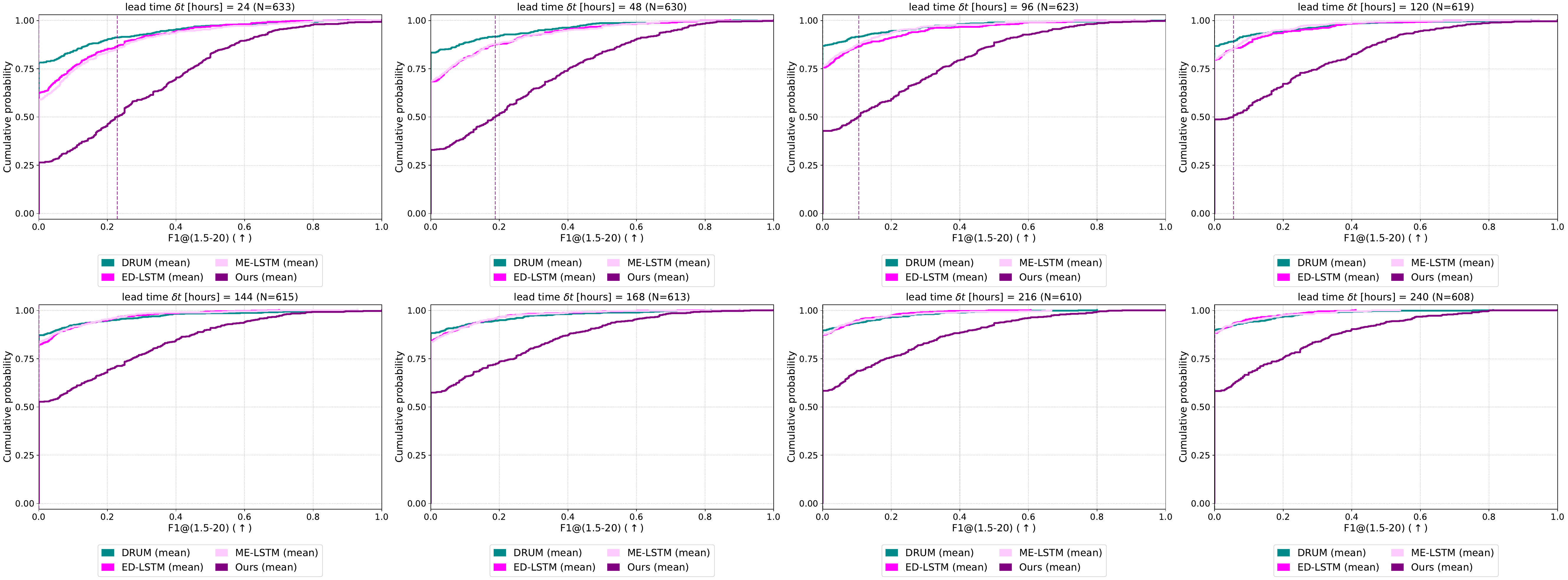}
  \caption{Cumulative distribution function (CDF) of station-level F1@(1.5-20) on GRDC observations (N=758), across lead times. Dashed vertical lines shows each model's media.}
  \label{fig:appendix_obs_grdc_cdf_f1}
\end{figure}

\begin{figure}[h!]
  \centering
  \includegraphics[width=0.98\textwidth]{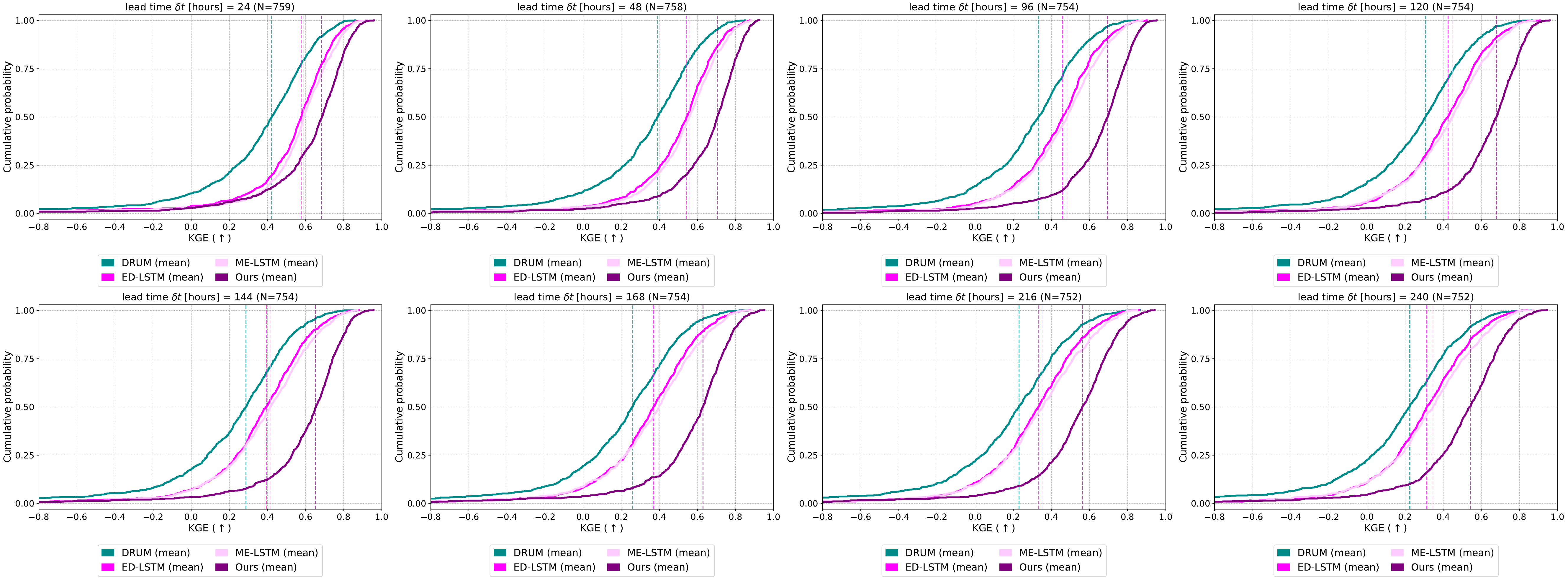}
  \caption{Cumulative distribution function (CDF) of station-level KGE on GRDC observations (N=758), across lead times. Dashed vertical lines shows each model's media.}
  \label{fig:appendix_obs_grdc_cdf_kge}
\end{figure}

\begin{figure}[h!]
  \centering
  \includegraphics[width=0.98\textwidth]{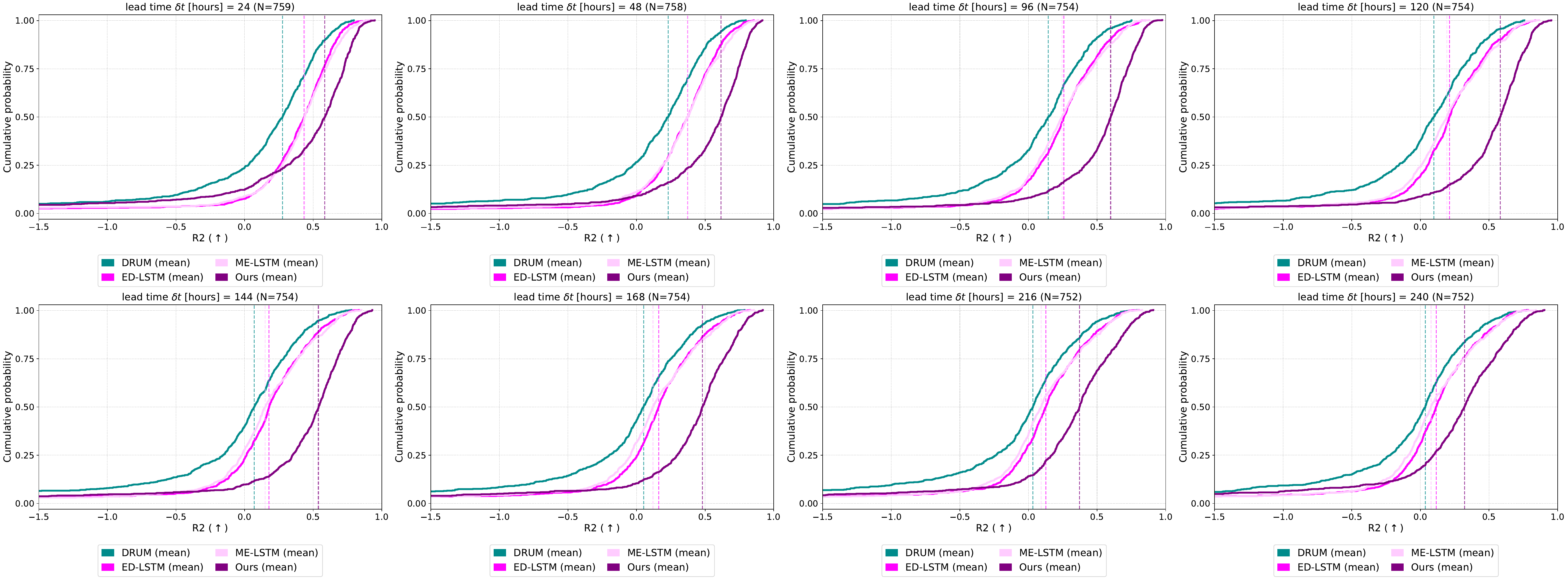}
  \caption{Cumulative distribution function (CDF) of station-level R2 on GRDC observations (N=758), across lead times. Dashed vertical lines shows each model's media.}
  \label{fig:appendix_obs_grdc_cdf_r2}
\end{figure}

\clearpage
\newpage

\subsection{Experiments on CAMELS Observations}

\begin{table}[!h]
  \caption{Results on CAMELS observations. \gray{($\pm$)} denotes the standard deviation for 3 runs.}
  \label{table:appendix_reanalysis_baselines_1}
  \centering
  \small
  \tabcolsep=3.0pt\relax
  \setlength\extrarowheight{0pt}
  \begin{tabular}{c r *{8}l}
  \toprule
  & & & \multicolumn{3}{c}{Validation (2022-2023)} & & \multicolumn{3}{c}{Test (2024)} \\
    \midrule
    & Model & & RMSE {\scriptsize($\downarrow$)} & R2 {\scriptsize($\uparrow$)} & KGE {\scriptsize($\uparrow$)} & & RMSE {\scriptsize($\downarrow$)} & R2 {\scriptsize($\uparrow$)} & KGE {\scriptsize($\uparrow$)} \\
    \midrule

    \multirow{3}{*}{\rotatebox[origin=c]{90}{\tiny{Deterministic}}} & ED-LSTM & & 6.0542{\scriptsize\color{gray}{$\pm$0.0082}} & 0.2705{\scriptsize\gray{$\pm$0.0012}} & 0.3401{\scriptsize\gray{$\pm$0.0214}} & & 6.8658{\scriptsize\gray{$\pm$0.1326}} & 0.0544{\scriptsize\gray{$\pm$0.0072}} & 0.0906{\scriptsize\gray{$\pm$0.0136}} \\
    
    & ME-LSTM & & 5.8810{\scriptsize\color{gray}{$\pm$0.0206}} & 0.2993{\scriptsize\gray{$\pm$0.0044}} & 0.3550{\scriptsize\gray{$\pm$0.0083}} & & 6.9621{\scriptsize\gray{$\pm$0.0874}} & 0.0500{\scriptsize\gray{$\pm$0.0100}} & 0.1186{\scriptsize\gray{$\pm$0.0123}} \\
    
    & Ours & & 4.6795{\scriptsize\color{gray}{$\pm$0.0681}} & 0.5603{\scriptsize\gray{$\pm$0.0095}} & 0.5890{\scriptsize\gray{$\pm$0.0038}} & & 5.8406{\scriptsize\gray{$\pm$0.1311}} & 0.3485{\scriptsize\gray{$\pm$0.0202}} & 0.4162{\scriptsize\gray{$\pm$0.0284}} \\
 \midrule

    \multirow{4}{*}{\rotatebox[origin=c]{90}{\tiny{Probabilistic}}} & DRUM (mean) & & 7.0236 & 0.1514 & 0.2177 & & 7.3433 & 0.0115 & 0.0576 \\
    
    & ED-LSTM (mean) & & 6.4217 & 0.2004 & 0.2780 & & 7.0896 & 0.0452 & 0.0831 \\
    
    & ME-LSTM (mean) & & 6.0267 & 0.2682 & 0.3921 & & 6.9401 & 0.0305 & 0.1046 \\
    
    & Ours (mean) & & \textbf{4.3823} & \textbf{0.6343} & \textbf{0.7021} & & \textbf{5.5031} & \textbf{0.4222} & \textbf{0.5143} \\
  \bottomrule
\end{tabular}
\end{table}

\begin{table}[!h]
  \caption{Results for probabilistic models on CAMELS observations.}
  \label{table:appendix_reanalysis_baselines_2}
  \centering
  \small
  \tabcolsep=6.0pt\relax
  \setlength\extrarowheight{0pt}
  \begin{tabular}{*{6}c}
  \toprule
   \multicolumn{6}{l}{Results on the Validation set (2022-2023)} \\
    \midrule
     & & DRUM & ED-LSTM & ME-LSTM & Ours \\
    \midrule
  CRPS {\scriptsize($\downarrow$)} & & 2.7239 & 2.3110 & 2.1237 & \textbf{1.4840} \\
    \midrule
  \multicolumn{6}{l}{Results on the Test set (2024)} \\
    \midrule
    & & DRUM & ED-LSTM & ME-LSTM & Ours \\
    \midrule
    CRPS {\scriptsize($\downarrow$)} & & 2.3685 & 2.2683 & 2.2683 & \textbf{1.5678} \\
  \bottomrule
\end{tabular}
\end{table}

\begin{figure}[h!]
  \centering
  \includegraphics[width=0.99\textwidth]{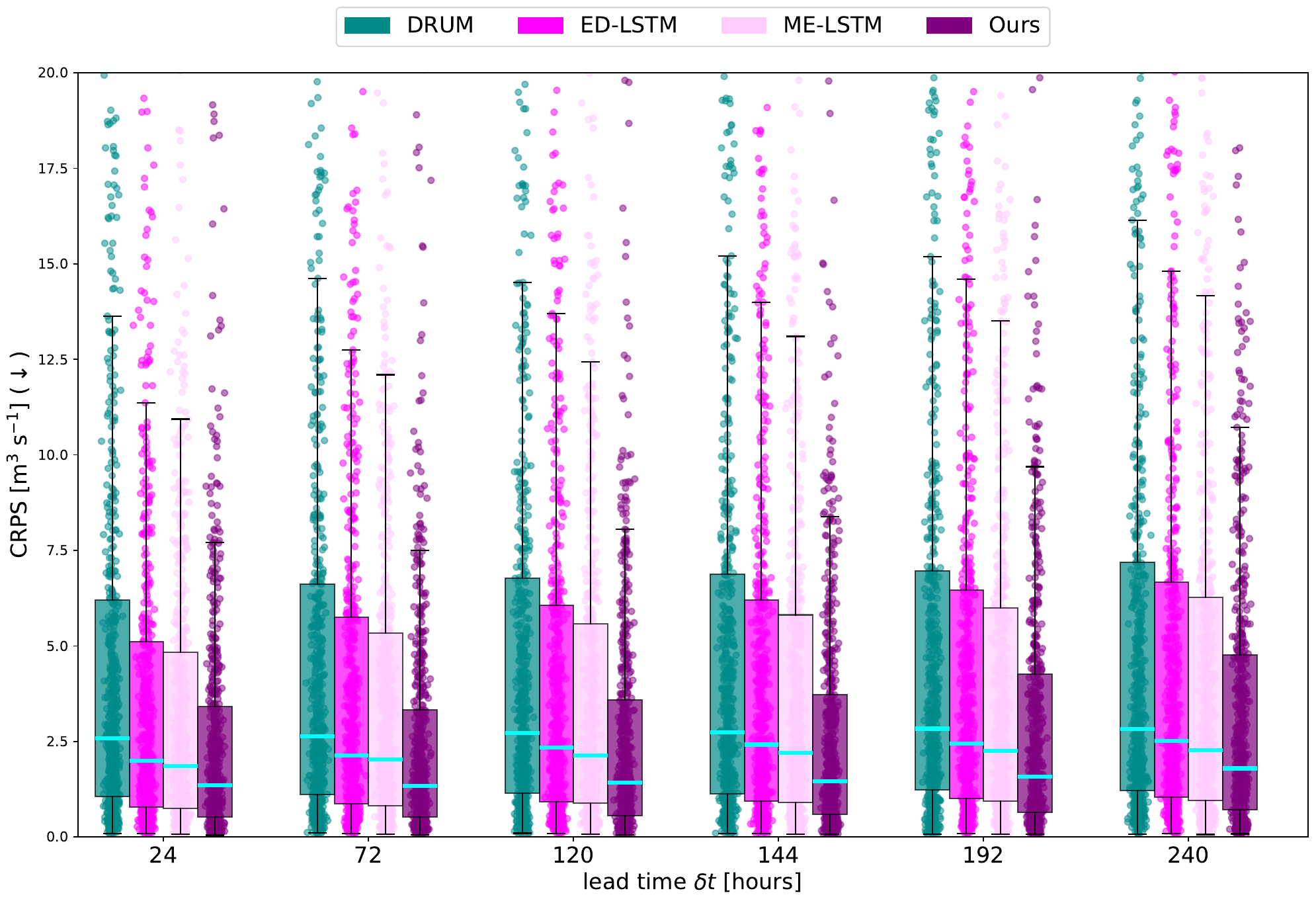}
  \caption{CRPS of the river discharge forecasting with different lead time on CAMELS observations (val set 2022-2023, temporally out-of-sample). Distribution quartiles are displayed in boxes, and the entire range excluding outliers is displayed in whiskers. The median score for the model is shown by the cyan line in the box.}
  \label{fig:appendix_obs_camels_crps}
\end{figure}

\begin{figure}[h!]
  \centering
  \includegraphics[width=0.9\textwidth]{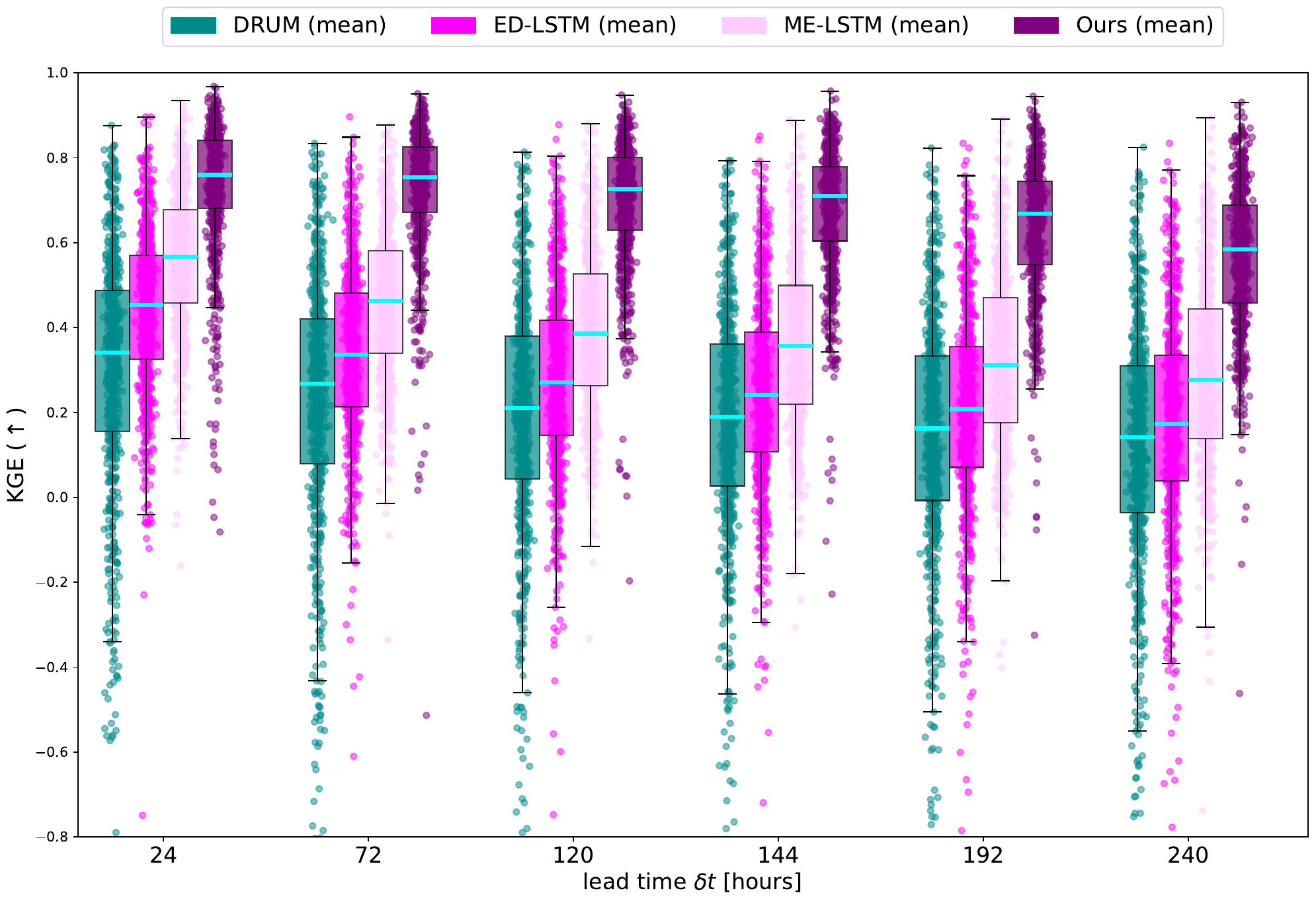}
  \caption{KGE of the river discharge forecasting with different lead time on CAMELS observations (val set 2022-2023, temporally out-of-sample). Distribution quartiles are displayed in boxes, and the entire range excluding outliers is displayed in whiskers. The median score for the model is shown by the cyan line in the box.}
  \label{fig:appendix_obs_camels_kge}
\end{figure}

\begin{figure}[h!]
  \centering
  \includegraphics[width=0.9\textwidth]{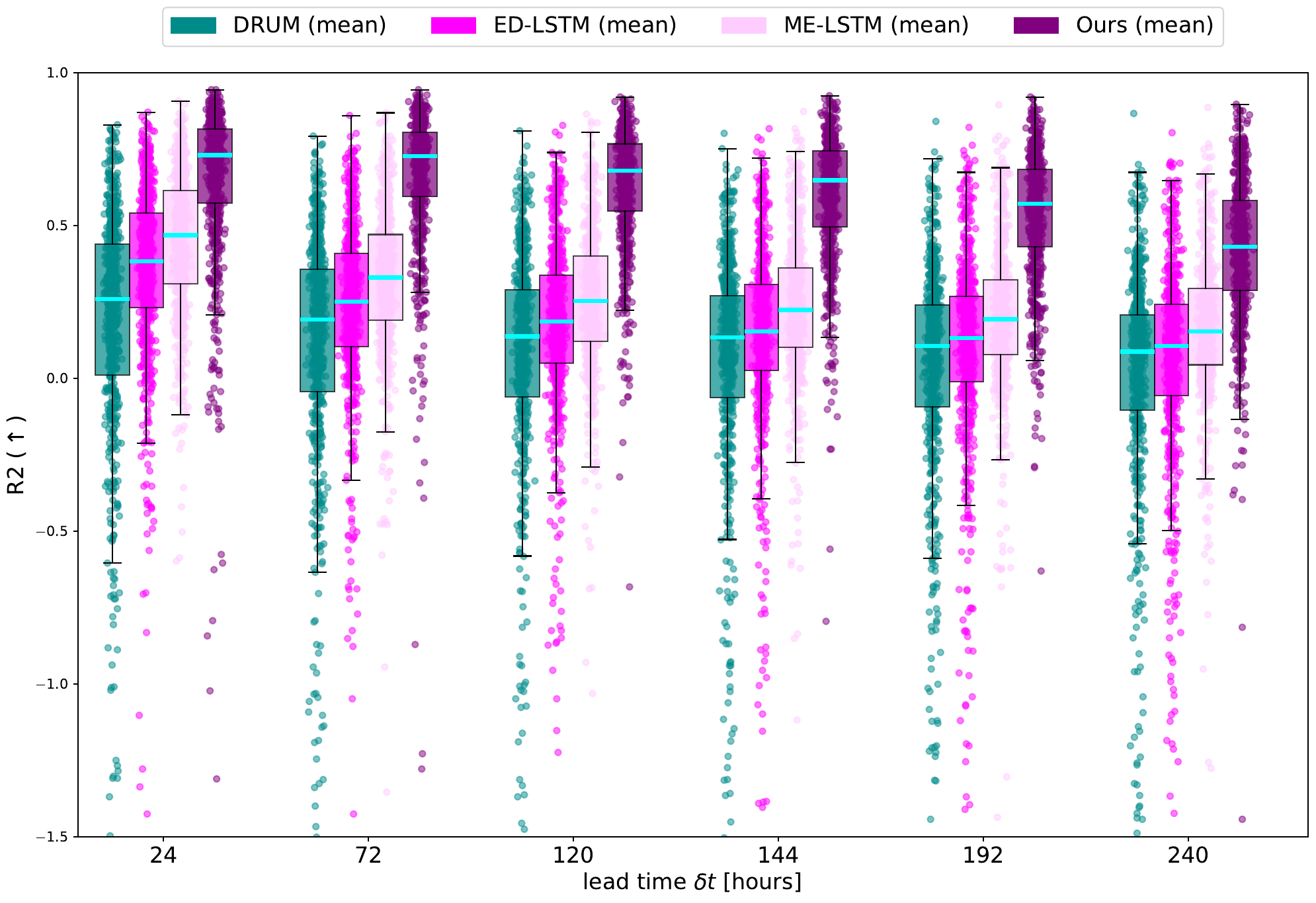}
  \caption{R2 of the river discharge forecasting with different lead time on CAMELS observations (val set 2022-2023, temporally out-of-sample). Distribution quartiles are displayed in boxes, and the entire range excluding outliers is displayed in whiskers. The median score for the model is shown by the cyan line in the box.}
  \label{fig:appendix_obs_camels_r2}
\end{figure}

\begin{figure}[h!]
  \centering
  \includegraphics[width=0.99\textwidth]{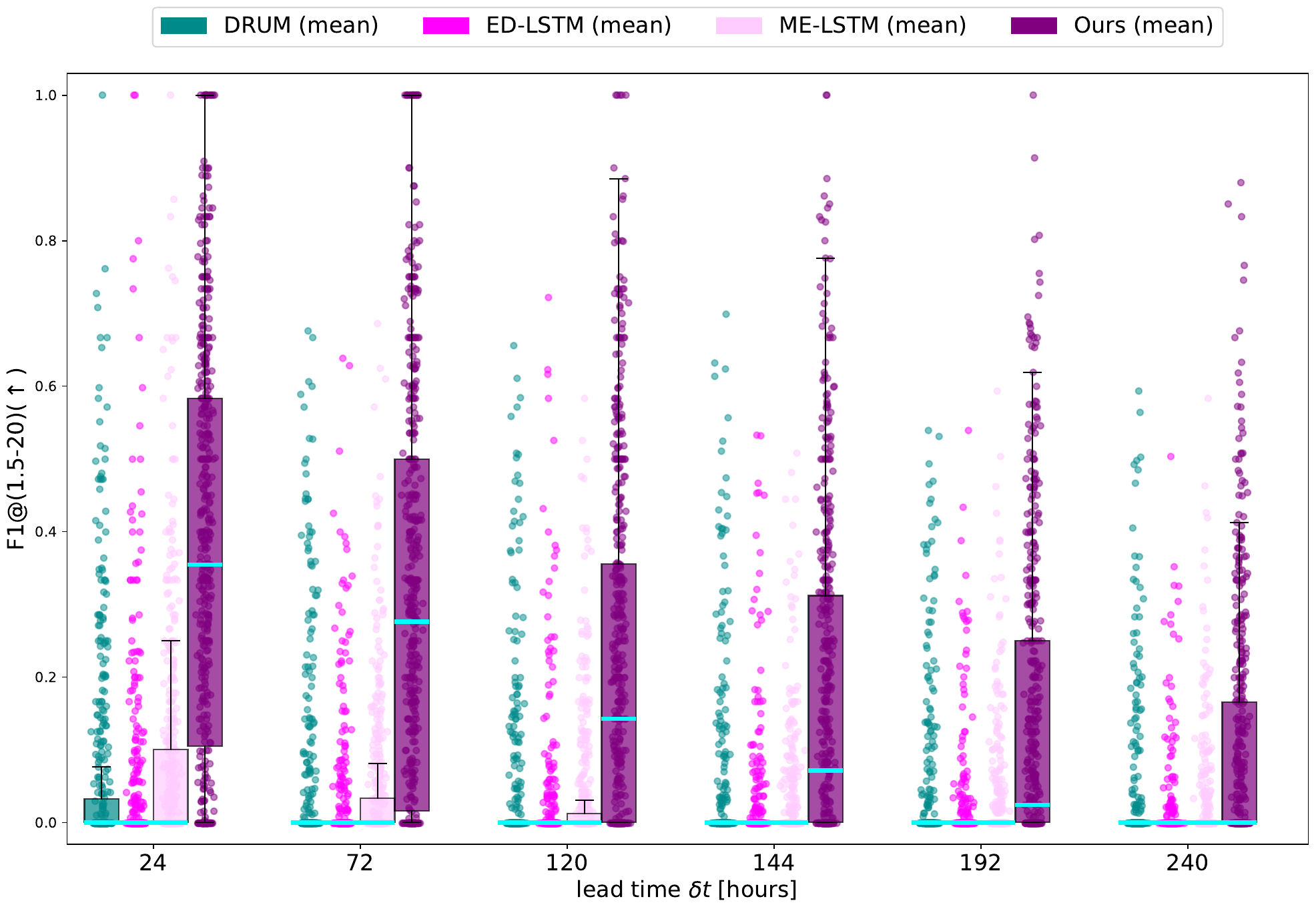}
  \caption{Mean F1@(1.5-20) of the river discharge forecasting with different lead time on CAMELS observations (val set 2022-2023, temporally out-of-sample). Distribution quartiles are displayed in boxes, and the entire range excluding outliers is displayed in whiskers. The median score for the model is shown by the cyan line in the box.}
  \label{fig:appendix_obs_camels_f1}
\end{figure}

\begin{figure}[h!]
  \centering
  \includegraphics[width=0.98\textwidth]{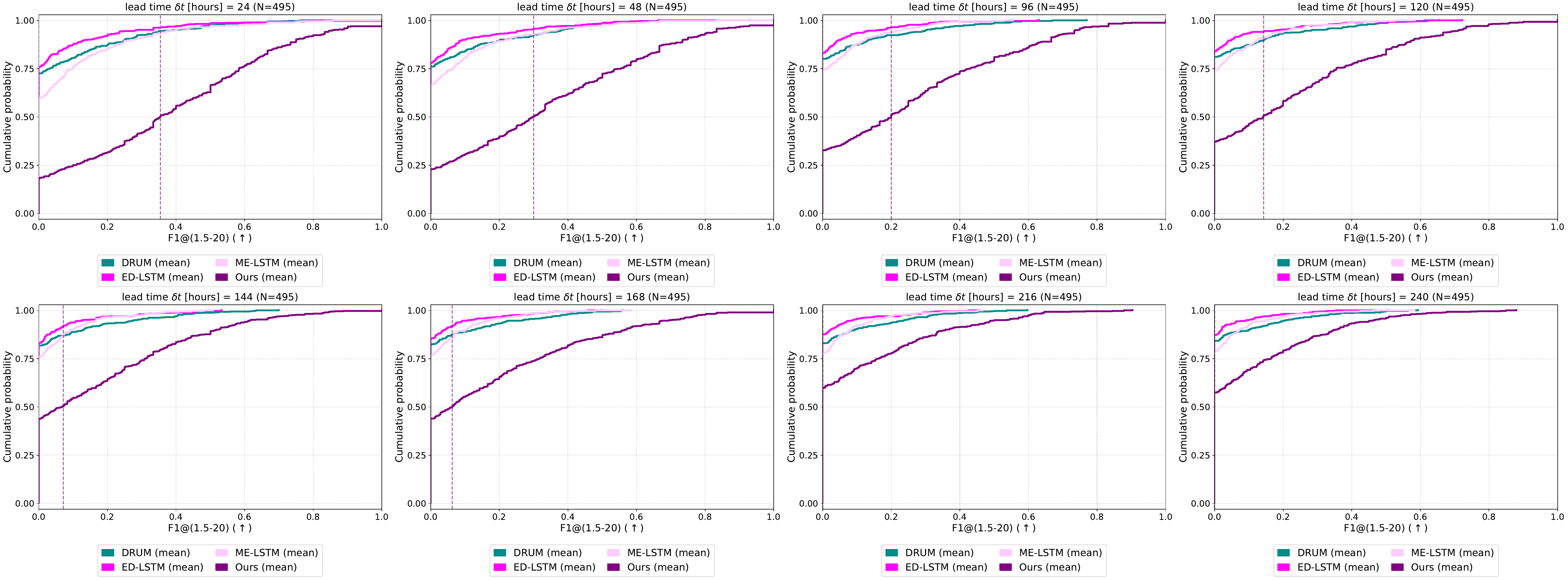}
  \caption{Cumulative distribution function (CDF) of station-level F1@(1.5-20) on CAMELS observations (N=715), across lead times. Dashed vertical lines shows each model's media.}
  \label{fig:appendix_obs_camels_cdf_f1}
\end{figure}

\begin{figure}[h!]
  \centering
  \includegraphics[width=0.98\textwidth]{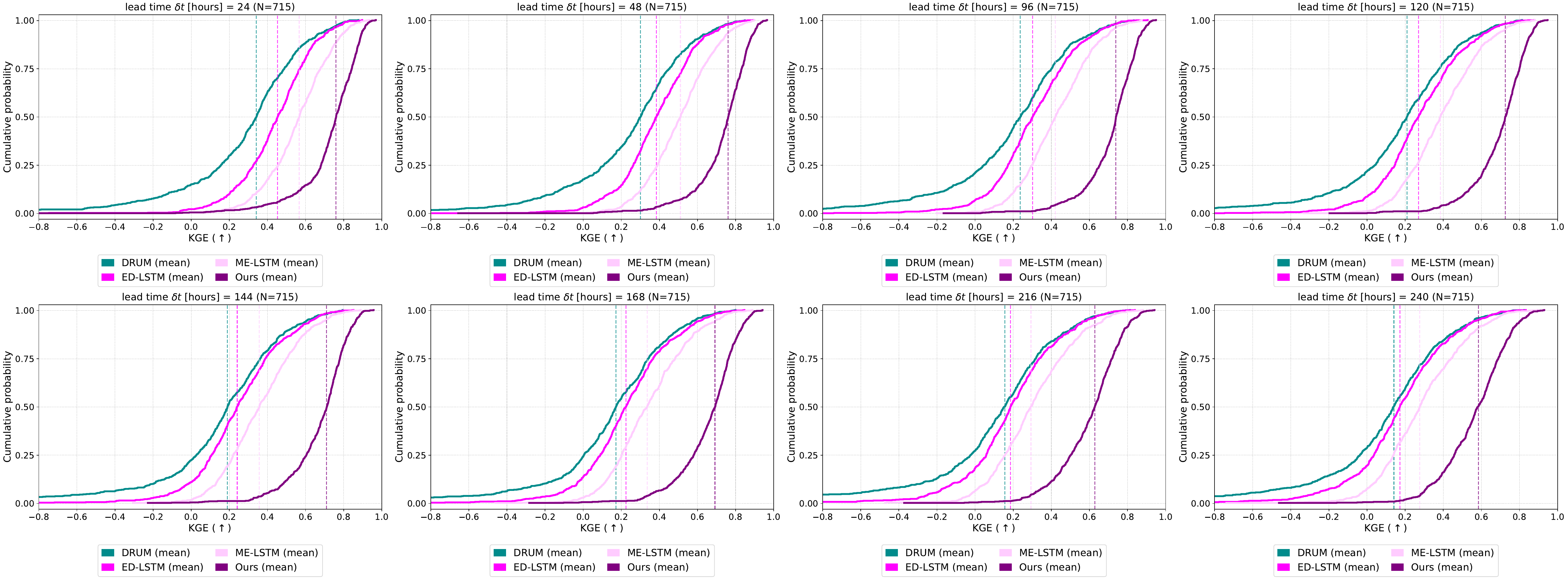}
  \caption{Cumulative distribution function (CDF) of station-level KGE on CAMELS observations (N=715), across lead times. Dashed vertical lines shows each model's media.}
  \label{fig:appendix_obs_camels_cdf_kge}
\end{figure}

\begin{figure}[h!]
  \centering
  \includegraphics[width=0.98\textwidth]{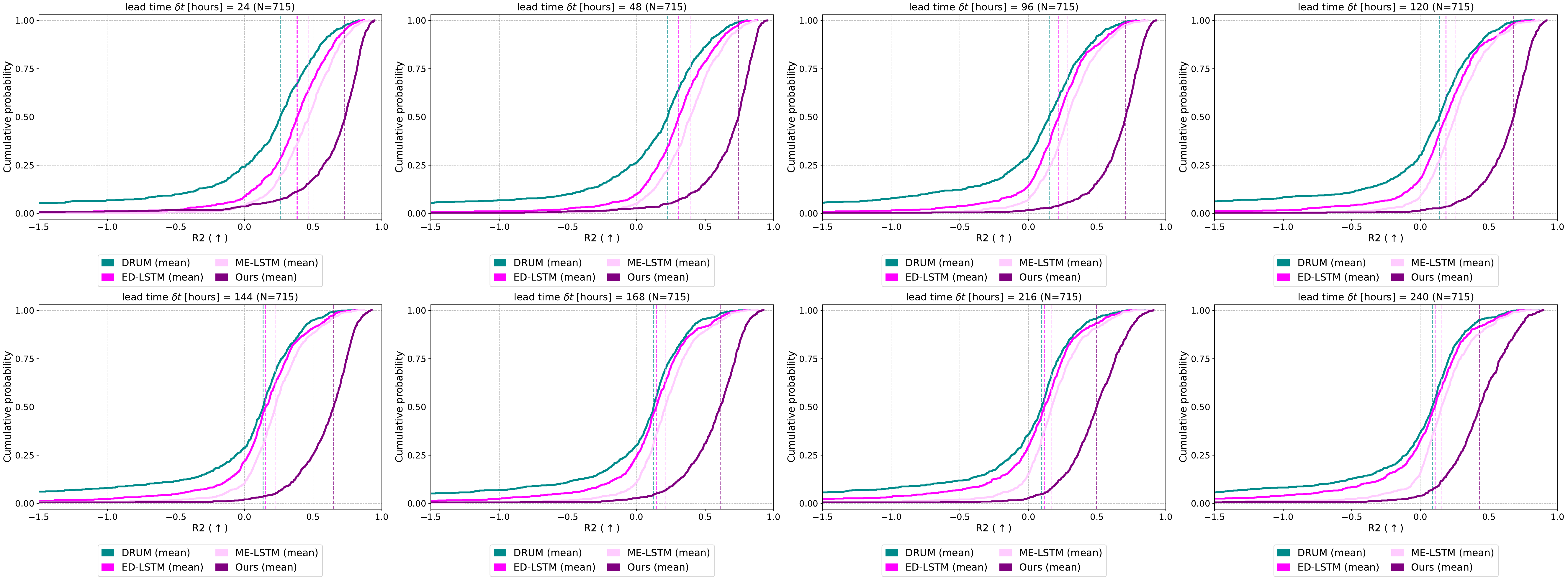}
  \caption{Cumulative distribution function (CDF) of station-level R2 on CAMELS observations (N=715), across lead times. Dashed vertical lines shows each model's media.}
  \label{fig:appendix_obs_camels_cdf_r2}
\end{figure}

\clearpage
\newpage

\section{Implementation details}
\label{sec:appendix_implementation_details}

\begin{figure}[!h]
  \centering
  \includegraphics[width=.98\textwidth]{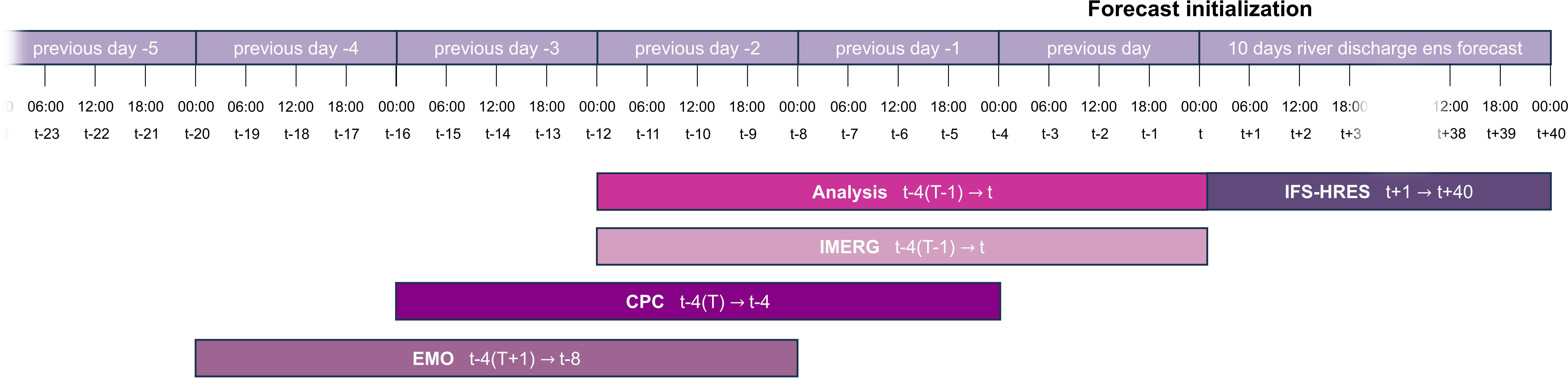}
  \caption{Overview of the input dynamic data to the model. We use $4$ daily time steps as input for the initial condition ($T=4$). For example, CPC data has $1$ day latency ($t-4$) so we take 4 data samples at $t-4$, $t-8$, $t-12$, and $t-16$, where $t$ is the 6-hourly time step. Note that we ignore here the $4$ hours latency for IMERG since we use the daily mean. In an operational setting, step $t$ daily averaged for IMERG can be obtained using the mean over 00:00\,UTC$\rightarrow\sim$20:00\,UTC.}\label{fig:appendix_input_data}
\end{figure}

\begin{figure}[!h]
  \centering
  \includegraphics[width=.98\textwidth]{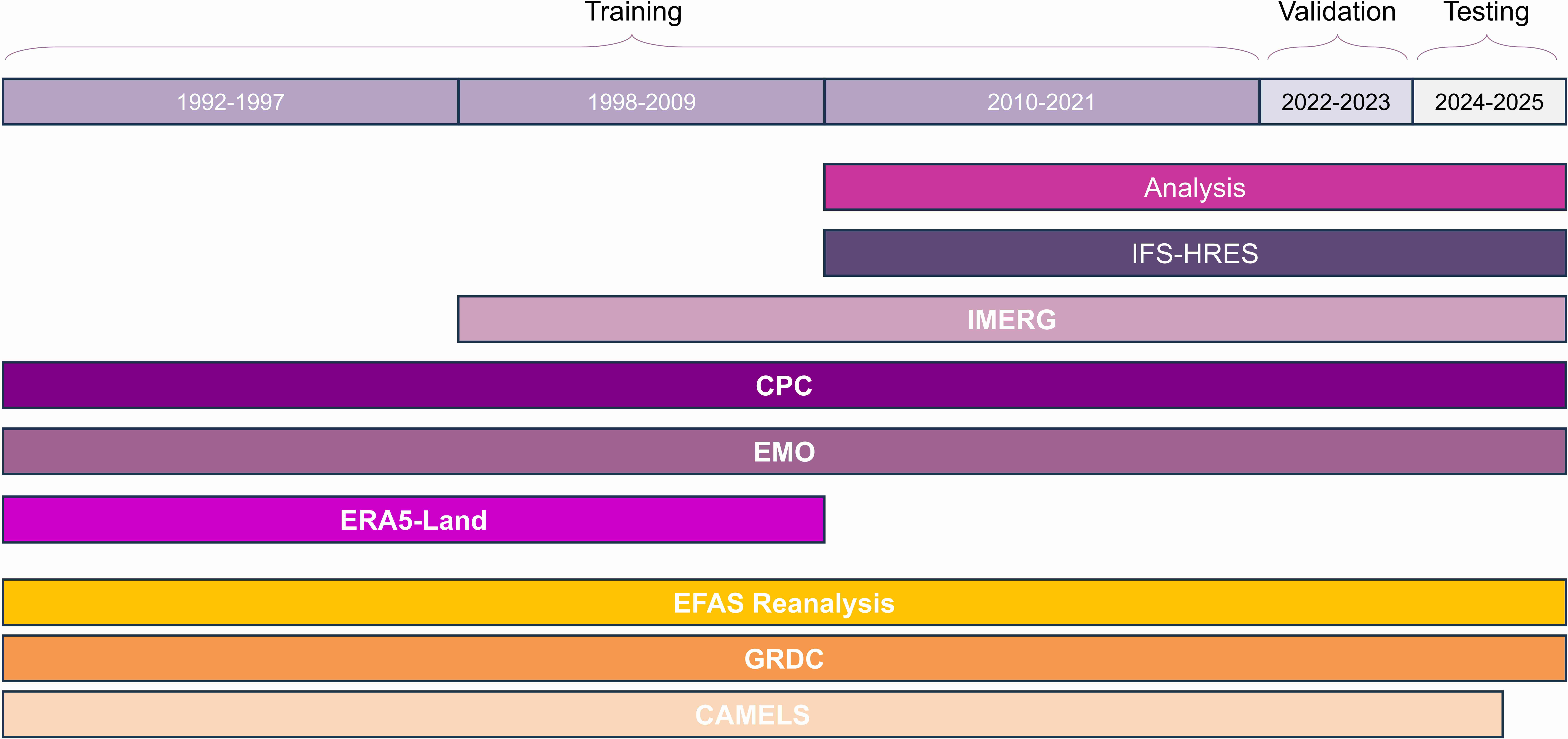}
  \caption{Timeframe and coverage of input dynamic datasets and target streamflow data. Unavailable sources are replaced with ERA5-Land.}\label{fig:appendix_years}
\end{figure}

To make the model compatible with an operational setting, we consider the real-time latency for each source of data. Fig.~\ref{fig:appendix_input_data} gives an overview of the input initial conditions and weather forecast.
We do not use ERA5-LAND for validating or testing. Instead IFS-HRES data is always used. 
The training, validation, and testing splits are shown in Fig.~\ref{fig:appendix_years} along with the timeframe and coverage of input dynamic datasets and target streamflow data. During training, ERA5-Land is used to replace dataset if they are unavailable, e.g., to replace Analysis and IFS-HRES before 2010 and IMERG before 1998. 
All input data are normalized with z-score pre-computed from the training set except CPC, IMERG, and "rg" variable from EMO which are normalized with the log1p transformation. 
Accumulated variables are accumulated every 6-hours to match the output temporal resolution. For training we used bfloat16 mixed floating point precision. Epistemic uncertainty is additionally incorporated as deep ensemble~\citepsupp{NIPS2017_9ef2ed4b} where we train $2$ randomly initialized models. 

We optimize the model gradually on maximum rollout steps of $40$ using the following function: 
\begin{equation}
    n_{rollout} = 1 + \left\lfloor \left( \min\left(1,\ \frac{e}{0.75E}\right) \right)\cdot \big(S - 0.0001\big) \right\rfloor\,,
\end{equation}
where $e$ be the current epoch, $n_{rollout}$ is the number of rollout steps to use at epoch $e$,  $E$ is the total epochs, and $S$ is maximum number of rollout steps. 

\clearpage
\newpage

In Table \ref{table:hyperparameter}, we highlight the main hyperparameters used in training and finetunning the model.

\begin{table}[!h]
\small
 \caption{Implementation details.}\label{table:hyperparameter}
 \centering
 \setlength\tabcolsep{2pt}
 \begin{tabular}{l c c}
 \toprule
 \multirow{2}{*}{Configuration} & Pre-training & Fine-tuning \\
   & (EFAS Reanalysis) & (GRDC/CAMELS) \\
 \midrule
 Optimizer & AdamW & AdamW \\
 Learning rate & 0.0003 & 0.0001 \\
 Minimum learning rate & 0.0001 & 0.0001\\
 Batch size ($B$) & 1 & 1 \\
 Learning rate scheduler & Cosine annealing & Cosine annealing \\
 Weight decay & 0.001 & 0.001 \\
 Training epochs & 20 & 20 \\
 Starting epoch for fine-tuning on IFS-HRES & 11 & 8 \\
 Warmup epochs & 0 & 1 \\
 Gradient clip & 1 & 1 \\
 \midrule
 Input encoder length ($T$) & 4 & 4 \\
 Lead time ($L$) & 40 & 40 \\
 Number of input points $P$ & 200,000 & 200,000$^*$  \\
 \midrule
 Embedding dimension for Analysis & 96 & 96 \\
 Embedding dimension for EMO & 64 & 64 \\
 Embedding dimension for CPC & 16 & 16 \\
 Embedding dimension for IMERG & 16 & 16 \\
 Number of encoder layers & 3 & 3 \\
 Hidden dimension in encoder block ($K$) & 192 & 192 \\
 Depth of encoder layers & [2, 2, 2] & [2, 2, 2] \\
 Curves in encoder layers & \{Sweep\_H, Sweep\_V, & \{Sweep\_H, Sweep\_V,  \\
 & Gilbert, Gilbert trans\} & Gilbert, Gilbert trans\} \\
 Grouping size in encoder block & [(4, 200000), (2, 200000), & [(4, 200000), (2, 200000), \\
 & (1, 200000)] & (1, 200000)] \\
 Dropout in encoder block & 0.1 & 0.1 \\
 D\_state in encoder block & 8 & 8 \\
 D\_conv in encoder block & 3 & 3 \\
 Head dimension in encoder block & 32 & 32 \\
 Chunk size in encoder block & 256 & 256 \\
 MLP ratio in encoder block & 1 & 1 \\
 \midrule
 Hidden dimension in forecast block ($K$) & 192 & 192\\
 Embedding dimension for IFS-HRES ($K_{HRES}$) & 64 & 64 \\
 Number of forecast layers & 1 & 1 \\
 Depth of forecast layers & [2] & [2] \\
 Curves in forecast layers & \{Sweep\_H, Sweep\_V\} & \{Sweep\_H, Sweep\_V\} \\
 Grouping size in forecast block & [(1, 200000)]  & [(1, 200000)] \\
 Dropout in forecast block & 0.1 & 0.1 \\
 D\_state in forecast block & 8 & 8 \\
 D\_conv in forecast block & 3 & 3 \\
 Head dimension in forecast block & 32 & 32 \\
 Chunk size in forecast block & 256 & 256 \\
 Chunk size for routing module & 64 & 64 \\
 MLP ratio in encoder block & 2 & 2 \\
 \midrule
 Hidden dimension in forecasting head & 64 & 64 \\
 Dropout in head & 0.1 & 0.1 \\
 \midrule

 Dimension for noise vector & 32 & 32 \\
 Standard dviation for the noise vector & 0.5 & 0.5 \\
\bottomrule
\multicolumn{3}{l}{\footnotesize{$^*$}For GRDC/CAMELS fine-tuning, we only compute the loss where observations are available.} \\
\end{tabular}
\end{table}

\clearpage
\newpage

\begin{figure}[!h]
  \centering
  \includegraphics[width=.8\textwidth]{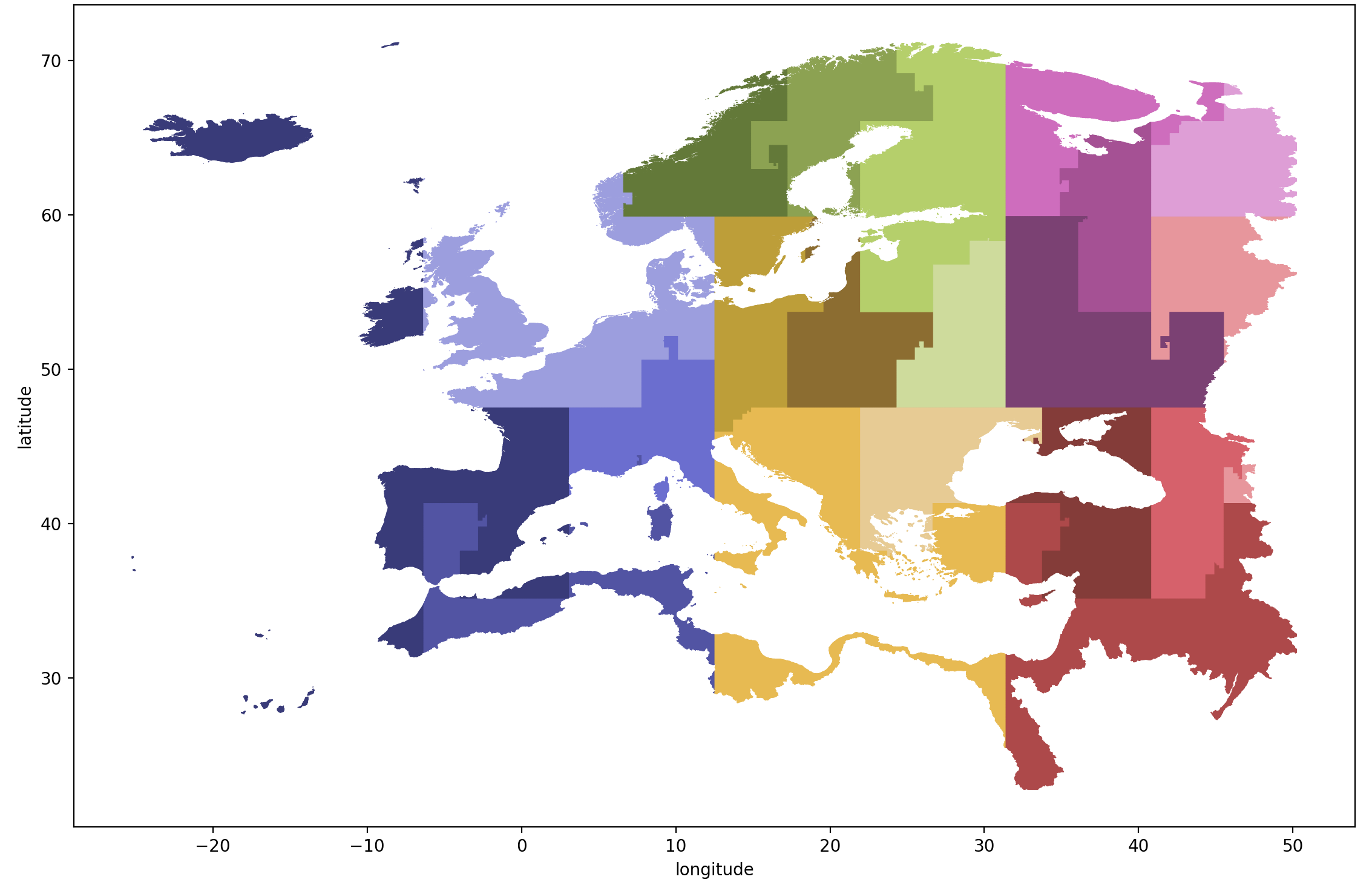}
  \caption{A simplified view of splitting along Gilbert space-filling curve over the EFAS domain. Each split has $200,000$ points. The model runs over these splits and the final results are stitch together to form the prediction over EFAS domain with $5.2$ million points. 
  }\label{fig:appendix_split_curves_over_europe}
\end{figure}

During training and inference, fitting all $\sim 5.2$ million input points of the full European EFAS domain into memory is impractical. Following \citetsupp{RiverMamba}, we order the points in the domain along a Gilbert space-filling curve and partition the curve it into sub-curves of $200,000$ points each. A simplified version of the splits is shown in Fig.~\ref{fig:appendix_split_curves_over_europe}.

\subsection{Computational time}

Training the models on the Mid European domain was done on clusters with NVIDIA A100 $\sim$64\,GB GPUs. The final probabilistic model trained over Europe on a cluster with NVIDIA H200 and NVIDIA B200 GPUs. Pre-training on EFAS reanalysis took about 27 hours on 32 GPUs (8 nodes). CRPS was computed for each node with 4 GPUs and the total loss was averaged across nodes. Fine-tuning on observational GRDC/CAMELS data took about 19.5 hours on 32 GPUs. In Table \ref{table:computational_time}, we report the inference time (seconds) w.r.t. the number of input points for our model. We use one NVIDIA A40 GPU for all runs.

\begin{table}[h!]
  \caption{Inference time in seconds.} \label{table:computational_time}
  \centering
  \tabcolsep=4.5pt\relax
  \setlength\extrarowheight{0pt}
  \begin{tabular}{l*{8}l}
  
    \toprule
      10K & 20K & 40K & 80K & 160K & 300K & 600K \\
    \midrule
    0.0154 & 0.301 & 0.591 & 1.172 & 2.347 & 4.395 & 8.913 \\
    \bottomrule
  \end{tabular}
\end{table}

\subsection{Serialization}
Serialization fixes the spatio-temporal scanning path over the points fed into each Mamba block. Following the space-filling curve concept~\citepsupp{peano1990courbe} in RiverMamba, we define a bijective map $\xi: \ZZ^3 \rightarrow \NN$ that assigns every point a unique
64-bit integer index according to its order along a curve. Sorting points by this index gives the \emph{serialization}, while the inverse $\xi^{-1}: \NN \rightarrow \ZZ^3$ recovers each point's original position, i.e.\ the \emph{deserialization}. 
For simplicity, curves are defined on the 2D PlateCarree projection of the Earth. We consider the, Generalized Hilbert (Gilbert) curve~\citepsupp{hilbert1935stetige} together with Sweep and Zigzag curves in both vertical and horizontal directions.
\begin{figure}[!h]
  \centering
  \includegraphics[width=.9\textwidth]{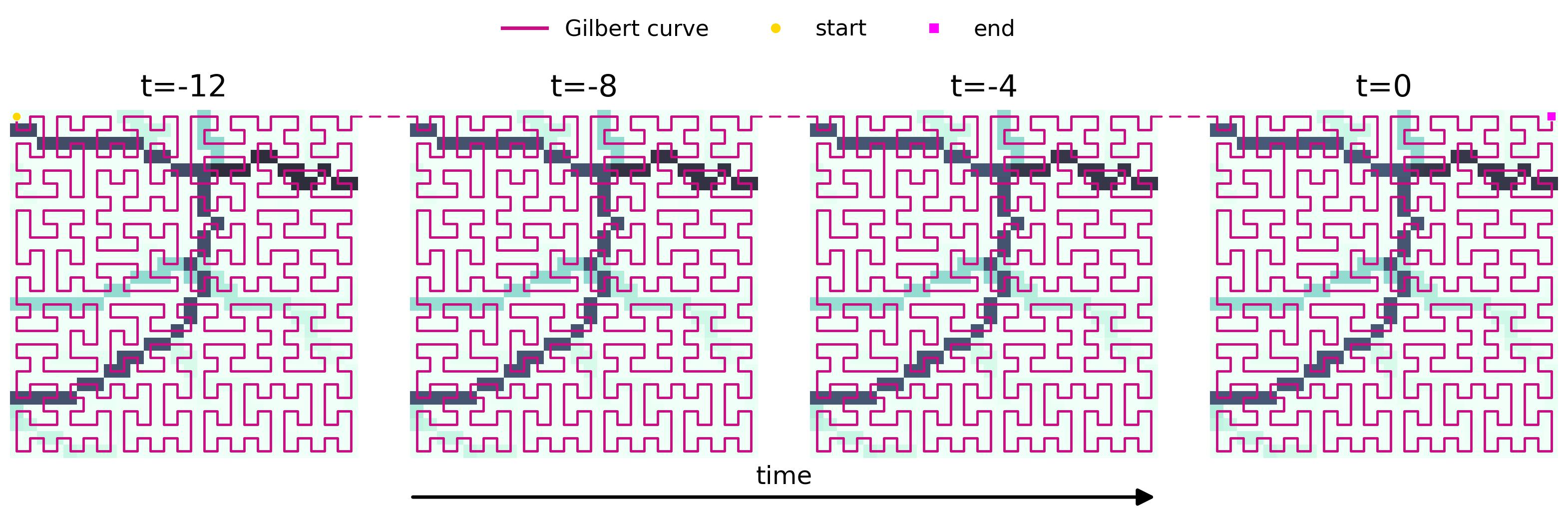}
  \caption{A spatio-temporal scans along $4$ frames in the encoder block using Gilbert serialization.}\label{fig:appendix_curves_blocks}
\end{figure}
We therefore give each encoder block its own curve, cycling through horizontal Sweep, vertical Sweep, Gilbert, and transposed Gilbert across successive blocks, with a block's spatial curve linked over time by joining its last point at $t$ to its first point at $t{+}1$ (Fig.~\ref{fig:appendix_curves_blocks}). Varying the curve per block enables the model to scan the sampled points from diverse spatial perspectives, capturing different contextual features.

\subsection{Location-aware Adaptive Normalization (LOAN)}
In order to condition the model on static attributes $\XALPHA$, the location-aware adaptive normalization layer (LOAN)~\citepsupp{LOAN} modulates the features $\X$ systematically within the encoder block:
\begin{align}
    \text{LOAN}(\X) = \left(\frac{\X}{\text{RMS}(\X)}\right) ~+~ \text{GELU}(\text{Linear}(\XALPHA)) \,, \label{eq:06}
\end{align}
where $\text{RMS}(\X) = \sqrt{\frac{1}{K}\sum_{k=1}^{K} x_{(k)}^2}$. A linear layer projects $\XALPHA \in \RR^{B, 1, P, V_{s}}$, with $V_{s}=64$ being the number of embedding channels, to $\RR^{B, 1, P, K}$, which is then broadcast along the temporal dimension to match $\X \in \RR^{B, T, P, K}$, the input to the LOAN layer.
$\text{RMS}(\X)$ is likewise broadcast across all $K$ channels.
For the first LOAN application within the encoder block only, we skip the RMS normalization (i.e. $\text{RMS}(\X)\!\to\!1$ in Eq.~\ref{eq:06}) and condition directly on the normalized input features $\X$.


\section{Evaluation Metrics}
\label{sec:Appendix_evaluation_metrics}

We evaluate the model performance across five standard metrics in hydrological modeling and flood forecasting. This includes Root Mean Square Error (RMSE), the Coefficient of Determination ($R^2$), Kling–Gupta Efficiency (KGE), the F1 score, and the Continuous Ranked Probability Score (CRPS). Individual metric calculation between observed river discharge ($X^{\text{obs}}$) and predicted river discharge ($X^{\text{pred}}$) for points $P$ are given below:

\textbf{Root Mean Square Error (RMSE).} computes the square root of the mean squared deviations between predicted and observed values:
\begin{equation}
\mathrm{RMSE} = \sqrt{ \frac{1}{P} \sum_{p=1}^{P} (X_{p}^{\text{obs}} - X_{p}^{\text{pred}})^2 } \,.
\end{equation}

\textbf{Coefficient of Determination (R2).} Measures a model's explanatory power relative to the mean of observed data. In hydrological applications, R2 has the same meaning as Nash–Sutcliffe Efficiency (NSE). Values close to 1 reflect near-perfect agreement, whereas negative values indicate that the model performs worse than the mean:
\begin{equation}
\text{R2} = 1 - \frac{\sum_{p=1}^{P} (X_p^{\text{obs}} - X_p^{\text{pred}})^2}{\sum_{p=1}^{P} (X_p^{\text{obs}} - \bar{X}^{\text{obs}})^2} \,.
\end{equation}

\textbf{Kling–Gupta Efficiency (KGE).} Is a combination of correlation, bias, and variability. Values close to 1 indicate high predictive accuracy, while scores below 0 indicate performance inferior to the observed mean:

\begin{equation}\label{eq:KGE}
\mathrm{KGE} = 1 - \sqrt{ (r - 1)^2 + (\beta - 1)^2 + (\gamma - 1)^2 } \,,
\end{equation}
\vspace{0.3cm}
\begin{equation}
\beta = \frac{\bar{X}^{\text{pred}}}{\bar{X}^{\text{obs}}}, \quad
\gamma = \frac{\text{CV}^{\text{pred}}}{\text{CV}^{\text{obs}}}, \quad
r = \text{Pearson correlation coefficient} \,,
\end{equation}
where $r$ denotes the Pearson correlation between observed and simulated discharge series, $\beta$ represents the bias ratio, and $\gamma$ is the variability ratio. The respective coefficients of variation are given by $\mathrm{CV}^{\text{obs}} = \sigma^{\text{obs}} / \bar{X}^{\text{obs}}$ and $\mathrm{CV}^{\text{pred}} = \sigma^{\text{pred}} / \bar{X}^{\text{pred}}$, with $\sigma^{\text{obs}}$ and $\sigma^{\text{pred}}$ are the standard deviations of observed and predicted streamflow.

\textbf{Continuous Ranked Probabilistic Score (CRPS).} Is used for probabilistic forecast. It measures the deviation for each ensemble member from the observations using mean absolute error in the first term and measures the spread of the ensemble members in the second term:
\begin{align}
    \text{fCRPS} =
    \frac{1}{P} \sum_{p=1}^{P}
    \left[
    \begin{aligned}
        &\frac{1}{M} \sum_{m=1}^{M} \left| X_p^{\text{pred}} - X_p^{\text{obs}} \right| - \frac{1}{2M(M-1)} \sum_{m,m'=1}^{M} \left| X_p^{\text{pred}} - X_p^{\text{obs}} \right|
    \end{aligned}
    \right] \,,
\end{align}
where $M$ is the number of ensemble members. During evaluation, we set $M=12$ ensemble members for all probabilistic models. 

\textbf{F1 Score.} Represents the harmonic mean of precision and recall. It can serve as a robust benchmark for imbalanced prediction tasks such as detecting less frequent flood events:

\begin{equation}\label{eq:F1}
\mathrm{F1} = 2 \cdot \frac{\mathrm{Precision} \cdot \mathrm{Recall}}{\mathrm{Precision} + \mathrm{Recall}} \,.
\end{equation}

These metrics are calculated independently for each point's time series before being aggregated to determine the final score.
For RMSE, R2, KGE, and CRPS, we always report the median for all points, while for F1 score, we report the mean over all points. all scores are first aggregated per lead time forecast e.g., from $t=0$ to $t=40$ and then averaged to get the final score for all lead times. Categorical metrics, such as the F1 score, evaluate event detection performance across varied return intervals. For instance, when streamflow exceeds a 2-year flood threshold on a given day, an accompanying forecast above that threshold is marked as a true positive. Unless noted otherwise, all reported F1 scores reflect averages taken across return periods 1.5 to 20 years.

\section{Floods return periods}
\label{sec:Appendix_return_periods}

The return period (or recurrence interval) estimates how often a hydrological flood event of a given magnitude occurs on average. Here we apply it to flood frequency. A 2-year flood, for instance, has a 50\% annual exceedance probability (AEP). Return period ($RP$) is simply the inverse of AEP: $RP = \frac{1}{AEP}$.

These return periods are commonly used to set discharge thresholds for flood warnings (e.g., triggering an alert once discharge exceeds the 2-year level). Depending on context, a high return period (e.g., 20 years) does not necessarily mean flooding will occur but it reflects the statistical rarity in streamflow magnitude. We therefore treat return period as a proxy for hydrological extremity, which we call flood severity.

Following the EFAS approach, we define thresholds for return periods of 1.5, 2, 5, 10, 20, 50, 100, 200, and 500 years. For EFAS reanalysis, return levels are estimated by fitting a Gumbel distribution (L-moments method) to annual maxima. We use the pre-computed return period data from the Copernicus Emergency Management Service (\colorh{\url{https://confluence.ecmwf.int/display/CEMS/Auxiliary+Data}}). Figs.~\ref{fig:Appendix_hydrograph_grdc} and \ref{fig:CAMELS_CH_2063} show the flood thresholds. For the GRDC and CAMELS observational datasets, return periods are computed per station using each station's full available record. We did not compute return periods for the ML reforecasts, as this would require a long reforecast climatology to fit an extreme value distribution.

\section{Baselines}

\subsection{Climatology}
\label{sec:Appendix_baselines_climatology}

Following \citetsupp{GloFAS_Forecast} and \citetsupp{RiverMamba}, we compute the climatological baseline using the long-term river discharge record for the years 1992 to 2025. For each day of the year, climatology was computed over a 31-day centered moving window, from which we derived 11 fixed percentiles at 10\% intervals. Climatology can be used as a probabilistic baseline with 11 ensemble members (11 percentiles).

\subsection{Persistence}
\label{sec:Appendix_baselines_persistence}

Persistence is defined as the daily river discharge of EFAS reanalysis from the day preceding the day on which the forecast was issued, i.e., for a forecast starting at 00:00 UTC $t$, the persistence is defined as the river discharge one day before at 00:00 UTC $t-4$. 
The prediction remains constant across all lead-time steps. Note that this baseline has an advantage over the deep learning models because it has access to the values of the river discharge, while deep learning models need to estimate the magnitude of discharge from initial conditions and weather forecast.

\subsection{EFAS Forecast}
\label{sec:Appendix_baselines_efas_forecast}

Operational forecast from EFAS v5.0 was obtained from the ECMWF Early Warning Data Store (EWDS) \colorh{\url{https://doi.org/10.24381/cds.9f696a7a}}. 
These operational real-time data are provided by the Copernicus Emergency Management Service (CEMS), which is managed and developed by the European Commission's Joint Research Centre. EFAS forecasts are generated by forcing the LISFLOOD hydrological model~\citepsupp{LISFLOOD} with meteorological data from the ECMWF ensemble forecast (ENS). 
The deterministic EFAS (control run) does not use perturbation, while the probabilistic version has $51$ ensemble members via perturbation. We used the high-resolution EFAS v5.0 forecast from 29 May 2024 to the end of 2025 for the deterministic control run and data from 01 June 2024 to 31 August 2024 for the probabilistic forecast. 
We compare to this baseline in Sec.~\ref{sec:results_obs} and Sec.~\ref{sec:appendix_comparison_efas}.

\subsection{RiverMamba}
\label{sec:Appendix_rivermamba_baseline}

We adapted RiverMamba from~\citetsupp{RiverMamba}. The baseline relies on Mamba~\citepsupp{mamba} as a State Space Model (SSM) to effectively capture spatio-temporal relations in large-scale river networks through scanning on space-filling curves. The RiverMamba baseline has no explicit
routing mechanism inside the model. It relies mainly on the spatio-temporal modeling of SSM without considering local drain direction. 
We followed~\citetsupp{RiverMamba} and the official implementation from \colorh{\url{https://github.com/HakamShams/RiverMamba_code}} to define the main hyper-parameters. RiverMamba was only trained over Mid European domain for 16 lead time. Since RiverMamba has to build a dedicated forecasting layer for each step, we used hidden dimension of 128 for the encoder and forecasting layers to reduce overfitting and computations. For optimization, we pretrained for 20 epochs on reanalysis and fine-tune for 10 epochs on observations. We compare to this baseline in Sec.~\ref{sec:results_reanalysis} and \ref{sec:appendix_rivermamba}.

\subsection{DRUM}
\label{sec:Appendix_drum}

This is a probabilistic and diffusion-based baseline adapted from \citetsupp{DRUM_2025}. DRUM trains a conditional and an unconditional noise-prediction network jointly for classifier-free guidance and samples forecasts via DDIM. 
We used the official repository from \colorh{\url{https://github.com/ozg2021/DRUM/tree/main}} to train the model. The model is trained on single-step nowcasting and 40 forecast steps at 6-hourly resolution using denoising jointly in one forward pass. We used 1200 diffusion steps (linear $\beta$ schedule, $\beta_1=10^{-4}$, $\beta_T=0.02$), 120 step DDIM, hidden size of 256, and time embedding dimension of 128.  For optimization, we used AdamW with $10^{-4}$ learning rate and $10^{-2}$ weight decay for 200 epochs on EFAS reanalysis and for 50 epochs for fine-tuning on observation.

\subsection{ED-LSTM (Encoder-Decoder LSTM}
\label{sec:Appendix_ed_lstm}

This is a probabilistic baseline from Google's operational flood forecasting system~\citepsupp{Google_nature}. We followed the implementation in NeuralHydrology (\colorh{\url{https://github.com/neuralhydrology/neuralhydrology}}). In ED-LSTM, a bidirectional hindcast LSTM encodes the meteorological forcing initial conditions, and its final hidden and cell states pass through a learnable handoff network to initialize a unidirectional forecast LSTM that decodes the forecast horizon. 
The probabilistic forecast is generated via a Mixed Asymmetric Laplace (CMAL) with $K{=}3$ components. The model is trained on a single-step nowcasting and 40 forecast steps at 6-hourly resolution. We used a hidden size of 256 for both LSTMs and a state-handoff network with hidden layers $[512, 256]$. For optimization, we used Adam with $10^{-3}$ learning rate for 35 epochs on EFAS reanalysis and 20 epochs for fine-tuning on observations.

\subsection{ED-LSTM (Mean Embedding LSTM)}
\label{sec:Appendix_mean_embedding}

This is the new probabilistic baseline from Google's operational flood forecasting system adapted from \citetsupp{Google_neu}. We followed the implementation in NeuralHydrology (\colorh{\url{https://github.com/neuralhydrology/neuralhydrology}}).
The probabilistic forecast is generated via a Mixed Asymmetric Laplace (CMAL) with $K{=}3$ components. The model is trained on a single-step nowcasting and 40 forecast steps at 6-hourly resolution. We used a hidden size of 256 for both LSTMs, output dropout of 0.4, and an initial forget-gate bias of 3. For optimization, we used Adam with $5\times10^{-4}$ learning rate for 35 epochs on EFAS reanalysis and 20 epochs for fine-tuning on observation.

\section{Dataset}
\label{sec:Appendix_dataset}

\subsection{EFAS reanalysis river discharge data}
\label{sec:Appendix_EFAS_data}

\begin{figure}[!h]
  \centering
  \includegraphics[width=.98\textwidth]{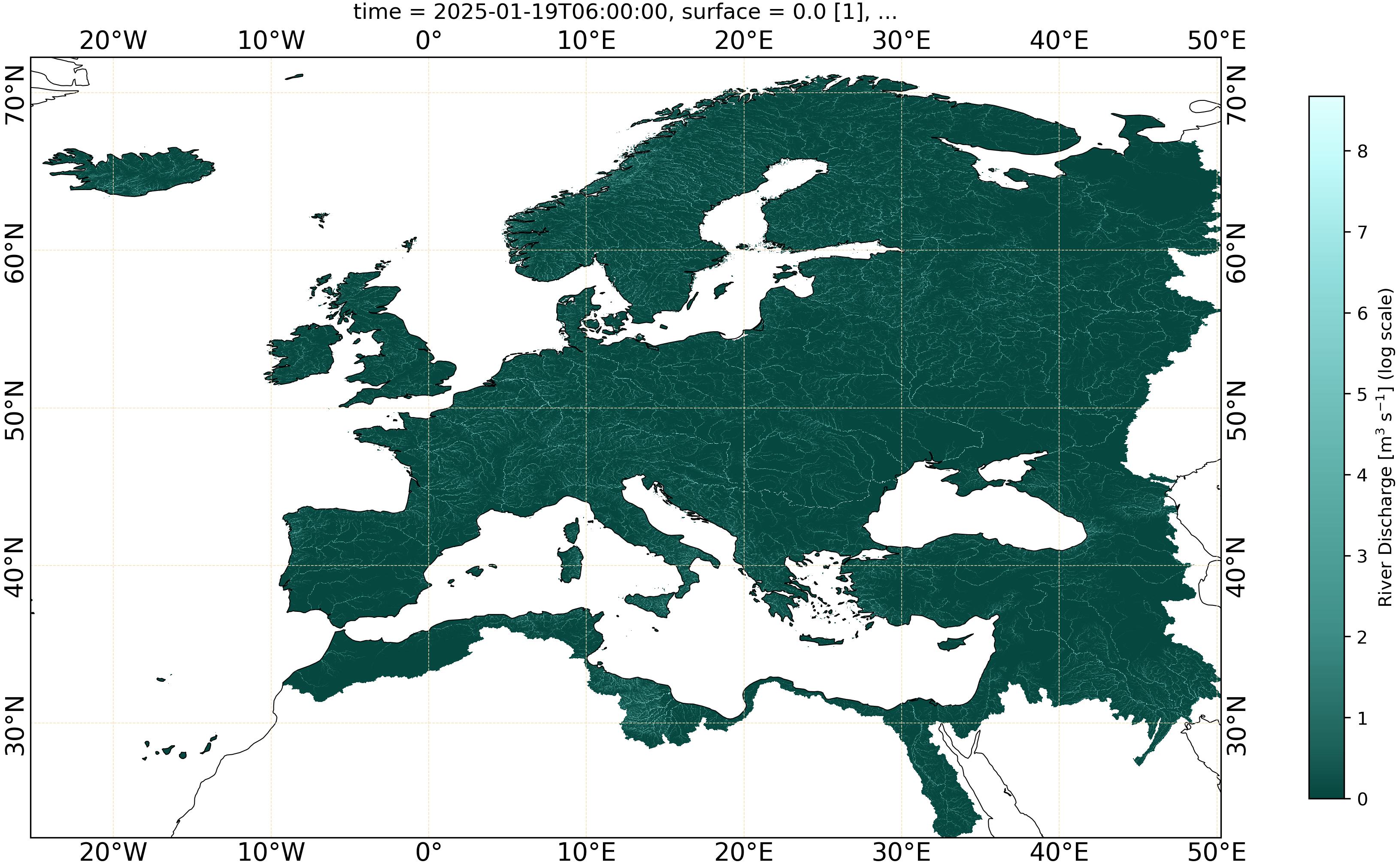}
  \caption{Overview of the EFAS domain.}\label{fig:efas_domain}
\end{figure}

The European Flood Awareness System (EFAS)~\citesupp{EFAS} provides consistent representation of hydrological variables across Europe with sub-daily temporal resolution and high spatial resolution of $0.017^\circ$ ($1$-arcminute $\sim1.5$~km at EFAS latitudes) compared to $0.05^\circ$ in the Global Flood Awareness System (GloFAS)~\citepsupp{GloFAS_Reanalysis} (Fig.~\ref{fig:efas_domain}). 
The system is fully operational and developed by the European
Commission’s Joint Research Centre (JRC) while operated by ECMWF under the Copernicus Emergency Management Service (CEMS). It delivers river discharge reanalysis from $1992$ to present covering the European domain ($72^\circ$\,N-$24^\circ$\,S, $35^\circ$\,W-$74^\circ$\,E) on the regular latitude-longitude grid. The reanalysis is generated using the LISFLOOD hydrological and kinematic wave equations for river routing~\citepsupp{LISFLOOD} to simulate realistic river discharge (dis06, in m$^{3}$ s$^{-1}$) across Europe. The sub-daily EFAS reanalysis discharge data represents the mean value of river discharge for the previous $6$ hours in UTC as a reference time. 

\begin{figure}[!h]
  \centering
  \includegraphics[width=.5\textwidth]{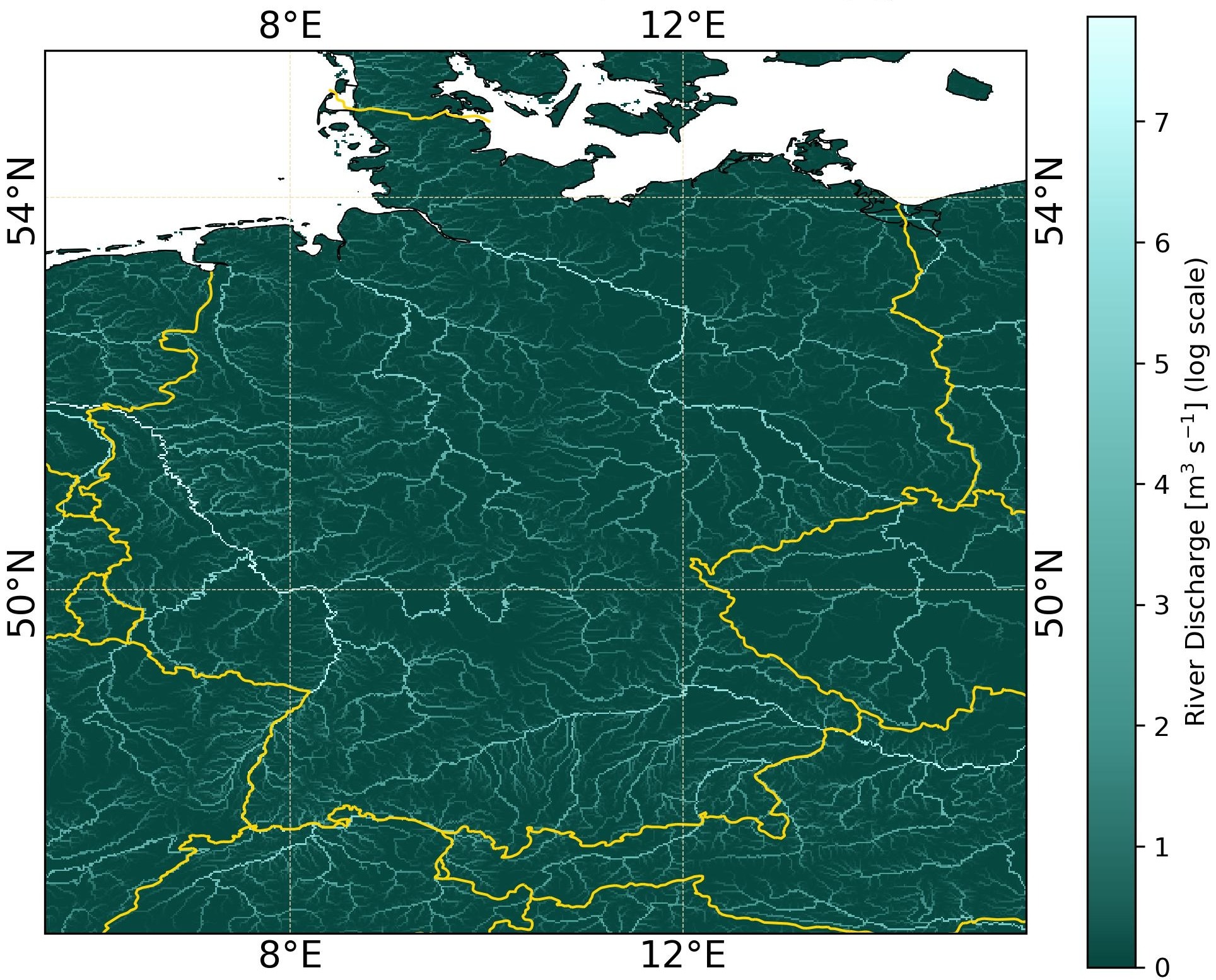}
  \caption{Overview of the Mid European domain as a subset of EFAS domain. We used this domain for the ablation studies.}\label{fig:efas_domain_germany}
\end{figure}

Fig.~\ref{fig:efas_domain} shows the full domain of EFAS which we used for the main experiments and Fig.~\ref{fig:efas_domain_germany} shows the domain we defined over Mid Europe for the ablations and other experiments on Mid Europe. The full domain has an image size of $2970 \times 4530 $ and $\sim5.2$ million points over land, while the domain over Mid Europe has $540 \times 600$ and around $\sim281K$ points over land.
In our work, we used the latest version v5.0 of EFAS and we only used the river discharge "dis06" as a target variable for the forecast (no data from EFAS is used as initial condition or input to the model in our work). 
The dataset is publicly available on Climate Data Store and Early Warning Data Store (EWDS) \colorh{\url{https://doi.org/10.24381/cds.e3458969}}. 

\subsection{Observational river discharge data}

No streamflow data were used as inputs to our model. They were only used as targets for fine-tuning and evaluation. Note that we fine-tuned on each observational dataset individually. In the following, we describe GRDC and CAMELS observational datasets that we used in our study.

\subsubsection{GRDC}
\label{sec:Appendix_GRDC_data}

Observed river discharge is obtained from the Global Runoff Data Centre (GRDC) which provides quality controlled discharge measurements from stations worldwide. The GRDC Data is given as daily time series and it is publicly available at \colorh{\url{https://grdc.bafg.de/}}. Every GRDC daily value is timestamped at 00:00 at the start of the day (left-labeled). Following \citetsupp{RiverMamba} and \citetsupp{EFAS}, we discarded catchments with a drainage area under $150$ km$^2$, leaving $2{,}158$ GRDC stations after matching them to the EFAS drainage network (Sec.~\ref{sec:Appendix_EFAS_data}). This avoids large mismatches between the drainage area reported by GRDC and the one provided by the LISFLOOD EFAS static layers. Each station was geo-located by projecting it onto the EFAS grid, evaluating its $9$ nearest grid cells, and keeping the cell with the highest KGE. Stations whose matched cell still differed from the reported drainage area by more than $20\%$ and scored below $0.4$ KGE were then dropped. This left $1{,}971$ stations for evaluation across the EFAS domain. Over this final set, the median drainage-area mismatch fell to $2.4\%$ (interquartile range $0.75\%$–$6.05\%$). Discharge values are reported daily in m$^3$ s$^{-1}$ and were converted from local time to 00:00 UTC via linear interpolation. For evaluation, the originally left-labeled GRDC series were converted to right-labeled series so they align temporally with the right-labeled EFAS simulation output. F1 scores for the GRDC evaluation are computed over detection windows synchronized between model predictions and observations.

\begin{figure}[!h]
\centering
\includegraphics[width=.99\textwidth]{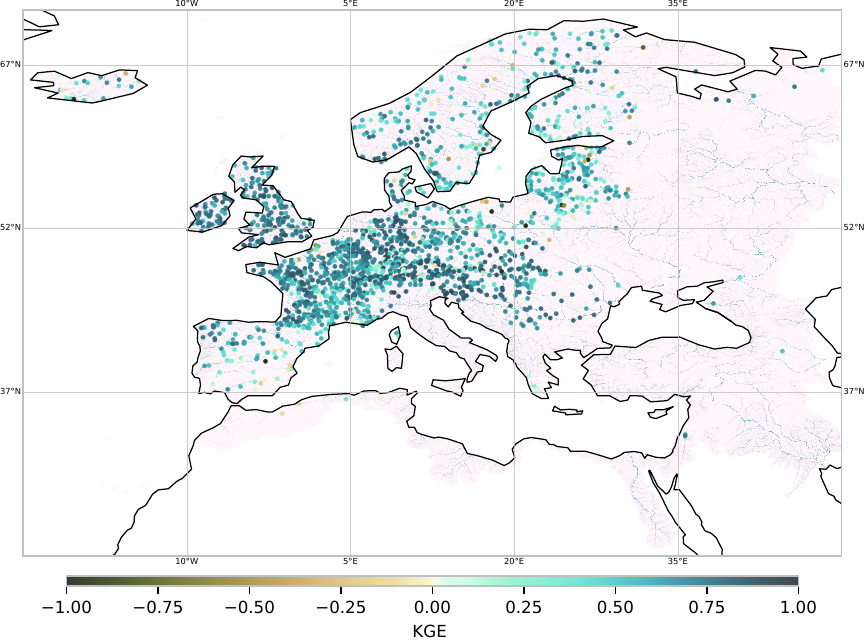}

\caption{Locations of the $1{,}971$ selected GRDC stations used to train and evaluate the model. Color encodes the KGE of the EFAS reanalysis discharge against the GRDC observations at each station.}\label{fig:GRDC_stations}
\end{figure}

\begin{figure}[!h]
  \centering
  \includegraphics[width=.99\textwidth]{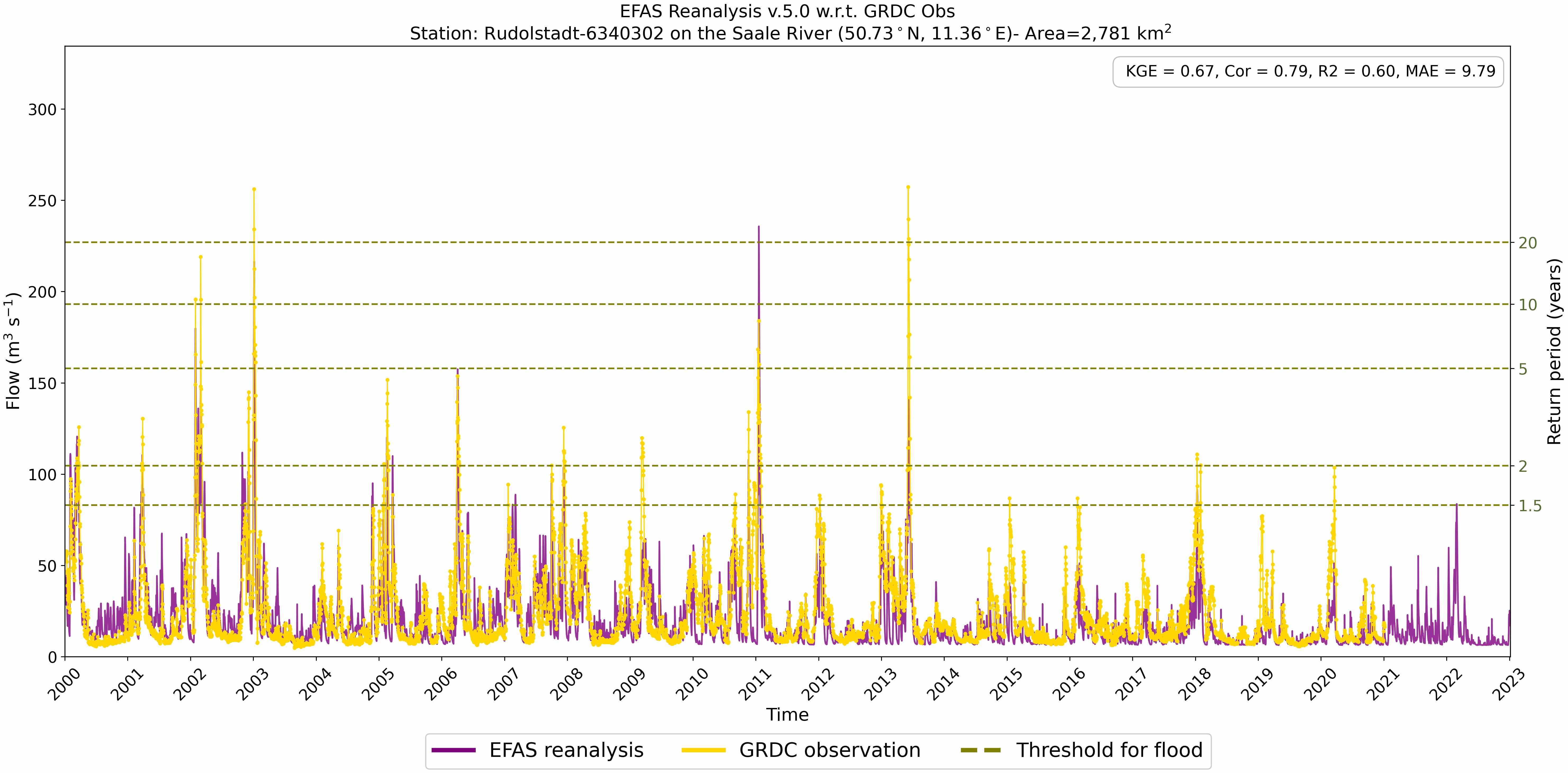}
  \caption{Hydrograph for EFAS reanalysis (purple line) from 1 January 2000 to 31 December 2022 and observations (gold line), for a gauging station on the Saale River (Rudolstadt, GRDC-6340302). The top-right box displays summary statistics from the reanalysis's evaluation against the observations.}\label{fig:Appendix_hydrograph_grdc}
\end{figure}
Fig.~\ref{fig:GRDC_stations} gives an overview of the KGE values for EFAS reanalysis data against the GRDC observation at the locations of all $1{,}971$ selected stations. In general, there is a good agreement between EFAS and GRDC data over Europe, with a median KGE at $0.65$ and an interquartile range of $0.50$ to $0.75$. Fig.~\ref{fig:Appendix_hydrograph_grdc} shows an exemplar hydrograph for a gauged station. 

It should be noted that, unlike the GRDC observations, the EFAS reanalysis represents naturalized flow and does not account for anthropogenic influences like dams, reservoirs, irrigation withdrawals, and other water-management practices.

\subsubsection{CAMELS}
\label{sec:Appendix_CAMELS_data}

In addition to GRDC, we processed the CAMELS (Catchment Attributes and MEteorology for Large-sample Studies) as a separate dataset. 
CAMELS dataset comes from different sources from various national agencies and research projects.
All CAMELS datasets have daily data and only few have hourly data e.g., CAMELS-DE \citepsupp{camels_de}. To make the dataset compatible, we use the daily river discharge from all CAMELS datasets.

We combine $10$ national CAMELS datasets, following the same matching procedure used for GRDC (Sec.~\ref{sec:Appendix_GRDC_data}). Each station was discarded if its drainage area was below $150$ km$^2$. For matching we projected the points onto the EFAS grid and evaluated its $9$ nearest grid cells, keeping the cell with the highest KGE, and then dropped if its matched cell still differed from the reported drainage area by more than $20\%$ and scored below $0.4$ KGE. We excluded the area matching procedure for CAMELS-ES because reanalysis has very bad quality there where EFAS reanalysis shows poor agreement with observations in Spain. This left $2{,}215$ CAMELS stations across the ten datasets which are listed in Table~\ref{tab:camels_datasets}, with a median drainage-area mismatch of $3.6\%$ (interquartile range $1.2\%$–$10.6\%$).

\begin{table}[!h]
  \centering
  \caption{CAMELS datasets used in this work. Shown are the source of data, observation period, and the number of stations retained after matching to the EFAS drainage network.}
  \label{tab:camels_datasets}
  \begin{tabular}{*{6}l}
    \toprule
    Dataset & Source & Country & Start & End & \# Stations \\
    \midrule
    CAMELS-CH  & \citesupp{camels_ch} & Switzerland    & 1981 & 2020 & 136 \\
    CAMELS-CZ  & \citesupp{camels_cz} & Czechia        & 1978 & 2024 & 173 \\
    CAMELS-DE  & \citesupp{camels_de} & Germany        & 1951 & 2020 & 605 \\
    CAMELS-DK  & \citesupp{camels_dk} & Denmark        & 1989 & 2019 & 54  \\
    CAMELS-ES  & \citesupp{camels_es} & Spain          & 1991 & 2020 & 244 \\
    CAMELS-FI  & \citesupp{camels_fi} & Finland        & 1961 & 2023 & 234 \\
    CAMELS-FR  & \citesupp{camels_fr} & France         & 1970 & 2021 & 368 \\
    CAMELS-GB  & \citesupp{camels_gb} & Great Britain  & 1970 & 2022 & 332 \\
    CAMELS-LUX & \citesupp{camels_lux} & Luxembourg     & 2004 & 2021 & 25  \\
    CAMELS-SE  & \citesupp{camels_se} & Sweden         & 1961 & 2020 & 44  \\
    \midrule
    Total\textsuperscript{*}      &  &              &      &      & 2{,}206 \\
    \bottomrule
  \end{tabular}
  \\[2pt]
  \footnotesize\textsuperscript{*}Row counts sum to 2{,}215; 9 grid cells are shared between neighboring countries' stations and are counted once in the combined mask used for training/evaluation.
\end{table}

\begin{figure}[!h]
  \centering
  \includegraphics[width=.99\textwidth]{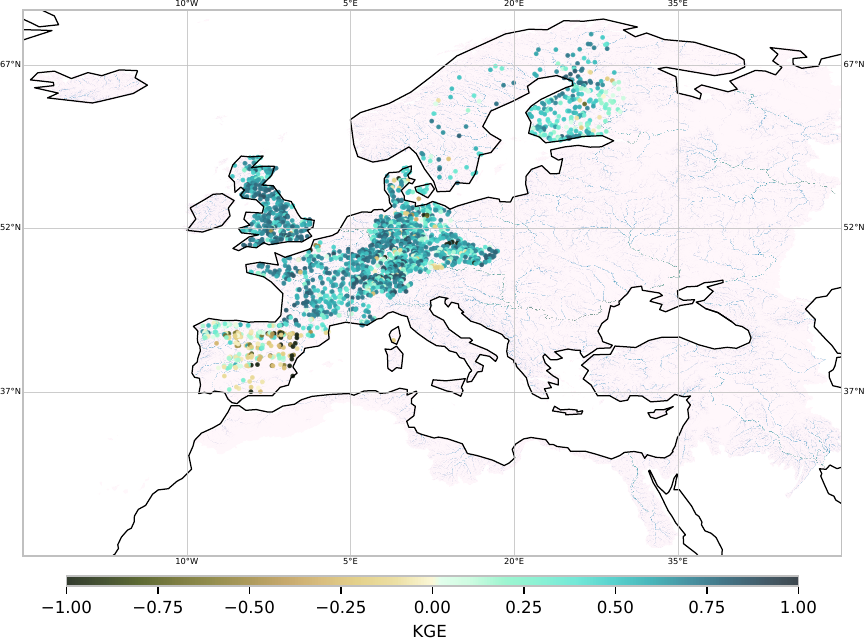}
  \caption{Locations of the $2{,}240$ selected CAMELS stations used to train and evaluate the model. Color encodes the KGE of the EFAS reanalysis discharge against the CAMELS observations at each station.}\label{fig:CAMELS_stations}
\end{figure}

\begin{figure}[!h]
  \centering
  \includegraphics[width=.99\textwidth]{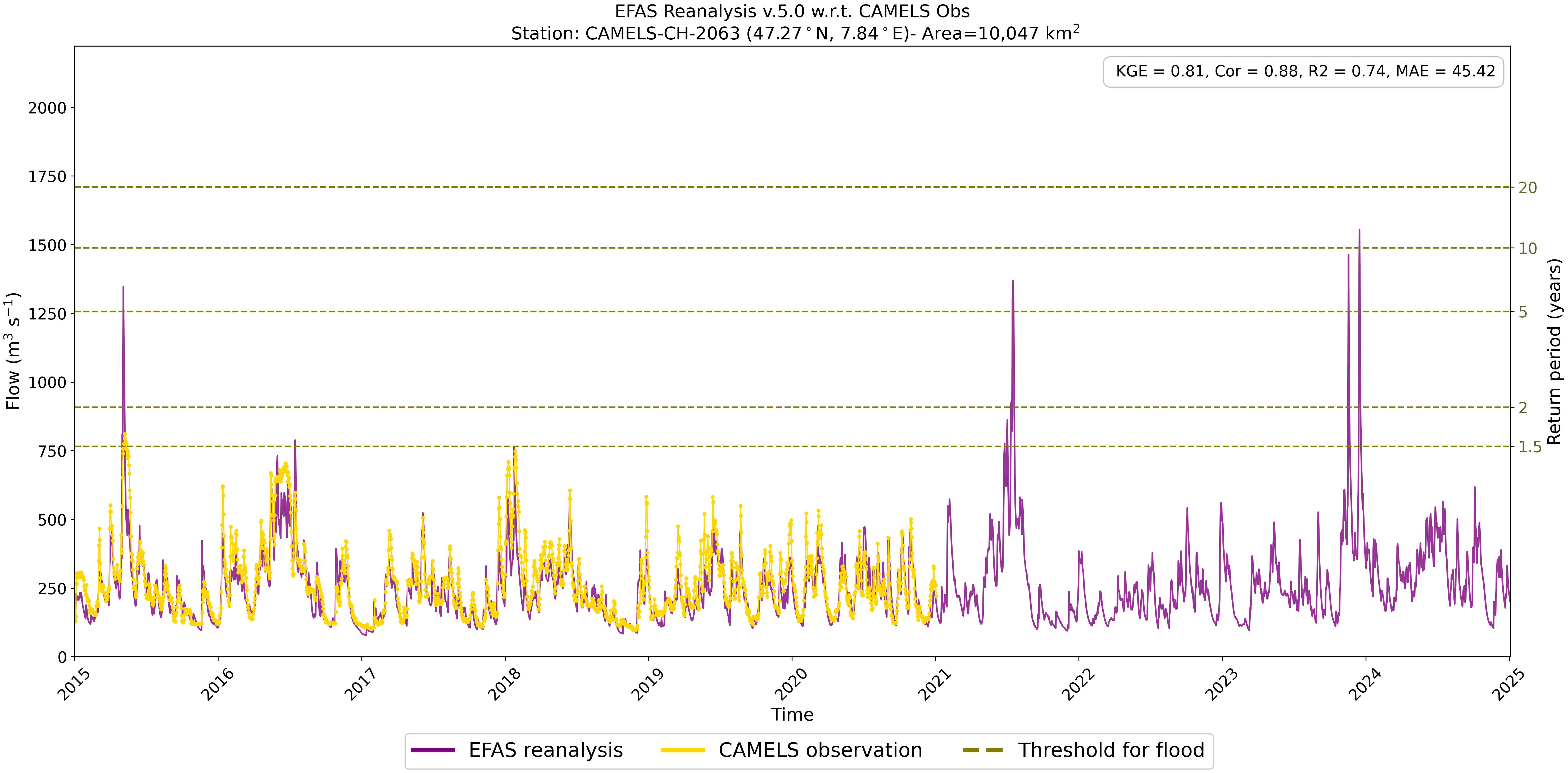}
  \caption{Hydrograph for EFAS reanalysis (purple line) from 1 January 2015 to 31 December 2024 and observations (gold line), for a gauging station in Switzerland (CAMELS-CH-2063). The top-right box displays summary statistics from the reanalysis's evaluation against the observations.}\label{fig:CAMELS_CH_2063}
\end{figure}

Fig.~\ref{fig:CAMELS_stations} maps the KGE between EFAS reanalysis and CAMELS observations at all $2{,}215 $ stations. 
In general, there is a good agreement between EFAS and CAMELS data over Europe (except Spain), with a median KGE at $0.54$ and an interquartile range of $0.18$ to $0.69$. 
Fig.~\ref{fig:CAMELS_CH_2063} shows a representative hydrograph for one gauged station.

\subsection{IFS-HRES data}
\label{sec:Appendix_IFS_HRES_data}

We get operational weather forecast data from the ECMWF Archive Catalogue (\colorh{\url{https://www.ecmwf.int/en/forecasts/dataset/operational-archive}}).
The forecast is produced by the deterministic forecast of the ECMWF Integrated Forecast System (IFS) High Resolution (HRES) atmospheric model.
The forecast runs once per day at $00$:$00$ UTC and up to $10$ days lead time with $6$-hourly temporal resolutions and $41$ time steps (step $0$ is analysis, see Sec.~\ref{sec:Appendix_analysis_data}).
We processed data from $2007$ to $2025$. The data before $2007$ were replaced by ERA5-Land data (Sec.~\ref{sec:Appendix_ERA5_Land_data}).
The data has at $3$ spatial resolutions, ~$25$ KM from $2007$-$01$-$01$ to $2010$-$01$-$26$, ~$16$ KM from $2010$-$01$-$27$ to $2016$-$03$-$08$, and ~$9$ KM from $2016$-$03$-$09$ to $2026$-$01$-$01$ at $0.1^\circ \times 0.1^\circ$ resolution.
To match the resolution of EFAS ($1$ arcminute), we precompute the nearest neighbored point from IFS-HRES to EFAS points for these $3$ different resolutions. During training/inference, we used the precomputed nearest neighborhood mappings to get IFS-HRES forecast data at EFAS resolution.
The variables include $6$ instantaneous variables and $8$ accumulated variables.
Details about the processed variables are available in Table \ref{table:era5-land_ifs}.

\subsection{Analysis data}
\label{sec:Appendix_analysis_data}

Near real-time operational analysis data are used as initial condition and obtained from step $0$ of IFS-HRES (Sec.~\ref{sec:Appendix_IFS_HRES_data}).
Analysis data only provide instantaneous variables, to have the accumulated variables for analysis, we use the forecast of the previous day at step $4$.
The processed variables are the same as IFS-HRES (Sec.~\ref{sec:Appendix_IFS_HRES_data}) and described in Table \ref{table:era5-land_ifs}.

\subsection{ERA5-Land reanalysis data}
\label{sec:Appendix_ERA5_Land_data}

ERA5-Land reanalysis provides a reconstruction of the historical land surface state \citepsupp{ERA5-Land}.
We do not use ERA5-Land as initial condition. ERA5-Land is only used to substitute the forecast and analysis of IFS-HRES data (Sec.~\ref{sec:Appendix_IFS_HRES_data}) for the years $1992$-$2006$ in the initital phase of the training.
We processed the data from the Copernicus Climate Change Service (C3S) Climate Data Store (CDS) (\colorh{\url{https://doi.org/10.24381/cds.e2161bac}}). 
ERA5-Land is provided at $0.1^\circ \times 0.1^\circ$ at regular latitude and longitude (Plate Carr\'ee projection). 
To match the resolution of EFAS ($1$ arcminute), we precompute the nearest neighbored point from ERA5-Land to EFAS points and during training/inference, we used the precomputed nearest neighborhood mapping to get ERA5-Land data at EFAS resolution.
The processed variables are the same as IFS-HRES (Sec.~\ref{sec:Appendix_IFS_HRES_data}) and described in Table \ref{table:era5-land_ifs}.
More details about ERA5-Land can be found in \citesupp{ERA5-Land}. 

\begin{table}[h!]
\small
 \caption{Details about the processed variables from the ECMWF Integrated Forecast System (IFS) High Resolution (HRES) atmospheric model (Sec.~\protect\ref{sec:Appendix_IFS_HRES_data}) and ERA5-Land reanalysis (Sec.~\protect\ref{sec:Appendix_ERA5_Land_data}). IFS-HRES accumulated state variables start at $00$:$00$ UTC of the forecast time and ends at the end of the forecast. ERA5-Land accumulation starts at $00$:$00$ UTC previous day and ends at $00$:$00$ UTC current day.}
 \label{table:era5-land_ifs}
 \centering
 \setlength\tabcolsep{2.5pt}
 \begin{tabular}{*{5}l}
    \toprule
        Variable & Long name & Unit & Height & Surface parameters  \\
    \midrule
        d2m    & 2m dewpoint temperature & K & 2m & instantaneous\\
        e      & total evaporation & m of water equivalent& surface & accumulated\\
        ro   & runoff & m & surface \& subsurface & accumulated\\
        sd     & snow depth & m of water equivalent & surface & accumulated\\
        sf     & snowfall & m of water equivalent & surface & accumulated\\
        smlt   & snowmelt & m of water equivalent & surface & accumulated\\
        sp     & surface pressure & Pa & surface & instantaneous \\
        ssr    & surface net solar radiation & J/m$^{2}$ & surface & accumulated\\
        str    & surface net thermal radiation & J/m$^{2}$ & surface & accumulated\\        
        swvl1  & volumetric soil water & m$^{3}$/m$^{3}$ & soil layer (0 - 7 cm) & instantaneous \\
        t2m   & 2m temperature & K & 2m & instantaneous \\
        tp    & total precipitation & m & surface & accumulated \\
        u10   & 10 metre U wind component & m/s & 10m & instantaneous \\
        v10   & 10 metre V wind component & m/s & 10m & instantaneous \\

\bottomrule
\end{tabular}
\end{table}

\subsection{EMO data}

European Meteorological Observations (EMO) are gridded high-resolution data obtained from historical and real-time ground weather stations across Europe. EMO data are used originally to provide the EFAS with near real-time initial condition. We also use EMO as initial condition for our model with $2$ days latency to mimic the operational setting (EMO data is updated annually for public use but real-time EMO data can be obtained with a close collaboration with ECMWF). We processed the data daily at $1$ arcminute spatial resolution for the years $1990$ to $2025$. The processed variables are described in Table \ref{table:emo}. The data for can be obtained from \colorh{\url{https://jeodpp.jrc.ec.europa.eu/ftp/jrc-opendata/CEMS-EFAS/meteorological_forcings/EMO-1arcmin/}}.

\begin{table}[h!]
\small
 \caption{Details about the processed variables from EMO data \protect\citepsupp{thiemig2022emo5}.}
 \label{table:emo}
 \centering
 \setlength\tabcolsep{2.5pt}
 \begin{tabular}{*{5}l}
    \toprule
        Variable & Long name & Unit & Height & Surface parameters  \\
    \midrule
        pd   & water vapour pressure & hPa & - & instantaneous \\
        pr    & total precipitation & mm & surface & accumulated \\
        rg   & downward surface solar radiation & J/m$^{2}$ & surface & accumulated \\
        tn   & minimum air temperature & C$^{\circ}$ & 2m & instantaneous \\
        tx   & maximum air temperature & C$^{\circ}$ & 2m & instantaneous \\
        ws   & wind speed & m/s & 10m & instantaneous \\
\bottomrule
\end{tabular}
\end{table}

\subsection{IMERG data}

We obtain observational precipitation estimates from NASA’s Integrated Multi-satellitE Retrievals for GPM (IMERG V07) product~\citepsupp{IMERG}.
IMERG data are used only as initial condition.
We use the near real-time early run data of processed mean daily precipitation at $0.1^\circ \times 0.1^\circ$ spatial resolution.
To match the resolution of EFAS data, we regridded IMERG data onto the EFAS domain using nearest point algorithm (nearest\_s2d) implemented by~\citetsupp{xESMF}.
The daily mean rate in mm/day is derived by computing the mean precipitation rate in mm/hour for every grid cell of the respective day, and then multiplying the result by $24$ to get mm/day.
IMERG is available from the year $1998$. Since we start the training from year $1992$, we substitute IMERG data for the years $1992$-$1998$ with the total precipitation (tp) of ERA5-Land (Sec.~\ref{sec:Appendix_ERA5_Land_data}) converted from m to accumulated mm/day.
The IMERG data for operational forecast can be retrieved from \colorh{\url{https://psl.noaa.gov/data/gridded/data.cpc.globalprecip.html}} with $\sim4$ hours latency.

\subsection{CPC data}

We obtained the observational gauged-based precipitation data from the National Oceanic and Atmospheric Administration (NOAA), Climate Prediction Center (CPC).
The data are accumulated daily at $0.5^\circ \times 0.5^\circ$ spatial resolution.
To match the resolution of EFAS data, we regridded CPC data onto the EFAS domain using nearest point algorithm (nearest\_s2d) implemented by~\citetsupp{xESMF}.
CPC data are only used as initial condition. The CPC data for operational use can be retrieved from \colorh{\url{https://psl.noaa.gov/data/gridded/data.cpc.globalprecip.html}} with $\sim1$ day latency.
More details about the data can be found in~\citetsupp{xie2007gauge, chen2008assessing, xie2010cpc}.

\subsection{Static data}

In addition to AlphaEarth/LISFLOOD static features, we also use the Cartesian coordinates for the points on the WGS-84 ellipsoid as positional encoding in the way as~\citetsupp{RiverMamba}.

\subsubsection{AlphaEarth}

\begin{figure}[!h]
  \centering
  \includegraphics[width=.98\textwidth]{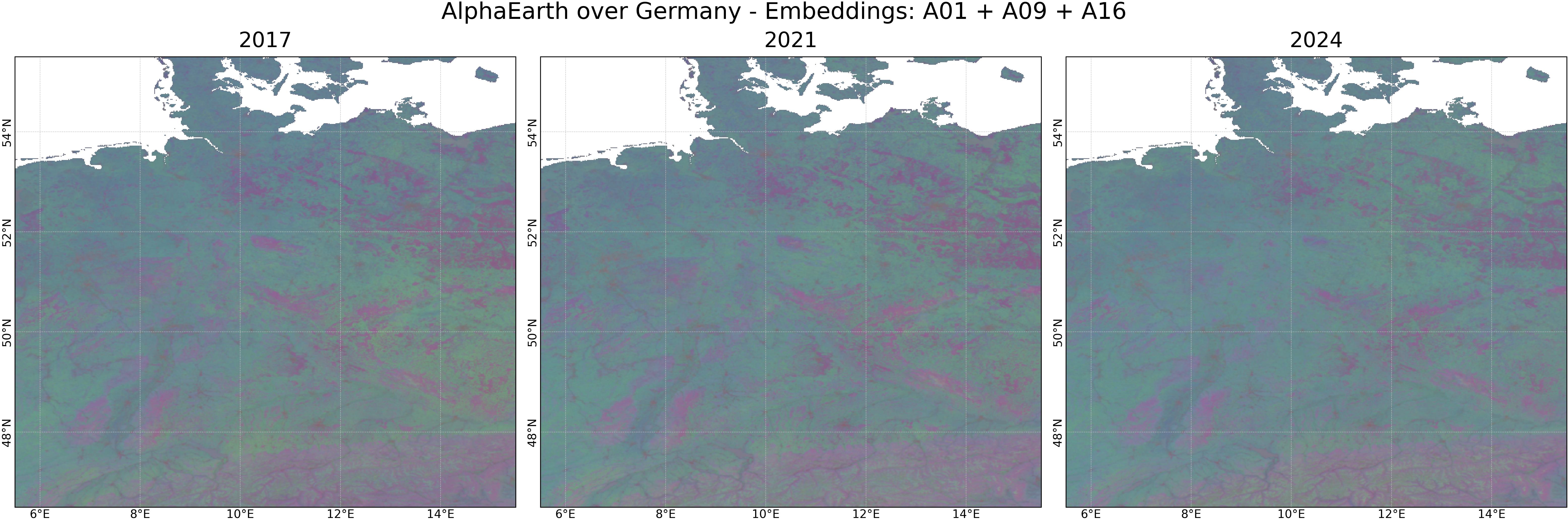}
  \caption{AlphaEarth Embedding over Germany for the years $2017$, $2021$, and $2024$. Shown are Bands; A$01$, A$09$, and A$16$ as an RGB image.}\label{fig:alpha_earth_2}
\end{figure}

\begin{figure}[!h]
  \centering
  \includegraphics[width=.98\textwidth]{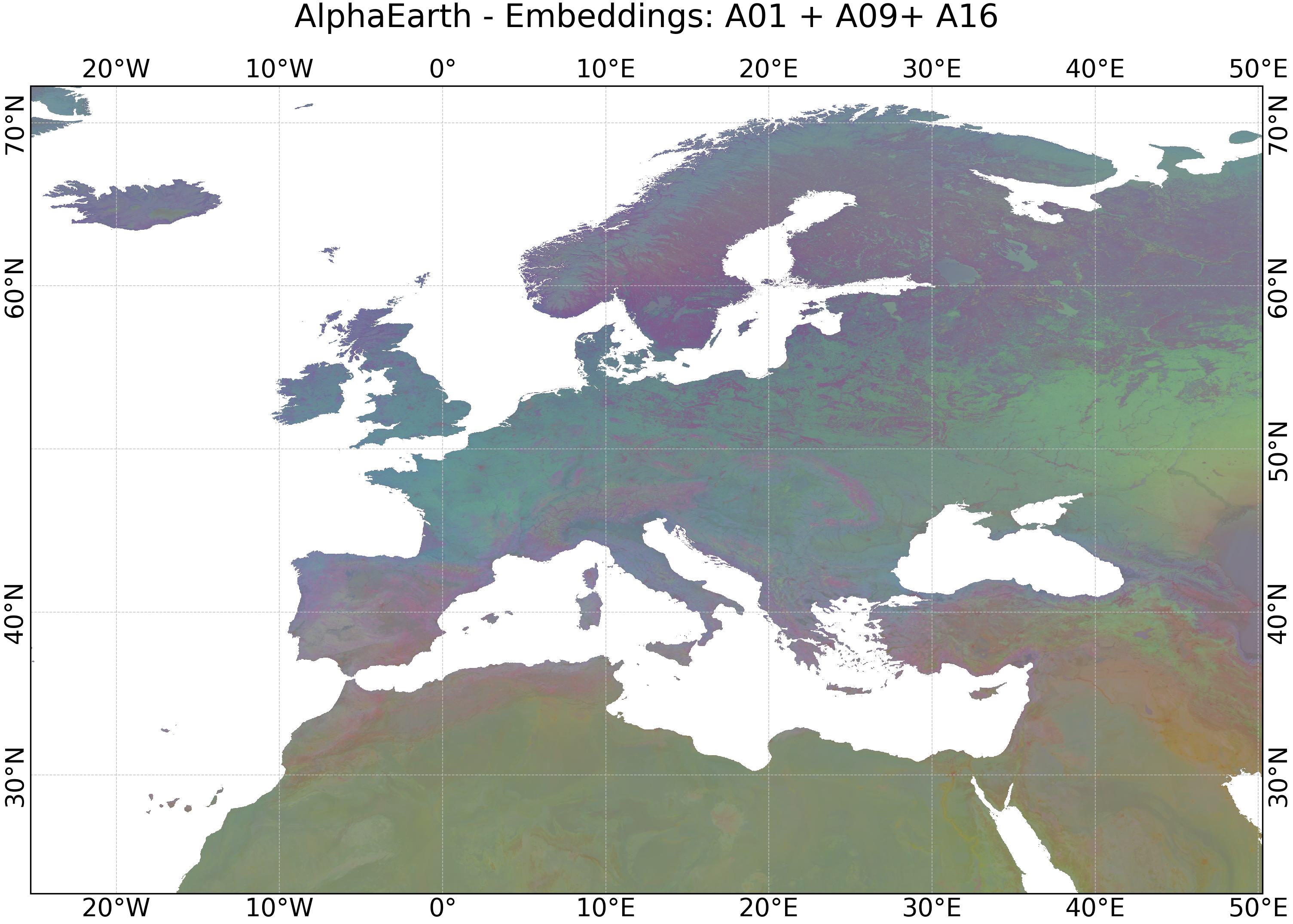}
  \caption{AlphaEarth Embedding over EFAS domain. Shown are Bands; A$01$, A$09$, and A$16$ as an RGB image.}\label{fig:alpha_earth_1}
\end{figure}

AlphaEarth embeddings provide annual representation of the Earth surface at very high resolution ($10 \times 10$ meter squares) with $64$ bands (A$01$-A$64$). The embeddings are produced by Google DeepMind and obtained by training a foundation model on multi-source Earth observations \citepsupp{AlphaEarth}. To match the resolution of EFAS ($1$ arcminute), we first preprocessed the AlphaEarth data on Google Engine by taking the mean embeddings at $1$ KM resolution (processed in small tiles). Then we generate a mosaic of these tiles and map it to EFAS grid via bilinear interpolation. Since the data only span the years $2017$ to $2024$ (at the date of processing), we average over the time dimension to get the static embeddings. Fig.~\ref{fig:alpha_earth_2} shows the embeddings for $3$ years over Germany and Fig.~\ref{fig:alpha_earth_1} shows the processed data over EFAS domain. The raw data can be obtained from \colorh{\url{https://developers.google.com/earth-engine/datasets/catalog/GOOGLE_SATELLITE_EMBEDDING_V1_ANNUAL?utm_source=deepmind.google&utm_medium=referral&utm_campaign=gdm&utm_content}}.

\subsubsection{LISFLOOD}

\label{sec:Appendix_LISFLOOD_static_data}

We used LISFLOOD static maps from~\citetsupp{LISFLOOD_static}. The data are provided at the same resolution of EFAS domain ($1$ arcminute). For the experiments, we used $119$ time-invariant variables covering $7$ main categories and excluding reservoirs, lakes and some static water demand variables. The EFAS LISFLOOD static maps can be obtained from the Joint Research Centre Data Catalogue at \colorh{\url{https://data.jrc.ec.europa.eu/dataset/f572c443-7466-4adf-87aa-c0847a169f23}}.

\section{Code and data availability}
\label{sec:appendix_code_data_availability}
We will release the datasets over mid Europe, reforecast, and the code upon publication. The source code will include instructions on how to access and use the data. 
GRDC data that we used in this study is available after signing a license agreement with the data provider at \url{https://grdc.bafg.de/}. For CAMELS data, see Sec.~\ref{sec:Appendix_CAMELS_data} in the Appendix. 
Raw ECMWF-IFS HRES weather forecasts and analysis can be downloaded from ECMWF archive catalogue (MARS).
The raw River discharge and related historical data from the European Flood Awareness System (EFAS) can be downloaded from the Early Warning Data Store (EWDS) at \url{https://doi.org/10.24381/cds.e3458969}~\citepsupp{EFAS}. 
The raw LISFLOOD static and parameter maps can be downloaded from the European Commission, Joint Research Centre (JRC) at \url{https://data.jrc.ec.europa.eu/dataset/f572c443-7466-4adf-87aa-c0847a169f23}~\citepsupp{LISFLOOD_static}. 
The raw EMO data can be downloaded from the European Commission, Joint Research Centre (JRC) at \url{EMO http://data.europa.eu/89h/0bd84be4-cec8-4180-97a6-8b3adaac4d26}. 
The raw CPC data can be obtained from the National Oceanic and Atmospheric Administration (NOAA), Climate Prediction Center (CPC) at \url{https://psl.noaa.gov/data/gridded/data.cpc.globalprecip.html}.
IMERG data can be downloaded from NOAA at \url{https://psl.noaa.gov/data/gridded/data.cpc.globalprecip.html}. 
The raw ERA5-Land reanalysis data can be downloaded from NOAA at Copernicus Climate Change Service (C3S), Climate Data Store (CDS) at \url{https://doi.org/10.24381/cds.e2161bac}~\citepsupp{ERA5-Land}.

{
\bibliographystylesupp{iclr2027_conference}
\bibliographysupp{ref}
}

\end{document}